\documentclass[10pt,twocolumn,letterpaper]{article}

\usepackage{cvpr}              

\usepackage{algorithm}
\usepackage{algorithmic}
\usepackage{amsmath}
\usepackage{amssymb}
\usepackage{mathtools}
\usepackage{amsthm}
\DeclareMathOperator*{\argmax}{arg\,max}
\DeclareMathOperator{\sign}{sign}  
\newtheorem{definition}{Definition}
\newtheorem{theorem}{Theorem} 
\usepackage{subcaption}
\usepackage{etoc}
\etocdepthtag.toc{mtchapter}
\etocsettagdepth{mtchapter}{subsection}
\etocsettagdepth{mtappendix}{none}
\usepackage{tabularx}

\usepackage{colortbl} 
\usepackage{xcolor}   

\usepackage{microtype}
\usepackage{graphicx}
\usepackage{booktabs}
\usepackage{balance}
\definecolor{cvprblue}{rgb}{0.21,0.49,0.74}
\usepackage[pagebackref,breaklinks,colorlinks,allcolors=cvprblue]{hyperref}

\def\paperID{*****} 
\def\confName{CVPR}
\def\confYear{2026}

\newcommand{\epsball}{\mathcal{B}_\epsilon}

\title{A Unified Perspective on Adversarial Membership Manipulation in Vision Models}

\author{
Ruize Gao$^{1}$, Kaiwen Zhou$^{2,3}$, Yongqiang Chen$^{3}$, Feng Liu$^{4}$\\
$^{1}$National University of Singapore, CNRS@CREATE\quad
$^{2}$Knowin AI\\
$^{3}$The Chinese University of Hong Kong\quad
$^{4}$The University of Melbourne\\
\texttt{ruizegao@u.nus.edu}\hspace{0.5cm}
\texttt{kwzhou@cse.cuhk.edu.hk}
\\
\texttt{\normalsize{\url{https://github.com/Sjtubrian/Adversarial_Membership_Manipulation}}}
}

\begin{document}
\maketitle

\begin{abstract}
Membership inference attacks (MIAs) aim to determine whether a specific data point was part of a model’s training set, serving as effective tools for evaluating privacy leakage of vision models. However, existing MIAs implicitly assume honest query inputs, and their adversarial robustness remains unexplored. We show that MIAs for vision models expose a previously overlooked adversarial surface: adversarial membership manipulation, where imperceptible perturbations can reliably push non-member images into the “member’’ region of state-of-the-art MIAs. In this paper, we provide the first unified perspective on this phenomenon by analyzing its mechanism and implications. We begin by demonstrating that adversarial membership fabrication is consistently effective across diverse architectures and datasets. We then reveal a distinctive geometric signature—a characteristic gradient-norm collapse trajectory—that reliably separates fabricated from true members despite their nearly identical semantic representations. Building on this insight, we introduce a principled detection strategy grounded in gradient-geometry signals and develop a robust inference framework that substantially mitigates adversarial manipulation. Extensive experiments show that fabrication is broadly effective, while our detection and robust inference strategies significantly enhance resilience. This work establishes the first comprehensive framework for adversarial membership manipulation in vision models.
\end{abstract}

\begin{figure*}[!t]
    \begin{center}
    {\includegraphics[width=0.99\textwidth]{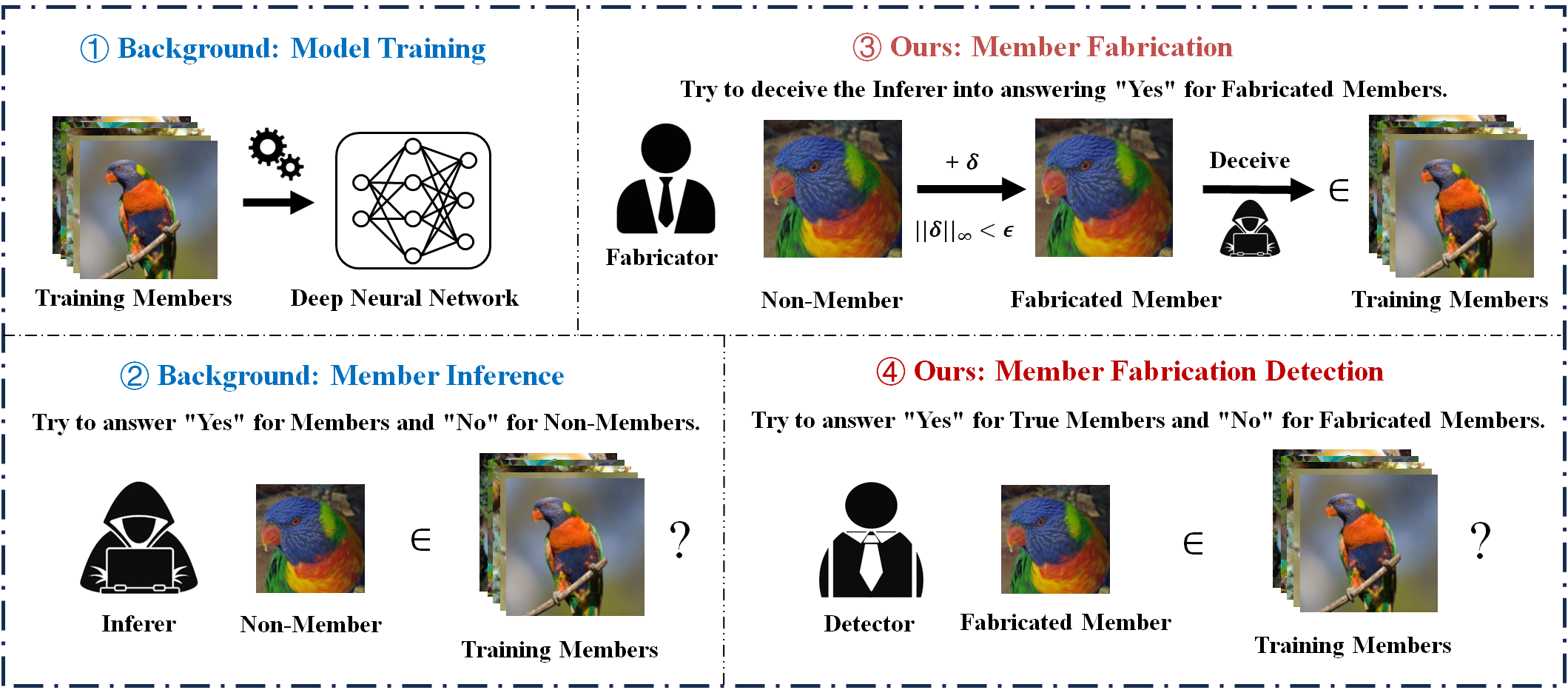}}
    \caption{\footnotesize Overview of the Background and Our Proposed Research Problems.}
    \label{flowchart}
    \end{center}
    \vspace{-1em}
\end{figure*}

\section{Introduction}

\textbf{Membership inference in vision models.}
As large-scale vision models become widely deployed in sensitive domains such as medical imaging, surveillance, and autonomous driving, concerns about whether these models inadvertently leak their training data have grown substantially. \emph{Membership Inference Attacks} (MIAs) aim to determine whether a particular data point was part of a model's training set—a problem first studied in genomic data analysis~\citep{Homer2008Resolving} and later extended to machine learning models. Modern MIAs~\citep{watson2021importance, ye2022enhanced, carlini2022membership, zarifzadeh2024low} operate through carefully designed test statistics, ranging from per-sample loss~\citep{Yeom2018Privacy} and shadow-model logits~\citep{shokri2017membership}, to likelihood-based estimates and calibrated ratio tests~\citep{carlini2022membership}. These methods now form the backbone of privacy auditing pipelines and are widely used to assess memorization and information leakage in large-scale visual learning systems.

\vspace{0.2em}
\noindent\textbf{A missing robustness dimension.}
Despite significant progress in attack design and statistical modeling, existing MIAs implicitly assume that the query image is \emph{benign}. This assumption contrasts sharply with the adversarial learning literature, where imperceptible perturbations are known to drastically alter model predictions. Yet, no prior study has asked a critical question: \emph{Are MIAs themselves robust when adversaries manipulate their query inputs?} This question is crucial because MIAs are not merely academic tools—they underpin practical auditing tasks such as copyright verification, dataset provenance tracking, and regulatory compliance for “right-to-be-forgotten’’ policies. If an adversary can adversarially manipulate MIA queries, then the entire auditing process risks producing misleading or even legally invalid results. Although RMIA~\citep{zarifzadeh2024low} briefly discusses robustness against out-of-distribution (OOD) non-members, stating that “a strong MIA should be able to rule out all non-members,” it does not consider \emph{adversarially fabricated} in-distribution queries, which expose a much deeper vulnerability. Several representative scenarios are illustrated in Appendix \ref{Asec:examples}.

\vspace{0.2em}
\noindent\textbf{A previously overlooked adversarial surface.}
We identify and formalize this new threat surface as \emph{adversarial membership manipulation}, where imperceptible input perturbations reliably push non-member images into the “member’’ region of existing MIAs. Unlike traditional adversarial attacks that alter class predictions, this attack alters a model’s \emph{training-set attribution}. It directly undermines the integrity of privacy audits, test-set decontamination checks, and forensic model accountability.
Compared with prior output-space perturbation defenses such as \textsc{MemGuard}~\citep{jia2019memguard}, which modify model logits to protect privacy, our study focuses on \emph{input-space manipulations}. These two directions are orthogonal: \textsc{MemGuard} defends the model, whereas we show that the auditing mechanism itself can be attacked.

\vspace{0.2em}
\noindent\textbf{A unified geometric perspective.}
Through theoretical analysis and empirical validation, we find that these perturbed non-members, \emph{fabricated members}, share a distinct optimization geometry: their input-gradient norms progressively collapse as perturbations drive them toward regions of abnormally high confidence that MIAs interpret as evidence of training-set membership. This \emph{gradient-norm collapse trajectory} serves as a reliable geometric fingerprint separating fabricated from true members, even when their semantic embeddings are nearly identical. This discovery uncovers a fundamental vulnerability in current MIAs: their reliance on semantic confidence signals (e.g., loss, likelihood ratio) makes them susceptible to perturbations that exploit the adversarial geometry of the input space.

Building on these insights, we propose the first unified framework for analyzing and mitigating adversarial membership manipulation in vision models:
\begin{itemize}[leftmargin=1.5em]
\item \emph{Member Fabrication Attack} (\textbf{MFA}): an input-space attack that fabricates membership by imperceptibly perturbing non-members to deceive existing MIAs (\S\ref{subsec:mfa}).
\item \emph{Member Fabrication Detection} (\textbf{MFD}): a detection method that leverages gradient-geometry statistics to distinguish fabricated from true members (\S\ref{subsec:mfd}).
\item \emph{Adversarially Robust MIAs} (\textbf{AR-MIAs}): an inference scheme that incorporates gradient-geometry statistics into existing MIAs to defend against fabrication (\S\ref{subsec:armia}).
\end{itemize}

\vspace{0.2em}
\noindent\textbf{Contributions.}  
We summarize our contributions as follows:
\begin{itemize}[leftmargin=1.5em]
\item \textbf{I.} We present the identification of a previously undocumented adversarial vulnerability, \emph{adversarial membership manipulation}, in vision-model auditing.
\item \textbf{II.} We provide a unified geometric explanation grounded in the \emph{gradient-norm collapse} phenomenon, revealing how adversarial perturbations drive non-members into high-confidence, low-gradient regions.
\item \textbf{III.} We propose principled detection and robust-inference strategies that improve resilience while remaining compatible with existing MIA designs.
\end{itemize}

\begin{figure*}[!t]
    \begin{center}
    {\includegraphics[width=0.16\textwidth]{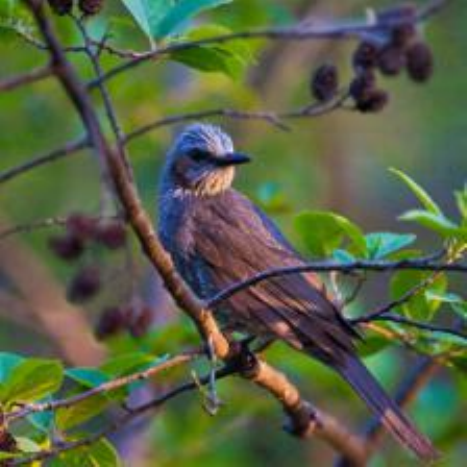}}
    {\includegraphics[width=0.16\textwidth]{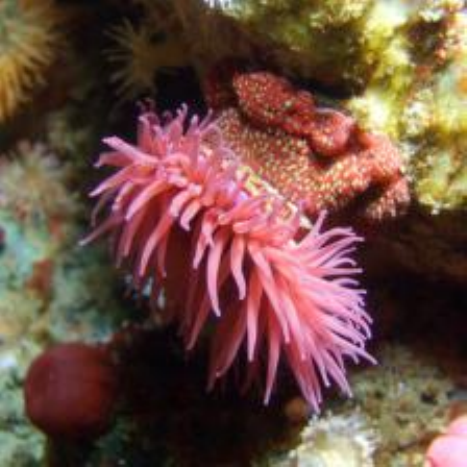}}
    {\includegraphics[width=0.16\textwidth]{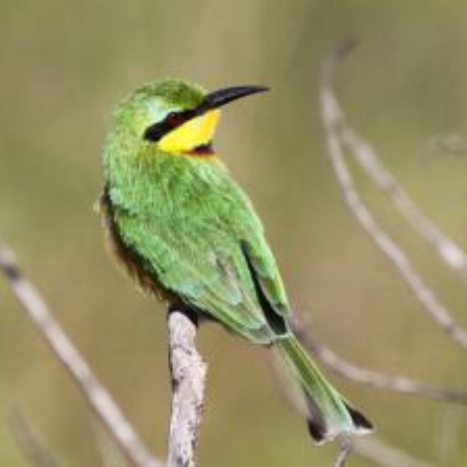}}
    {\includegraphics[width=0.16\textwidth]{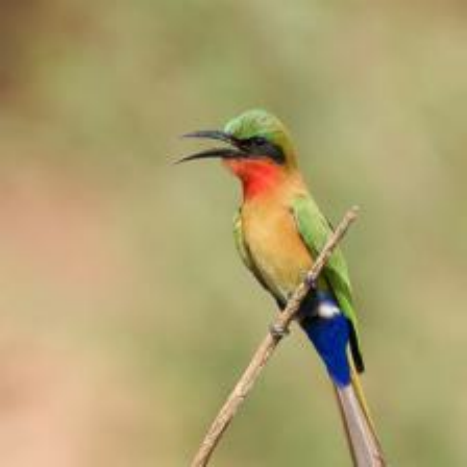}}
    {\includegraphics[width=0.16\textwidth]{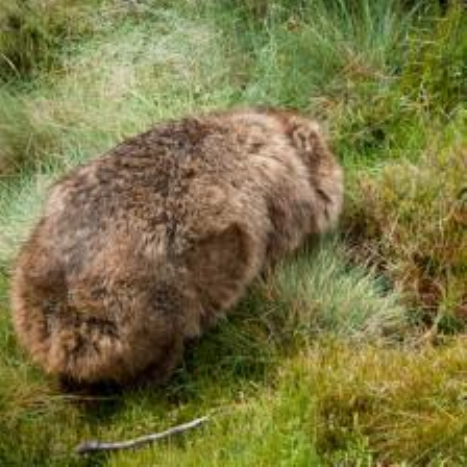}}
    {\includegraphics[width=0.16\textwidth]{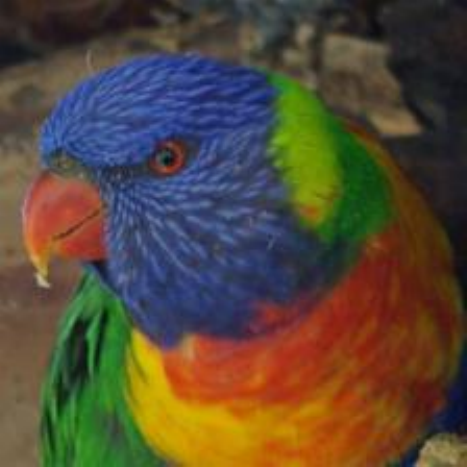}}
    {\includegraphics[width=0.16\textwidth]{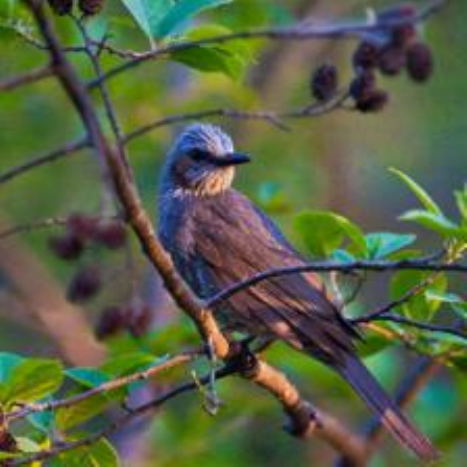}}
    {\includegraphics[width=0.16\textwidth]{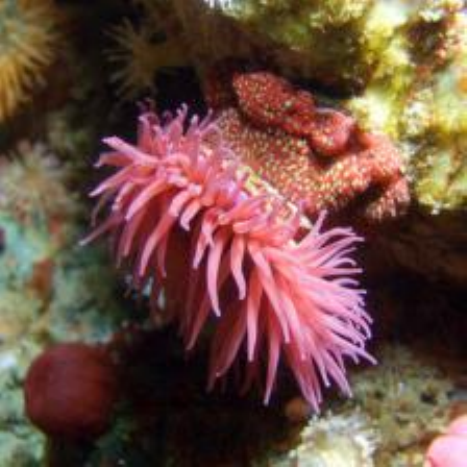}}
    {\includegraphics[width=0.16\textwidth]{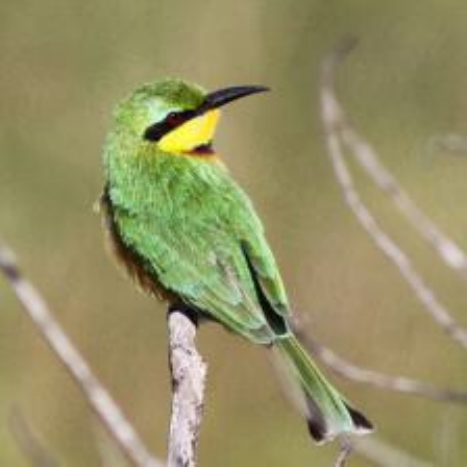}}
    {\includegraphics[width=0.16\textwidth]{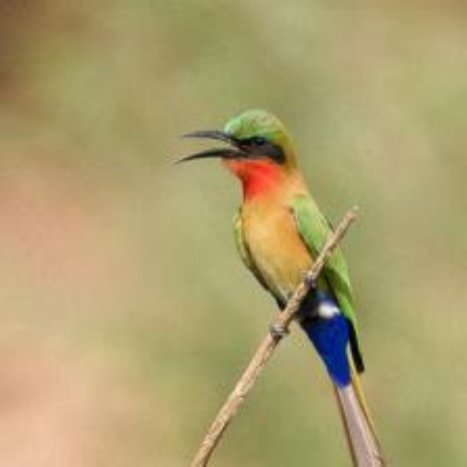}}
    {\includegraphics[width=0.16\textwidth]{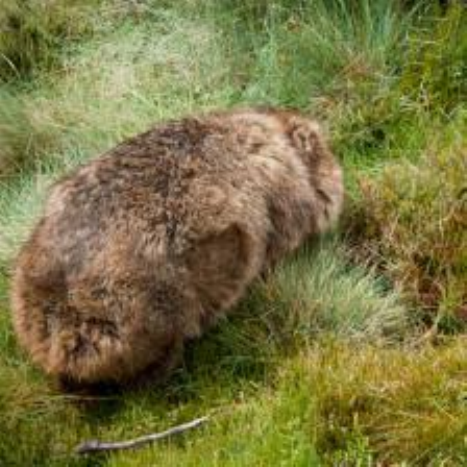}}
    {\includegraphics[width=0.16\textwidth]{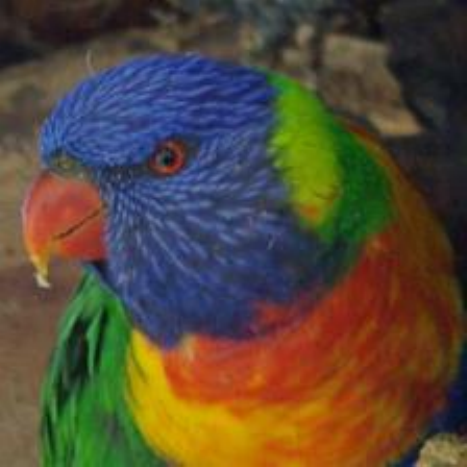}}
    \caption{\footnotesize Imperceptible Adversarial Perturbations on  \textbf{ImageNet-100}. The first row are the original non-members, and the second row are the corresponding perturbed fabricated members. We used $\epsilon = 2/255$ for $\epsball[x]$ here. The perturbations are extremely imperceptible to the human eye, which demonstrates that the Member Fabrication Attack (\textbf{MFA}) can be successful with only the addition of very small perturbations.
    }
    \label{main_adv_show_3}
    \end{center}
    \vspace{-1em}
\end{figure*}

\section{Preliminary}
\label{sec:preliminaries}

This section formalizes the setting of image classification, membership inference, and adversarial perturbations. We follow the standard \emph{security-game formulation} from the privacy and security literature~\cite{shokri2017membership,ye2022enhanced,carlini2022membership}, and extend it to vision models to establish a rigorous foundation for adversarial membership manipulation.

\subsection{Classification}
We focus on \emph{image classification}, one of the most common tasks in machine learning.
A neural network $f_\theta: \mathcal{X}\to[0,1]^n$ maps an image $x\in\mathcal{X}$ to a probability distribution over $n$ classes.
Let $p_y=f_\theta(x)_y$ denote the probability assigned to the true label $y$, also referred to as the model’s \emph{confidence} on $x$. Given a dataset $D$ sampled from distribution $\mathbb{D}$, we denote $f_\theta\gets\mathcal{T}(D)$ to indicate that parameters $\theta$ are obtained by training algorithm $\mathcal{T}$ (e.g., SGD):
\begin{equation}
\theta_{i+1} = \theta_i - \eta \sum_{(x,y)\in B} \nabla_\theta \ell(f_\theta(x),y),
\label{eq:grad_descent}
\end{equation}
where $\eta$ is the learning rate, $B$ is a minibatch, and $\ell$ is the cross-entropy loss.

\subsection{Membership Inference Attacks}
\label{background_MIAs}

\emph{Membership Inference Attacks} (MIAs) aim to determine whether a specific data point was part of a model’s training dataset.
They are a cornerstone problem in the \emph{machine learning security and privacy} literature~\cite{Homer2008Resolving,shokri2017membership,Yeom2018Privacy,ye2022enhanced,carlini2022membership,zarifzadeh2024low}, providing essential tools for auditing potential privacy leakage of deep models.
To ensure comparability and conceptual alignment with prior studies, we follow the canonical \emph{security-game formulation} used in this domain.

\begin{definition}[Membership Inference Game]
\label{def:MIA-game}
The game involves two participants: a \emph{Challenger} $\mathbf{C}$ and an \emph{Inferer} $\mathbf{I}$, and proceeds as follows:
\begin{enumerate}
    \item $\mathbf{C}$ samples a training dataset $D \leftarrow \mathbb{D}$ and trains a model $f \leftarrow \mathcal{T}(D)$ on $D$.
    \item $\mathbf{C}$ flips a bit $b$: if $b = 0$, it samples a fresh challenge point $(x, y) \leftarrow \mathbb{D}$ such that $(x, y) \notin D$; otherwise, it selects $(x, y) \leftarrow D$.
    \item $\mathbf{C}$ sends $(x, y)$ to $\mathbf{I}$.
    \item $\mathbf{I}$ has query access to $\mathbb{D}$ and $f$, and outputs a prediction $\hat{b} \leftarrow \mathbf{I}_{\mathbb{D}, f}(x, y)$.
    \item $\mathbf{I}$ succeeds if $\hat{b} = b$ and fails otherwise.
\end{enumerate}
\end{definition}
\vspace{0.2em}
\noindent\textbf{Knowledge.} The Inferer has black-box access to the model’s outputs (e.g., confidence vectors or logits) and can query from $\mathbb{D}$ to obtain additional non-member samples~\cite{shokri2017membership}.

\vspace{0.2em}
\noindent\textbf{MIAs as Hypothesis Testing.} Existing MIAs~\citep{Yeom2018Privacy, shokri2017membership, watson2021importance, ye2022enhanced, carlini2022membership, zarifzadeh2024low} can be framed as \emph{hypothesis tests}, where the design of test statistic is key to distinguishing members from non-members. For example, the \emph{loss attack} \citep{Yeom2018Privacy} uses the sample loss as the test statistic, while the \emph{Likelihood Ratio Attack} (LiRA)~\citep{carlini2022membership} uses the likelihood ratio (\Cref{app_related_MIAs}). Given a threshold~$\tau$, the Inferer predicts membership if the statistic exceeds~$\tau$:
\begin{align}
\label{MIA_base}
I(x, y) = \mathbf{1}[S(x, y) > \tau],
\end{align}
where~$\mathbf{1}$ denotes the indicator function, $S$ is the test statistic, and~$\tau$ is a tunable decision threshold.

\subsection{Adversarial Examples}
\label{subsec:adversarial_examples}

Szegedy et al.~\citep{szegedy2013intriguing} first pointed out the existence of adversarial examples: given a valid input $x$ with its true label $y$ and a trained classifier $f$, it is often possible to find another input $x'$ such that $f(x') \neq y$ yet $x$ and $x'$ are close according to some distance metric. In this paper, we focus on the $\ell_{\infty}$ distance metric. The example $x'$ is called the adversarial example. Subsequent works have refined adversarial attack strategies, such as \emph{Fast Gradient Sign Method (FGSM)} \citep{goodfellow2014explaining}, \emph{Projected Gradient Descent Attack} (PGD) \citep{Madry18PGD}, \emph{Carlini and Wagner attack} (CW) \citep{carlini2017towards}, \emph{AutoAttack} \citep{croce2020reliable}, and \emph{Minimum-Margin Attack} (MM) \citep{gao2022fast}. Among these, \citet{gao2022fast} summarized that the technical strategies in adversarial attacks should serve to generate the most adversarial example:
\begin{definition}[The most adversarial example]
Given a natural example $x$ with its true label $y$ and, the most adversarial example $x^\ast$ within $\epsball[x]$ is defined as:
\begin{align}
\label{mostadv:def_app}
{\forall} x^{\prime} \in \epsball[x],
x^\ast = \argmax_{x^{\prime}}{-(z_y{(x')} - \max_{i \neq y}{z_i{(x')}})},
\end{align}
where $\epsball[x] = \{x^{\prime} \mid d_{\infty}(x,x')\le\epsilon\}$ is the closed ball of radius $\epsilon>0$ centered at $x$; $z_y{(x')} = {f(x')}_y$; $z_i{(x')} = {f(x')}_i$.
\end{definition}

\begin{figure}[!t]
        {\includegraphics[width=0.235\textwidth]{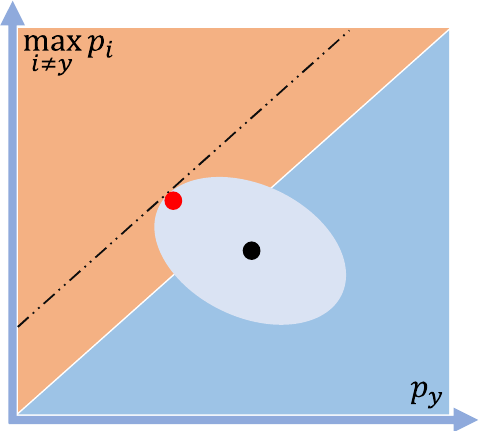}}
        {\includegraphics[width=0.235\textwidth]{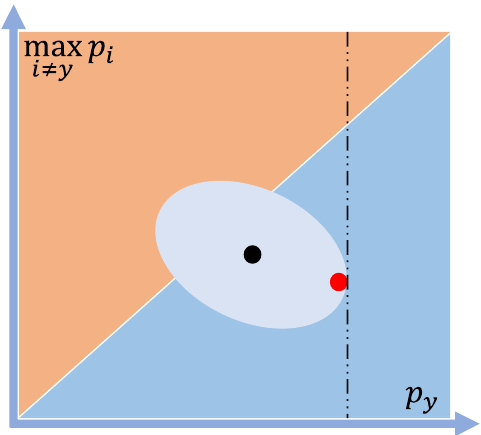}}
        \caption{\footnotesize Objective of adversarial attacks (left) vs.\ \textbf{MFA} (right). The black and red dots denote the original input and the perturbed sample within the $\epsilon$-ball (gray region). Adversarial attacks push inputs into the \emph{misclassification region} (orange), where $\max_{i\neq y} p_i > p_y$. In contrast, \textbf{MFA} drives inputs into high-confidence regions.}
        \label{motivation_adv_fabri}
    \vspace{-1em}
\end{figure}

\section{Method}
\label{sec:method}

This section presents a unified framework for analyzing and mitigating \emph{adversarial membership manipulation} in vision models. 
We begin by formalizing how imperceptible perturbations can fabricate membership and deceive existing attacks (\S\ref{subsec:mfa}), 
then identify a consistent geometric signature that enables reliable detection (\S\ref{subsec:mfd}), 
and finally incorporate this geometric insight into a robust inference mechanism that enhances resilience (\S\ref{subsec:armia}). All definitions in this section follow the standard security-game formulation widely adopted in prior security literature~\cite{shokri2017membership,ye2022enhanced,carlini2022membership}.

\subsection{Member Fabrication Attack (MFA)}
\label{subsec:mfa}
We formalize how imperceptible input perturbations can \emph{fabricate} membership and thereby deceive existing MIAs. 

\begin{definition}[Member Fabrication Attack]
\label{def:MF-game}
The game involves three participants: a Challenger~$\mathbf{C}$, a Fabricator~$\mathbf{F}$, and an Inferer~$\mathbf{I}$, and proceeds as follows:
\begin{enumerate}
    \item $\mathbf{C}$ samples a training dataset $D \leftarrow \mathbb{D}$ and trains a model $f \leftarrow \mathcal{T}(D)$ on $D$.
    \item $\mathbf{C}$ flips a bit $b$: if $b = 0$, it samples a fresh challenge point $(x, y) \leftarrow \mathbb{D}$ such that $(x, y) \notin D$; otherwise, it selects $(x, y) \leftarrow D$.
    \item When $b = 0$, $\mathbf{F}$ introduces an adversarial perturbation~$\delta$ to $x$, where $\|\delta\|_{\infty} \leq \epsilon$, resulting in a perturbed example $x' = x + \delta$; otherwise, $x' = x$.
    \item $\mathbf{F}$ sends the perturbed $(x', y)$ to $\mathbf{I}$.
    \item $\mathbf{I}$ has query access to $\mathbb{D}$ and $f$, and outputs a prediction $\hat{b} \leftarrow \mathbf{I}_{\mathbb{D}, f}(x', y)$.
    \item $\mathbf{I}$ succeeds if $\hat{b} = b$, and fails otherwise.
\end{enumerate}
\end{definition}

\noindent In this setting, the Inferer represents to an auditor or privacy verifier, while the Fabricator represents an adversary (or a malicious data provider) who slightly perturbs query inputs before the audit to manipulate membership outcomes. Several representative scenarios are illustrated in Appendix~\ref{Asec:examples}.

\vspace{0.2em}
\noindent\textbf{Fabricator's objective.}  
The Fabricator\footnote{We assume white-box access to construct the strongest adversary. When such access is unavailable, existing black-box techniques \citep{guo2019simple,ilyas2018black,bhambri2019survey} can be naturally incorporated into our framework to realize black-box variants.} aims to deceive the Inferer into predicting membership ($\hat b=1$) for a perturbed non-member $(x',y)$ while keeping $\delta$ imperceptible (see \Cref{main_adv_show_3}). In loss-based MIAs \citep{Yeom2018Privacy}, this corresponds to finding a point within the $\ell_\infty$ ball that maximizes model confidence $p_y$:
\begin{align}
\label{mostmember:def}
{\forall} x^{\prime} \in \epsball[x], \quad \bar{x} = \argmax_{x'} (p_y(x')),
\end{align}
where \( \epsball[x] = \{x' \mid d_{\infty}(x, x') \leq \epsilon\} \) is the closed ball of radius \( \epsilon \) centered at \( x \).

\vspace{0.2em}
\noindent\textbf{On the Transferability of MFA.}
We find that the definition in \Cref{mostmember:def} enables fabricated members with high $p_y$ to deceive not only the loss attack but also a wide range of existing MIAs. This transferability arises because many inference strategies—despite differing in implementation—share a common reliance on the model's confidence on the true label (or monotonic transformations thereof). For example, likelihood-ratio methods such as LiRA \citep{carlini2022membership} operate on log-likelihoods that increase with $p_y$, and even density- or calibration-based statistics like RMIA \citep{zarifzadeh2024low} tend to grow as $p_y$ increases. As a result, a perturbation that raises $p_y$ consistently pushes the input into the “member” region across these attacks, explaining the strong cross-attack transferability observed in our experiments.

\vspace{0.2em}
\noindent\textbf{The Difference from Adversarial Attacks.}
As illustrated in \Cref{motivation_adv_fabri}, conventional adversarial attacks seek to \emph{minimize the classification margin} between the correct and the most competitive incorrect class, forcing the model to misclassify. In contrast, \textbf{MFA} manipulates a model’s \emph{training-set attribution} by \emph{maximizing confidence} on the true label, thereby pushing non-members into regions that MIAs interpret as “members’’ with high confidence—undermining the reliability of privacy audits and model accountability.

\vspace{0.2em}
\noindent\textbf{Strategies for Finding Fabricated Members.}  
Existing adversarial attacks such as PGD perform \emph{loss minimization} to reduce model confidence on the true label.  
To fabricate membership, we invert this process and perform \emph{confidence ascent} instead—maximizing $p_y$ within the perturbation bound.  
A straightforward inversion of PGD \citep{Madry18PGD}, however, suffers from instability due to its fixed step size and gradient oscillation near high-confidence regions.  
Although AutoAttack~\citep{croce2020reliable} adopts an adaptive halving heuristic, its coarse, trigger-based rule proved less effective in our setting. To overcome these issues, we propose a momentum-based cosine-annealed ascent scheme that produces smoother and more stable optimization.  
At each iteration $k$, the update is:
\begin{align}
m_{k+1} &= \beta m_k + (1-\beta)\nabla_{x_k}\ell(f(x_k),y), \\
x_{k+1} &= \Pi_{\mathcal{B}_\epsilon[x]}\!\left(x_k - \alpha_k \sign(m_{k+1})\right),
\end{align}
where $\Pi_{\mathcal{B}_\epsilon[x]}(\cdot)$ projects onto the $\ell_\infty$-ball, $\beta$ is a momentum factor, and $\alpha_k$ follows a cosine-decay schedule \citep{loshchilov2016sgdr}
$\alpha_k = \alpha_0 \tfrac{1+\cos(\pi k/N)}{2}$ for gradual annealing. By progressively decaying the step size according to a smooth, cosine-based function, our method ensures more stable and controlled updates, preventing large, abrupt changes that could destabilize the optimization. This smooth adjustment of the step size, coupled with momentum also allows for fine-tuning during the optimization process, enhancing the \textbf{MFA}'s ability to escape local minima and find more optimal solutions. The quantified results of ablation study in \Cref{tab:ablation_study} demonstrate our superiority, and the pseudo-code for our method is provided in \Cref{alg:Alg__maxconf}. 
\begin{algorithm}[!h]
\footnotesize
\caption{\sc Member Fabrication Attack (MFA)}
\label{alg:Alg__maxconf}
\begin{algorithmic}[1]
\STATE \textbf{Input:} non-member $(x,y)$, model $f$, loss function $\ell$, iterations $N$, perturbation bound $\epsilon$, initial step size $\alpha_0$, momentum factor $\beta$
\STATE $x_0\!\gets\!x$, $m_0\!\gets\!0$, $\bar{x}\!\gets\!x$
\FOR{$k=0$ \textbf{to} $N-1$}
  \STATE $\alpha_k=\alpha_0(1+\cos(\pi k/N))/2$
  \STATE $m_{k+1}=\beta m_k+(1-\beta)\nabla_{x_k}\ell(f(x_k),y)$
  \STATE $x_{k+1}=\Pi_{\mathcal{B}_\epsilon[x]}(x_k-\alpha_k\sign(m_{k+1}))$
  \IF{$\ell(f(x_{k+1}),y)<\ell(f(\bar{x}),y)$}
    \STATE $\bar{x}=x_{k+1}$
  \ENDIF
\ENDFOR
\STATE \textbf{Return:} fabricated member $\bar{x}$
\end{algorithmic}
\end{algorithm}
\subsection{Member Fabrication Detection}
\label{subsec:mfd}

Given the adversarial vulnerabilities that imperceptible perturbations can fabricate membership and deceive existing MIAs, we next ask a critical question: \emph{Can we reliably detect such fabricated members?} To formalize this, we introduce the problem of \emph{Member Fabrication Detection} (\textbf{MFD}), 
which extends the membership inference game by incorporating a new participant, the \emph{Detector}~$\mathbf{T}$\footnote{We assume white-box access to construct the strongest detector. When such access is restricted, we show how finite-difference gradient estimation yields slightly weaker but still effective black-box variants in Appendix~\ref{finite_difference}.}, responsible for distinguishing true members from fabricated ones.

\begin{definition}[Member Fabrication Detection]
\label{def:MF-detection}
The game involves four participants: a Challenger $\mathbf{C}$, a Fabricator $\mathbf{F}$, an Inferer $\mathbf{I}$, and a Detector $\mathbf{T}$, and proceeds as follows:
\begin{enumerate}
    \item $\mathbf{C}$ samples a training dataset $D \leftarrow \mathbb{D}$ and trains a model $f \leftarrow \mathcal{T}(D)$ on $D$.
    \item $\mathbf{C}$ flips a bit $b$: if $b=0$ it samples a fresh challenge $(x,y)\leftarrow\mathbb{D}$ with $(x,y)\notin D$; otherwise it selects $(x,y)\leftarrow D$.
    \item When $b=0$, $\mathbf{F}$ perturbs $x$ with $\delta$ satisfying $\|\delta\|_{\infty}\le\epsilon$, yielding $x'=x+\delta$; otherwise $x'=x$.
    \item $\mathbf{F}$ sends the perturbed $(x',y)$ to $\mathbf{I}$.
    \item $\mathbf{I}$ has query access to $\mathbb{D}$ and $f$, and outputs a prediction $\hat b \leftarrow \mathbf{I}_{\mathbb{D},f}(x',y)$.
    \item If $\hat b=1$, $\mathbf{T}$ has query access to $\mathbb{D}$ and $f$, and outputs $\bar b \leftarrow \mathbf{T}_{\mathbb{D},f}(x',y)$.
    \item $\mathbf{T}$ succeeds if $\bar b = b$, and fails otherwise.
\end{enumerate}
\end{definition}

\begin{figure}[!t]
    \begin{center}
    {\includegraphics[width=0.235\textwidth]{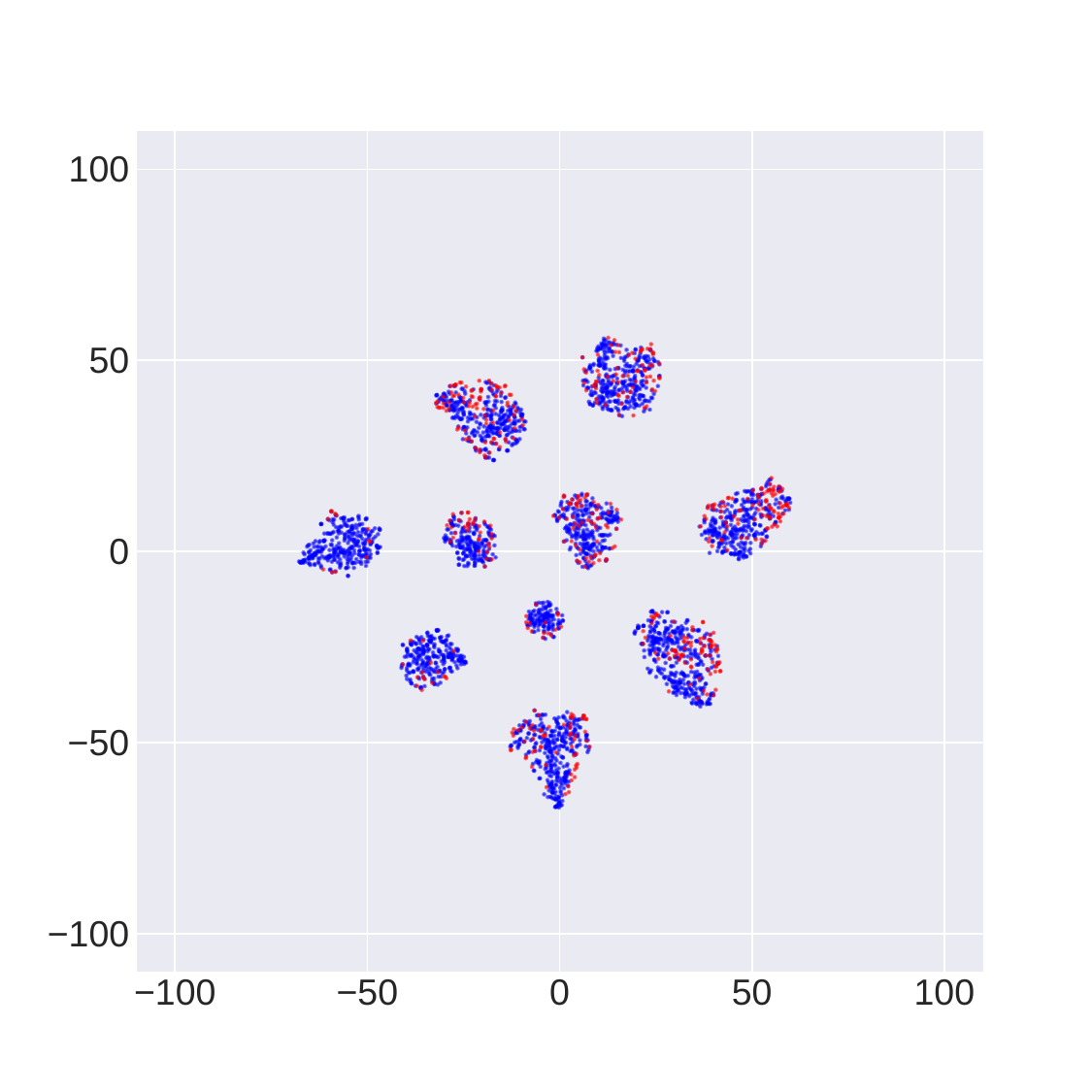}}
    {\includegraphics[width=0.235\textwidth]{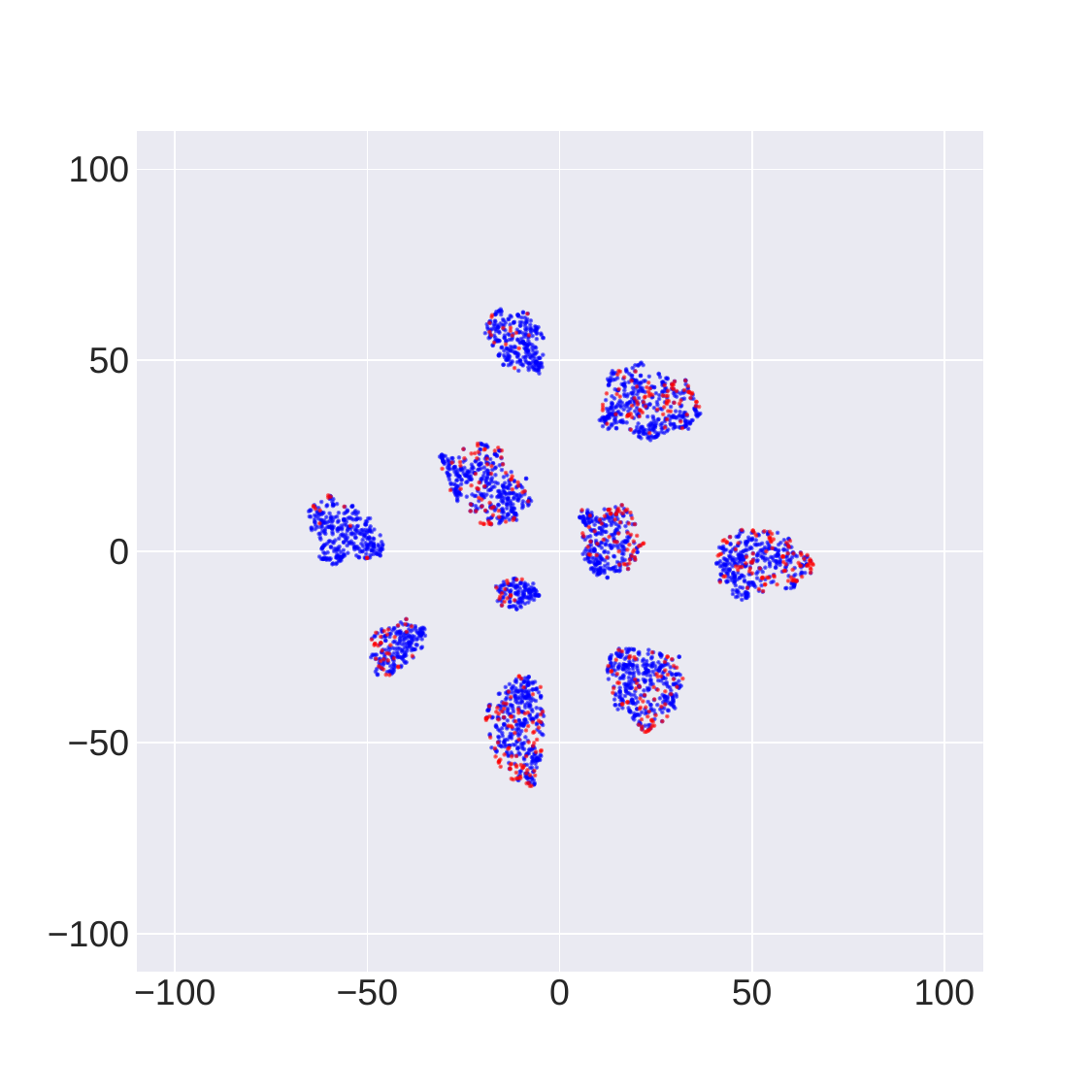}}
    \caption{\footnotesize  Visualization of the Distribution of Fabricated and True Members in Different Semantic Feature Spaces Using t-SNE \citep{maaten2008visualizing}. The two subfigures represent the semantic features at the penultimate and antepenultimate layers, with perturbation constrained to \(|\delta|_{\infty} \leq 4/255\). Red and blue dots denote true and fabricated members, respectively. The high degree of overlap of the red dots and blue dots suggests that semantic features alone are insufficient to distinguish between them.}
    \label{tsne_show}
    \end{center}
    \vspace{-2em}
\end{figure}

\vspace{0.2em}
\noindent\textbf{Limitations of feature-based detectors.}  
Existing adversarial detection methods—such as those based on \emph{Mahalanobis distance}~\cite{lee2018simple} and 
\emph{Local Intrinsic Dimensionality (LID)}~\cite{ma2018characterizing}—are ineffective for this task.
These methods assume semantic feature shifts between normal and adversarial samples.  
However, fabricated members are purposefully optimized to remain semantically indistinguishable from true members.  
As shown in \Cref{tsne_show}, their embeddings overlap almost perfectly, leading to near-identical feature distributions.
\begin{figure}[!h]
        \begin{center}
        {\includegraphics[width=0.47\textwidth]{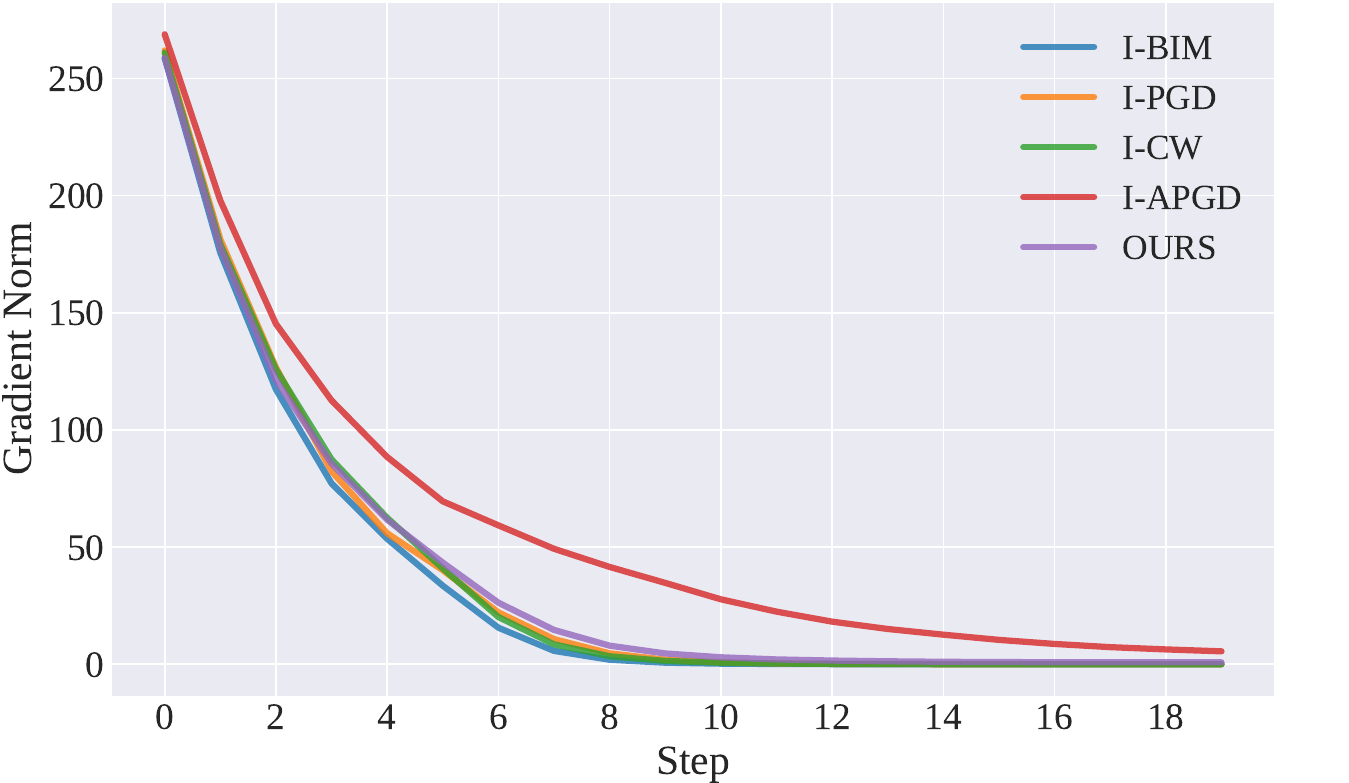}}
        \caption{\footnotesize Decay of Gradient Norm with Respect to Input Across Steps. As the steps increases, the gradient norm with respect to the input progressively diminishes. For clarity, a large epsilon ball ($\|\delta\|_{\infty} \leq 8/255$) is selected, alongside a small initial step size of $1/24 \times (8/255)$, across 20 steps.}
        \label{Comparison_adv}
        \end{center}
    \vspace{-2em}
\end{figure}
\begin{figure*}[!t]
    \begin{center}
    {\includegraphics[width=0.32\textwidth]{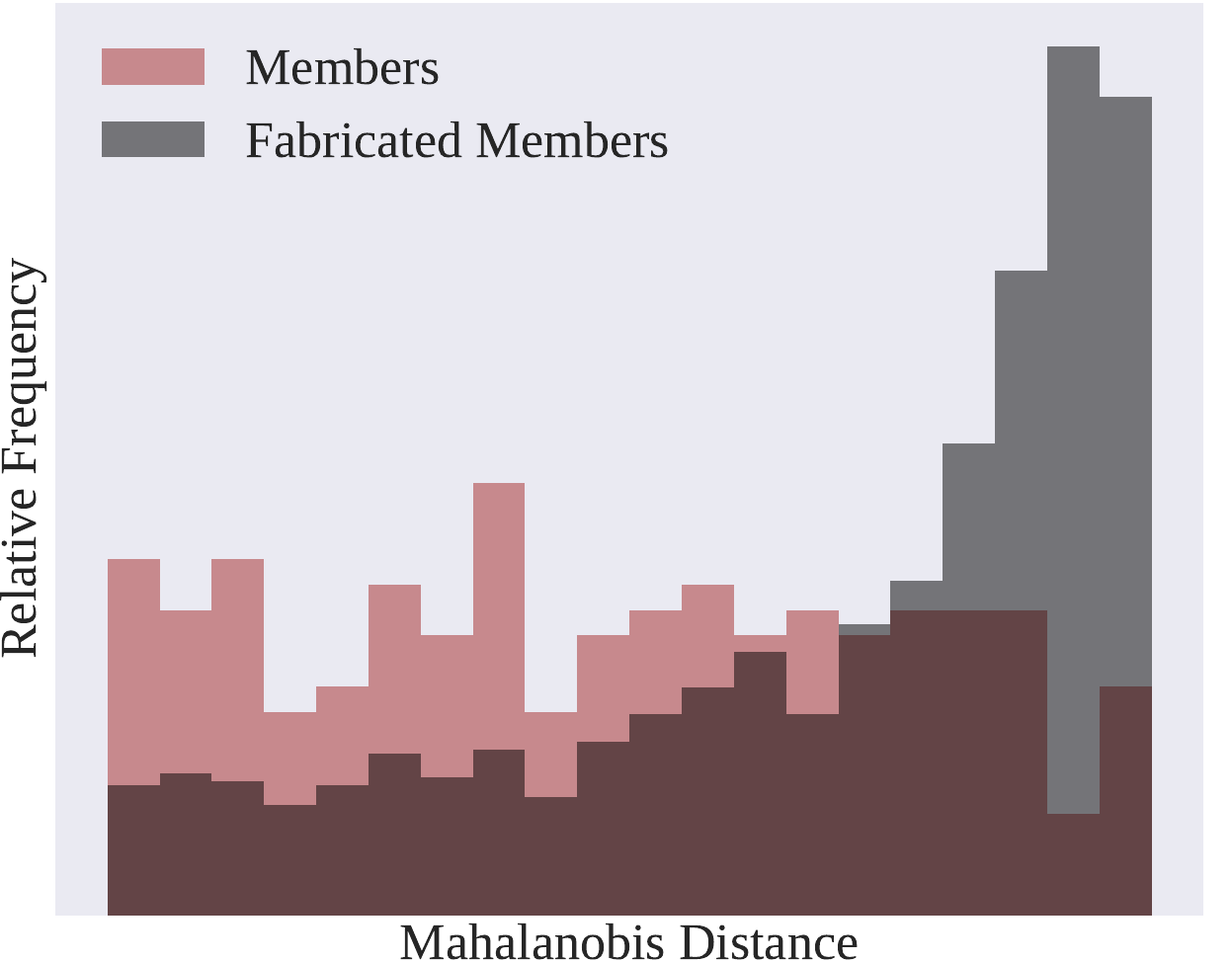}}
    {\includegraphics[width=0.32\textwidth]{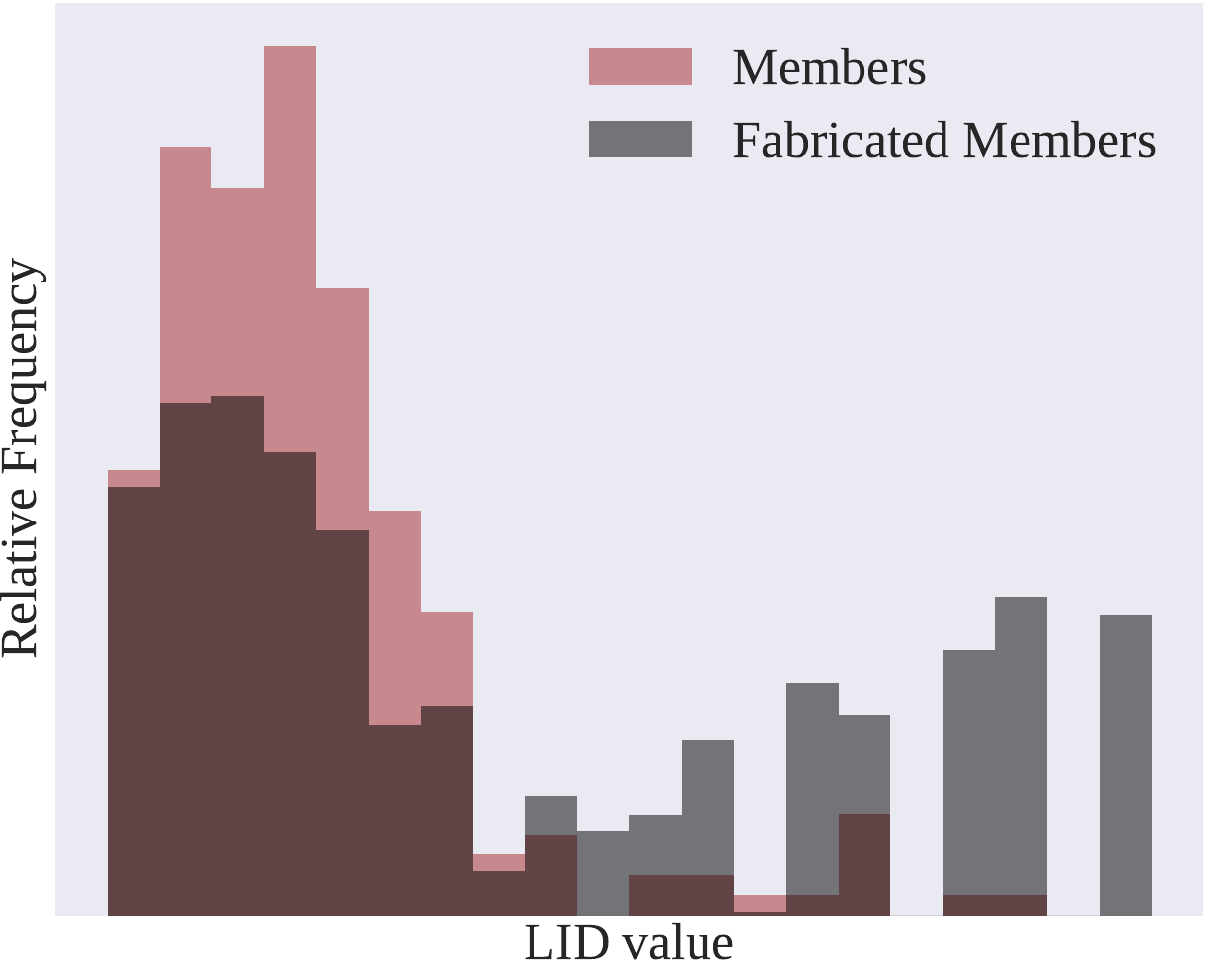}}
    {\includegraphics[width=0.32\textwidth]{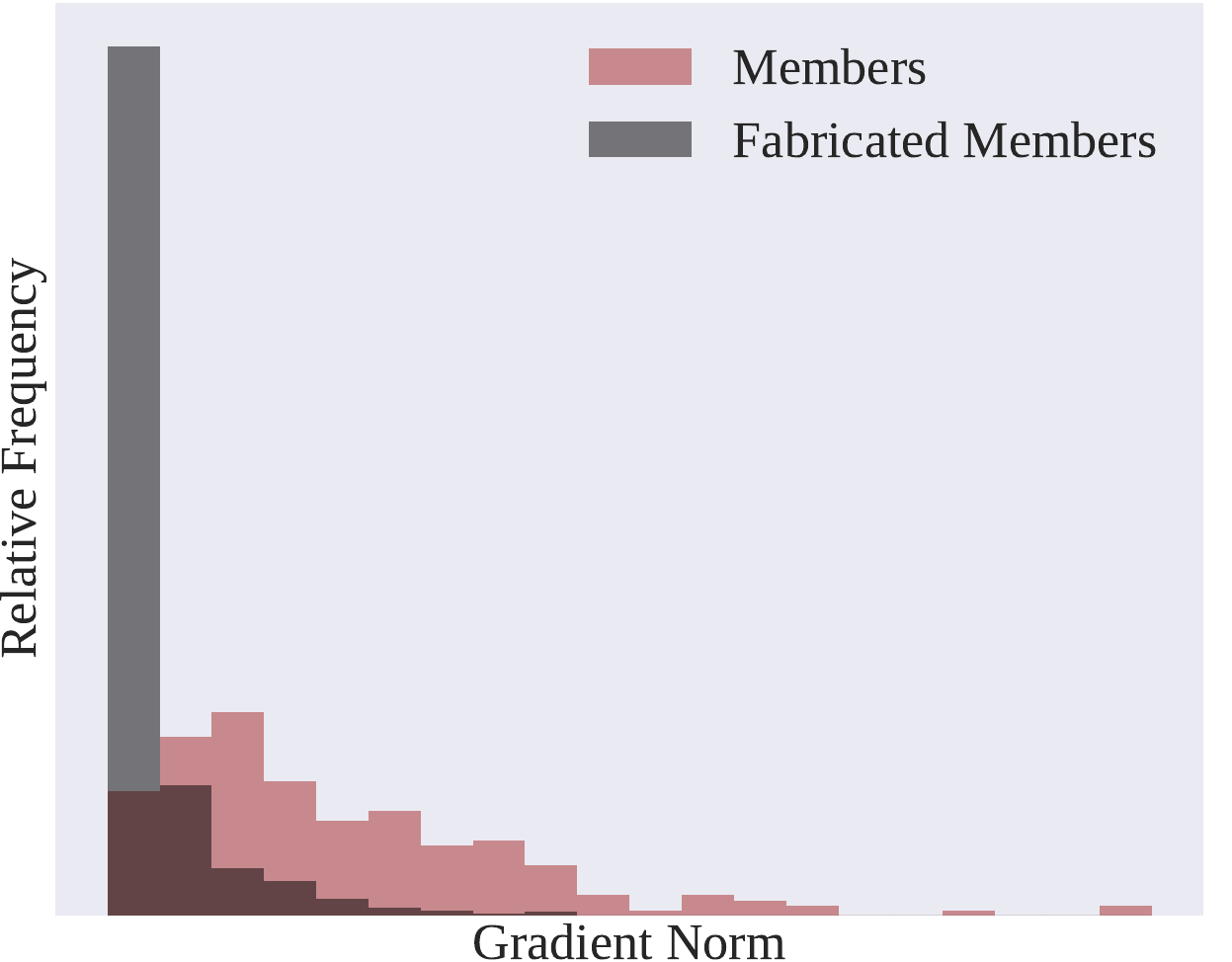}}
    \caption{\footnotesize Distribution of fabricated and true members across different detection strategies. The first plot (left) shows the Mahalanobis distance, the second plot (middle) shows the Local Intrinsic Dimensionality (LID) values, and the third plot (right) presents the gradient norm with respect to the input. We use the relative frequency within each membership class as the vertical axis. The gradient norm in the third plot demonstrates superior distinguishability between fabricated and true members, where the other two show limited differentiation. The perturbation is constrained to $\|\delta\|_{\infty} \leq 4/255$ here.}
    \label{detection_distribution}
    \end{center}
    \vspace{-2em}
\end{figure*}

\vspace{0.2em}
\noindent\textbf{Empirical observation and geometric interpretation.}  
Although fabricated and true members are visually and semantically similar, we observe a consistent divergence in their local optimization geometry. During fabrication, the gradient norm $\|\nabla_x \ell(f(x),y)\|$ tends to decrease over optimization steps as the perturbation drives $x$ toward higher-confidence regions. As shown in \Cref{Comparison_adv}, the gradient norm with respect to the input progressively diminishes as the number of steps increases. In practice, this behavior leads fabricated members to generally exhibit smaller gradient norms than true members. Moreover, this gap is not solely due to confidence: even within the same $p_y$ bins, fabricated samples still exhibit smaller gradient norms than true members (see Appendix~\ref{app_confidence_match}). As shown in \Cref{detection_distribution}, the gradient norm provides stronger distinguishability between fabricated and true members than Mahalanobis and LID statistics based on semantic features. This behavior is also supported by the following local result.

\begin{theorem}[Local Gradient-Norm Decrease Under Small-Step Fabrication]
\label{thm:gd_norm}
Assuming that the $\epsilon$-ball is sufficiently small such that the local curvature of $\ell \circ f$ around $x$ can be well approximated by its second-order Taylor expansion, after taking one step of signed gradient descent with respect to the input sample
\begin{equation}
    x^{\prime} = x - \alpha \cdot \sign(\nabla_{x}\ell(f(x),y)),
\end{equation}
there exists an $\alpha$ such that the following approximately holds:
\begin{equation}
    \|\nabla_{x^{\prime}} \ell(f(x^{\prime}), y)\| < \|\nabla_{x} \ell(f(x), y)\|.
\end{equation}
\end{theorem}

The proof is given in Appendix~\ref{app_proof}. The theorem provides a local explanation for why gradient decrease can occur under small-step fabrication, which is consistent with the empirical behavior observed in \Cref{Comparison_adv}. This \emph{gradient-norm collapse} therefore serves as a useful geometric signature of adversarial membership manipulation. Notably, this property also grants \textbf{MFD} strong resilience against \emph{adaptive} \textbf{MFA} that is explicitly aware of the detection mechanism\footnote{Appendix~\ref{app_adaptive_MFA} provides a concise analysis showing that adaptive fabrication in \textbf{MFA} encounters an inherent trade-off: increasing attack efficacy inevitably amplifies the gradient-norm signal, thus making evasion harder.}.

\vspace{0.2em}
\noindent\textbf{Gradient-norm as a detection statistic.}  
The above finding implies that fabricated members reside in low-gradient, high-confidence basins, whereas true members occupy regions of moderate gradient strength.  
This motivates the following gradient-based decision rule:
\begin{equation}
\label{eq:detection_method}
\mathbf{T}(x,y) = 
\mathbf{1}\!\left[\|\nabla_x \ell(f(x),y)\| \le \tau'\right].
\end{equation}

\begin{figure*}[!t]
    \begin{center}
    \begin{minipage}{0.33\textwidth}
        \centering
        \includegraphics[width=\textwidth]{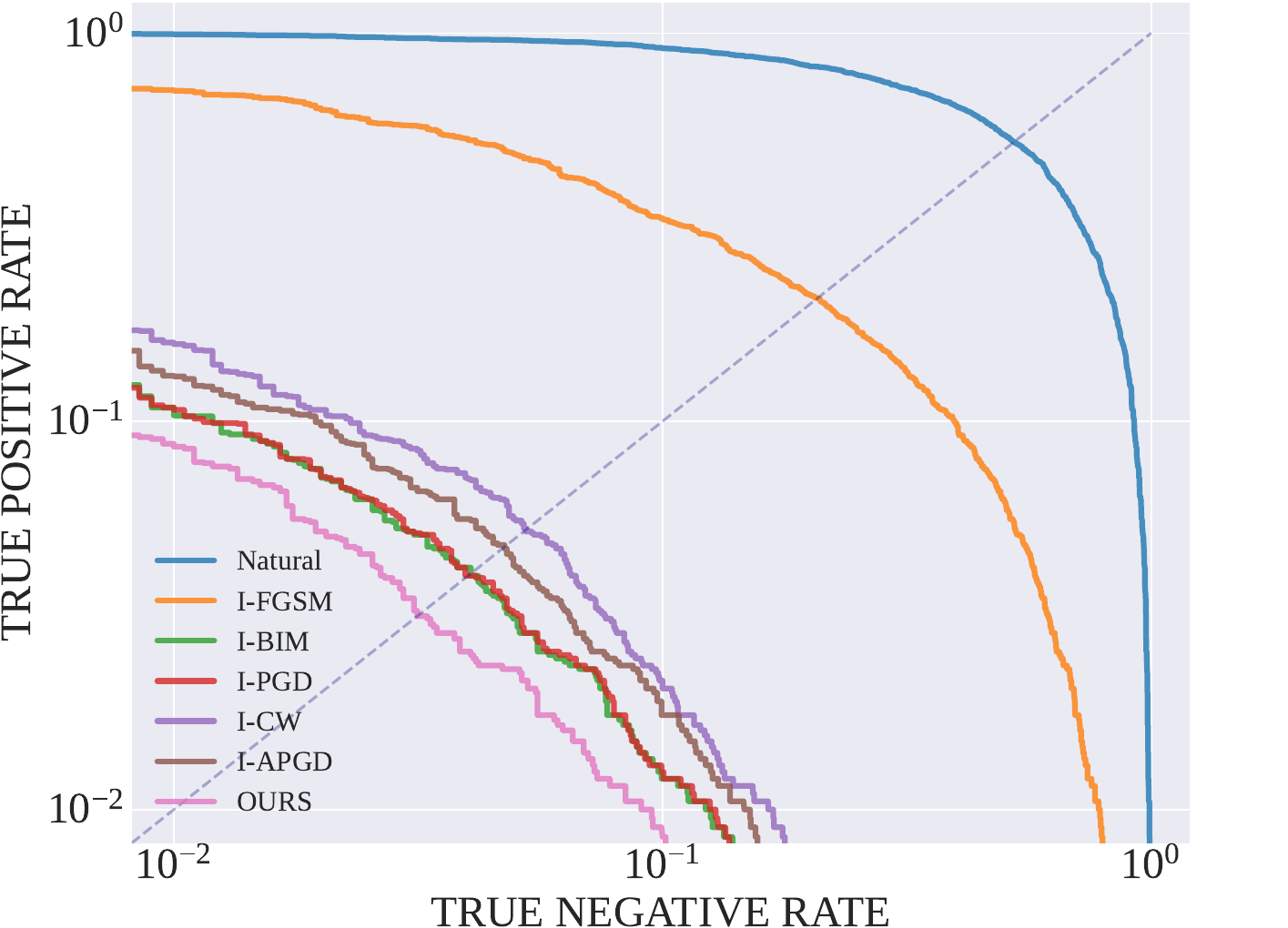}
        \subcaption{Our \textbf{MFA} vs. Baselines}
    \end{minipage}
    \hfill
    \begin{minipage}{0.33\textwidth}
        \centering
        \includegraphics[width=\textwidth]{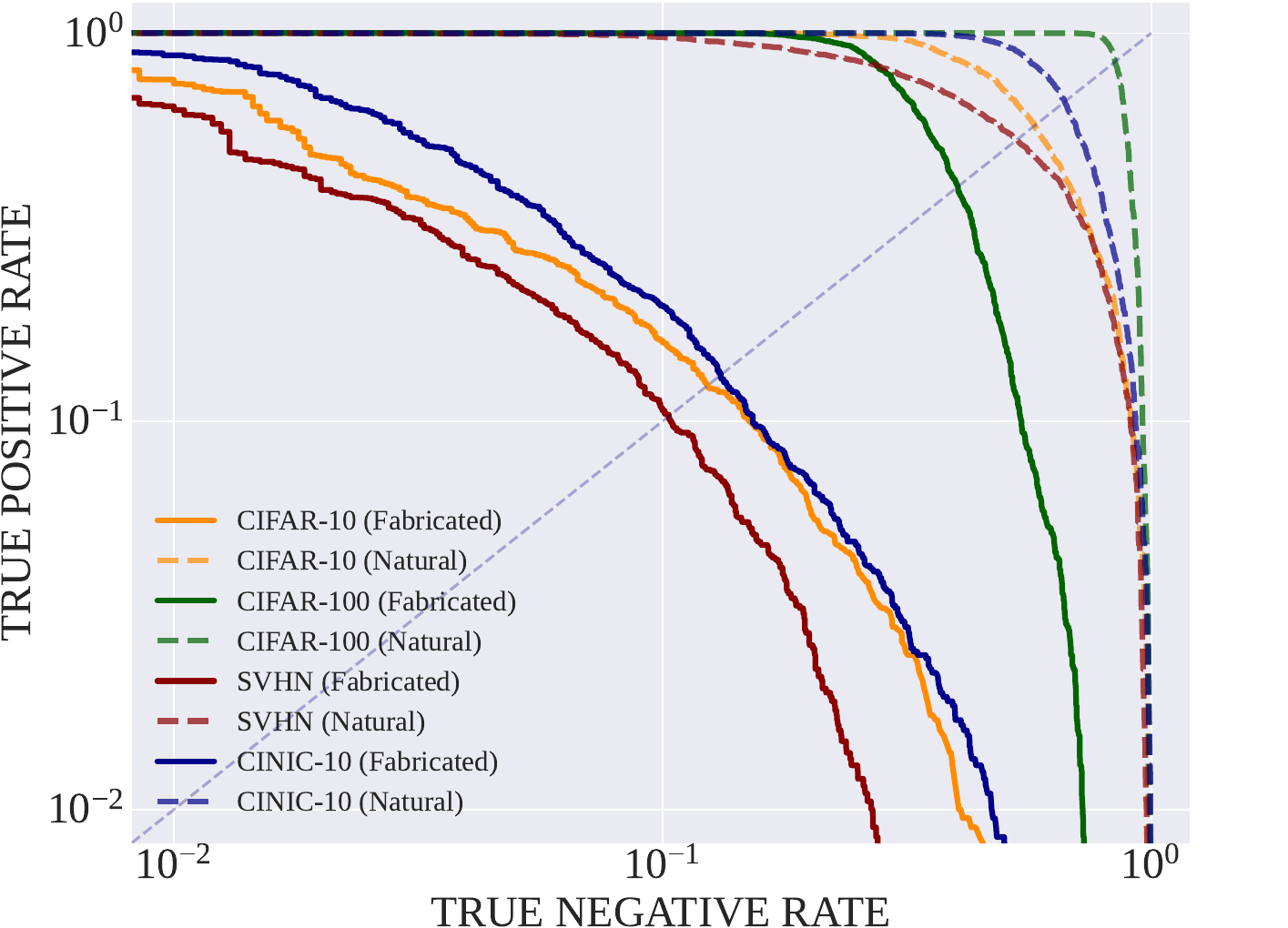}
        \subcaption{Our \textbf{MFA} across Datasets for Loss Attack}
    \end{minipage}
    \hfill
    \begin{minipage}{0.33\textwidth}
        \centering
        \includegraphics[width=\textwidth]{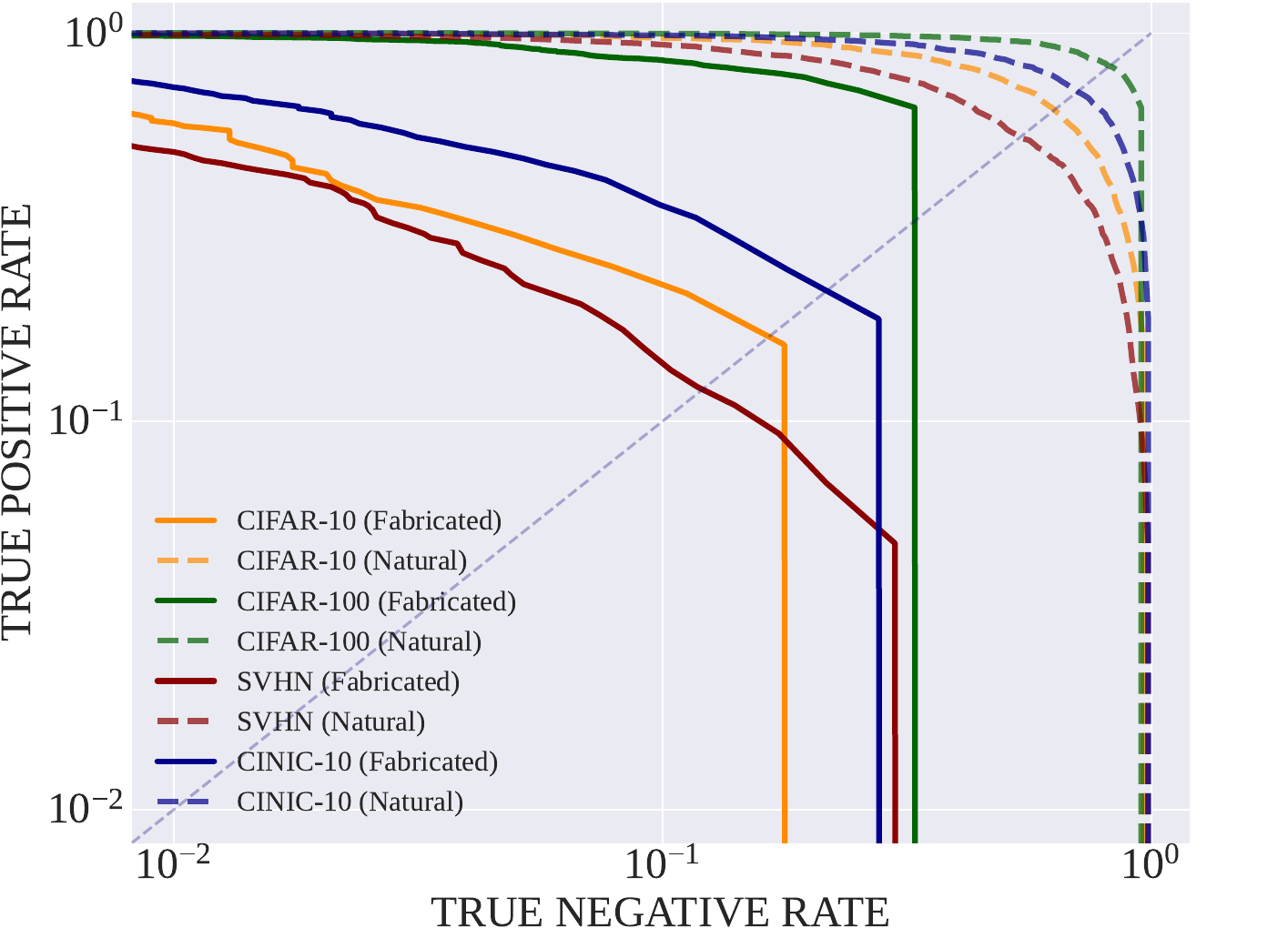}
        \subcaption{Our \textbf{MFA} across Datasets for Attack R}
    \end{minipage}

    \vspace{0.5em}

    \begin{minipage}{0.33\textwidth}
        \centering
        \includegraphics[width=\textwidth]{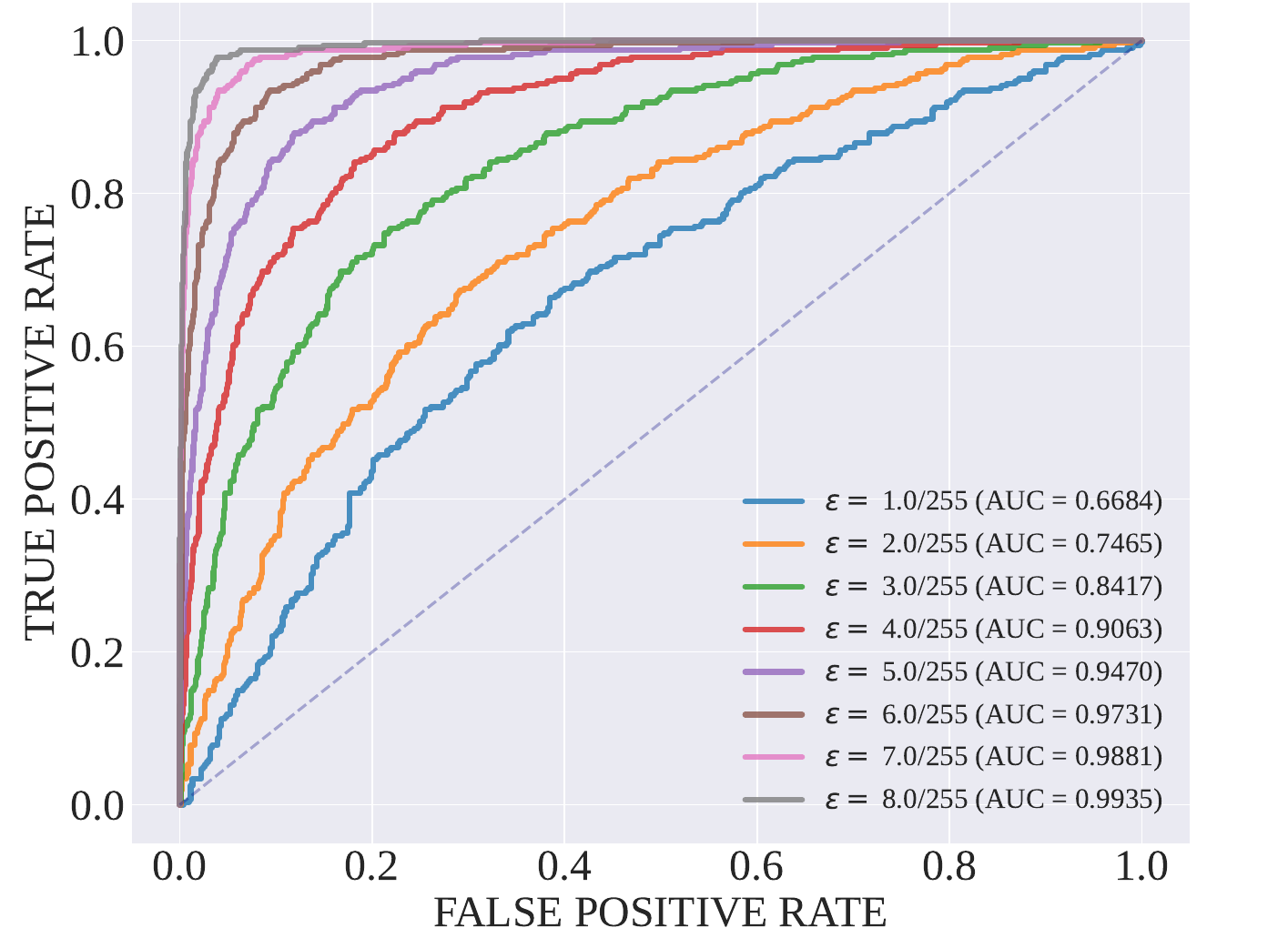}
        \subcaption{Our \textbf{MFD} across $\epsilon$ on \textbf{CINIC-10}}
    \end{minipage}
    \hfill
    \begin{minipage}{0.33\textwidth}
        \centering
        \includegraphics[width=\textwidth]{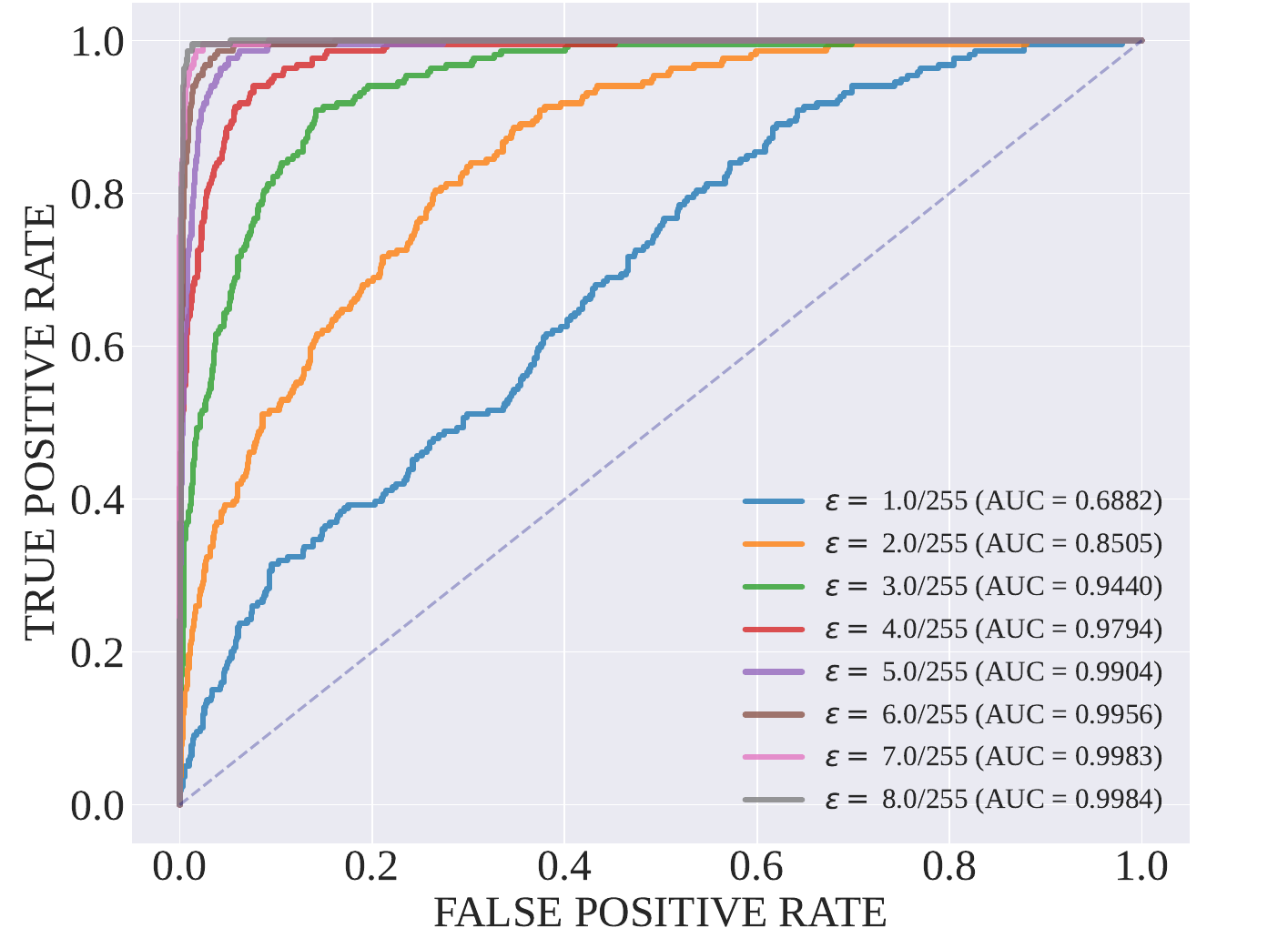}
        \subcaption{Our \textbf{MFD} across $\epsilon$ on \textbf{SVHN}}
    \end{minipage}
    \hfill
    \begin{minipage}{0.33\textwidth}
        \centering
        \includegraphics[width=\textwidth]{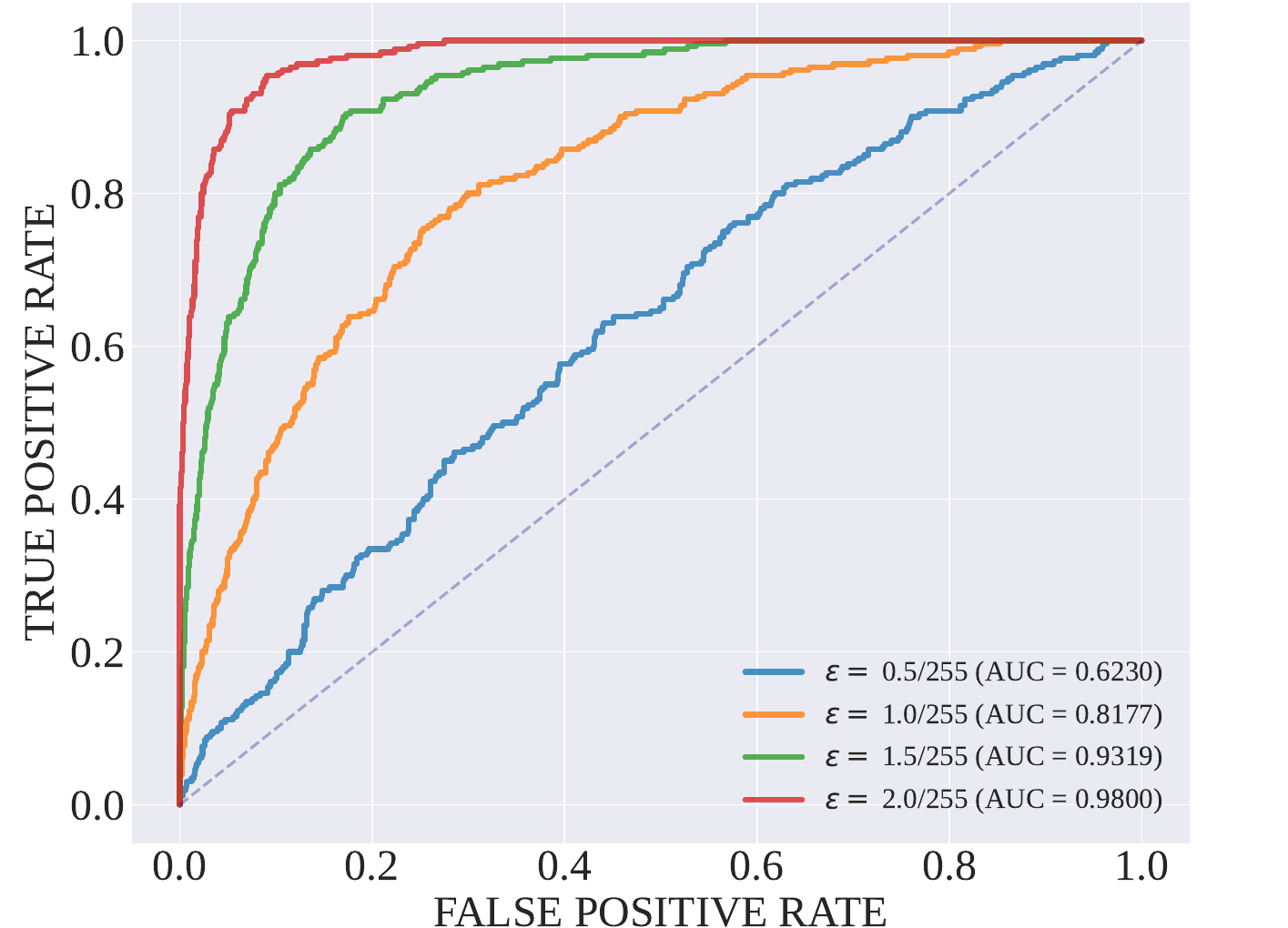}
        \subcaption{Our \textbf{MFD} across $\epsilon$ on \textbf{ImageNet-100}}
    \end{minipage}

    \begin{minipage}{0.33\textwidth}
        \centering
        \includegraphics[width=\textwidth]{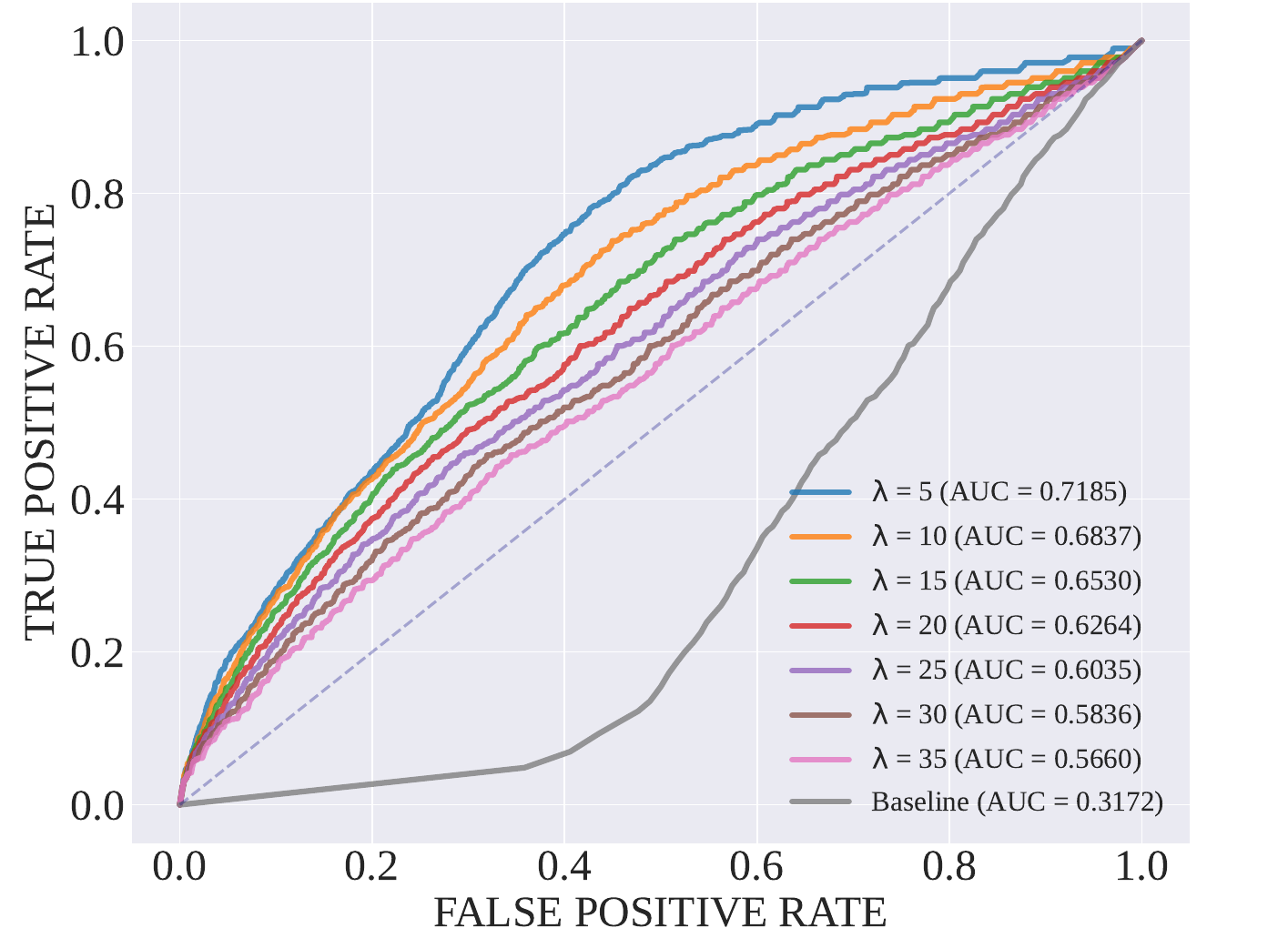}
        \subcaption{Our \textbf{AR-Attack R} across $\lambda$ on \textbf{SVHN}}
    \end{minipage}
    \hfill
    \begin{minipage}{0.33\textwidth}
        \centering
        \includegraphics[width=\textwidth]{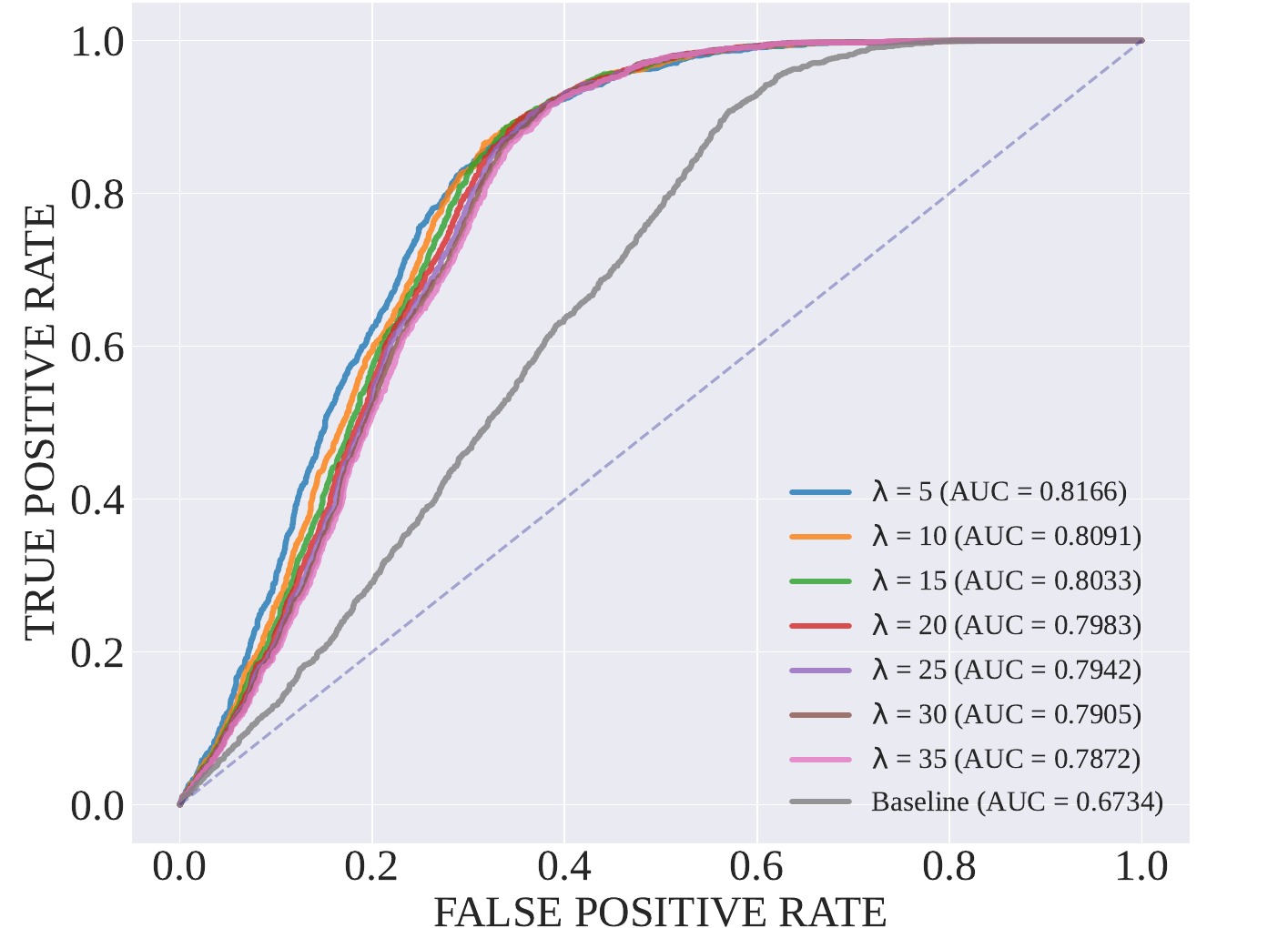}
        \subcaption{Our \textbf{AR-LiRA} across $\lambda$ on \textbf{CINIC-10}}
    \end{minipage}
    \hfill
    \begin{minipage}{0.33\textwidth}
        \centering
        \includegraphics[width=\textwidth]{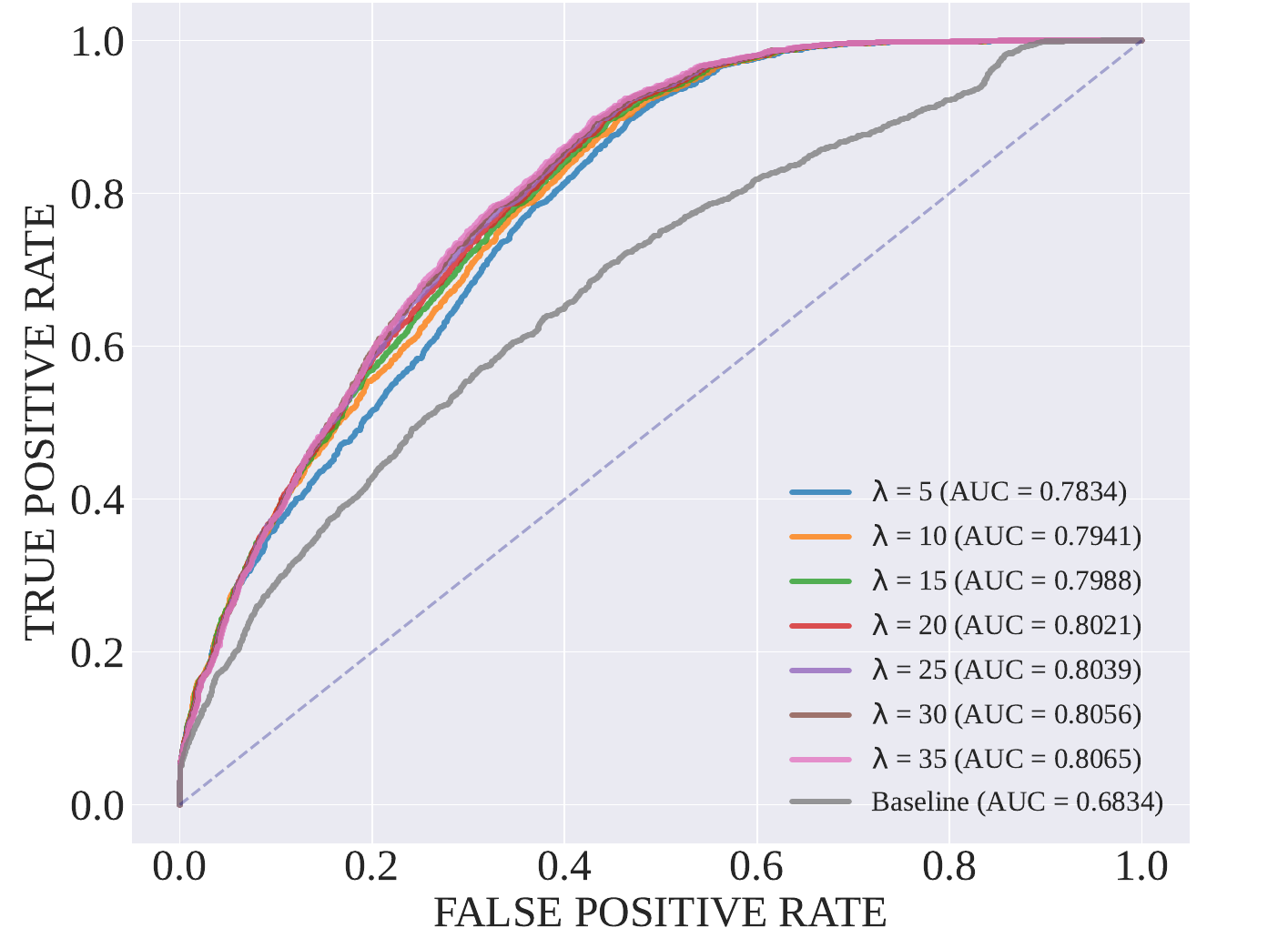}
        \subcaption{Our \textbf{AR-RMIA} across $\lambda$ on \textbf{CIFAR-10}}
    \end{minipage}

    \caption{\footnotesize Representative Experimental Results. Subfigures (a-c) demonstrate the superior performance of (\textbf{MFA}) compared to baselines, highlighting its effectiveness across different MIAs and datasets. Subfigures (d-f) show the effectiveness of (\textbf{MFD}) across varying perturbation levels \(\epsilon\) and datasets. Subfigures (g-i) illustrate the effectiveness of our Adversarially Robust Strategies in combination with different strong MIAs across multiple datasets.}
    \label{main_experiments}
    \end{center}
    \vspace{-2em}
\end{figure*}

\subsection{Adversarially Robust MIAs}
\label{subsec:armia}

While \textbf{MFD} introduces a separate detector after inference, a more practical strategy is to embed robustness directly into the inference process itself. We thus extend the setting to an \emph{Adversarial Membership Inference Game}, where the Inferer itself becomes aware of potential fabrication and adjusts its decision process accordingly.

\begin{definition}[Adversarial Membership Inference Game]
\label{def:adv_MIA-game}
The game involves three participants: a Challenger \(C\), an Inferer \(I\), and a Fabricator \(F\), proceeds as follows:
\begin{enumerate}
    \item $\mathbf{C}$ samples a training dataset $D \leftarrow \mathbb{D}$ and trains a model $f \leftarrow \mathcal{T}(D)$ on $D$.
    \item $\mathbf{C}$ flips a bit $b$: if $b=0$ it samples a fresh challenge $(x,y)\leftarrow\mathbb{D}$ with $(x,y)\notin D$; otherwise it selects $(x,y)\leftarrow D$.
    \item When \(b = 0\), $\mathbf{F}$ with probability \(Pr\) perturbs $x$ with $\delta$ satisfying $\|\delta\|_{\infty}\le\epsilon$, yielding $x'=x+\delta$; otherwise $x'=x$.
    \item $\mathbf{F}$ sends the perturbed $(x',y)$ to $\mathbf{I}$.
    \item $\mathbf{I}$ has query access to $\mathbb{D}$ and $f$, and outputs a bit $\hat{b} \leftarrow I(x', y)$.
    \item $\mathbf{I}$ succeeds if $\hat{b} = b$, and fails otherwise.
\end{enumerate}
\end{definition}

The probability $\Pr$ is set to $50\%$ by default, and the Inferer is granted white-box access to $f$, similar to the Detector in \textbf{MFD}, to evaluate both semantic and geometric cues. The objective is to jointly distinguish \emph{true members} from both \emph{non-members} and \emph{fabricated members}.

Given that the statistics \(S\) used in existing strong MIAs can effectively distinguish true members from non-members (where \(S\) is relatively small for non-members), and the gradient norm used in our detection strategy can effectively distinguish true members from fabricated members (where the gradient norm is relatively small for fabricated members), we consider combining both approaches. This involves constructing weights based on the gradient norm and incorporating them with the statistics from existing strong MIAs to enhance the adversarial robustness of these MIAs.

However, previous studies \citep{nasr2019comprehensive} show that non-members tend to have larger gradient norms than members. Therefore, when constructing weights based on the gradient norm, we must avoid situations where the gradient norm for certain non-members becomes too large, thereby skewing the overall statistic. To address this, we use the \(\tanh\) function combined with a parameter \(\lambda\) to design the following statistic:
\begin{align}
\label{weight_definition}
w(x, y) &= \tanh \left( \lambda \cdot \|\nabla_{x} \ell(f(x), y)\| \right),
\end{align}
where $\lambda$ is a hyper-parameter, the optimal value of $\lambda$ may vary across different datasets and evaluation metrics. In practice, $\lambda$ can be calibrated offline on a shadow model using non-members and simulated fabricated queries, following standard calibration practice in membership inference. The \(\tanh\) function constrains the range of the weight, ensuring that excessively large gradient norms of certain non-members do not diminish the discriminative power of their original test statistic, thereby preventing the overall weighted statistic from becoming disproportionately large. The definition of \Cref{weight_definition} for its simplicity in achieving our goal, and we also encourage practitioners to adjust based on actual needs. Then, the statistic for \textbf{AR-MIAs} is:
\begin{align}
\label{robust_MIA_base}
I(x, y) &= \mathbf{1}\left[ w(x, y) \cdot S(x, y) > \tau \right]
\end{align}
In our experiments, we demonstrate the significant performance improvements achieved by combining our strategies with existing strong MIAs that Attack R, LiRA, and RMIA.

\begin{table}[t]
\centering
\scriptsize
\caption{
Ablation study of \textbf{MFA} with four step size strategies.
}
\vspace{-6pt}
\renewcommand{\arraystretch}{1.1}
\setlength{\tabcolsep}{5pt}
\rowcolors{2}{}{white} 
\label{tab:ablation_study}
\begin{tabularx}{\linewidth}{
>{\centering\arraybackslash}p{0.180\linewidth}|
>{\centering\arraybackslash}p{0.15\linewidth}
>{\centering\arraybackslash}p{0.15\linewidth}
>{\centering\arraybackslash}p{0.15\linewidth}
>{\centering\arraybackslash}p{0.155\linewidth}}
\toprule[1.5pt]
\rowcolor[HTML]{D9D9D9}
\textbf{Dataset} & \textbf{I} \citep{Madry18PGD} & \textbf{II} \citep{croce2020reliable} & \textbf{III} & \textbf{IV} \textcolor{red}{[OURS]} \\
\midrule[0.1pt]
\midrule[0.1pt]
\rowcolor[HTML]{F2F2F2}
\multicolumn{5}{c}{\textbf{Error Area} $(\|\delta\|_{\infty} \leq 4.0/255)$} \\
\midrule[0.1pt]
\midrule[0.1pt]
CIFAR-10 & 0.9399 & 0.9369 & 0.9444 & \textcolor{red}{0.9451} \\
CIFAR-100 & 0.5917 & 0.5880 & 0.6120 & \textcolor{red}{0.6136} \\
SVHN & 0.9548 & 0.9560 & 0.9612 & \textcolor{red}{0.9618} \\
CINIC-10 & 0.9198 & 0.9162 & 0.9318 & \textcolor{red}{0.9324} \\
\midrule[0.1pt]
\midrule[0.1pt]
\rowcolor[HTML]{F2F2F2}
\multicolumn{5}{c}{\textbf{Equal Error Rate (EER)} $(\|\delta\|_{\infty} \leq 4.0/255)$} \\
\midrule[0.1pt]
\midrule[0.1pt]
CIFAR-10 & 86.75\% & 86.30\% & 87.55\% & \textcolor{red}{87.60\%} \\
CIFAR-100 & 58.05\% & 57.85\% & 59.75\% & \textcolor{red}{59.80\%} \\
SVHN & 88.60\% & 88.95\% & 89.60\% & \textcolor{red}{89.70\%} \\
CINIC-10 & 85.25\% & 85.00\% & 86.75\% & \textcolor{red}{86.85\%} \\
\bottomrule[1.5pt]
\end{tabularx}
\vspace{-2em}
\end{table}

\section{Experimental Results}
We conduct extensive experiments to demonstrate the effectiveness of our methods. Due to space constraints, we present representative results in this section. For evaluation metrics, implementation details, and comprehensive experimental results, please refer to Appendix \ref{app:evaluation_metrics}, \ref{app_expsetup} and \ref{app_addexp}.

\vspace{0.3em}
\noindent\textbf{Experimental Setup.}
We evaluate our method across \textbf{CIFAR-10/100} \citep{krizhevsky2009learning}, \textbf{SVHN} \citep{krizhevsky2009learning}, \textbf{CINIC-10} \citep{darlow2018cinic}, and \textbf{ImageNet-100} \citep{deng2009imagenet}, using ResNet-18 \citep{he2016deep} and Wide-ResNet-50-2 \citep{zagoruyko2016wide}. We compare against representative MIAs, including the basic loss attack \citep{Yeom2018Privacy}, \textbf{Attack R} \citep{ye2022enhanced}, \textbf{LiRA} \citep{carlini2022membership}, and \textbf{RMIA} \citep{zarifzadeh2024low}. For comparison with \textbf{MFA}, we additionally invert the perturbation direction of standard adversarial attacks as baselines (Appendix~\ref{app_implementation}).\footnote{Appendix~\ref{app_defense_results} further evaluates our framework under standard MI-defenses, e.g. $\ell_1$-reg and distillation, demonstrating its broad applicability.}
As \textbf{MFD} and \textbf{AR-MIAs} are first introduced in this work and have no direct baselines, we only demonstrate the effectiveness across diverse settings, establishing them as initial benchmarks for future research.

\vspace{0.3em}
\noindent\textbf{Evaluation Metrics.} We evaluate \textbf{MFA} by measuring the Inferer’s performance after exposure to fabricated members. In practice, we found that the ROC curve does not clearly distinguish between different \textbf{MFA} methods. As an alternative, we suggest using the TNR-TPR curve, combined with a logarithmic scale, to enhance the clarity of comparisons. We employ two primary evaluation metrics: \textbf{Error Area} (i.e., 1 - \textbf{AUC}) and \textbf{Equal Error Rate} (\textbf{EER}). For MFD and AR-MIA, we retain standard ROC-based evaluation (\textbf{AUC}, \textbf{TPR@FPR}). 
Further details are provided in Appendix~\ref{app:evaluation_metrics}.

\vspace{0.3em}
\noindent\textbf{Experiments of \textbf{MFA}.} Following \Cref{def:MF-game}, we conduct experiments on a balanced testing dataset consisting of 50\% fabricated members and 50\% true members. In the TNR-TPR curve, TNR and TPR represent the Inferer's ability to correctly identify true members (membership label 1) and fabricated members (membership label 0) generated by the Fabricator. A better \textbf{MFA} should result in the Inferer with lower TPR and TNR values, causing the curves to approach the lower-left corner of the TNR-TPR plot, along with a larger \textbf{Error Area} and \textbf{EER}. \Cref{main_experiments}(a-c) demonstrate the superior performance of our \textbf{MFA} compared to baselines, showing the curves of \textbf{MFA} approaching the lower-left corner. More settings and quantitative comparison results can be found in Appendix \Cref{Fabric_figure_4_attack,Fabric_figure_cifar10,Fabric_figure_cifar100,Fabric_figure_cinic,Fabric_figure_svhn,Fabric_figure_Imagenet} and \Cref{tab:fabric_error_area,tab:fabric_error_area_more_MIAs,tab:fabric_eer_more_MIAs,tab:fabric_eer}.

\vspace{0.3em}
\noindent\textbf{Ablation study of \textbf{MFA}.}
We further conduct an ablation study to evaluate the impact of step-size scheduling in \Cref{alg:Alg__maxconf}.
Four strategies are compared: 
\textbf{(I)} Fixed Step Size (I-PGD) \citep{Madry18PGD}, 
\textbf{(II)} Halving Heuristic (APGD) \citep{croce2020reliable}, 
\textbf{(III)} Cosine Annealing, and 
\textbf{(IV)} Cosine Annealing with Momentum (ours). 
As shown in \Cref{tab:ablation_study}, strategy (IV) consistently achieves the best performance on both \textbf{Error Area} and \textbf{EER}, confirming the superiority of the adopted schedule.

\vspace{0.3em}
\noindent\textbf{Experiments of \textbf{MFD}.} Following \Cref{def:MF-detection}, for a balanced testing dataset of 50\% fabricated members and 50\% true members, we first use the loss attack and set the threshold when FPR = 10\% to select the classified members. We then perform \textbf{MFD} on the selected data and compute the FPR and TPR in the ROC curve. FPR and TPR here represent the Detector's ability to correctly identify true members (membership label 1) and fabricated members (membership label 0). An effective \textbf{MFD} should result in a Detector with a low FPR, high TPR, and high AUC value, causing the curves to approach the upper-left corner of the FPR-TPR plot. \Cref{main_experiments}(d-f) show the effectiveness of our \textbf{MFD} across varying perturbation levels \(\epsilon\) and datasets. More settings can be found in Appendix \Cref{Detection_figure_cifar10,Detection_figure_cifar100,Detection_figure_cinic,Detection_figure_svhn,Detection_figure_Imagenet}.

\vspace{0.3em}
\noindent\textbf{Experiments of \textbf{AR-MIAs}.}
Following \Cref{def:adv_MIA-game}, we conducted experiments on a balanced testing dataset containing 50\% true members, 25\% fabricated members, and 25\% non-members. In the ROC curve, FPR and TPR represent the Inferer's ability to correctly identify true members (membership label 1) and non-members or fabricated members (both labeled 0). An effective \textbf{AR-MIA} should result in an Inferer with a low FPR, high TPR, and high AUC value, causing the curves to approach the upper-left corner of the FPR-TPR plot. We select a wide range and granularity of $\lambda$ in \Cref{weight_definition}. \Cref{main_experiments} (g-i) illustrate the effectiveness of our Adversarially Robust Strategies in combination with different strong MIAs across multiple datasets. More settings and quantitative comparison results can be found in Appendix \Cref{Robust_MIA_cifar10,Robust_MIA_cifar100,Robust_MIA_svhn,Robust_MIA_cinic} and \Cref{tab:robust_MIA_Attack_R,tab:robust_MIA_LiRA,tab:robust_MIA_RMIA}.

\section{Conclusion}
We present the first unified framework for \emph{adversarial membership manipulation} in vision models.
By extending the security-game formulation to include a \emph{Fabricator} and leveraging gradient geometry, we revealed the phenomenon of \emph{gradient-norm collapse} as a universal signature of fabricated members.
Our \textbf{MFA}, \textbf{MFD}, and \textbf{AR-MIA} together form a principled pipeline for analyzing and mitigating this vulnerability.
This work bridges adversarial robustness and privacy auditing, reframing membership inference as a geometry-sensitive vision problem and establishing adversarial reliability as a new frontier for trustworthy evaluation across large-scale vision models and future AI systems.

\clearpage

{
    \small
    \bibliographystyle{ieeenat_fullname}
    \bibliography{main}
}

\clearpage

\appendix

\section{Related Works}

\subsection{Privacy Risks in Machine Learning}
\emph{Machine learning} (ML) models, particularly deep neural networks, have become integral to advancements in various high-stakes domains such as healthcare, genomics, and image recognition. These models' exceptional capacity to detect intricate patterns from large datasets has significantly advanced fields like healthcare, genomics, image recognition, network inference, and autonomous decision making~\citep{miotto2018deep,xiong2015human,he2016deep,huang2024scalable,shao2026hats}.However, this proficiency is accompanied by notable privacy concerns, as these models can inadvertently memorize sensitive training data, posing significant privacy risks~\citep{nasr2018machine, Salem2019ML, song2021systematic, choquette2021label}. Privacy breaches have emerged as a pervasive issue in modern ML, prompting calls for robust user data protection measures~\citep{fredrikson2015model, tramer2016stealing, shokri2017membership, ganju2018property}. In response, legislative bodies have enacted stringent privacy laws, such as the \emph{General Data Protection Regulation} (GDPR) in the European Union, the \emph{California Consumer Privacy Act} (CCPA) in the United States, and the \emph{Personal Information Protection and Electronic Documents Act} (PIPEDA) in Canada, which legally mandate data privacy safeguards. Within this context, \emph{Membership Inference Attacks} (MIAs) have gained prominence as tools for auditing the degree of information leakage from ML models with respect to their training datasets~\citep{shokri2017membership, Yeom2018Privacy, ye2022enhanced, carlini2022membership, zarifzadeh2024low}.

\subsection{Membership Inference Attacks}
\label{app_related_MIAs}
\emph{Membership Inference Attacks} (MIAs) aim to predict whether a specific training example was included in the training set of a given model. Existing MIAs~\citep{Yeom2018Privacy, shokri2017membership, watson2021importance, ye2022enhanced, carlini2022membership, zarifzadeh2024low} can be naturally framed as hypothesis testing, where the Inferer attempts to distinguish between two hypotheses concerning the presence of a target sample $(x, y)$ in the training data of a target model $f$. The hypothesis testing framework can be formalized by defining two hypotheses: \textbf{(I). Null hypothesis $H_0$ :} The target sample $(x, y)$ was not part of the training data of the model $f$; \textbf{(II). Alternative hypothesis $H_1$ :} The target sample $(x, y)$ was part of the training data of the model $f$. In such a hypothesis testing problem, the design of test statistics plays a critical role in distinguishing between members and non-members. For example, the \emph{loss attack}~\citep{Yeom2018Privacy} uses the sample loss as the test statistic, while the \emph{Likelihood Ratio Attack} (LiRA)~\citep{carlini2022membership} uses the likelihood ratio. By setting an appropriate threshold for these statistics, MIAs determine whether a given data point is a member of the training set. Specifically, the decision rule can be written as:
\begin{align}
\label{MIA_base_app}
I(x, y) = \mathbf{1}[S(x, y) > \tau]
\end{align}
where \( \mathbf{1} \) is the indicator function, \( \tau \) is a tunable decision threshold, and \( S \) is the test statistic. In this context, the Inferer makes a prediction based on whether the test statistic \(S(x, y)\) exceeds the threshold \(\tau\). Early methods, such as the loss attack~\citep{Yeom2018Privacy}, distinguished between members and non-members by setting a loss threshold. The insight here is that data points with lower loss values are more likely to belong to the training set:
\begin{align}
\label{loss_equation_app}
I(x, y) = \mathbf{1}[-\ell(x, y) > \tau] = \mathbf{1}[\log(p_y) > \tau],
\end{align}
where \( l(\cdot, \cdot) \) is the cross-entropy loss function and \( p_y \) is the softmax probability of the true label \( y \). The shadow model-based attack~\citep{shokri2017membership} involves training similar shadow models to simulate the target model's behavior. Building on these early approaches, subsequent works expanded these techniques to more extensive scenarios~\citep{Leino2020Stolen, nasr2019comprehensive, sablayrolles2019white, song2021systematic, hisamoto2020membership, chen2021machine, choquette2021label, li2021membership}. Recent advancements have incorporated the concept of sample hardness, with approaches like \citet{watson2021importance} adjusting the loss based on sample difficulty. Later studies~\citep{ye2022enhanced, carlini2022membership, zarifzadeh2024low} further refined these techniques. Among them, the \emph{Likelihood Ratio Attack} (LiRA)~\citep{carlini2022membership} represents the state-of-the-art approach, offering superior performance when sufficient shadow models are available.

\vspace{0.2em}
\noindent\textbf{Likelihood Ratio Attack} In this section, we introduce \emph{Likelihood Ratio Attack} LiRA. LiRA exploits the behavioral differences between two types of shadow models for each sample: those trained with the target sample (IN-Models) and those trained without it (OUT-Models). The Inferer trains multiple shadow models for each target sample, dividing them into two groups: one group is trained with the target sample, and the other is not. The objective of LiRA is to determine whether the target model belongs to the IN-Models group or the OUT-Models group. We consider distributions over models:
\begin{equation*}
\mathbb{Q}_{\text{out}}(x, y) = \{f \gets \mathcal{T}(D \setminus \{(x,y)\}) \mid D \gets \mathbb{D}\},
\end{equation*}
\begin{equation*}
\mathbb{Q}_{\text{in}}(x, y) = \{f \gets \mathcal{T}(D \cup \{(x,y)\}) \mid D \gets \mathbb{D}\}.
\end{equation*}
where \( \mathbb{Q}_{\text{out}}(x, y) \) and \( \mathbb{Q}_{\text{in}}(x, y) \) represent models trained excluding or including the target sample $(x, y)$, respectively. The Inferer's task is to determine whether the model \( f \) was drawn from \( \mathbb{Q}_{\text{out}}(x, y) \) or \( \mathbb{Q}_{\text{in}}(x, y) \). In both cases, we can apply the Neyman-Pearson lemma \citep{neyman1933ix} to establish that the most powerful test at a fixed false-positive rate is achieved by thresholding the \emph{likelihood ratio} between the two hypotheses. Specifically, the likelihood ratio is given by:
\begin{equation}
    \label{eq:likelihood_test_general}
    \Lambda(f; x, y) = \frac{p(\text{observed data} \mid H_1)}{p(\text{observed data} \mid H_0)}.
\end{equation}
For sample-tailored MIAs, this becomes:
\begin{equation}
    \label{eq:likelihood_test_sample}
    \Lambda(f; x, y) = \frac{p(f \mid \mathbb{Q}_{\text{in}}(x, y))}{p(f \mid \mathbb{Q}_{\text{out}}(x, y))},
\end{equation}
where $p(f \mid \mathbb{Q}_{b}(x, y))$ denotes the probability density function (PDF) of the model $f$ under the distribution $\mathbb{Q}_{b}(x, y)$. However, directly computing these likelihood ratios is generally intractable because the distributions $\mathbb{Q}_{\text{in}}$ and $\mathbb{Q}_{\text{out}}$ are not analytically known. \citet{carlini2022membership} provided a more tractable one-dimensional statistic: the logit-scaled predicted confidence of the model $f$ on the sample $(x, y)$, denoted as
\begin{equation}
\phi(p_y(x)) = \log \left( \frac{p_y(x)}{1 - p_y(x)} \right).
\label{logit_scaling}
\end{equation}
\citet{carlini2022membership} defined approximate distributions $\tilde{\mathbb{Q}}_{\text{in}}$ and $\tilde{\mathbb{Q}}_{\text{out}}$ as the distributions of $\phi(p_y(x))$ when $(x, y)$ is included or excluded from the training data. The likelihood ratio then simplifies to:
\begin{equation}
    \Lambda(f; x, y) = \frac{p(\phi(p_y(x)) \mid \tilde{\mathbb{Q}}_{\text{in}})}{p(\phi(p_y(x)) \mid \tilde{\mathbb{Q}}_{\text{out}})}.
\end{equation}
\noindent This reduction to a one-dimensional statistic allows for efficient computation of the likelihood ratio with $f$. Specifically, the decision rule in LiRA can be written as:
\begin{align}
\label{MIA_LiRA_app}
I(x, y) = \mathbf{1}[\Lambda(f; x, y) > \tau] = \mathbf{1}[\frac{p(\phi(p_y(x)) \mid \tilde{\mathbb{Q}}_{\text{in}})}{p(\phi(p_y(x)) \mid \tilde{\mathbb{Q}}_{\text{out}})} > \tau]
\end{align}
\vspace{0.2em}
\noindent\textbf{Assumptions of LiRA.} Despite the superior performance, the effectiveness of LiRA relies on several assumptions:

\vspace{0.2em}
\noindent\textbf{I. Dependence on shadow models.} LiRA needs to train a substantial number of shadow models to accurately approximate the distributions of IN-Models and OUT-Models. 

\vspace{0.2em}
\noindent\textbf{II. Knowledge of the training details.} LiRA assume that Inferer possesses detailed knowledge of the target model’s training environment, including hyperparameters and architectural specifics.

\subsection{Overfitting Drives MIA Effectiveness}
A key factor influencing the success of MIAs is the overfitting, a phenomenon well-documented in existing literature~\citep{shokri2017membership, Yeom2018Privacy, Salem2019ML, Leino2020Stolen, chen2020gan}. Overfitting arises when a model learns to recognize not just general patterns but also noise or overly specific features from its training data, thus enhancing performance on seen data while impairing its ability to generalize to unseen data. This issue is especially prominent in complex models and when training datasets are insufficiently representative of the broader data distribution~\citep{bishop2006pattern}. Deep learning models are particularly vulnerable to this issue due to their extensive parameterization, which often leads to the memorization of intricate details from the training data~\citep{song2017machine, carlini2019secret, murakonda2020ml, zhang2021understanding}. These models, when trained on non-representative datasets, exhibit discrepancies in behavior between training and non-training data, and MIAs exploit these disparities. Theoretical work has shown that overfitting directly contributes to these behavioral differences, which MIAs can detect and use to infer whether a given data point was included in the training set~\citep{bentley2020quantifying}.

\subsection{Adversarial Attacks.}

Recent studies have increasingly highlighted the vulnerability of neural networks to adversarial attacks—slightly altered inputs that are strategically crafted to induce misclassification~\citep{carlini2017adversarial, kurakin2016adversarial, wang2019convergence, zhang2020dual}. These attacks pose a serious threat to security-critical systems, such as autonomous driving and medical diagnostics~\citep{gao2021maximum, chen2015deepdriving, ma2021understanding, nguyen2015deep, szegedy2013intriguing}. A seminal work by Szegedy et al.~\citep{szegedy2013intriguing} first pointed out the existence of adversarial examples: given a valid input $x$ with its true label $y$ and a trained classifier $f$, it is often possible to find another input $x'$ such that $f(x') \neq y$ yet $x$, $x'$ are close according to some distance metric. This insight laid the foundation for understanding how neural networks can be deceived by small but strategically crafted perturbations. In this paper, we focus on the $\ell_{\infty}$ distance metric, and the example $x'$ is referred to as the adversarial example.

\vspace{0.2em}
\noindent\textbf{Fast Gradient Sign Method (FGSM).}
Based on the study of \citet{szegedy2013intriguing}, \citet{goodfellow2014explaining} put forward the Fast Gradient Sign Method (FGSM): given an image $x$,  FGSM sets
\begin{align}
\label{Equ_FGSM}
x^{\prime} = x + \epsilon \cdot \sign(\nabla\ell(f(x),y)),
\end{align}
where $\epsilon$ is chosen to be sufficiently small to ensure that the difference is unnoticeable by humans; $y$ is the true label; $\ell$ is a loss function that measures the performance of the classifier. A common loss function $\ell$ in Eq.~(\ref{Equ_FGSM}) is the \emph{Cross-Entropy}~(CE) loss:
\begin{align}
\label{Equ_CEloss}
\textnormal{CE}(f, x, y) = -\log(p_y),
\end{align}
where $p$ is the softmax of logits of the model outputs. For each pixel of the image, FGSM uses the gradient of the loss function to determine in which direction the pixel's intensity should be increased or decreased and minimizes the loss function.

\vspace{0.2em}
\noindent\textbf{Projected Gradient Descent Attack (PGD).}
\citet{Madry18PGD} introduced a simple refinement of FGSM, which is the projected gradient descent (PGD) attack. Instead of taking a single step of size $\epsilon$ in the direction of the gradient sign, multiple smaller steps are taken in PGD (the result is clipped by the same $\epsilon$). Specifically, we start with setting $x^{0}=x$, and then in each iteration:
\begin{align*}
x^{\prime}_{(t+1)} = \Pi_{\mathcal{B}_\epsilon[x_{(0)}]}(x'_{(t)}+\alpha \sign(\nabla_{x'_{(t)}}\ell(f(x'_{(t)}),y)), 
\end{align*}
$t = 0, 1, 2,\ldots$, where 
\begin{align*}
\epsball[x] = \{x^{\prime} \mid d_{\infty}(x,x')\le\epsilon\},
\end{align*}
is the closed ball of radius $\epsilon>0$ centered at $x$; the  $x_{(0)}$ refers to the starting point which corresponds to the natural example $x$; $x^{(t)}$ is the adversarial example at step $t$; $\Pi_{\mathcal{B}_\epsilon[x^{(0)}]}(\cdot)$ is the projection function that projects the adversarial variant back to the $\epsilon$-ball centered at $x^{(0)}$ if necessary; the $L_{\infty}$ distance metric is $d_{\infty}(x,x^{\prime})=\|x-x^{\prime}\|_\infty$; and $\ell$ is \emph{cross entropy}~(CE) loss.

\vspace{0.2em}
\noindent\textbf{Carlini and Wagner attack (CW).}
\citet{carlini2017towards} observed the phenomenon of gradient vanishing in the widely used CE loss for potential failure. Motivated by this, \citep{carlini2017towards} replaced the CE loss with several possible choices. Among these choices, the widely used one for the untargeted attack is 
\begin{align}
\label{Equ_CWloss}
\textnormal{CW}(x, y) = -z_y(x^{\prime})+\max_{i\neq y}z_i(x^{\prime}).
\end{align}
where $z$ is the logits of the model outputs.

\vspace{0.2em}
\noindent\textbf{AutoAttack and Minimum Margin attack (MM).}
\citet{croce2020reliable} claimed that the fixed step size and the lack of diversity in attack methods are the main reasons for the limitations of previous studies, and they put forward an ensemble of diverse attacks called AutoAttack. \citet{gao2022fast} argued that the high computational cost of \emph{AutoAttack} is unnecessary for identifying \emph{the most adversarial example}: 
\begin{definition}[The most adversarial example]
Given a natural example $x$ with its true label $y$ and, the most adversarial example $x^\ast$ within $\epsball[x]$ is defined as:
\begin{align}
\label{mostadv:def}
{\forall} x^{\prime} \in \epsball[x],
x^\ast = \argmax_{x^{\prime}}{-(z_y{(x')} - \max_{i \neq y}{z_i{(x')}})},
\end{align}
where $\epsball[x] = \{x^{\prime} \mid d_{\infty}(x,x')\le\epsilon\}$ is the closed ball of radius $\epsilon>0$ centered at $x$; $z_y{(x')} = {f(x')}_y$; $z_i{(x')} = {f(x')}_i$.
\end{definition}

\subsection{Traditional Adversarial Data Detection}
In addition to enhancing model robustness through improved adversarial training~\citep{Chen_2020_CVPR, wang2019convergence, wu2020adversarial, zhang2020geometry}, recent research has focused on the detection of adversarial data. Most of these approaches rely on features extracted from \emph{deep neural networks} (DNNs) and aim to train classifiers that distinguish between natural and adversarial data. Several recent advancements in adversarial data detection include the use of a cascade detector based on the \emph{Principal Component Analysis} (PCA) projection of activations~\citep{li2017adversarial}, detection subnetworks based on activations~\citep{metzen2017detecting}, and logistic regression detectors that utilize \emph{Kernel Density} (KD) and \emph{Bayesian Uncertainty} (BU) features~\citep{grosse2017statistical}. Other methods include an augmented neural network detector that employs statistical measures, a learning framework designed to address previously unexplored vulnerabilities in models~\citep{rouhani2017curtail}, and a characterization of adversarial data based on \emph{local intrinsic dimensionality} (LID)~\citep{ma2018characterizing}. Furthermore, generative classifiers based on Mahalanobis distance scores have also been proposed as effective tools for detecting adversarial data~\citep{lee2018simple}.

\section{Proof of Theorem \ref{thm:gd_norm}}
\label{app_proof}
Assume that the $\epsilon$-ball is small such that the local curvature of $\ell \circ f$ around $x$ can be well approximated by its second-order Taylor approximation, that is, there exists a Hessian matrix $\mathbf{H}$ and a gradient vector $\mathbf{g}=\nabla_{x}\ell(f(x),y)$ such that $\forall x' \in \mathcal{B}_\epsilon[x]$,
\[
\ell(f(x'),y) \approx \ell(f(x),y) + \mathbf{g}^T(x'-x) + \frac{1}{2}(x' - x)^T \mathbf{H} (x' - x).
\]

\noindent After taking one step of signed gradient descent with respect to the input sample:
\begin{equation}\label{app_pgd_step}
    x^{\prime} = x - \alpha \cdot \sign(\nabla_{x}\ell(f(x),y)),
\end{equation}
\noindent We analyze the change of gradient norm in the Taylor approximation.
\[
\begin{multlined}
\|\nabla_{x^{\prime}} \ell(f(x^{\prime}), y)\|^2 \approx{} \|\mathbf{H}(x' - x) + \mathbf{g}\|^2 \\
= \| \mathbf{g} \|^2 + 2 \mathbf{g}^T\mathbf{H}(x' - x) + (x' - x)^T\mathbf{H}^2(x' - x).
\end{multlined}
\]

\noindent Substituting \eqref{app_pgd_step}, we obtain
\[
\begin{multlined}
\|\nabla_{x^{\prime}} \ell(f(x^{\prime}), y)\|^2 = \| \nabla_{x}\ell(f(x),y) \|^2 \\
- 2 \alpha \cdot \mathbf{g}^T\mathbf{H} \sign(\mathbf{g}) + \alpha^2 \sign(\mathbf{g})^T\mathbf{H}^2\sign(\mathbf{g}).
\end{multlined}
\]

\noindent Thus, by letting 
\[
\alpha < \frac{2\mathbf{g}^T\mathbf{H} \sign(\mathbf{g})}{\sign(\mathbf{g})^T\mathbf{H}^2\sign(\mathbf{g})},
\]
we have the following inequality (approximately),
\begin{equation}
    \|\nabla_{x^{\prime}} \ell(f(x^{\prime}), y)\| < \|\nabla_{x} \ell(f(x), y)\|.
\end{equation}

\section{The Risks of Adversarial Vulnerability of MIAs in Practical Scenarios}
\label{Asec:examples}

MIAs are increasingly used in practical pipelines for privacy auditing~\citep{murakonda2020ml, song2020introducing} and model copyright verification~\citep{wang2024stronger}. However, our findings show that their adversarial vulnerability can lead directly to incorrect or even strategically manipulated conclusions. We highlight several representative scenarios where fabricated membership can create tangible risks.

\vspace{0.2em}
\noindent\textbf{I. Misleading copyright or dataset-provenance verification.}
When MIAs are used as third-party tools for detecting copyright infringement \citep{wang2024stronger}, their susceptibility to adversarial perturbations creates a vulnerability for malicious exploitation. Under this setting, an adversary could deliberately add imperceptible perturbations to their own private images to fabricate membership. A vulnerable MIA would then falsely conclude that the target vision model was trained on these manipulated inputs. Such fabricated evidence can be used to launch spurious copyright claims or regulatory complaints, even when no data misuse has occurred. This exposes a direct path for legal and financial exploitation if MIAs are trusted as forensic tools.

\vspace{0.2em}
\noindent\textbf{II. Manipulating MIA-based training data extraction and competitive auditing.}
Several recent pipelines use MIAs to infer the composition of a competitor’s training dataset or to verify compliance with data-deletion requests. In shared-data or competitive industrial settings, adversaries could upload imperceptibly perturbed images into a public or collaborative dataset. These fabricated members would cause MIA-based extraction tools to incorrectly infer that the model trained on these poisoned samples, thereby distorting the reconstructed training distribution. This undermines dataset-governance workflows and enables malicious parties to bias or sabotage competitors’ auditing systems.

\vspace{0.2em}
\noindent\textbf{III. Exploiting MIA vulnerabilities in defensive or adversarial intelligence settings.}
Because fabricated members follow a distinctive gradient-norm collapse, defenders could intentionally craft such samples to confuse attackers attempting to audit their models using MIAs. For example, an organization concerned about model-stealing or data-reconstruction attacks may release selectively fabricated samples in a public-facing API. Attackers relying on MIA-based analysis would misinterpret these fabricated members as genuine training samples, leading them to incorrectly infer the model’s training data or privacy weaknesses. This demonstrates that adversarial membership manipulation can be used not only offensively but also strategically in defensive contexts.

In all scenarios, the underlying risk stems from the same mechanism revealed in our analysis: adversarial perturbations can reliably push non-members into regions that MIAs interpret as strong evidence of membership. Without adversarially robust inference methods, MIAs cannot be reliably used in real-world privacy, copyright, or forensic workflows.

\section{Finite-Difference Gradient Estimation}
\label{finite_difference}

In the main paper, we assume white-box access when constructing the strongest Detector for \textbf{MFD}.  
This assumption allows direct computation of input gradients $\nabla_x \ell(f(x),y)$, which is essential for measuring the gradient-norm collapse phenomenon that distinguishes fabricated from true members.  
However, in many practical auditing scenarios, the model may only be accessible through black-box APIs that return confidence scores or class probabilities.

\vspace{0.2em}
\noindent\textbf{Black-box feasibility.}
Fortunately, the adversarial-robustness literature has established that input gradients can be reliably approximated in a black-box setting using finite-difference or score-based estimators such as NES (Natural Evolution Strategies) and SPSA.  
These methods exploit the key identity that for sufficiently small perturbation radius $\delta$,
\[
\nabla_x f(x) = 
\mathbb{E}_{u\sim \mathcal{N}(0,I)} 
\left[
\frac{f(x+\delta u)-f(x)}{\delta}u
\right],
\]
meaning that the directional response of the model to randomized perturbations already encodes information about the gradient.  
This enables gradient-norm estimation without requiring logits or intermediate activations—confidence scores alone are sufficient.

\vspace{0.2em}
\noindent\textbf{Finite-difference estimator.}
We follow the standard formulation of black-box gradient estimation in adversarial attacks and use the symmetric finite-difference estimator:
\[
\widehat{\nabla_x f(x)} \;=\; 
\frac{1}{N}
\sum_{i=1}^{N}
\frac{
    f(x+\delta u_i) - f(x-\delta u_i)
}{
    2\delta
} \, u_i,
\]
where $u_i$ are sampled from $\mathcal{N}(0,I)$ and normalized to unit norm.  
The estimated gradient norm is then computed as:
\[
\widehat{g}(x) \;=\;
\left\|
\widehat{\nabla_x \ell(f(x),y)}
\right\|.
\]

This procedure requires only \emph{model confidence queries} and therefore applies in the strictest black-box setting.

\vspace{0.2em}
\noindent\textbf{Practical configuration.}
For CIFAR-10 and ResNet-18, we find the following configuration offers a favorable trade-off between query cost and estimation accuracy:
\begin{itemize}[leftmargin=1.3em]
    \item number of directions $N = 100$,
    \item perturbation radius $\delta = 1\text{e-3}$,
    \item Gaussian $u_i$ followed by $\ell_2$ normalization,
    \item using the model's confidence score $p_y(x)$ in place of logits.
\end{itemize}
This yields a query-efficient estimate of the gradient norm that is sufficiently stable for MFD.

\vspace{0.2em}
\noindent\textbf{Effectiveness of black-box MFD.}
We evaluate the black-box variant of our MFD on CIFAR-10 with ResNet-18 under the same balanced evaluation protocol used in the main paper.  
While the AUC decreases from \textbf{0.9111} (white-box) to approximately \textbf{0.82} under finite-difference estimation, the detector remains highly effective and clearly separates fabricated from true members.  
This result demonstrates that our gradient-based detection principle generalizes beyond the white-box setting. Overall, this section provides a simple but representative black-box instantiation, showing that \textbf{MFD remains effective even without white-box access}, thus broadening the real-world applicability of our detection framework.

\section{Confidence-Matched Analysis of Gradient Norm}
\label{app_confidence_match}

A natural question is whether the smaller input-gradient norms of fabricated members are simply a byproduct of their higher target-class confidence. To examine this, we perform a confidence-matched comparison between true and fabricated members.

Specifically, we group samples according to their target-class confidence $p_y(x)$ and compare the input-gradient norm $\|\nabla_x \ell(f(x),y)\|$ within each confidence bin. If the gradient-based signal used by \textbf{MFD} were merely reflecting confidence, then the difference between true and fabricated members should largely disappear after conditioning on $p_y(x)$.

\begin{table}[h]
\centering
\small
\setlength{\tabcolsep}{15pt}
\begin{tabular}{c|cc}
\toprule
$p_y(x)$ bin & member & fabricated \\
\midrule
$[0.90,0.95]$ & $1.09 \pm 0.21$ & $\mathbf{0.73 \pm 0.18}$ \\
$[0.95,0.98]$ & $0.93 \pm 0.17$ & $\mathbf{0.58 \pm 0.15}$ \\
$[0.98,0.99]$ & $0.82 \pm 0.15$ & $\mathbf{0.49 \pm 0.14}$ \\
\bottomrule
\end{tabular}
\caption{Confidence-matched comparison of input-gradient norms. Fabricated samples consistently exhibit smaller gradient norms than true members within the same target-class confidence range.}
\label{tab:confidence_match}
\end{table}

As shown in \Cref{tab:confidence_match}, fabricated samples remain consistently lower in gradient norm than true members across all confidence bins. This indicates that the gradient-norm gap is not solely explained by confidence, but also reflects the geometric effect induced by the fabrication process itself. This observation supports the use of $\|\nabla_x \ell(f(x),y)\|$ as a detection statistic in \textbf{MFD}.

\section{Adaptive MFA}
\label{app_adaptive_MFA}

\vspace{0.2em}
\noindent\textbf{Adaptive fabrication via gradient-penalized optimization.}
A natural question is whether a Fabricator aware of \textbf{MFD} can jointly maximize fabrication success while suppressing the gradient-norm signal used for detection.  
To examine this, we consider an adaptive variant of MFA that augments the original objective with a penalty on the input-gradient magnitude:
\begin{equation}
\label{eq:adaptive_obj}
    \max_{\|\delta\|_\infty\le\epsilon}
    \Big( p_y(x+\delta) - \lambda_{\text{adv}} \cdot \|\nabla_x \ell(f(x+\delta),y)\| \Big),
\end{equation}
where $\lambda_{\text{adv}} \in \{0.05, 0.1, 0.5\}$ controls the strength of the penalty.
The rest of the settings follow the main experimental configuration: CIFAR-10, ResNet-18, LiRA as the reference MIA, and the same $\epsilon$ and optimization budget used in \S\ref{subsec:mfa}.  
This adaptive objective corresponds to a Fabricator attempting to counteract the ``gradient-norm collapse'' phenomenon formalized in Theorem~\ref{thm:gd_norm}.

\vspace{0.2em}
\noindent\textbf{Experimental behavior and the trade-off.}
We first validate the baseline (non-adaptive) MFA on CIFAR-10.  
Consistent with the results reported in the main paper, applying MFA increases LiRA's \emph{Error Area} from \textbf{0.2814} to \textbf{0.3523},  
and raises the \emph{Equal Error Rate} from \textbf{36.70\%} to \textbf{41.85\%}.  
Meanwhile, our \textbf{MFD} detects such fabricated members with a high \emph{AUC} of \textbf{0.9111}.  
These numbers establish a strong starting point for examining whether adaptive fabrication can reduce detectability without sacrificing attack strength.

\begin{itemize}[leftmargin=1.5em]
\item We then introduce the gradient penalty.  
With a mild penalty ($\lambda_{\text{adv}}=0.05$), we observe a marginal reduction in the MFD signal:  
the median gradient norm of fabricated samples increases slightly (by roughly 8\%), reducing the MFD AUC from \textbf{0.9111} to \textbf{0.8746}.  
However, fabrication effectiveness also decreases:  
LiRA's Error Area only rises to \textbf{0.3291} instead of \textbf{0.3523}, and the EER drops from \textbf{41.85\%} to \textbf{39.20\%}.  
This reflects a direct manifestation of Theorem~\ref{thm:gd_norm}: suppressing gradient reduction restricts the optimizer’s ability to move toward the high-confidence basin where MFA achieves maximal effect.

\item Increasing the penalty to $\lambda_{\text{adv}}=0.1$ produces a clearer shift.  
The gradient-norm suppression becomes stronger (roughly a 15\% increase in median gradient magnitude), and MFD AUC drops further to \textbf{0.8327}.  
Yet fabrication deteriorates significantly:  
LiRA's Error Area now only reaches \textbf{0.3048}, nearly erasing the gains achieved by non-adaptive MFA, and the EER falls to \textbf{37.05\%}, approaching the original no-attack baseline.  
This empirically confirms an intrinsic limitation: evading gradient-norm detection forces the optimization to remain in regions of weaker confidence ascent.

\item When the penalty is further strengthened to $\lambda_{\text{adv}}=0.5$,  
the Fabricator becomes dominated by the constraint.  
The gradient norm becomes visually indistinguishable from true members in most cases, reducing MFD AUC to \textbf{0.5914}.  
However, fabrication almost entirely collapses:  
LiRA's Error Area reaches only \textbf{0.2877} (barely above the unperturbed value of \textbf{0.2814}),  
and the EER drops to \textbf{36.82\%}, effectively neutralizing the attack.  
At this point, the Fabricator is unable to fool the Inferer while simultaneously masking the gradient-norm cue.
\end{itemize}
\vspace{0.2em}
\noindent\textbf{Interpretation through gradient-norm collapse.}
These observations align with the geometry predicted by Theorem~\ref{thm:gd_norm}.  
The theorem states that, under small $\epsilon$, a signed gradient step inevitably moves the input toward regions of lower input-gradient magnitude.  
Fabrication \emph{requires} traversing this trajectory into increasingly sharp, high-confidence basins where the loss landscape flattens, naturally shrinking the gradient norm.  
Adaptive attacks attempting to counteract this process must keep gradients artificially large, which fundamentally conflicts with the conditions needed to maximize membership likelihood.  
Thus, adaptive MFA faces a structural trade-off:  
\emph{reducing detectability directly undermines the optimization path that produces strong fabrication;  
strengthening fabrication unavoidably triggers gradient-norm collapse, making detection easier.}

\vspace{0.2em}
\noindent\textbf{Conclusion of adaptive analysis.}
Across all settings, we find that while adaptive fabrication can moderately reduce the MFD signal at small $\lambda_{\text{adv}}$,  
the cost is a proportional and sometimes severe degradation of fabrication effectiveness.  
Stronger penalties do suppress gradient signatures but simultaneously collapse the attack.  
This persistent and quantifiable trade-off confirms that \textbf{MFD} remains robust even against adaptive MFA,  
and that gradient-geometry signals form a structurally unavoidable barrier for adversarial manipulation.

\section{Behavior Under MI Defenses}
\label{app_defense_results}

MI defenses are designed to weaken the power of MIAs; from the perspective of adversarial membership manipulation, this actually creates a \emph{more permissive} environment for our framework: once the baseline auditor is weaker, it becomes easier to further degrade its reliability via \textbf{MFA}, while the geometry-driven components (\textbf{MFD}, \textbf{AR-MIAs}) remain effective because they directly exploit gradient behavior rather than raw membership signals. In this section we therefore provide a concise analysis under standard MI defenses on CIFAR-10 with ResNet-18, keeping all other settings identical to the main experiments and focusing on LiRA as a representative strong MIA. Concretely, we consider (i) $\ell_1$-regularized training with coefficient $\lambda_{\ell_1}=10^{-4}$, (ii) knowledge distillation with temperature $T=2$ and a student trained with a $0.7/0.3$ mix of soft and hard labels, and (iii) DP-SGD with clipping norm $C=1.0$ and noise multiplier $\sigma=1.0$.

\begin{itemize}[leftmargin=1.5em]
\item \textbf{MFA under MI defenses.}
For LiRA without MI defenses, our \textbf{MFA} with perturbation budget $\|\delta\|_{\infty} \le 4/255$ already induces a pronounced degradation, yielding an Error Area of $0.3523$ and an Equal Error Rate (EER) of $41.85\%$. When we retrain the same model under $\ell_1$-reg, distillation, or DP-SGD, the underlying LiRA auditor becomes less confident, and \textbf{MFA} becomes even more effective: across the three defenses, the Error Area increases into the $0.40$–$0.45$ range and the EER rises to roughly $48\%$–$52\%$ (i.e., about $15\%\uparrow$–$25\%\uparrow$ relative to the undefended case). This confirms the intuition that MI defenses, while reducing raw MIA accuracy, make it easier for a Fabricator to push non-members into regions where the auditor is systematically misled.

\item \textbf{MFD under MI defenses.}
Our detector \textbf{MFD} relies on the \emph{gradient-norm collapse} geometry rather than on any particular MIA score. On the undefended model, \textbf{MFD} achieves an AUC of $0.9111$ against fabricated members from \textbf{MFA} ($\|\delta\|_{\infty} \le 4/255$). Under $\ell_1$-reg, distillation, and DP-SGD, we observe only mild fluctuations, with AUC values remaining in the high $0.88$–$0.91$ range. In other words, although MI defenses significantly reduce the separability between true members and non-members for MIA itself, they do \emph{not} disrupt the gradient-geometry signal that \textbf{MFD} exploits; fabricated members still concentrate in low-gradient basins and remain reliably detectable.

\item \textbf{AR-MIAs under MI defenses.}
For adversarially robust LiRA (\textbf{AR-MIA} with the weighting in Eq.~\eqref{weight_definition}), we focus on the LiRA series and set $\lambda=10$. Under standard training, the base LiRA achieves an AUC of $0.6832$, while our adversarially robust variant improves this to $0.7937$, i.e., an absolute gain of about $0.11$ (approximately $15\%\uparrow$ in relative terms). When MI defenses are enabled, LiRA’s AUC drops substantially into the $0.55$–$0.58$ band, reflecting the intended regularization effect on memorization; nevertheless, the corresponding adversarially robust LiRA still achieves AUCs around $0.65$–$0.69$, preserving a similar absolute improvement (again on the order of $0.10$–$0.11$). Thus, even in the presence of MI defenses that inevitably weaken membership signals on true members, \textbf{AR-MIAs} continue to provide consistent gains over their non-robust counterparts and retain strong discriminative power against fabricated members.
\end{itemize}

\section{Discussion of Limitations on MFA}
\label{app:limitation}
Member Fabrication Attack, which introduces imperceptible adversarial perturbations to the data, encounters limitations in its application and practicality.

\vspace{0.2em}
\noindent\textbf{I. Practical Use Case Requirements.}
The primary limitation of Member Fabrication Attack is its dependency on specific use-case scenarios, such as those where adding noise to data before an inferer's access is feasible. In scenarios where the inferer can directly obtain the original data and conduct membership inference, Member Fabrication Attack cannot be effectively deployed.

\vspace{0.2em}
\noindent\textbf{II. Limited Application Beyond Image Datasets.}
The second limitation concerns its restricted applicability beyond image datasets. The specific characteristics of image data allow for the effective implementation of subtle perturbations without compromising data integrity. Extending this method to non-image datasets, such as text or tabular data, presents significant challenges. The concept of 'imperceptible' changes in these datasets demands a distinct approach. Adapting our methodology to accommodate these varied formats is a primary focus of our future work.

\section{T-SNE Visualization of Semantic Features}
\label{appdendix_tsne}
\begin{figure*}[!t]
    \begin{center}
    {\includegraphics[width=0.196\textwidth]{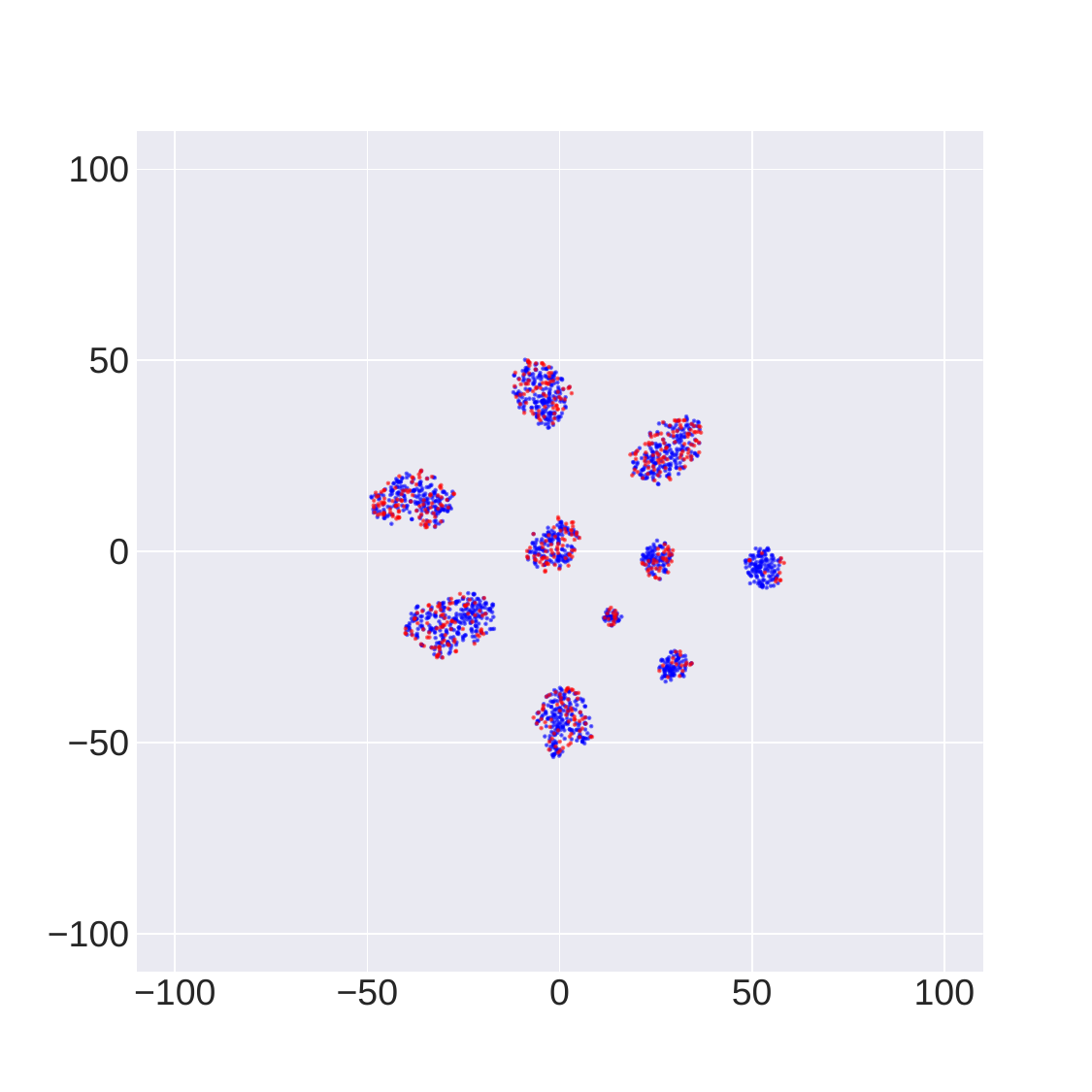}}
    {\includegraphics[width=0.196\textwidth]{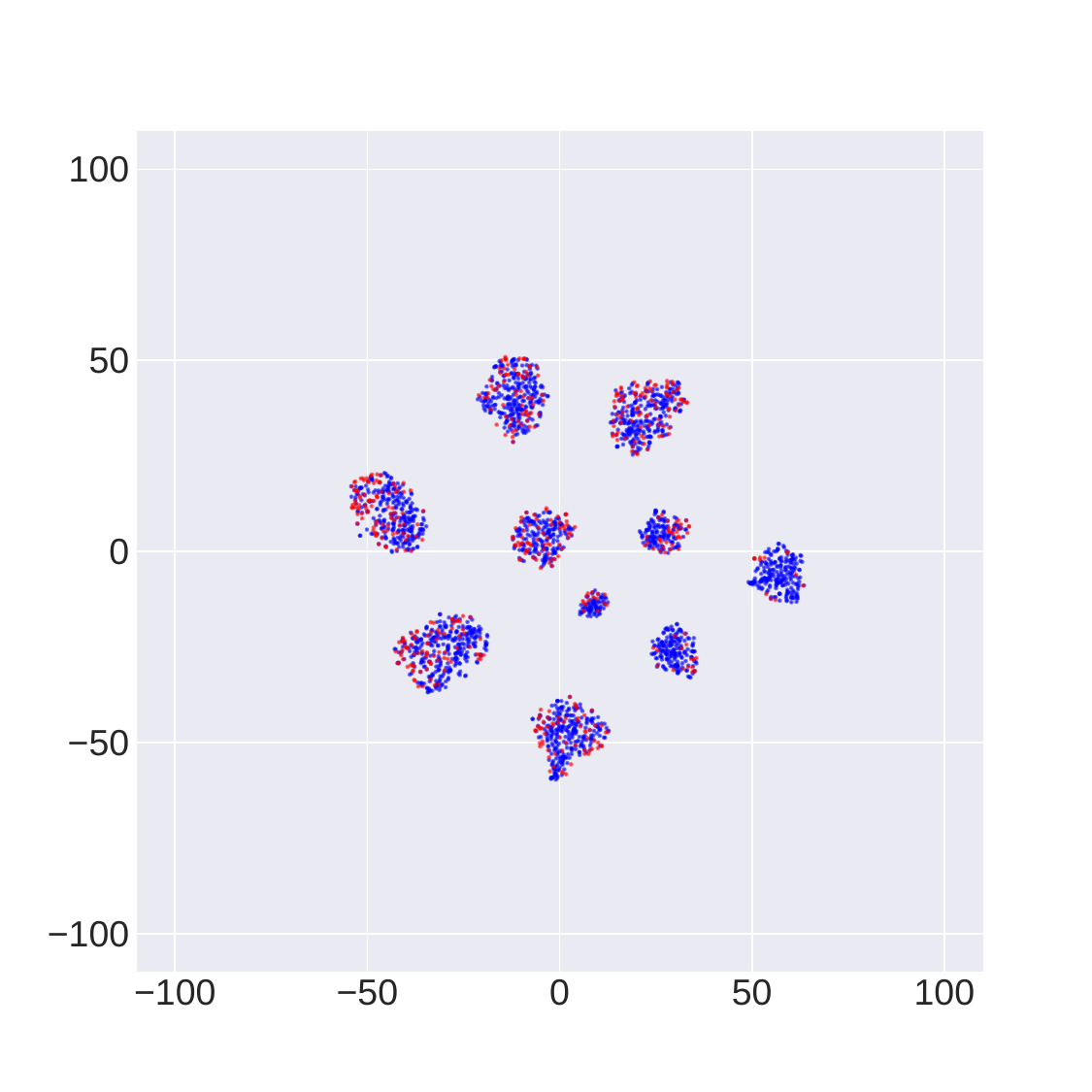}}
    {\includegraphics[width=0.196\textwidth]{TSNE_new/logit_4.pdf}}
    {\includegraphics[width=0.196\textwidth]{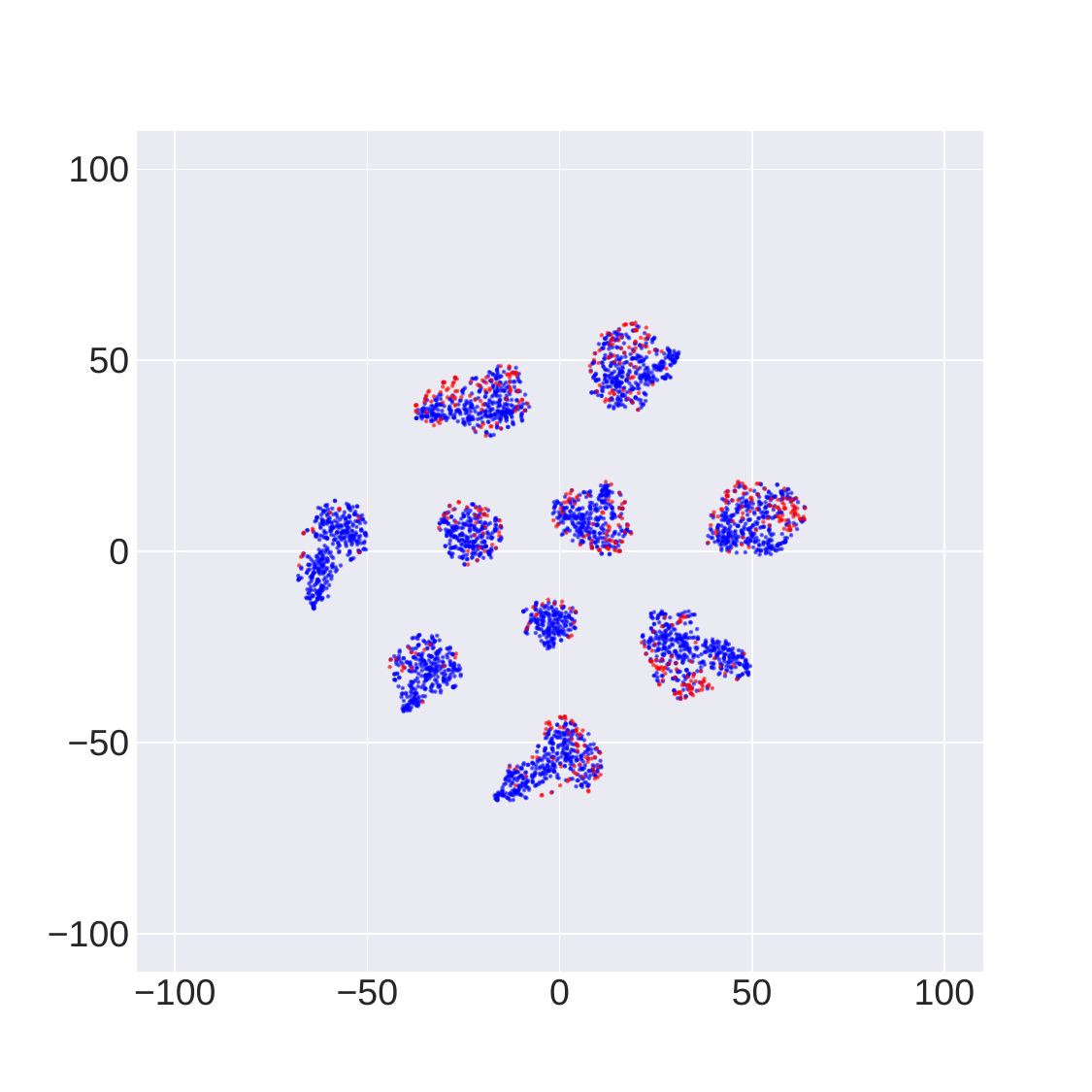}}
    {\includegraphics[width=0.196\textwidth]{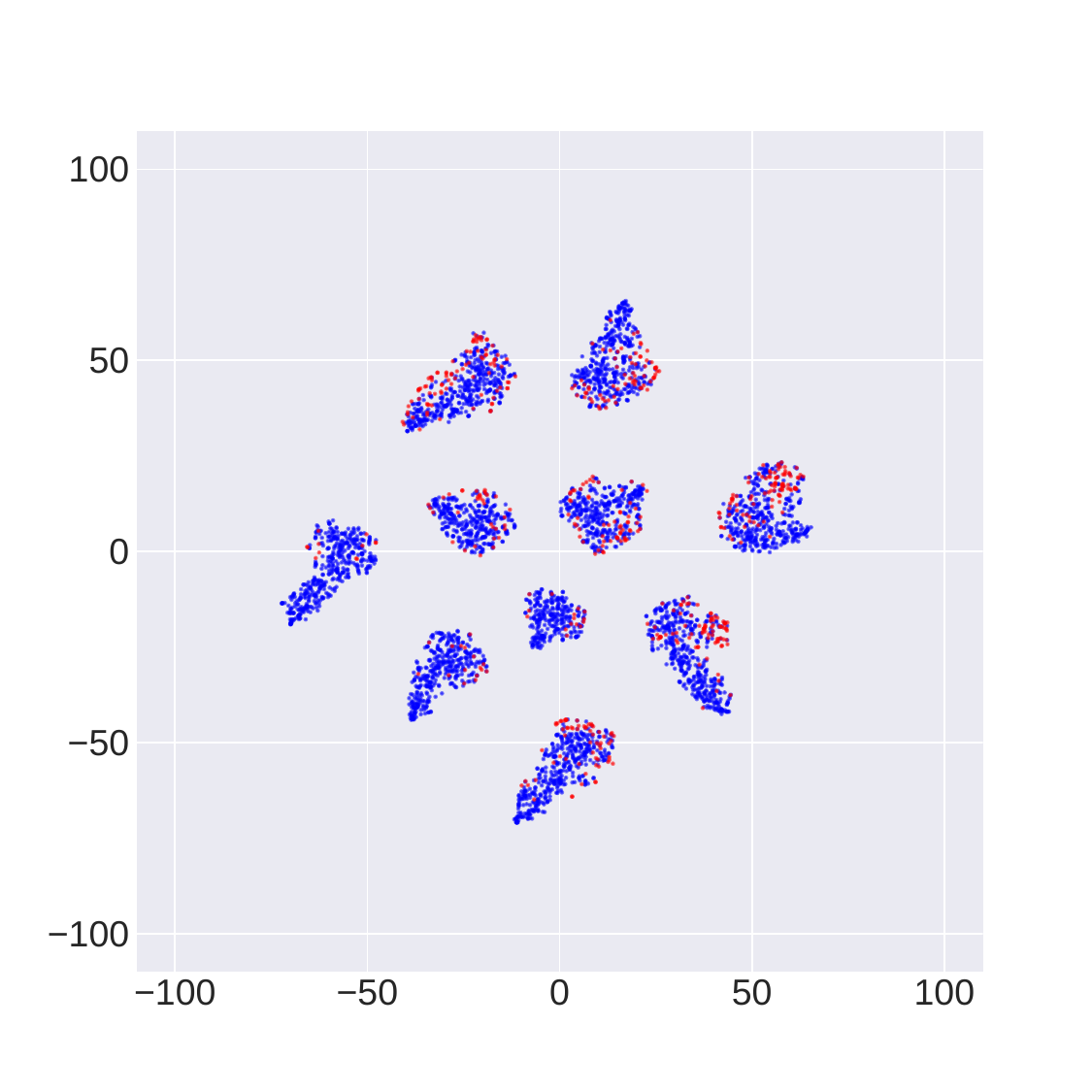}}
    {\includegraphics[width=0.196\textwidth]{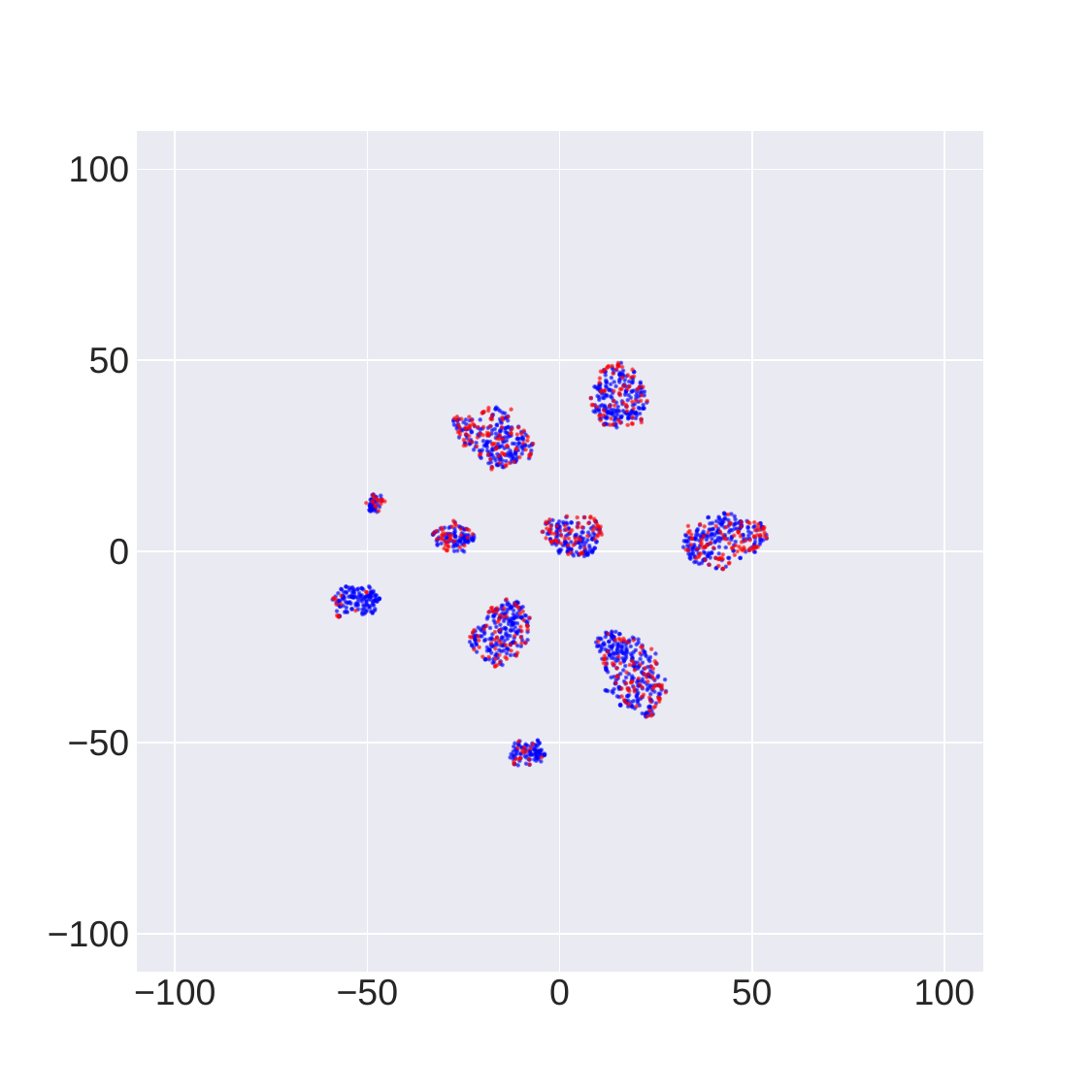}}
    {\includegraphics[width=0.196\textwidth]{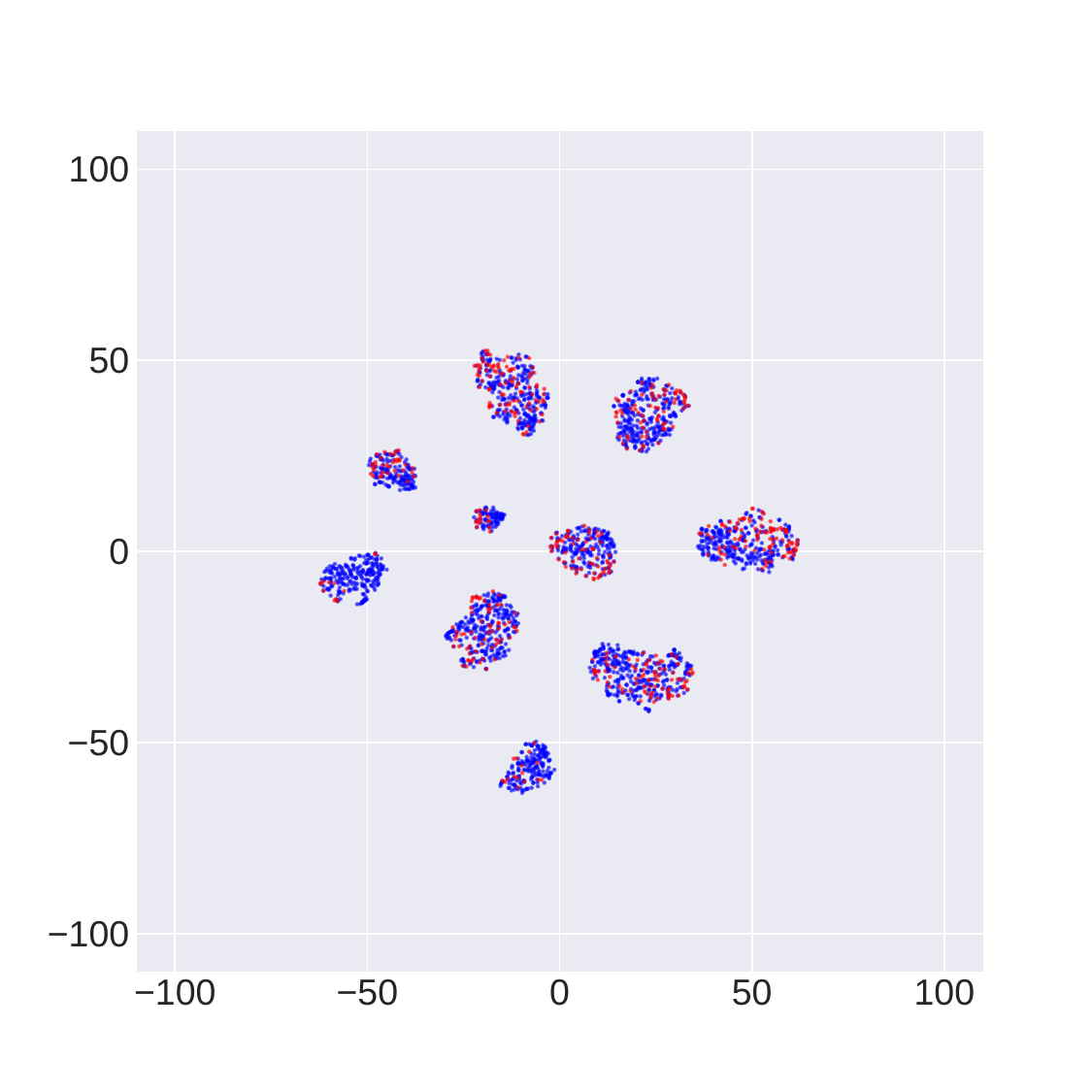}}
    {\includegraphics[width=0.196\textwidth]{TSNE_new/logit_more_4.pdf}}
    {\includegraphics[width=0.196\textwidth]{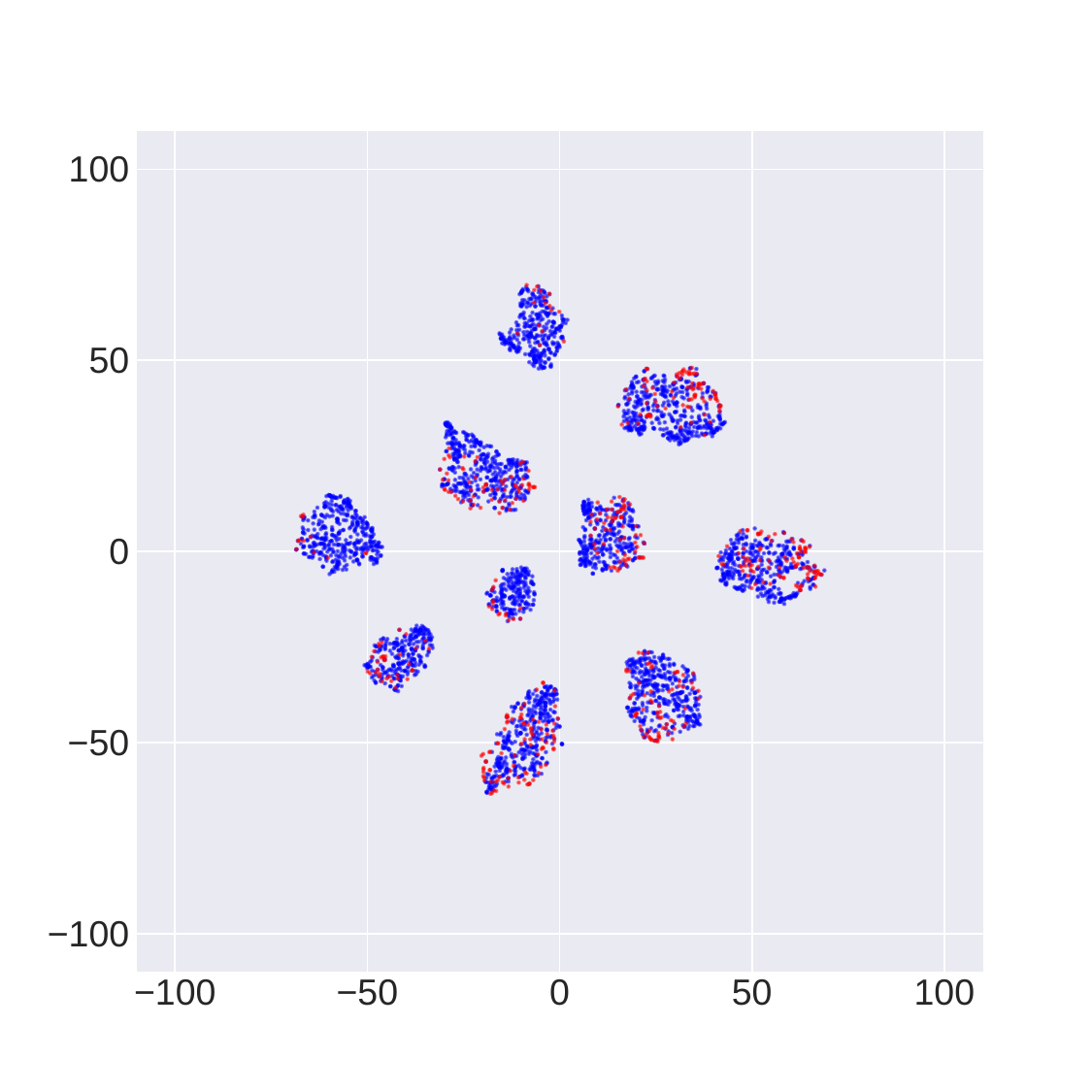}}
    {\includegraphics[width=0.196\textwidth]{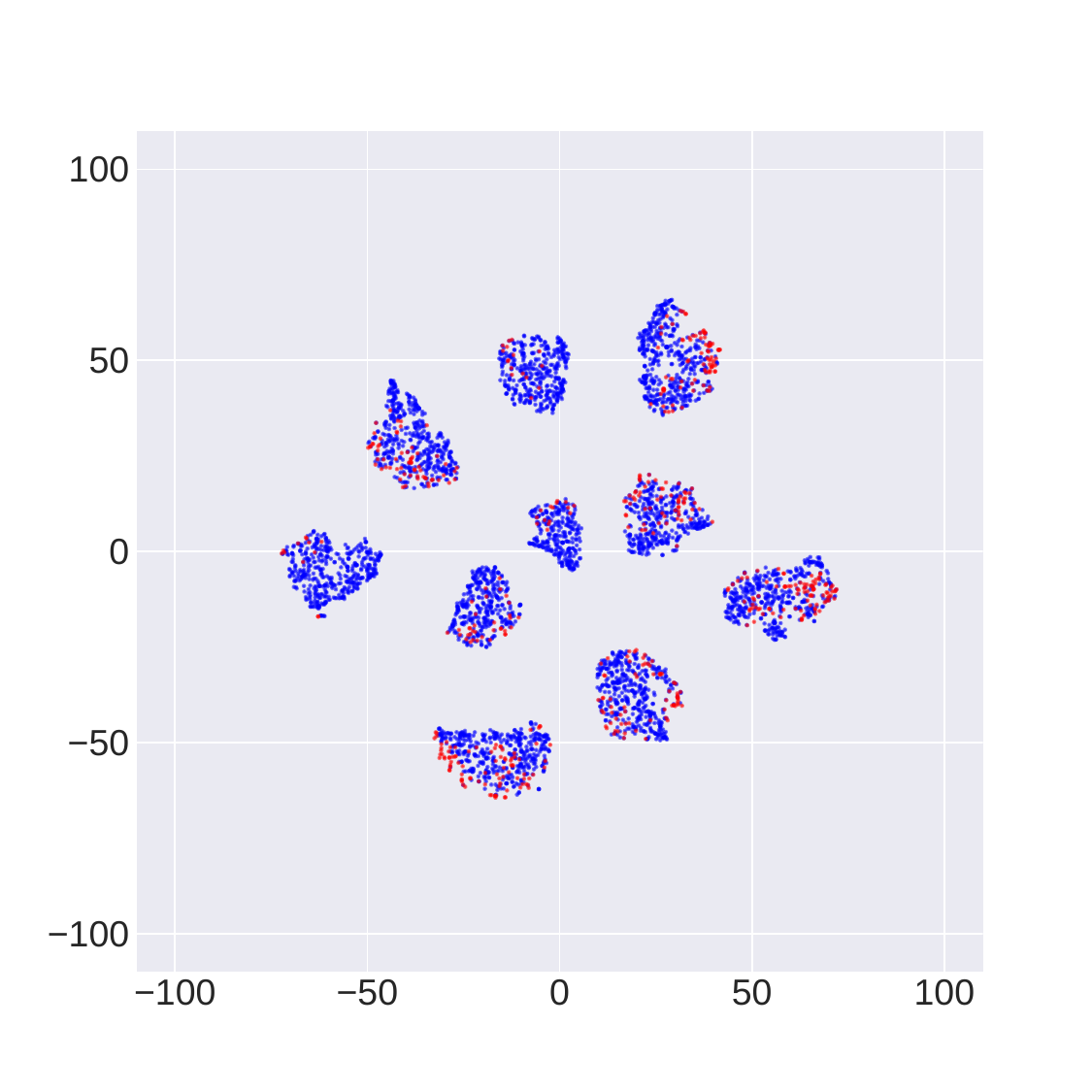}}
    \caption{\footnotesize  Visualization of the Distribution of Fabricated and True Members in Different Semantic Feature Spaces Using t-SNE \citep{maaten2008visualizing}. The first row displays the semantic features at the penultimate layer, with perturbation constrained to $(\|\delta\|_{\infty} \leq 2.0/255)$ to $(\|\delta\|_{\infty} \leq 6.0/255)$ from left to right. The second row shows the semantic features at the antepenultimate layer, with the same range of perturbations. Red dots represent true members' semantic features, and blue dots represent fabricated members. We observe a high degree of overlap between the distributions of fabricated and true members, indicating that semantic features alone are insufficient to distinguish between them.}
    \label{tsne_show_appendix}
    \end{center}
    \vspace{-1em}
\end{figure*}

The distribution of fabricated and true members in various semantic feature spaces is visualized using t-SNE \citep{maaten2008visualizing}. The first row of \Cref{tsne_show_appendix} presents the semantic features from the penultimate layer, with perturbations constrained within the range $(\|\delta\|_{\infty} \leq 2.0/255)$ to $(\|\delta\|_{\infty} \leq 6.0/255)$ from left to right. The second row shows the semantic features at the antepenultimate layer, with the same perturbation range. In these visualizations, red dots represent the true members' semantic features, while blue dots represent the fabricated members. The t-SNE visualization reveals a significant overlap between the distributions of fabricated and true members in the feature space, suggesting that semantic features alone cannot effectively distinguish between the two. This observation underscores the limitations of traditional methods based solely on semantic features for the detection task,

\section{The Visualization of Fabricated Members}

\begin{figure*}[!t]
    \begin{center}
    {\includegraphics[width=0.195\textwidth]{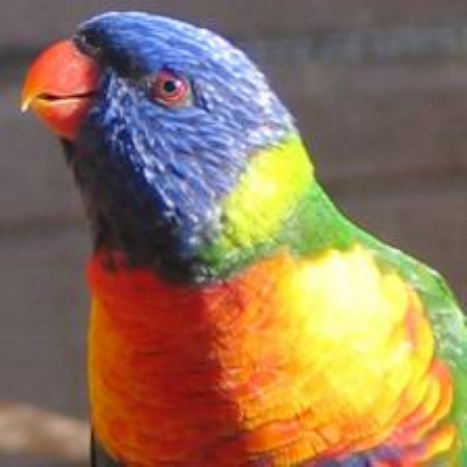}}
    {\includegraphics[width=0.195\textwidth]{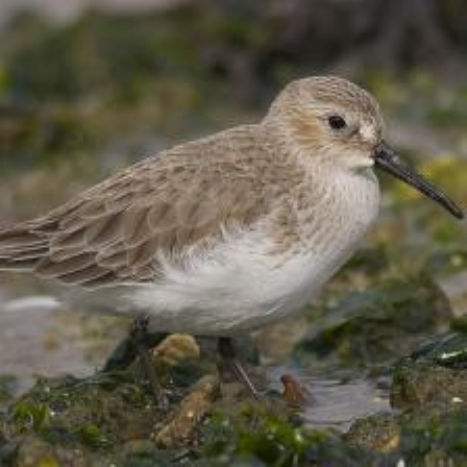}}    
    {\includegraphics[width=0.195\textwidth]{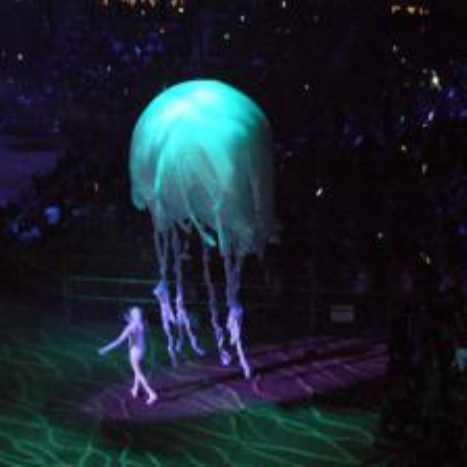}}
    {\includegraphics[width=0.195\textwidth]{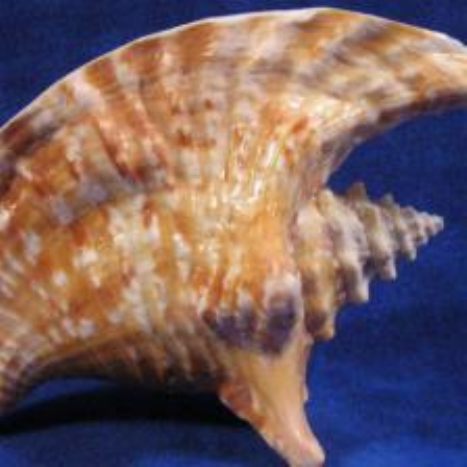}}
    {\includegraphics[width=0.195\textwidth]{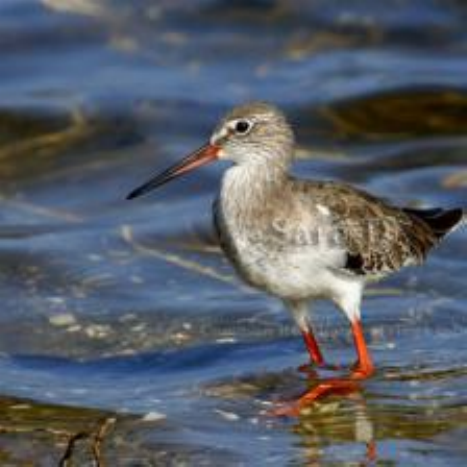}}
    {\includegraphics[width=0.195\textwidth]{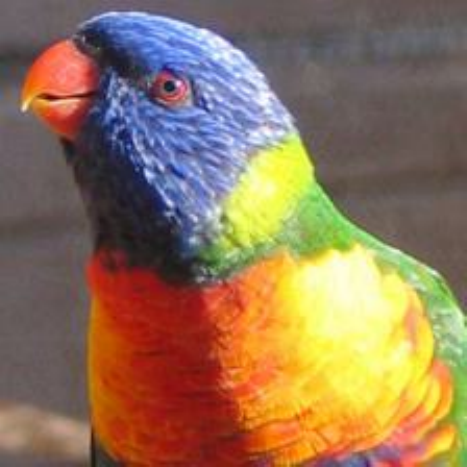}} 
    {\includegraphics[width=0.195\textwidth]{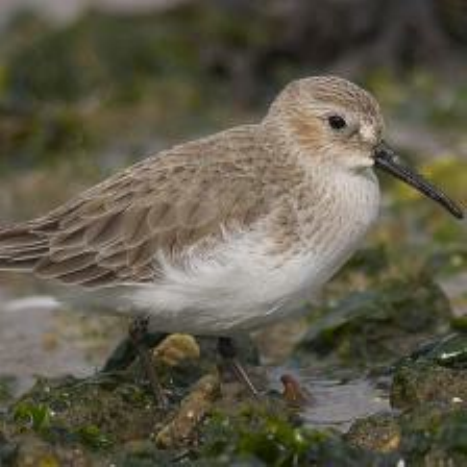}} 
    {\includegraphics[width=0.195\textwidth]{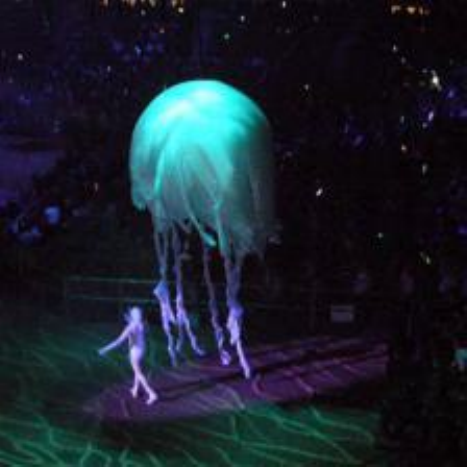}} 
    {\includegraphics[width=0.195\textwidth]{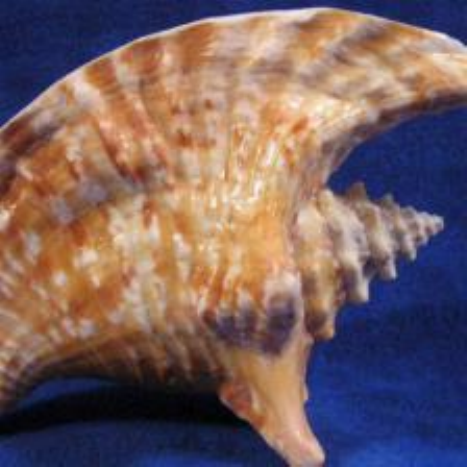}} 
    {\includegraphics[width=0.195\textwidth]{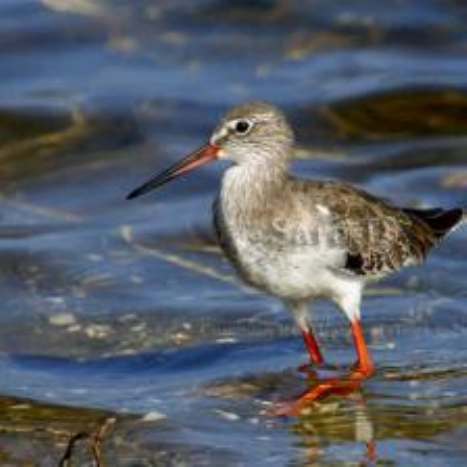}} 
    {\includegraphics[width=0.195\textwidth]{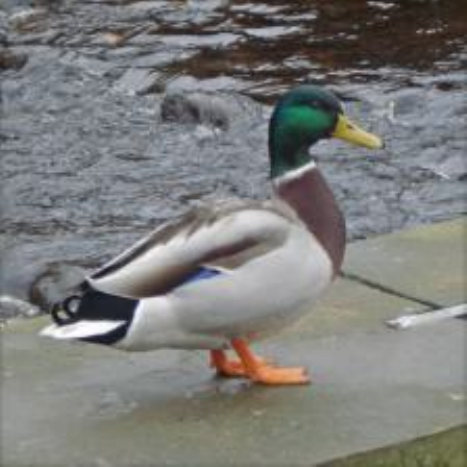}}
    {\includegraphics[width=0.195\textwidth]{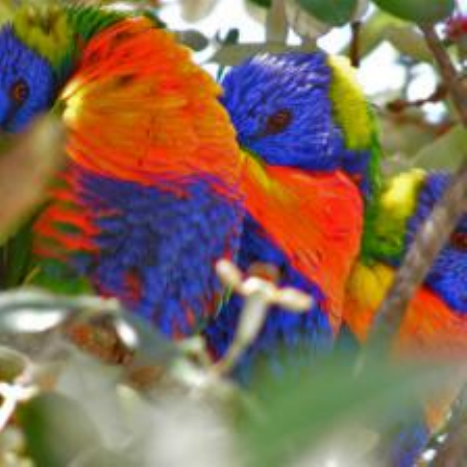}}    
    {\includegraphics[width=0.195\textwidth]{Imagenet100/native_pdf/native_8.pdf}}
    {\includegraphics[width=0.195\textwidth]{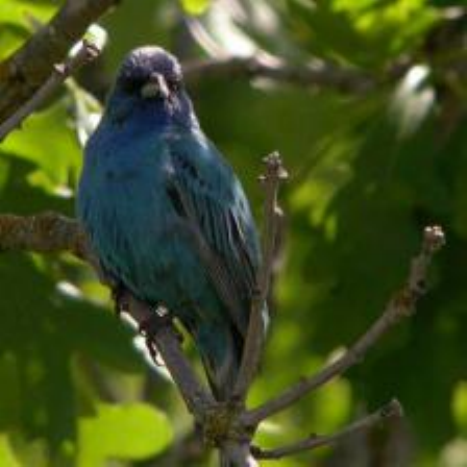}}
    {\includegraphics[width=0.195\textwidth]{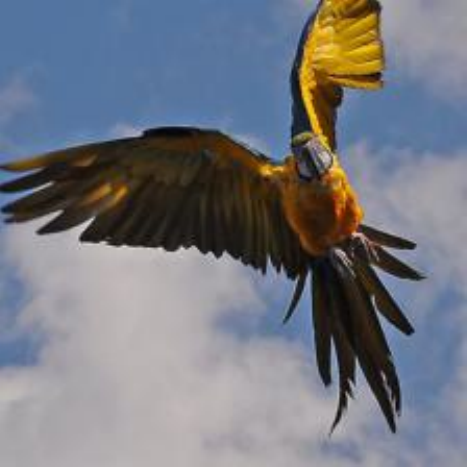}}
    {\includegraphics[width=0.195\textwidth]{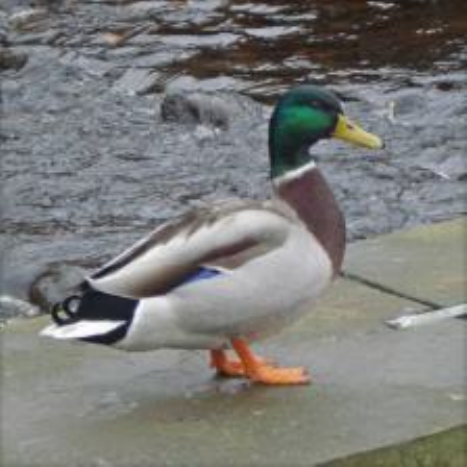}} 
    {\includegraphics[width=0.195\textwidth]{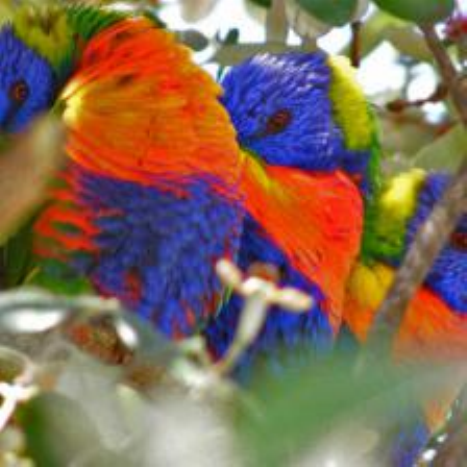}} 
    {\includegraphics[width=0.195\textwidth]{Imagenet100/adv_pdf/adv_8.pdf}} 
    {\includegraphics[width=0.195\textwidth]{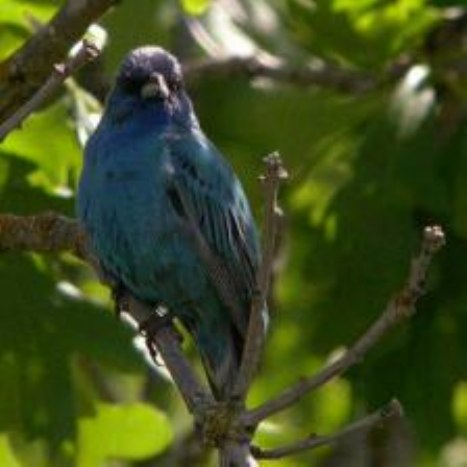}} 
    {\includegraphics[width=0.195\textwidth]{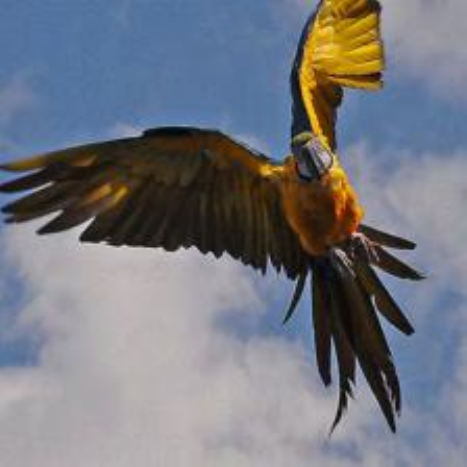}} 
    {\includegraphics[width=0.195\textwidth]{Imagenet100/native_pdf/native_11.pdf}}
    {\includegraphics[width=0.195\textwidth]{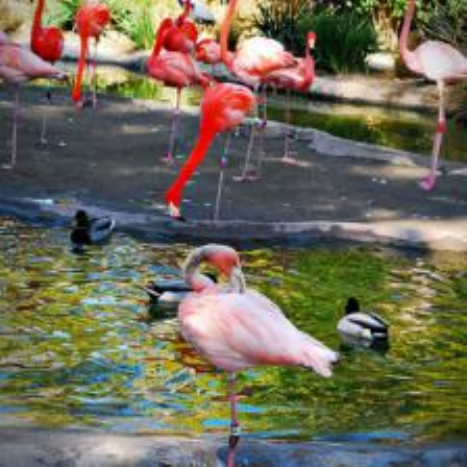}}    
    {\includegraphics[width=0.195\textwidth]{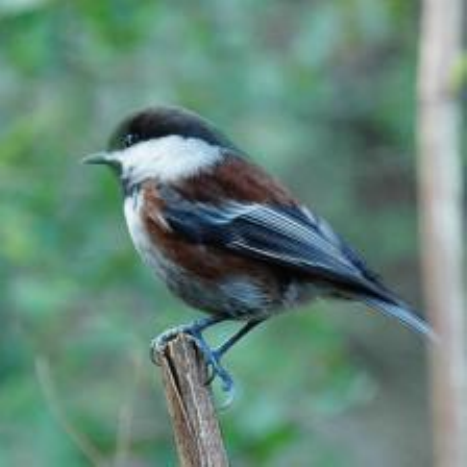}}
    {\includegraphics[width=0.195\textwidth]{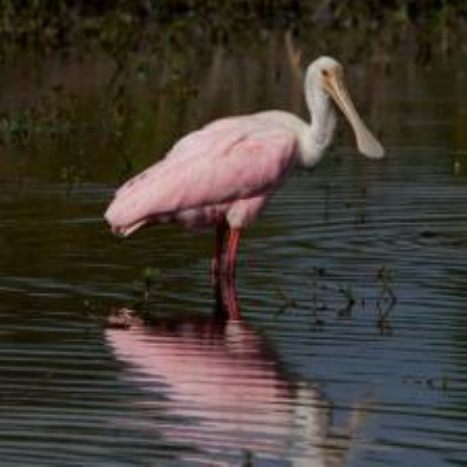}}
    {\includegraphics[width=0.195\textwidth]{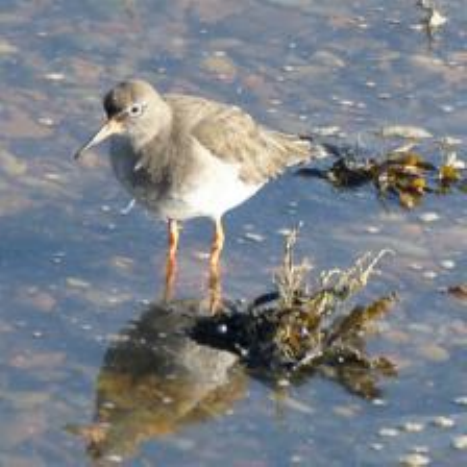}}
    {\includegraphics[width=0.195\textwidth]{Imagenet100/adv_pdf/adv_11.pdf}} 
    {\includegraphics[width=0.195\textwidth]{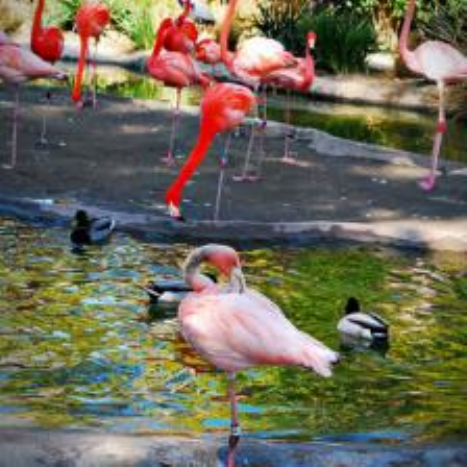}} 
    {\includegraphics[width=0.195\textwidth]{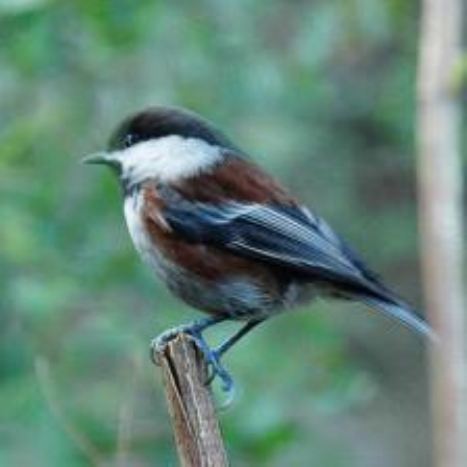}} 
    {\includegraphics[width=0.195\textwidth]{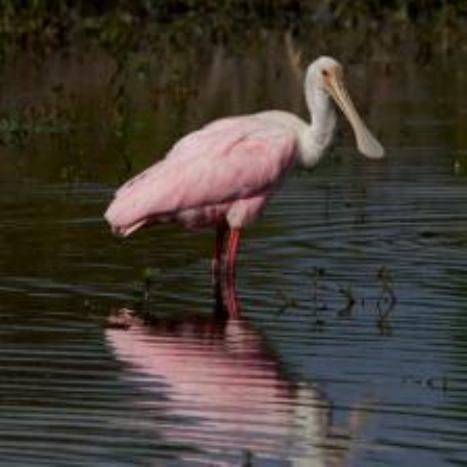}} 
    {\includegraphics[width=0.195\textwidth]{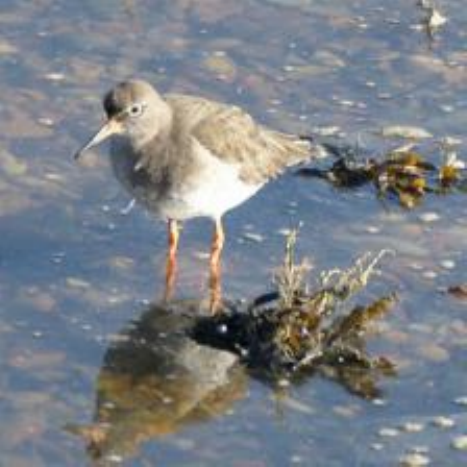}}     
    \caption{\footnotesize Imperceptible Adversarial Perturbations on \textbf{ImageNet-100}. For each pair, the top image is the original non-member, and the bottom image is the corresponding perturbed fabricated member, demonstrating that the perturbations are imperceptible to the human eyes. We used $\epsilon = 2/255$ for $\epsball[x]$ here.
    }
    \label{adv_show_1}
    \end{center}
    \vspace{-1em}
\end{figure*}

\begin{figure*}[!t]
    \begin{center}
    {\includegraphics[width=0.195\textwidth]{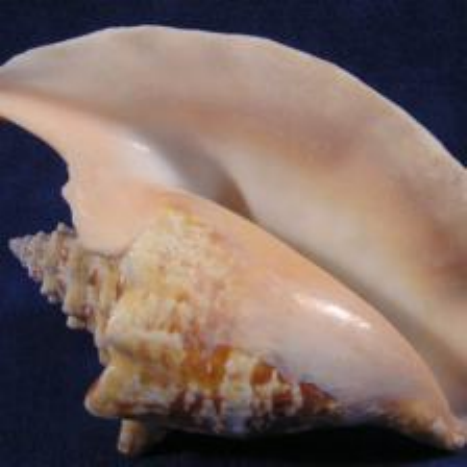}}
    {\includegraphics[width=0.195\textwidth]{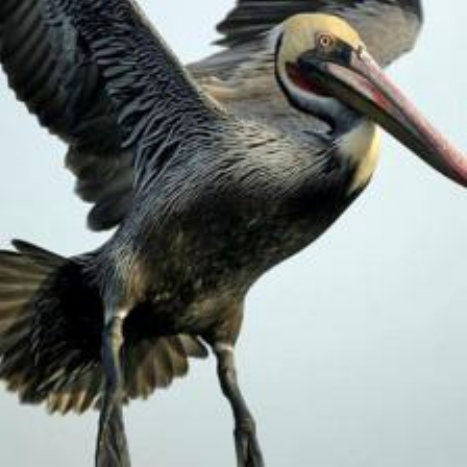}}    
    {\includegraphics[width=0.195\textwidth]{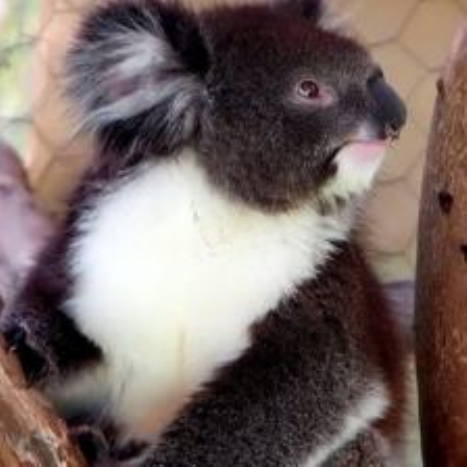}}
    {\includegraphics[width=0.195\textwidth]{Imagenet100/native_pdf/native_19.pdf}}
    {\includegraphics[width=0.195\textwidth]{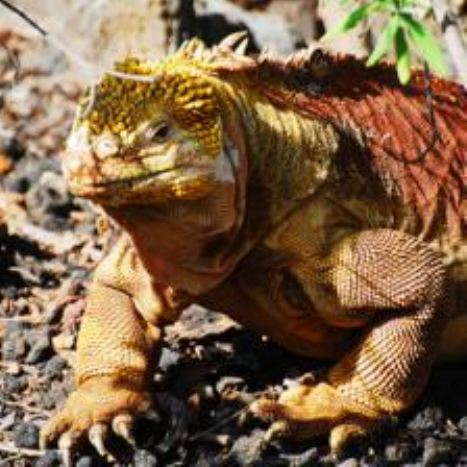}}
    {\includegraphics[width=0.195\textwidth]{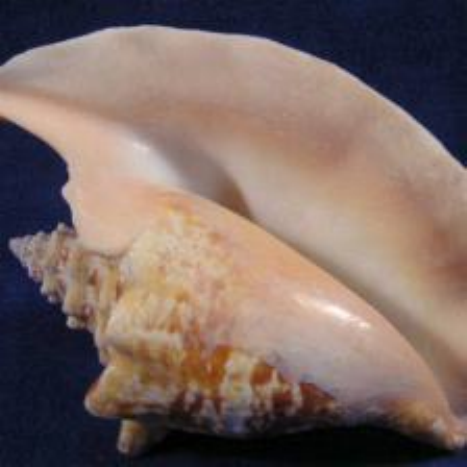}} 
    {\includegraphics[width=0.195\textwidth]{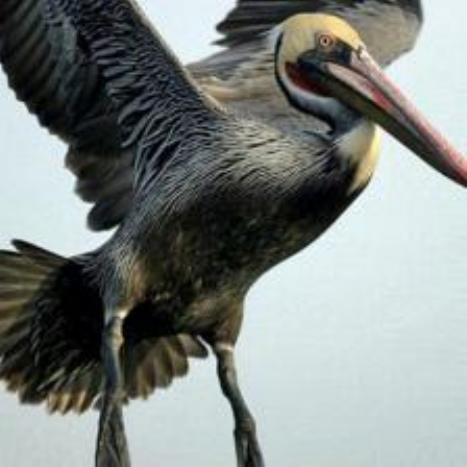}} 
    {\includegraphics[width=0.195\textwidth]{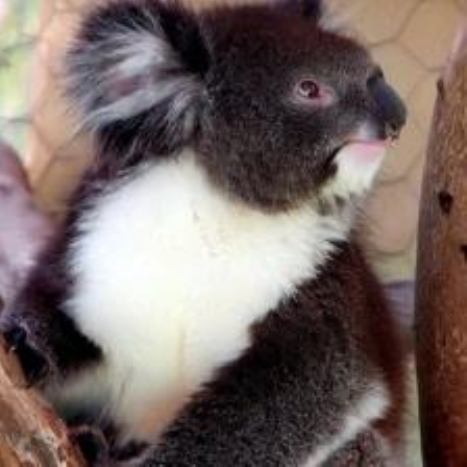}} 
    {\includegraphics[width=0.195\textwidth]{Imagenet100/adv_pdf/adv_19.pdf}} 
    {\includegraphics[width=0.195\textwidth]{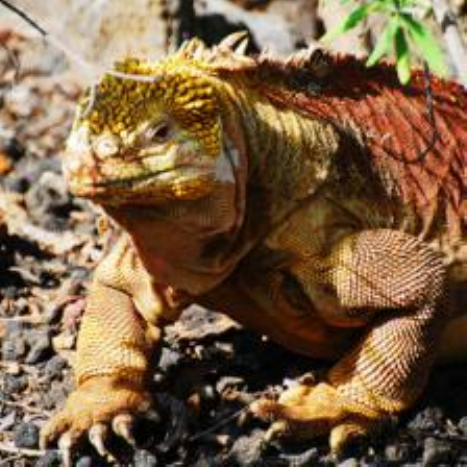}} 
    {\includegraphics[width=0.195\textwidth]{Imagenet100/native_pdf/native_21.pdf}}
    {\includegraphics[width=0.195\textwidth]{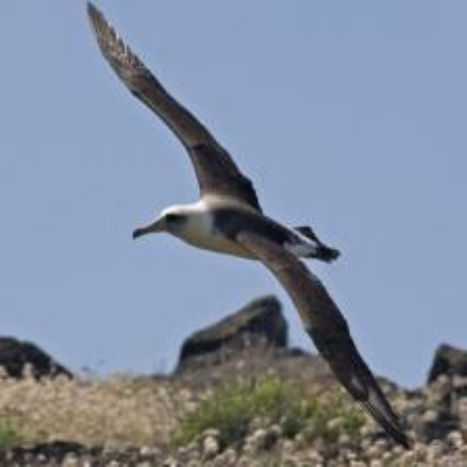}}    
    {\includegraphics[width=0.195\textwidth]{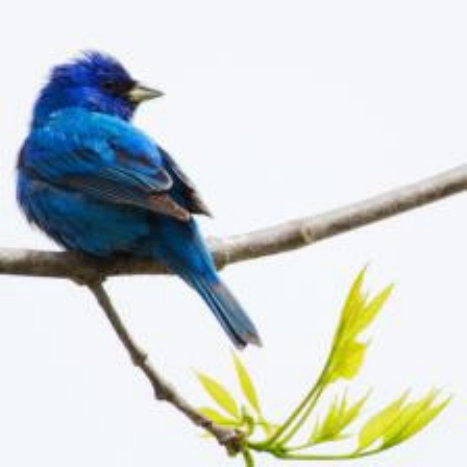}}
    {\includegraphics[width=0.195\textwidth]{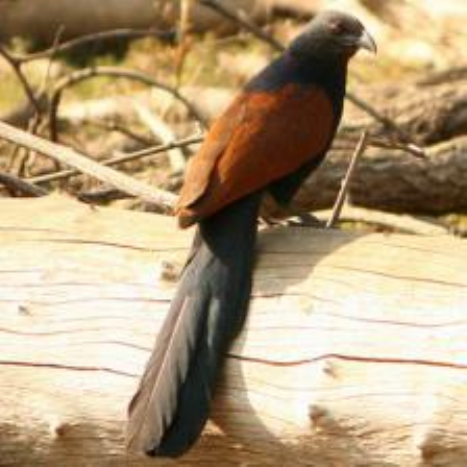}}
    {\includegraphics[width=0.195\textwidth]{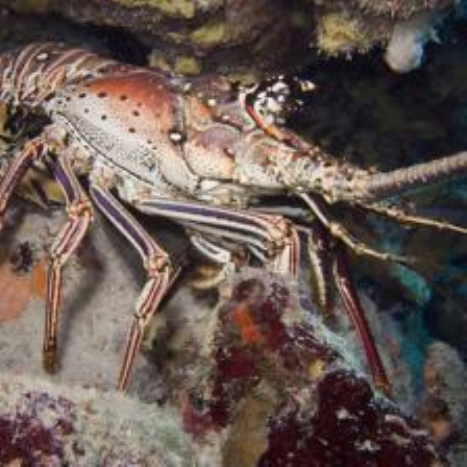}}
    {\includegraphics[width=0.195\textwidth]{Imagenet100/adv_pdf/adv_21.pdf}} 
    {\includegraphics[width=0.195\textwidth]{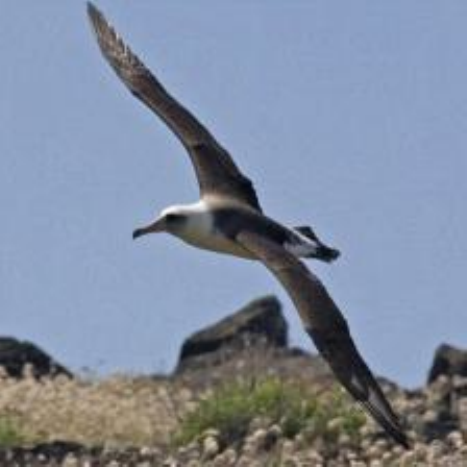}} 
    {\includegraphics[width=0.195\textwidth]{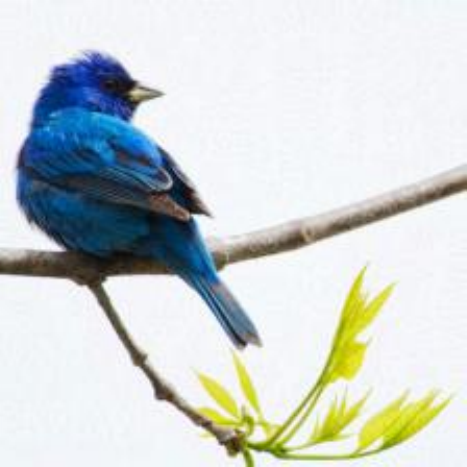}} 
    {\includegraphics[width=0.195\textwidth]{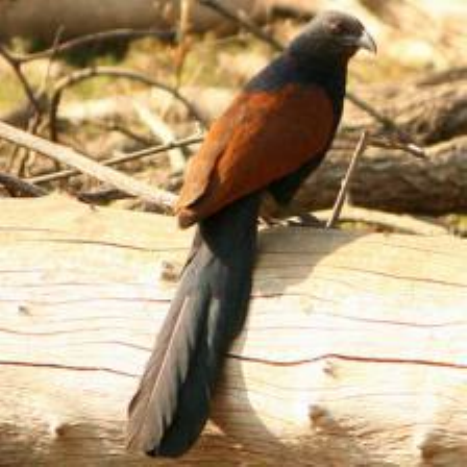}} 
    {\includegraphics[width=0.195\textwidth]{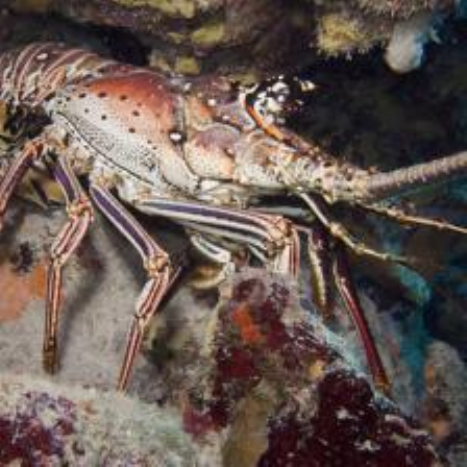}} 
    {\includegraphics[width=0.195\textwidth]{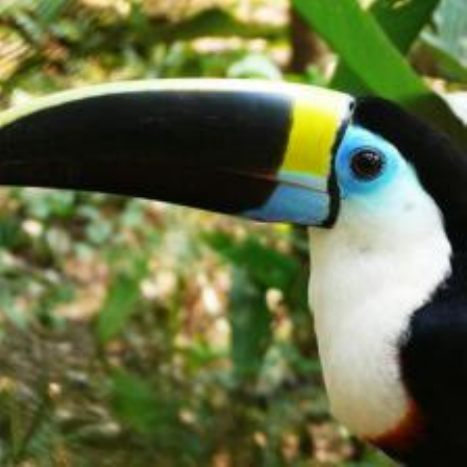}}
    {\includegraphics[width=0.195\textwidth]{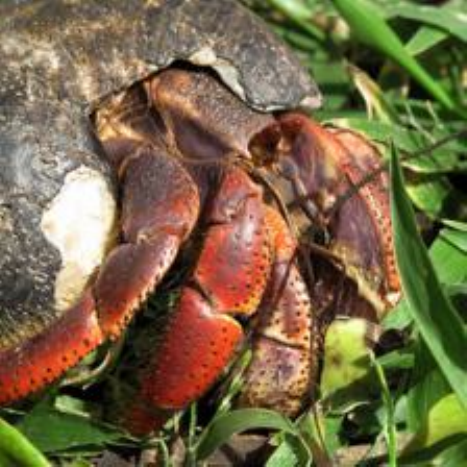}}    
    {\includegraphics[width=0.195\textwidth]{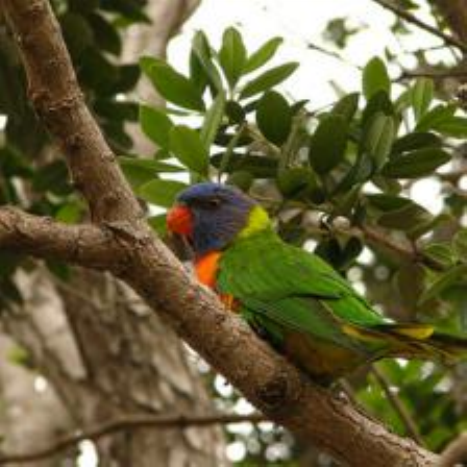}}
    {\includegraphics[width=0.195\textwidth]{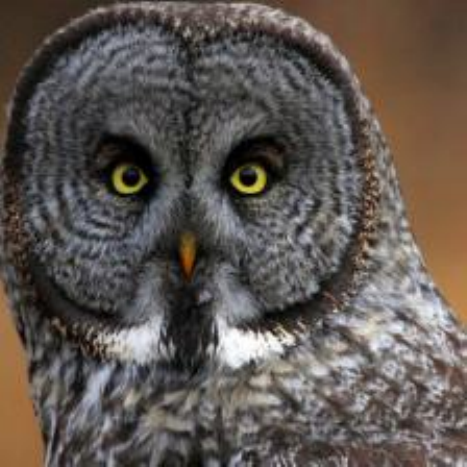}}
    {\includegraphics[width=0.195\textwidth]{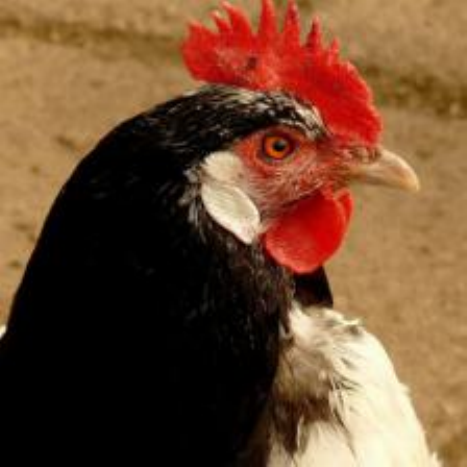}}
    {\includegraphics[width=0.195\textwidth]{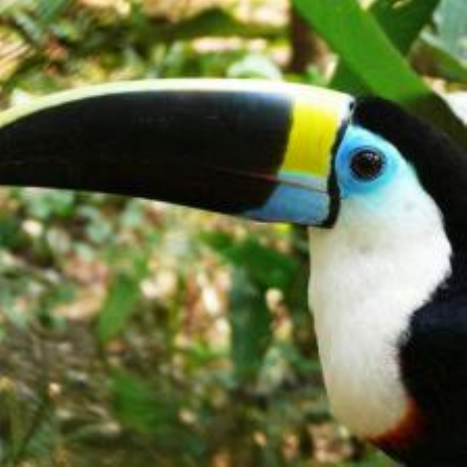}} 
    {\includegraphics[width=0.195\textwidth]{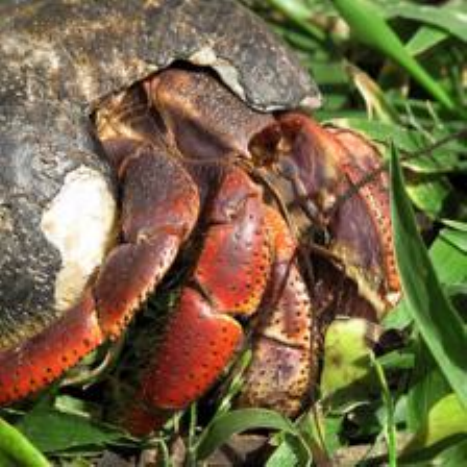}} 
    {\includegraphics[width=0.195\textwidth]{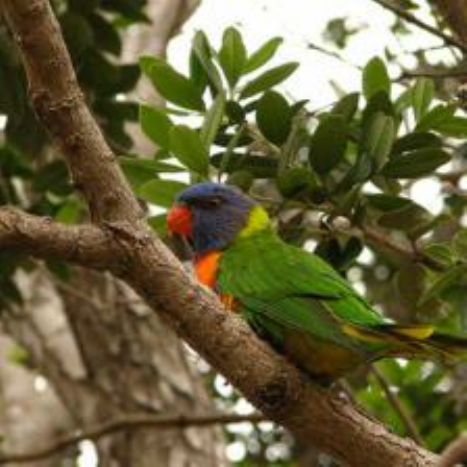}} 
    {\includegraphics[width=0.195\textwidth]{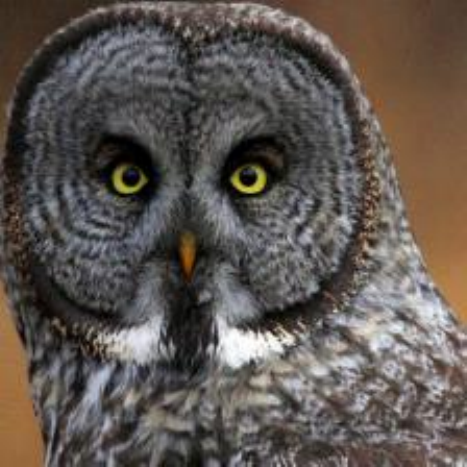}} 
    {\includegraphics[width=0.195\textwidth]{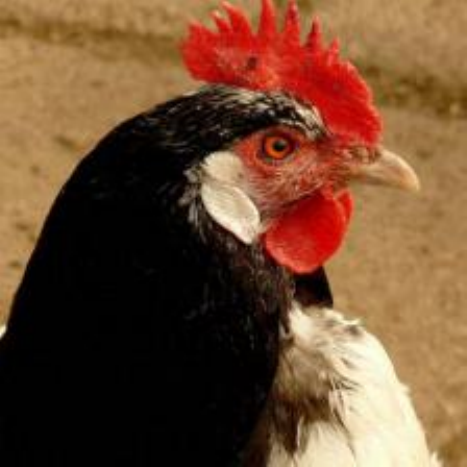}}     
    \caption{\footnotesize Imperceptible Adversarial Perturbations on \textbf{ImageNet-100}. For each pair, the top image is the original non-member, and the bottom image is the corresponding perturbed fabricated member, demonstrating that the perturbations are imperceptible to the human eyes. We used $\epsilon = 2/255$ for $\epsball[x]$ here.
    }
    \label{adv_show_2}
    \end{center}
    \vspace{-1em}
\end{figure*}

\begin{figure*}[!t]
    \begin{center}
    {\includegraphics[width=0.195\textwidth]{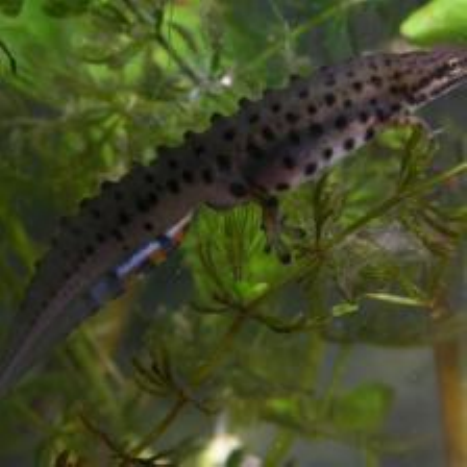}}
    {\includegraphics[width=0.195\textwidth]{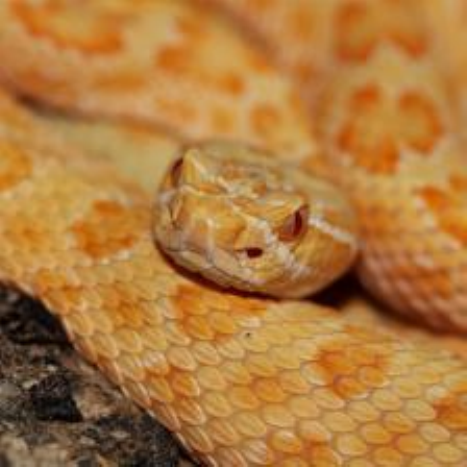}}    
    {\includegraphics[width=0.195\textwidth]{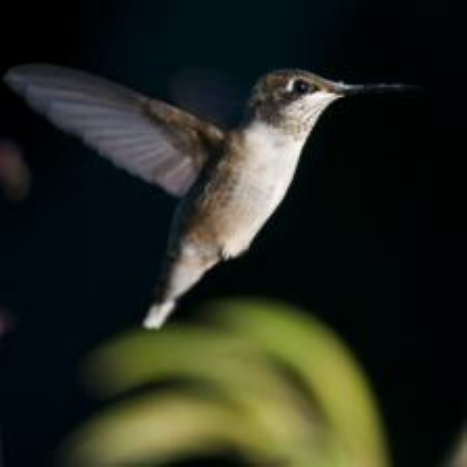}}
    {\includegraphics[width=0.195\textwidth]{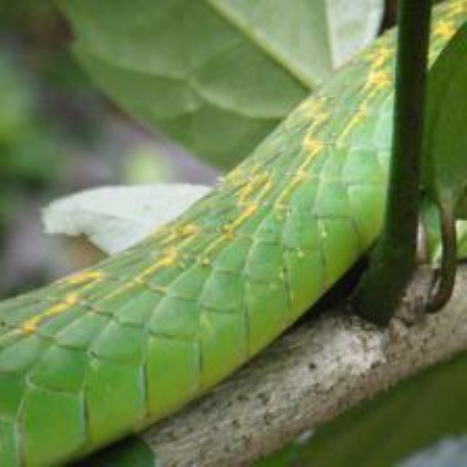}}
    {\includegraphics[width=0.195\textwidth]{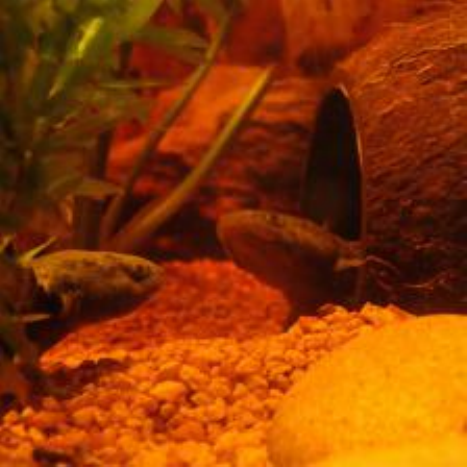}}
    {\includegraphics[width=0.195\textwidth]{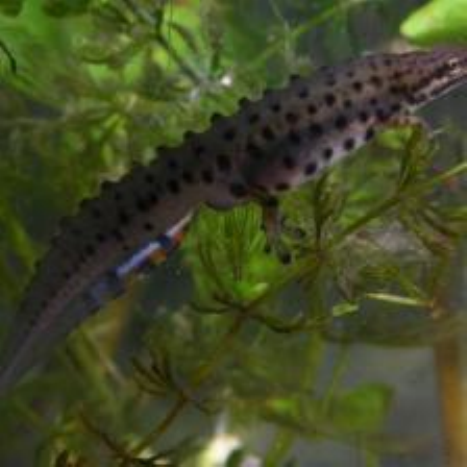}} 
    {\includegraphics[width=0.195\textwidth]{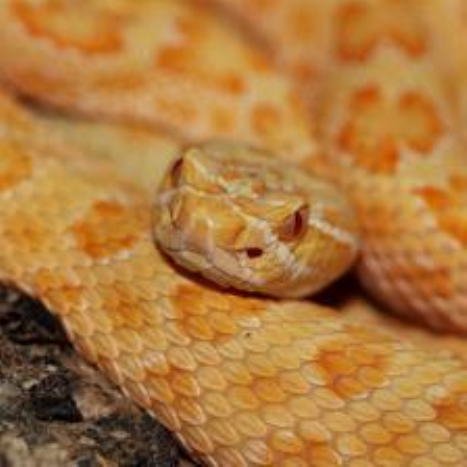}} 
    {\includegraphics[width=0.195\textwidth]{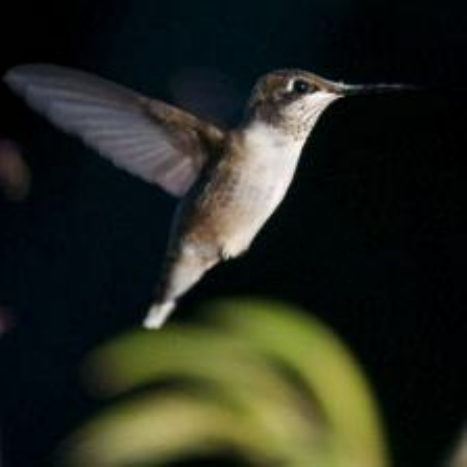}} 
    {\includegraphics[width=0.195\textwidth]{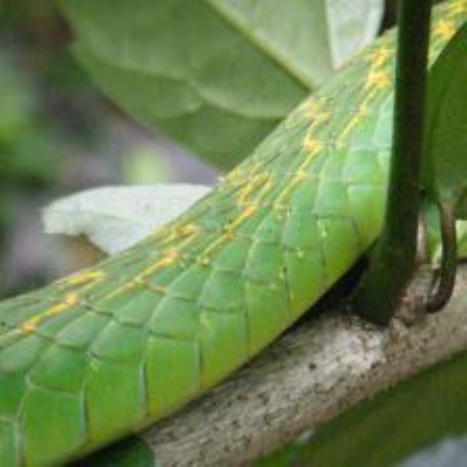}} 
    {\includegraphics[width=0.195\textwidth]{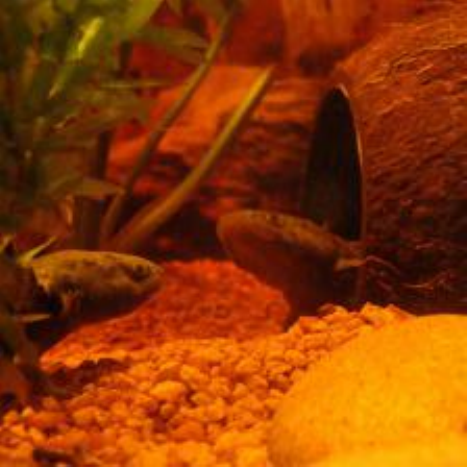}} 
    {\includegraphics[width=0.195\textwidth]{Imagenet100/native_pdf/native_36.pdf}}
    {\includegraphics[width=0.195\textwidth]{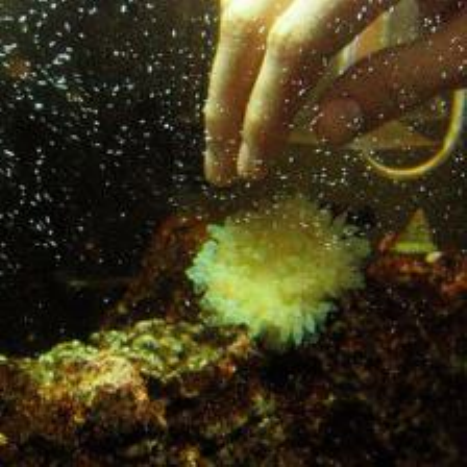}}    
    {\includegraphics[width=0.195\textwidth]{Imagenet100/native_pdf/native_38.pdf}}
    {\includegraphics[width=0.195\textwidth]{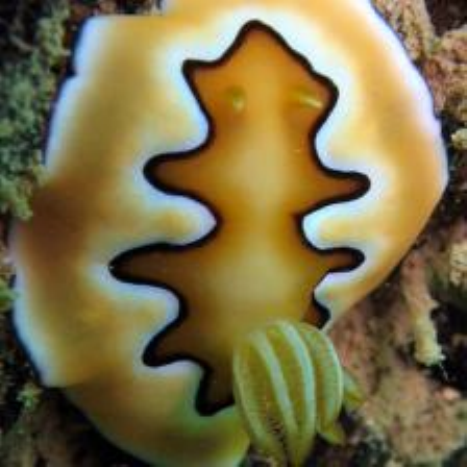}}
    {\includegraphics[width=0.195\textwidth]{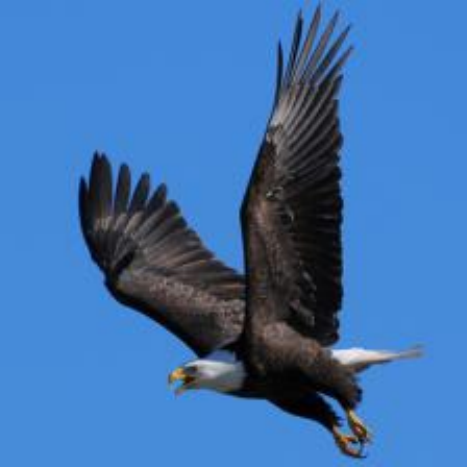}}
    {\includegraphics[width=0.195\textwidth]{Imagenet100/adv_pdf/adv_36.pdf}} 
    {\includegraphics[width=0.195\textwidth]{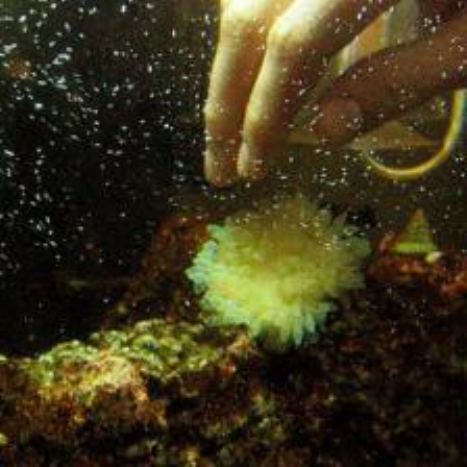}} 
    {\includegraphics[width=0.195\textwidth]{Imagenet100/adv_pdf/adv_38.pdf}} 
    {\includegraphics[width=0.195\textwidth]{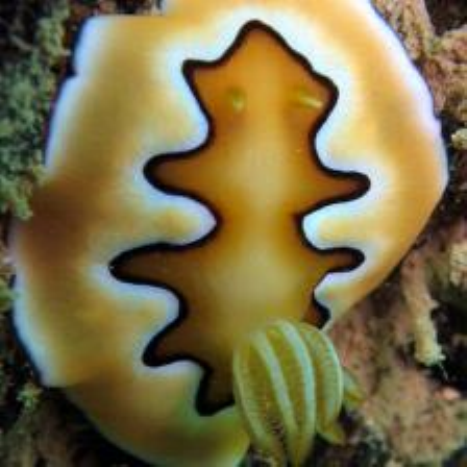}} 
    {\includegraphics[width=0.195\textwidth]{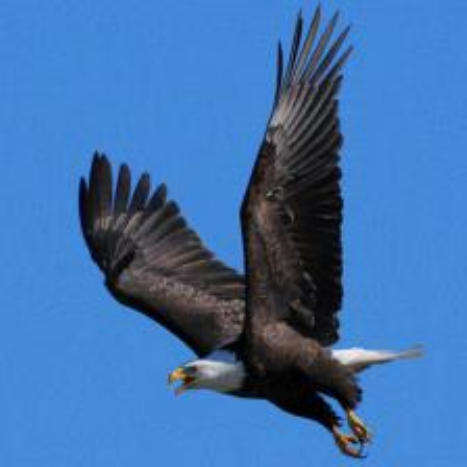}} 
    {\includegraphics[width=0.195\textwidth]{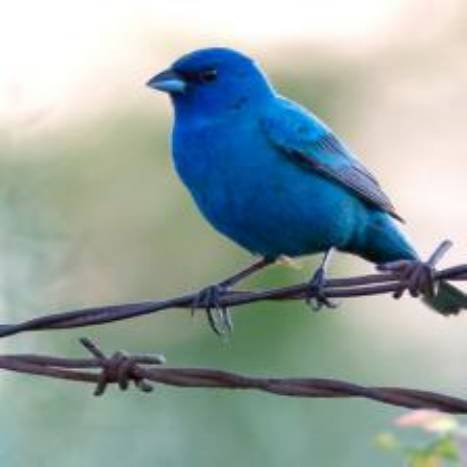}}
    {\includegraphics[width=0.195\textwidth]{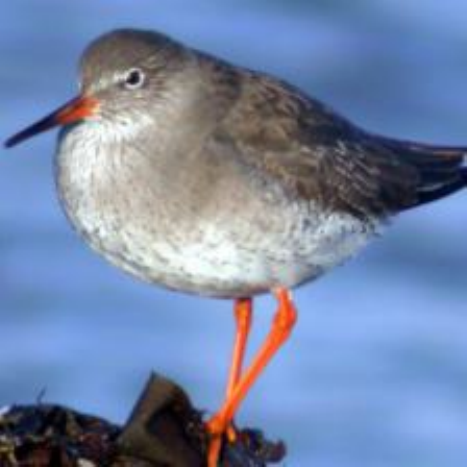}}    
    {\includegraphics[width=0.195\textwidth]{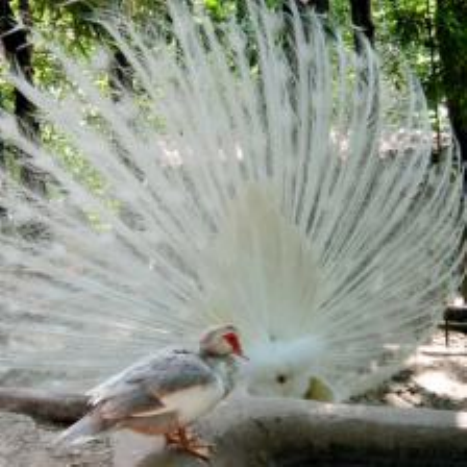}}
    {\includegraphics[width=0.195\textwidth]{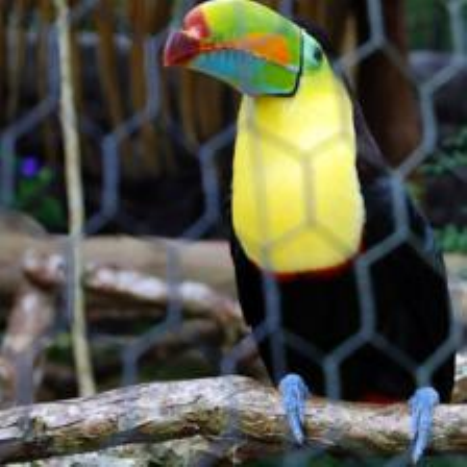}}
    {\includegraphics[width=0.195\textwidth]{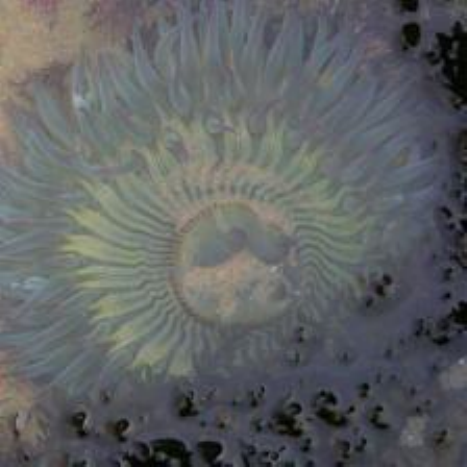}}
    {\includegraphics[width=0.195\textwidth]{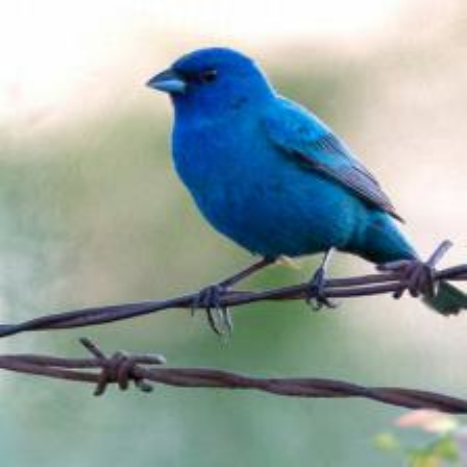}} 
    {\includegraphics[width=0.195\textwidth]{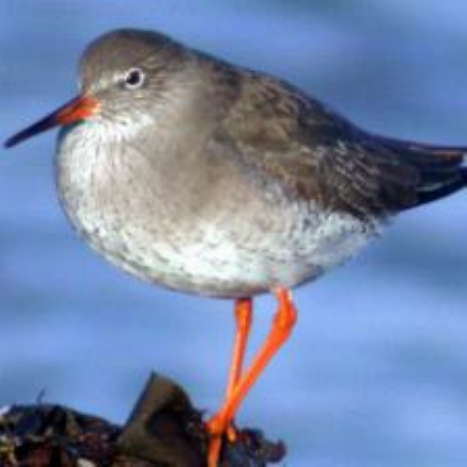}} 
    {\includegraphics[width=0.195\textwidth]{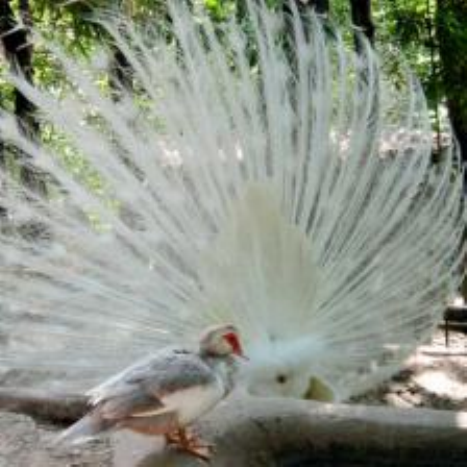}} 
    {\includegraphics[width=0.195\textwidth]{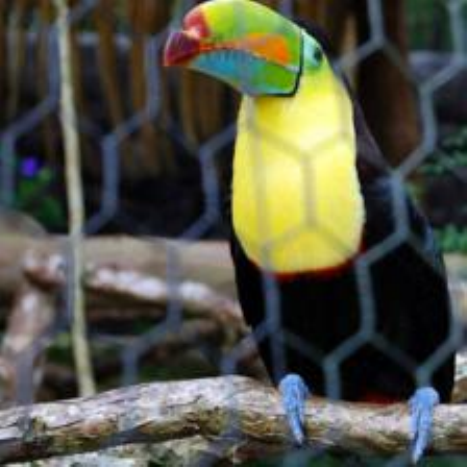}} 
    {\includegraphics[width=0.195\textwidth]{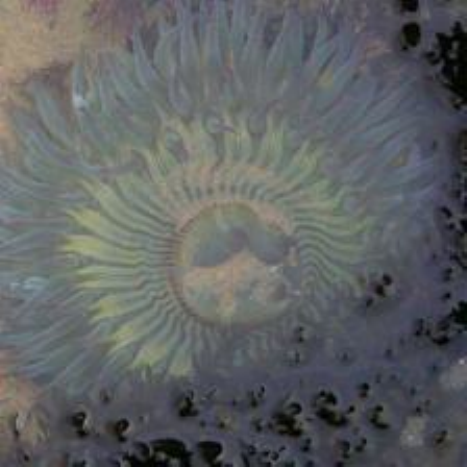}}     
    \caption{\footnotesize Imperceptible Adversarial Perturbations on \textbf{ImageNet-100}. For each pair, the top image is the original non-member, and the bottom image is the corresponding perturbed fabricated member, demonstrating that the perturbations are imperceptible to the human eyes. We used $\epsilon = 2/255$ for $\epsball[x]$ here.
    }
    \label{adv_show_3}
    \end{center}
    \vspace{-1em}
\end{figure*}

In \Cref{adv_show_1,adv_show_2,adv_show_3}, we demonstrate the visual quality of images after adding member fabrication perturbations. The perturbations are imperceptible adversarial changes applied to \textbf{ImageNet-100}. For each pair of images, the top image represents the original non-member, while the bottom image shows the corresponding perturbed fabricated member. We used $\epsilon = 2/255$ for $\epsball[x]$ in these examples. The perturbations are extremely subtle and nearly imperceptible to the human eye, illustrating that the Member Fabrication attack can be effective even with the addition of very small perturbations.

\section{Evaluation Metrics}
\label{app:evaluation_metrics}
\subsection{Evaluation Metrics of MIAs}

Several strategies have been proposed for evaluating the effectiveness of MIAs. In this section, we review two widely adopted evaluation metrics used to assess MIA performance in detail.

\vspace{0.2em}
\noindent\textbf{\emph{I. Receiver Operating Characteristic Curve (ROC Curve).}}  
To evaluate MIA performance comprehensively, it is essential to consider metrics that reflect an inferer's ability to make accurate predictions while minimizing false positives. The most commonly used approach is to examine the trade-off between the True Positive Rate (TPR) and False Positive Rate (FPR). An effective attack should maximize the TPR while minimizing the FPR. This trade-off is captured through the Receiver Operating Characteristic (ROC) curve, which plots TPR against FPR at various decision thresholds. Many prior studies report ROC curves and summarize their performance using the \textbf{Area Under the Curve (\textbf{AUC})}~\citep{Salem2019ML, melis2019exploiting, Leino2020Stolen, watson2021importance, ye2022enhanced, murakonda2020ml}. The \textbf{AUC} offers a single score that summarizes performance across all thresholds.

\vspace{0.2em}
\noindent\textbf{\emph{II. True Positive Rate at Low False Positive Rates (TPR @ FPR).}}  
While the ROC curve and \textbf{AUC} are useful, \citet{carlini2022membership} argue that these metrics may not fully capture the security risks associated with MIAs. In real-world scenarios, privacy breaches are often more consequential when they occur with a minimal number of false positives. Hence, they suggest that MIAs should be evaluated by focusing on TPR at low FPR values, as this regime better represents practical risk scenarios. This evaluation criterion provides a more robust measure of attack effectiveness, particularly in contexts where privacy violations must be detected with minimal false positives.

For a thorough assessment of performance, we adopt two primary evaluation metrics: \emph{Area Under the ROC Curve (\textbf{AUC})} and \emph{True Positive Rate at Low False Positive Rates (TPR @ FPR)}. To compute TPR values at specific FPR thresholds, we utilized numpy's interpolation methods, as the range of FPR values is often non-continuous in certain experimental settings. Our proposed adversarially robust MIAs also use these evaluation metrics.

\subsection{Evaluation Metrics of MFA}

We propose to assess the performance of \textbf{MFA} by evaluating the predictive performance of Inferer after they have been exposed to fabricated members. In practice, we found the conventional ROC curve, based on the standard False Positive Rate (FPR) and True Positive Rate (TPR), does not clearly distinguish between the different \textbf{MFA} methods. To address this limitation, we propose using the TNR-TPR curve, combined with a logarithmic scale, which enhances the clarity of comparisons. Additionally, we employ two primary evaluation metrics: \textbf{Error Area} and \textbf{Equal Error Rate} (\textbf{EER}). The \textbf{Error Area} (i.e., 1 - \textbf{AUC}) is defined as the complement of the Area Under the Curve (\textbf{AUC}) of the TNR-TPR curve, where a higher value indicates better \textbf{MFA} performance. The \textbf{Equal Error Rate} (\textbf{EER}) is the point at which the False Positive Rate (FPR) equals the False Negative Rate (FNR), providing a balanced measure of a method’s ability to correctly identify both true and fabricated members. In these evaluations, a better \textbf{MFA} should result in Inferer with lower TPR and TNR values, causing the curves to approach the lower-left corner of the TNR-TPR plot. Correspondingly, the \textbf{Error Area} should be larger, and the \textbf{Equal Error Rate} should also be higher.

\subsection{Evaluation Metrics of MFD}

We assess the performance of our Member Fabrication Detection using the Area Under the ROC Curve (\textbf{AUC}) as the primary evaluation metric. \textbf{AUC} provides a comprehensive measure of how well our detection method can distinguish fabricated members from true members across various thresholds. A higher \textbf{AUC} value indicates that our detection method is more effective at identifying fabricated members, with the optimal outcome being an \textbf{AUC} as close to 1 as possible.

\section{Experimental Details}
\label{app_expsetup}

\subsection{Datasets}

\vspace{0.2em}
\noindent\textbf{CIFAR-10}~\citep{krizhevsky2009learning} is a widely-used dataset consisting of 60,000 color images, each with a resolution of $32\times32\times3$. The images are evenly distributed across 10 distinct classes, which represent various objects such as airplanes, automobiles, birds, and more.

\vspace{0.2em}
\noindent\textbf{CIFAR-100}~\citep{krizhevsky2009learning} shares the same structure and resolution as \textbf{CIFAR-10}, with 60,000 color images of $32\times32\times3$. However, it poses a more complex classification challenge, as it is divided into 100 classes, each containing 600 images. 

\vspace{0.2em}
\noindent\textbf{SVHN}~\citep{svnh} is a real-world image dataset obtained from house numbers in Google Street View images. It consists of 99,289 color images in the training set and 26,032 color images in the test set, each with a resolution of $32\times32\times3$. The dataset is divided into 10 classes, representing the digits 0 through 9. \textbf{SVHN} is designed for digit recognition tasks in challenging, real-world scenarios with varying backgrounds, lighting conditions, and distortions.

\vspace{0.2em}
\noindent\textbf{CINIC-10} is an extension of the CIFAR dataset, designed to bridge the gap between \textbf{CIFAR-10} and more complex datasets like ImageNet. It contains 270,000 images with a resolution of $32\times32\times3$, with classes and data distribution similar to \textbf{CIFAR-10}. \textbf{CINIC-10} combines \textbf{CIFAR-10} images with a subset of ImageNet images, providing a larger and more diverse dataset.

\vspace{0.2em}
\noindent\textbf{ImageNet-100} is a subset of the larger ImageNet dataset, containing 100 classes from the original ImageNet hierarchy. The dataset is significantly more challenging due to its larger and more complex images, typically with a resolution of $224 \times 224$. The diversity of classes and image variations adds complexity to classification tasks.

\subsection{Implementation details}
\label{app_implementation}

\vspace{0.2em}
\noindent\textbf{Datasets, Model Structures and Membership Inference Attacks Used.} Our experiments span multiple datasets: \textbf{CIFAR-10} \citep{krizhevsky2009learning}, \textbf{CIFAR-100} \citep{krizhevsky2009learning}, \textbf{SVHN} \citep{krizhevsky2009learning}, \textbf{CINIC-10} \citep{darlow2018cinic}, and \textbf{ImageNet-100} \citep{deng2009imagenet}. We experiment with the first four datasets using the ResNet-18 architecture \citep{he2016deep}, and evaluate the larger \textbf{ImageNet-100} dataset with the Wide-ResNet-50-2 architecture \citep{zagoruyko2016wide}. We consider the following MIAs in this paper: the basic loss attack \citep{Yeom2018Privacy}, and strong existing baselines, including \textbf{Attack R} \citep{ye2022enhanced}, \textbf{LiRA} \citep{carlini2022membership}, and \textbf{RMIA} \citep{zarifzadeh2024low}.

\vspace{0.2em}
\noindent\textbf{Baselines Used for Comparison with \textbf{MFA}.} We invert the perturbation direction in traditional adversarial attacks as baselines for comparison with our \textbf{MFA}: \emph{(i)} \emph{Inverted FGSM} (I-FGSM)~\citep{goodfellow2014explaining}, \emph{(ii)} \emph{Inverted BIM} (I-BIM)~\citep{kurakin2016adversarial}, \emph{(iii)} \emph{Inverted PGD} (I-PGD)~\citep{Madry18PGD}, \emph{(iv)} \emph{Inverted Carlini and Wagner Attack} (I-CW)~\citep{carlini2017towards}, and \emph{(v)} \emph{Inverted APGD} (I-APGD)~\citep{croce2020reliable}. We exclude the full versions of AutoAttack and MM Attack~\citep{gao2022fast}, as they rely on adaptive decision-making methods specifically designed for traditional adversarial attacks and cannot generalize to our scenario.

\vspace{0.2em}
\noindent\textbf{Model Training Details.} Consistent random seeds and training settings are maintained across all experiments. For datasets \textbf{CIFAR-10} \citep{krizhevsky2009learning}, \textbf{CIFAR-100} \citep{krizhevsky2009learning}, \textbf{SVHN} \citep{krizhevsky2009learning} and \textbf{CINIC-10} \citep{darlow2018cinic}, we train the target model or shadow model using 20,000 samples to ensure consistency in data size. For \textbf{ImageNet-100} \citep{deng2009imagenet}, we train the target model using all samples in the original datasets. For testing MIAs, we select a testing set containing 2,000 member samples and 2,000 non-member samples. The models are trained using the Stochastic Gradient Descent (SGD) optimizer with momentum set to 0.9, a weight decay of $10^{-4}$, and a batch size of 128. The learning rate is initialized to $\tau = 0.1$ and follows a cosine annealing schedule, gradually decaying to zero over 100 epochs.

\vspace{0.2em}
\noindent\textbf{Shadow Models.} For \textbf{Attack R} \citep{ye2022enhanced}, we train 100 reference models (OUT-Models). For \textbf{LiRA} \citep{carlini2022membership} and \textbf{RMIA} \citep{zarifzadeh2024low}, we train 100 IN-Models and 100 OUT-Models for modeling $\tilde{\mathbb{Q}}_{\text{in/out}}$, and ensure that the same shadow models are used across different methods. Due to the high training cost of shadow models for \textbf{ImageNet-100} with the corresponding Wide-ResNet-50-2 architecture, we only conduct basic loss attacks on it. Note that the training samples for shadow models or for modeling $\tilde{\mathbb{Q}}_{\text{m/nm}}$ are disjoint from the testing data. 

\vspace{0.2em}
\noindent\textbf{Data Augmentation.} In training models on these datasets, common data augmentation techniques were applied to improve the generalization and robustness of the models. For the first four datasets (\textbf{CIFAR-10}, \textbf{CIFAR-100}, \textbf{SVHN}, and \textbf{CINIC-10}), typical augmentation strategies such as random horizontal flipping and random cropping (with padding) were used. For \textbf{ImageNet-100}, we employed more advanced augmentation methods, such as random resized cropping and random horizontal flipping, tailored to handle larger image resolutions. These augmentation methods are standard practices for enhancing model performance across various image classification tasks. 

\vspace{0.2em}
\noindent\textbf{Details of Fabricated Member Generation.} For generating the fabricated members, we set the $L_{\infty}$-norm bounded perturbation $\epsilon = [1/255,8/255]$ for datasets \textbf{CIFAR-10} \citep{krizhevsky2009learning}, \textbf{CIFAR-100} \citep{krizhevsky2009learning}, \textbf{SVHN} \citep{krizhevsky2009learning} and \textbf{CINIC-10} \citep{darlow2018cinic}, and set the $L_{\infty}$-norm bounded perturbation $\epsilon = [0.5/255,2/255]$ for \textbf{ImageNet-100} \citep{deng2009imagenet}; the maximum number of steps is $K = 100$; initial step size $\alpha = \epsilon / 4$, momentum factor $\beta = 0.75$, decay factor $\gamma = 0.9$.

\subsection{Required Resources}
We implement all methods on Python $3.7.3$ (Pytorch 1.13.1+cu117) with four NVIDIA RTX A5000 GPUs and an x86-64 CPU with 32 physical cores.

\section{Supplementary Experimental Results}
\label{app_addexp}
In this section, we present comprehensive supplementary experimental results detailing the performance of our method alongside various baselines. The results thoroughly demonstrate the effectiveness of our approaches. 

\begin{figure*}[!t]
    \begin{center}
    \begin{minipage}{0.49\textwidth}
        \centering
        \includegraphics[width=\textwidth]{crop_Figure/Fabric_figure/crop_loss_TPR_TNR_plot.pdf}
        \subcaption{\normalsize Loss Attack}
    \end{minipage}
    \hfill
    \begin{minipage}{0.49\textwidth}
        \centering
        \includegraphics[width=\textwidth]{crop_Figure/Fabric_figure/crop_attack_r_TPR_TNR_plot.pdf}
        \subcaption{\normalsize Attack R}
    \end{minipage}
    
    \begin{minipage}{0.49\textwidth}
        \centering
        \includegraphics[width=\textwidth]{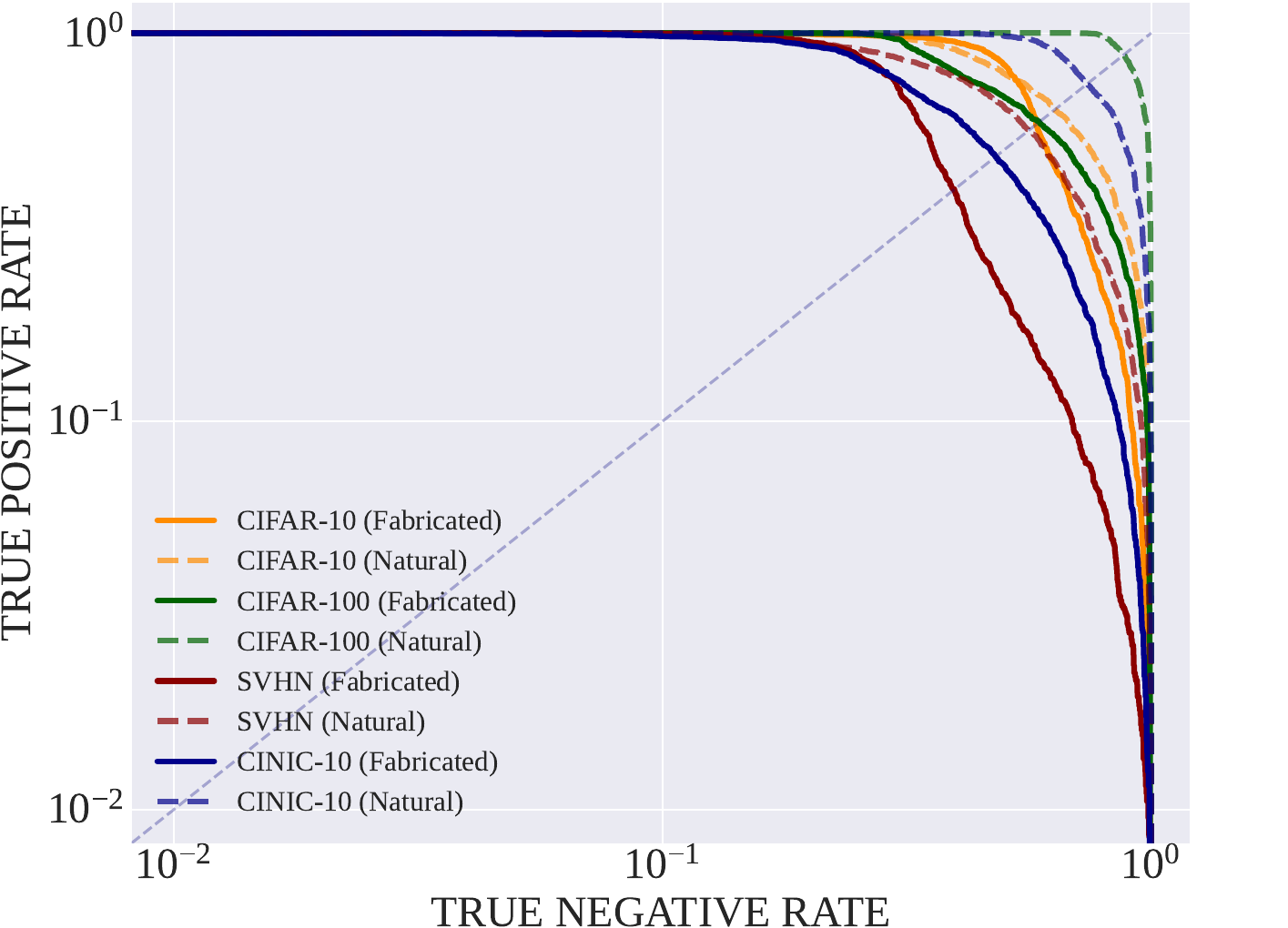}
        \subcaption{\normalsize LiRA}
    \end{minipage}
    \hfill
    \begin{minipage}{0.49\textwidth}
        \centering
        \captionsetup{font=Large}  
        \includegraphics[width=\textwidth]{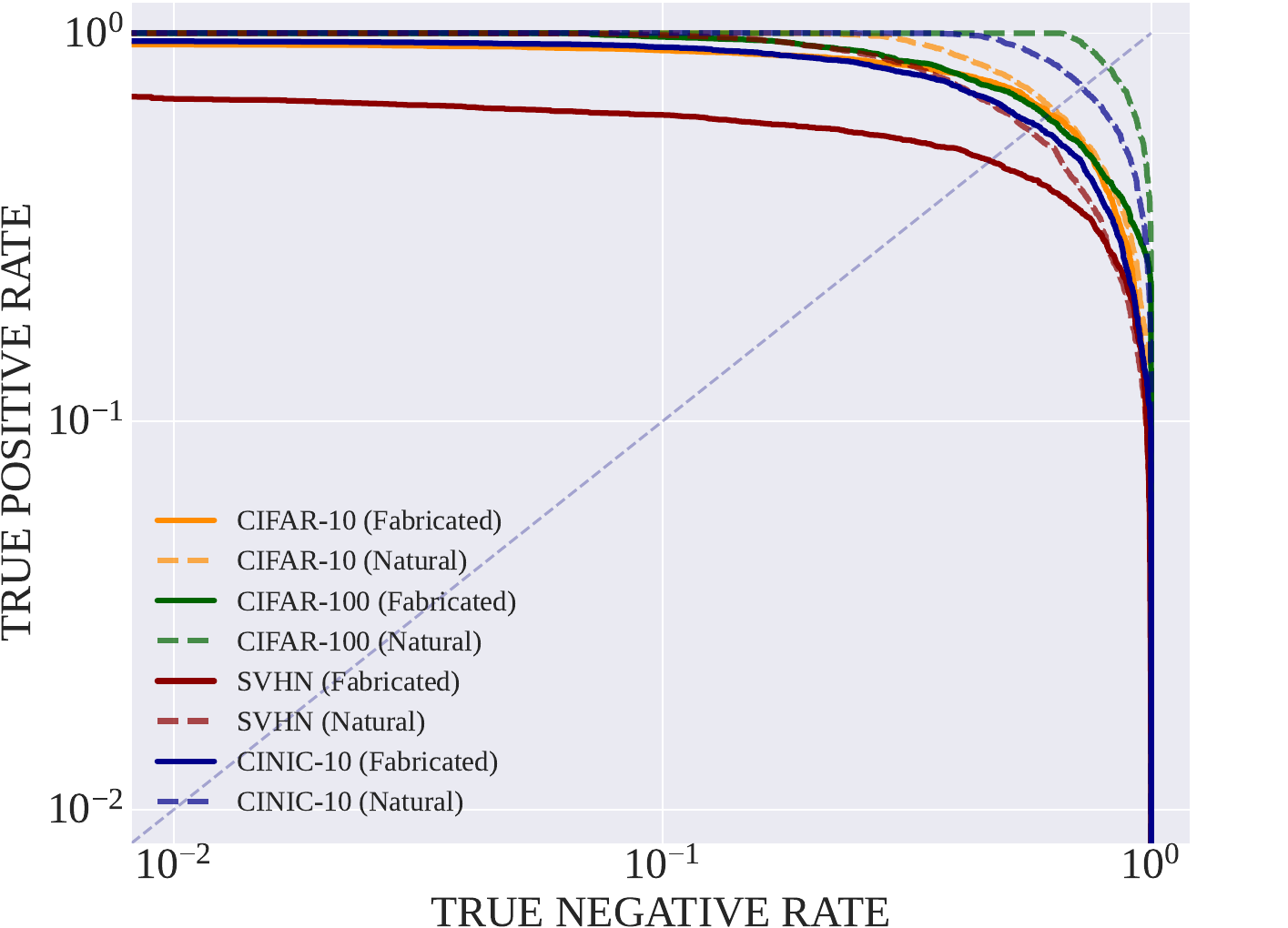}
        \subcaption{\normalsize RMIA}
    \end{minipage}
    \caption{\footnotesize Comparison of the \textbf{Error Area} Between Our Member Fabrication Attack and Baselines Across Diverse MIAs $(\|\delta\|_{\infty} \leq 4.0/255)$.}
    \label{Fabric_figure_4_attack}
    \end{center}
    \vspace{-1em}
\end{figure*}

\begin{figure*}[!t]
    \begin{center}
    \begin{minipage}{0.49\textwidth}
        \centering
        \includegraphics[width=\textwidth]{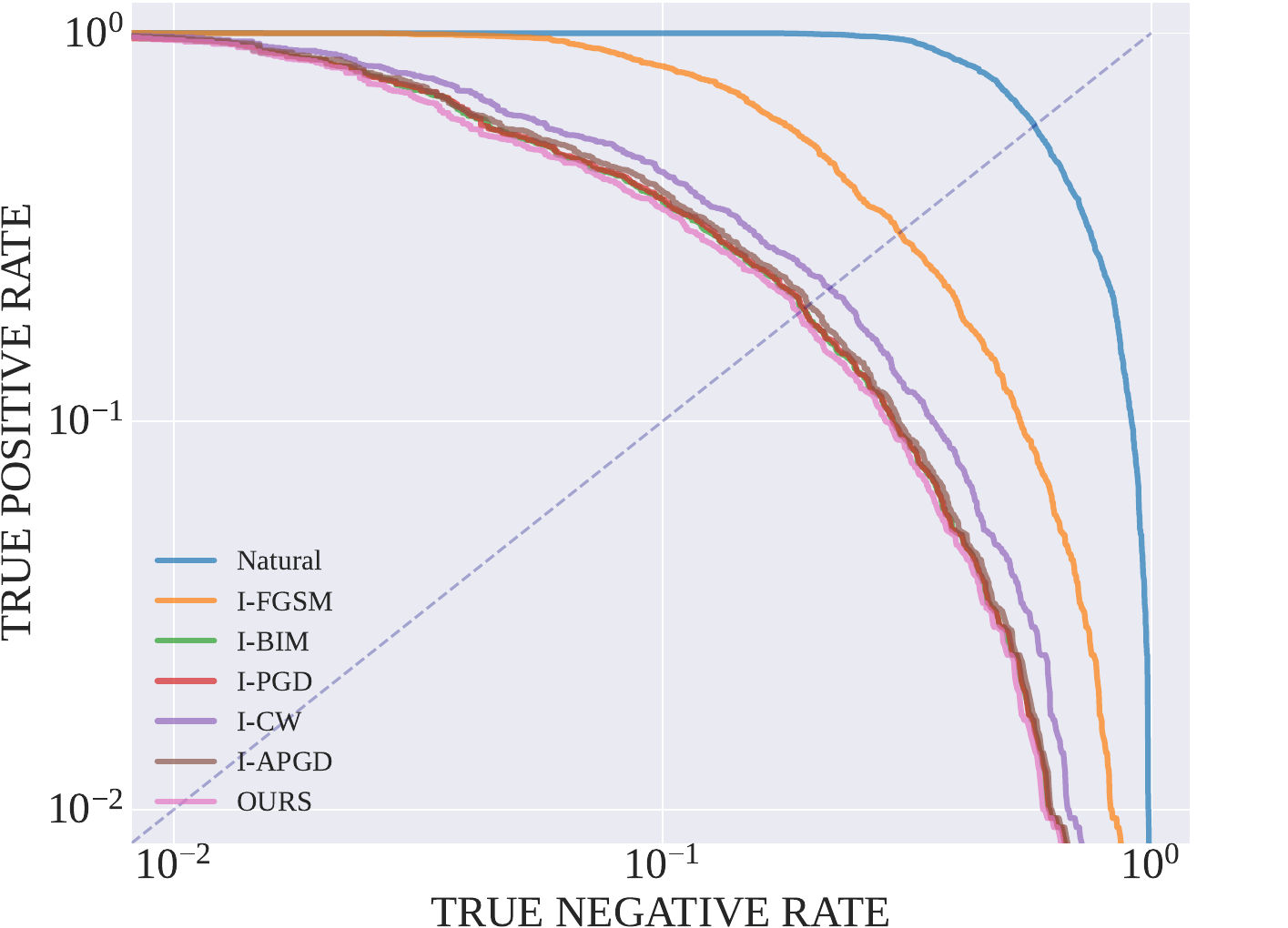}
        \subcaption{$\|\delta\|_{\infty} \leq 3.0/255$}
    \end{minipage}
    \hfill
    \begin{minipage}{0.49\textwidth}
        \centering
        \includegraphics[width=\textwidth]{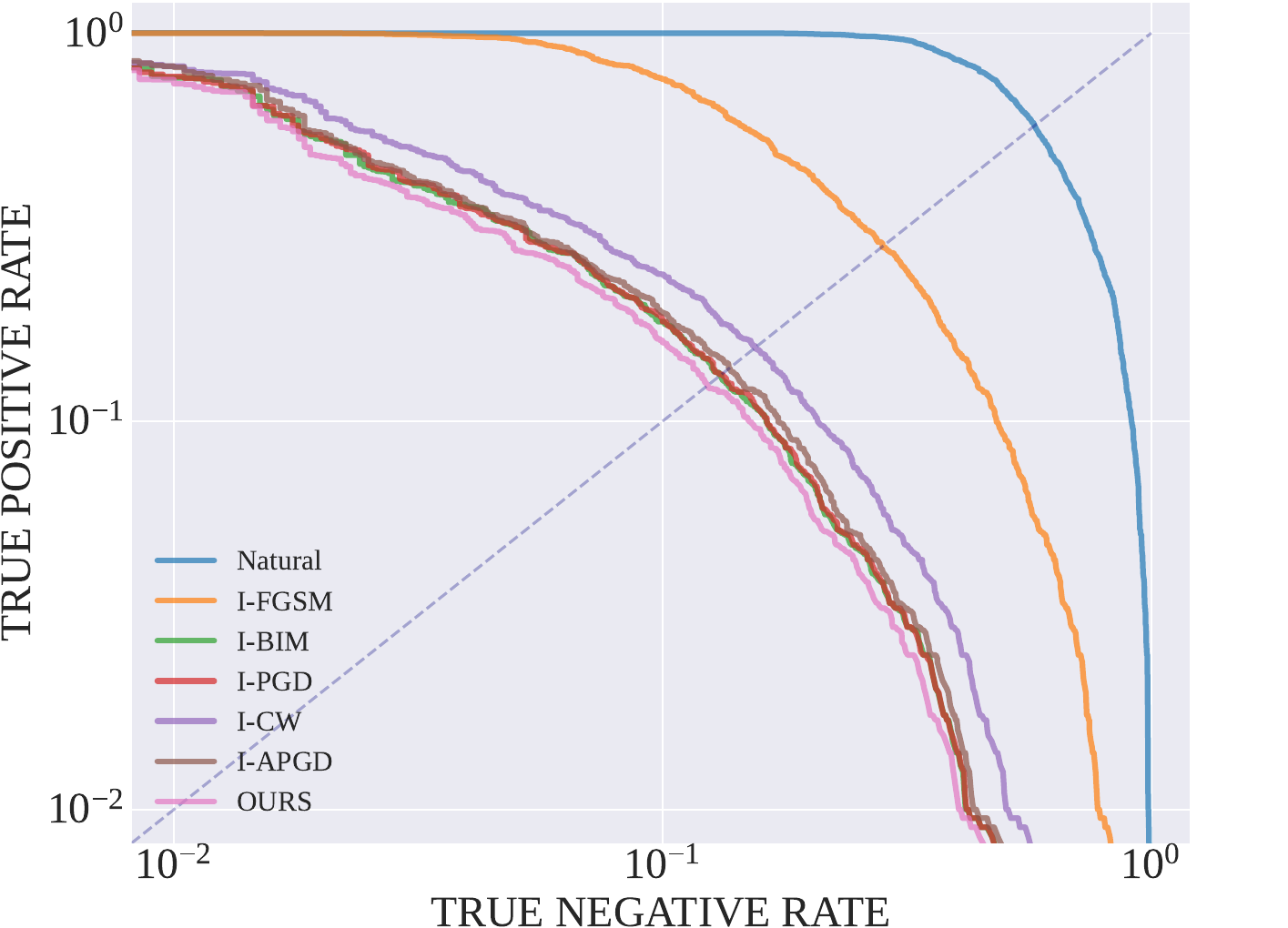}
        \subcaption{$\|\delta\|_{\infty} \leq 4.0/255$}
    \end{minipage}

    \begin{minipage}{0.49\textwidth}
        \centering
        \includegraphics[width=\textwidth]{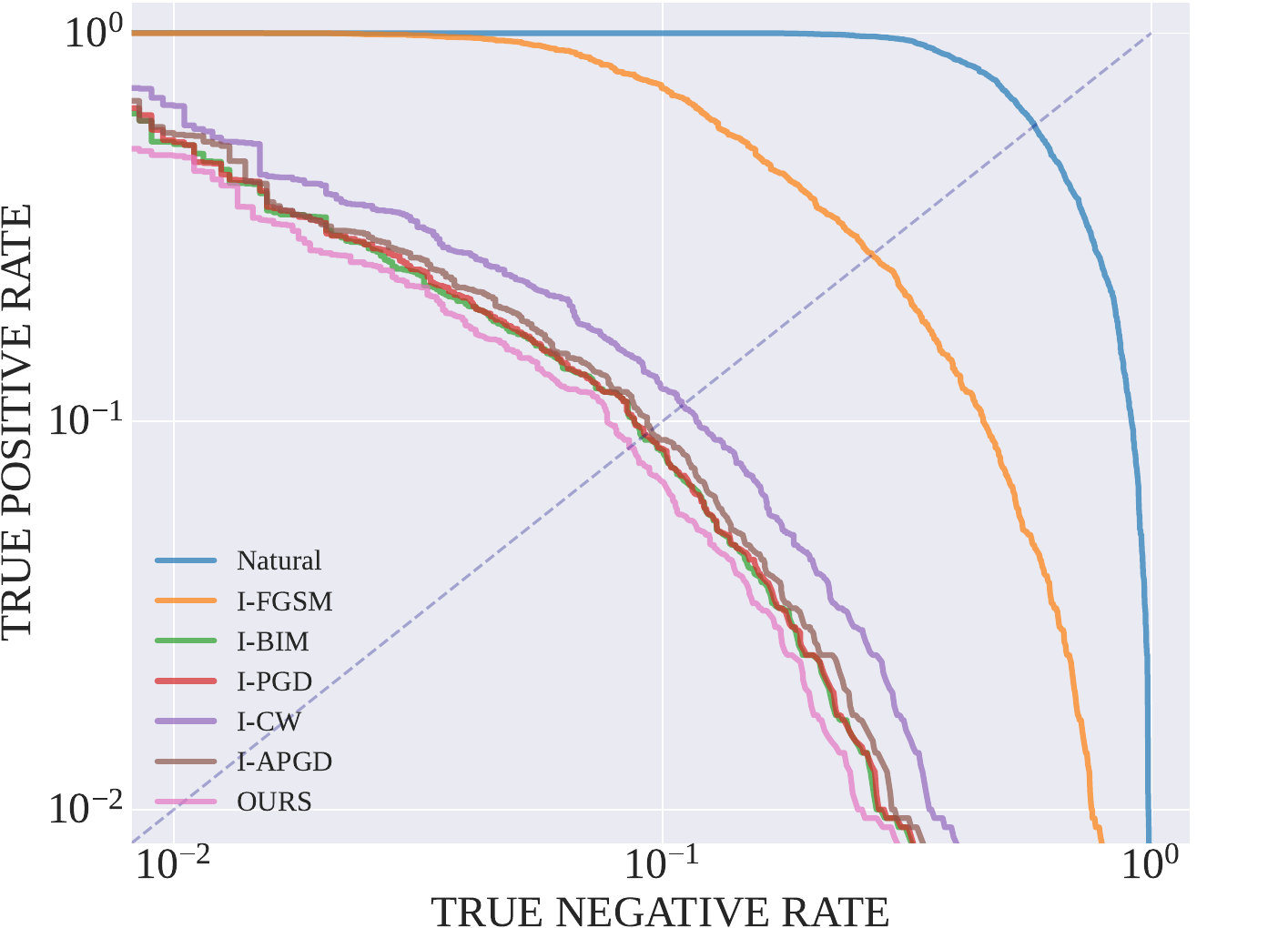}
        \subcaption{$\|\delta\|_{\infty} \leq 5.0/255$}
    \end{minipage}
    \hfill
    \begin{minipage}{0.49\textwidth}
        \centering
        \includegraphics[width=\textwidth]{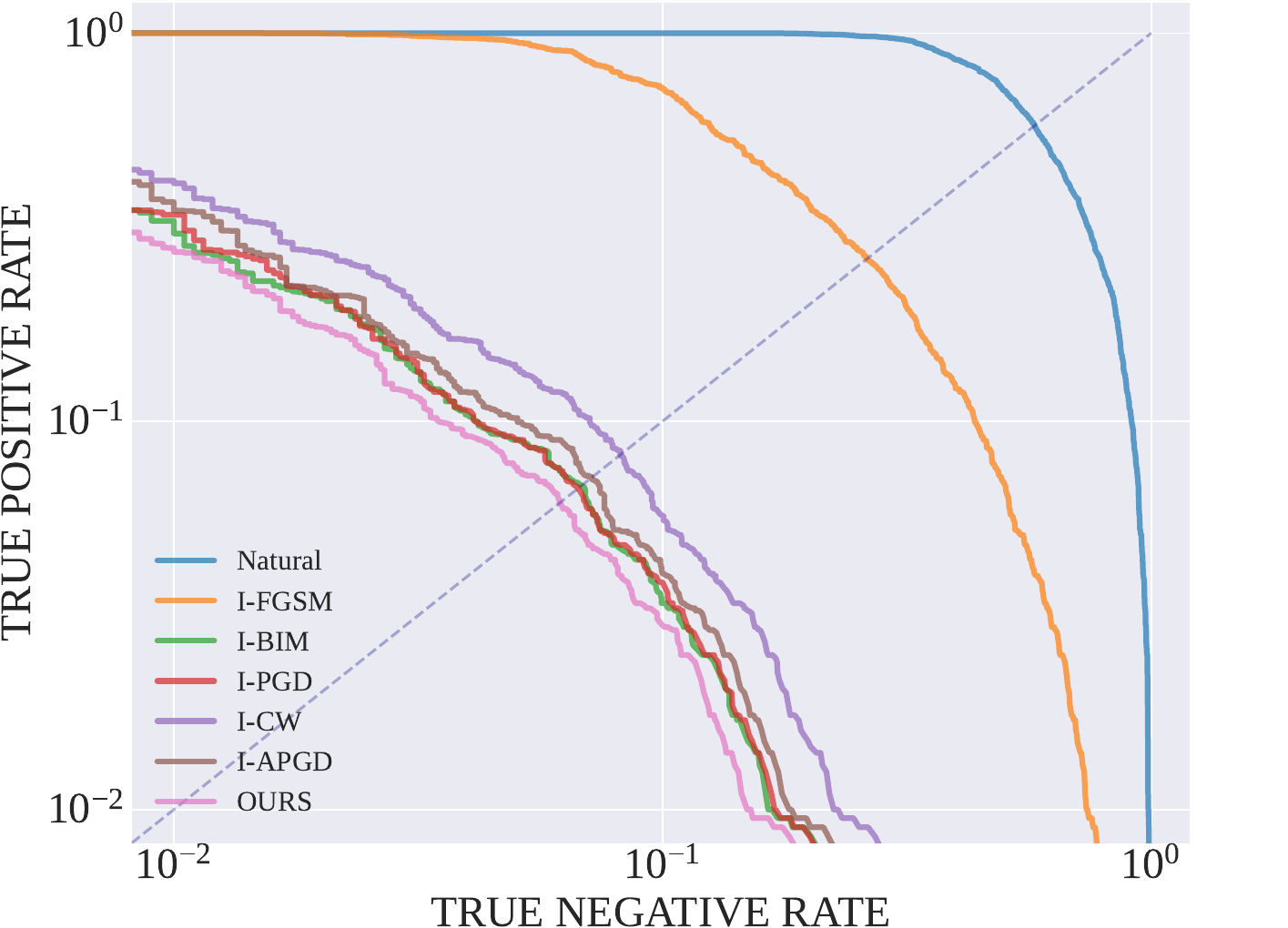}
        \subcaption{$\|\delta\|_{\infty} \leq 6.0/255$}
    \end{minipage}

    \begin{minipage}{0.49\textwidth}
        \centering
        \includegraphics[width=\textwidth]{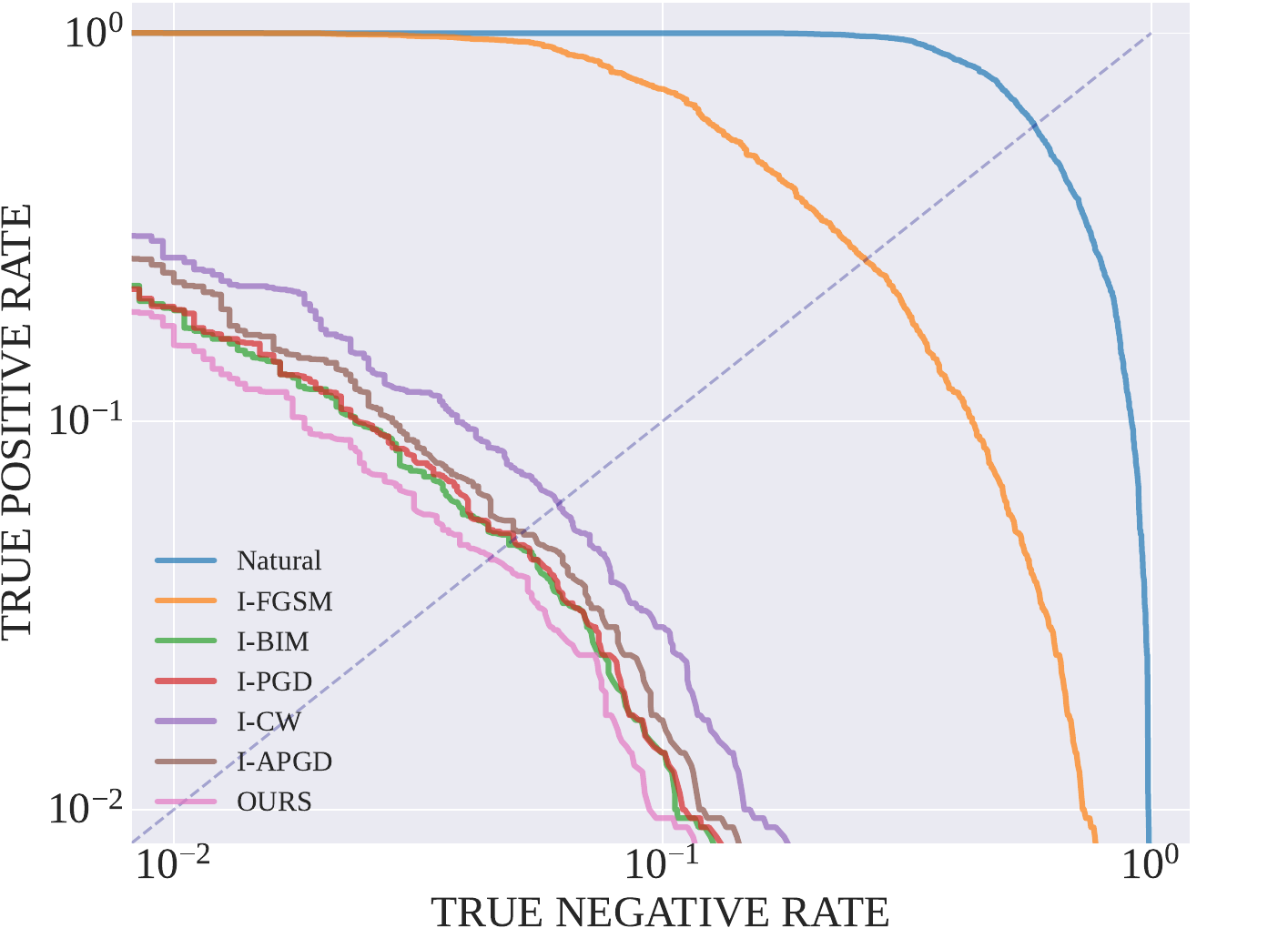}
        \subcaption{$\|\delta\|_{\infty} \leq 7.0/255$}
    \end{minipage}
    \hfill
    \begin{minipage}{0.49\textwidth}
        \centering
        \includegraphics[width=\textwidth]{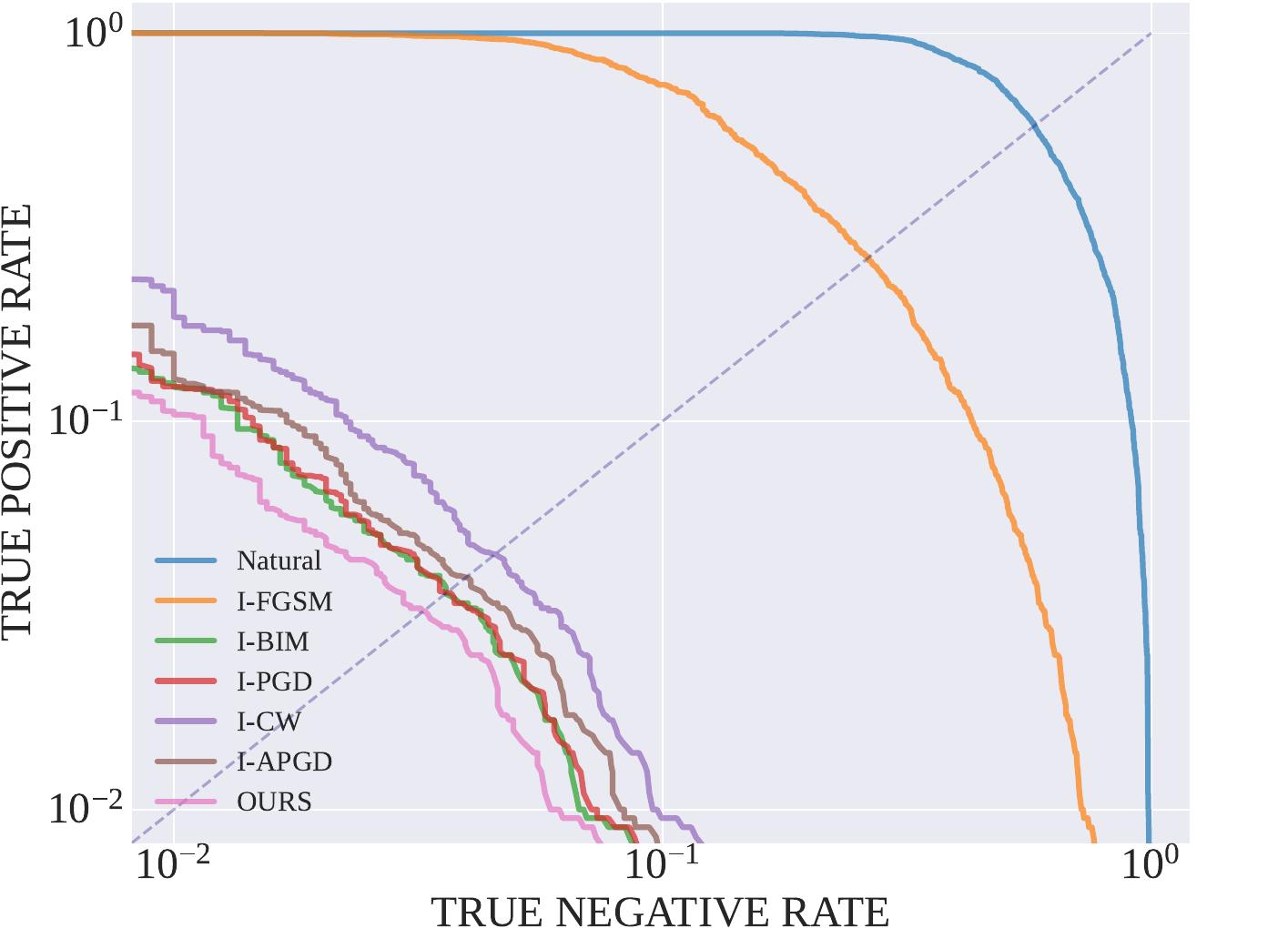}
        \subcaption{$\|\delta\|_{\infty} \leq 8.0/255$}
    \end{minipage}

    \caption{\footnotesize Comparison of the \textbf{Error Area} Between Our Member Fabrication Attack and Baselines Across Diverse Perturbation Bounds on \textbf{CIFAR-10}.}
    \label{Fabric_figure_cifar10}
    \end{center}
    \vspace{-1em}
\end{figure*}

\begin{figure*}[!t]
    \begin{center}
    \begin{minipage}{0.49\textwidth}
        \centering
        \includegraphics[width=\textwidth]{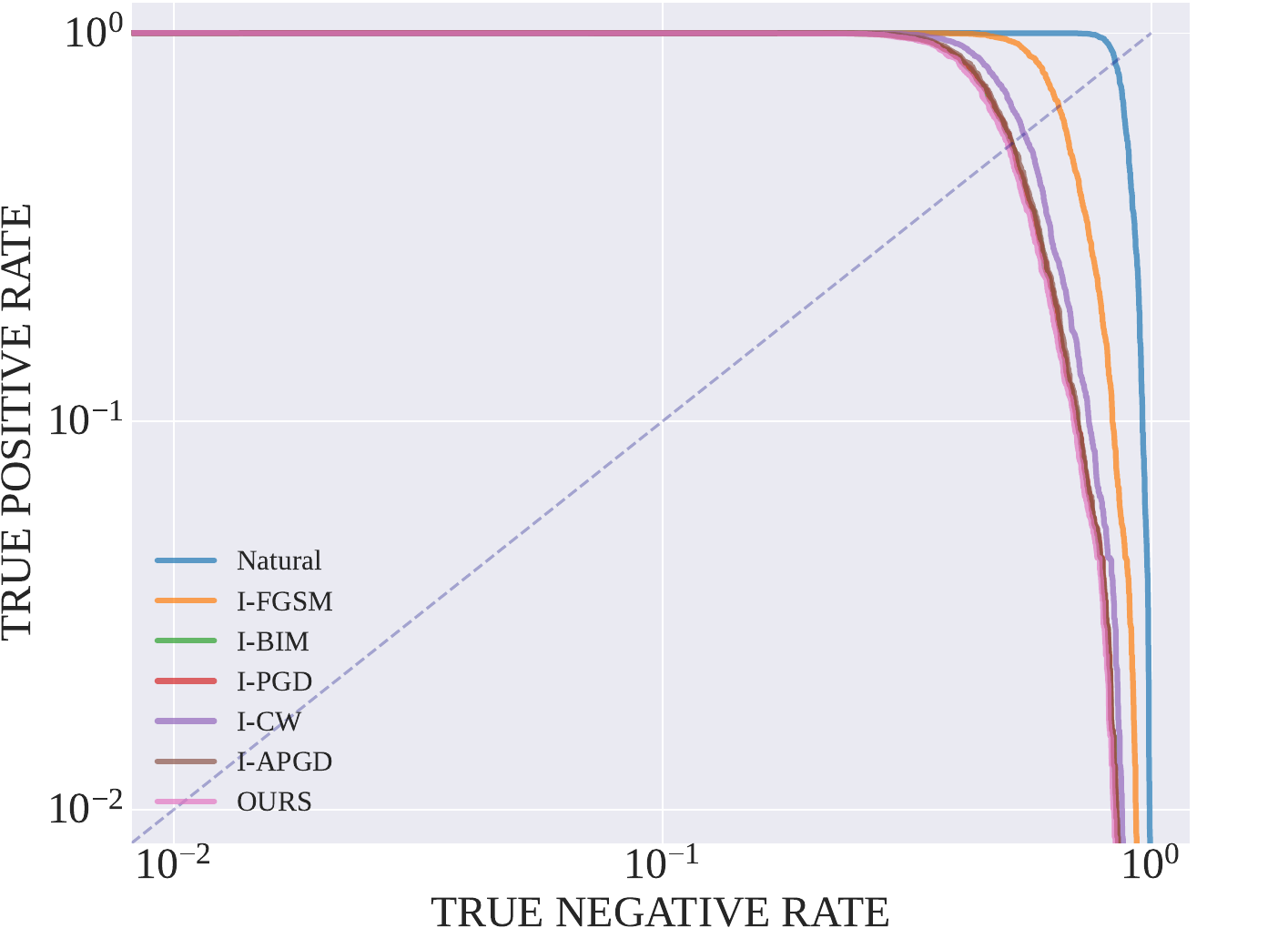}
        \subcaption{$\|\delta\|_{\infty} \leq 3.0/255$}
    \end{minipage}
    \hfill
    \begin{minipage}{0.49\textwidth}
        \centering
        \includegraphics[width=\textwidth]{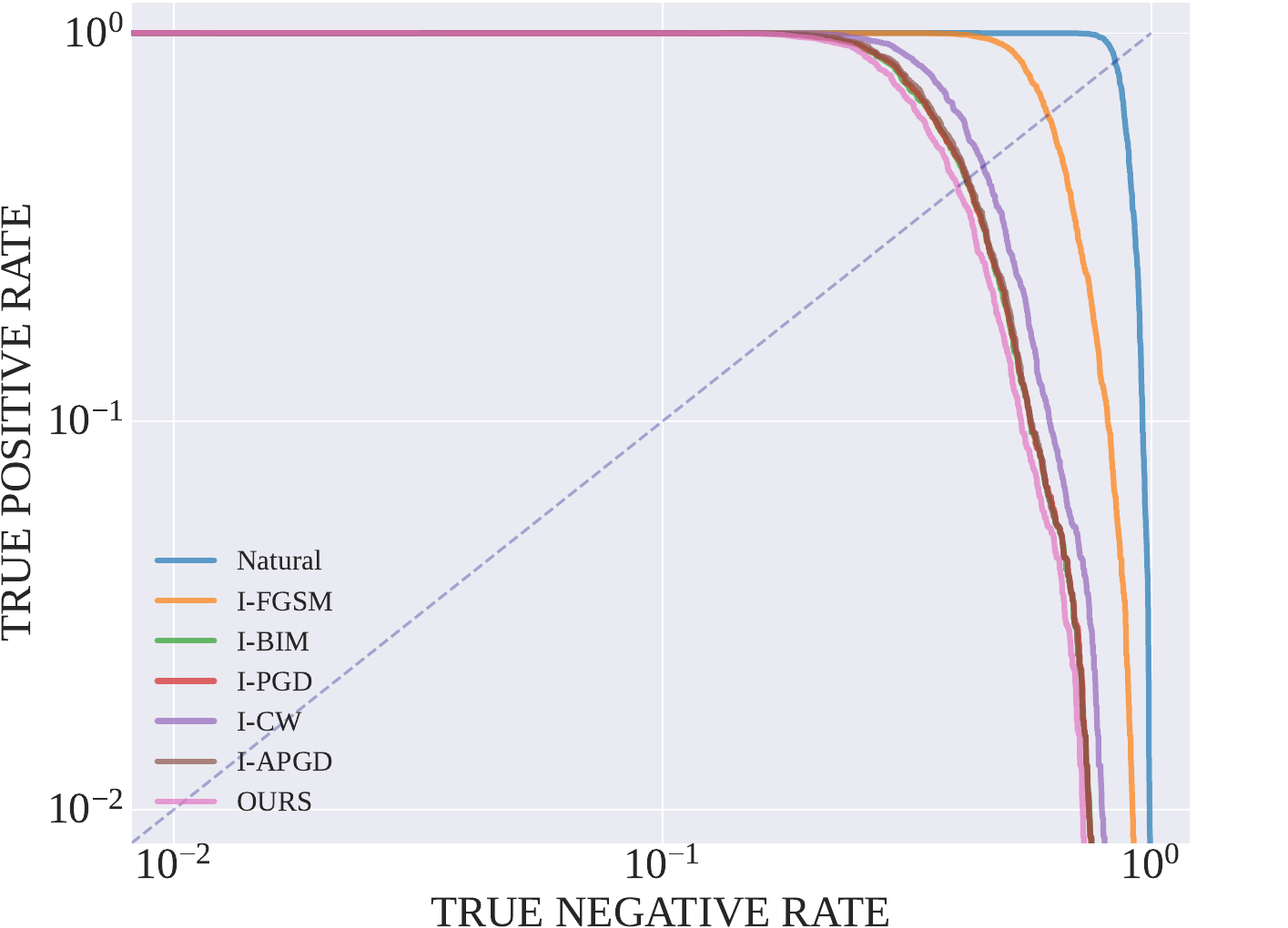}
        \subcaption{$\|\delta\|_{\infty} \leq 4.0/255$}
    \end{minipage}

    \begin{minipage}{0.49\textwidth}
        \centering
        \includegraphics[width=\textwidth]{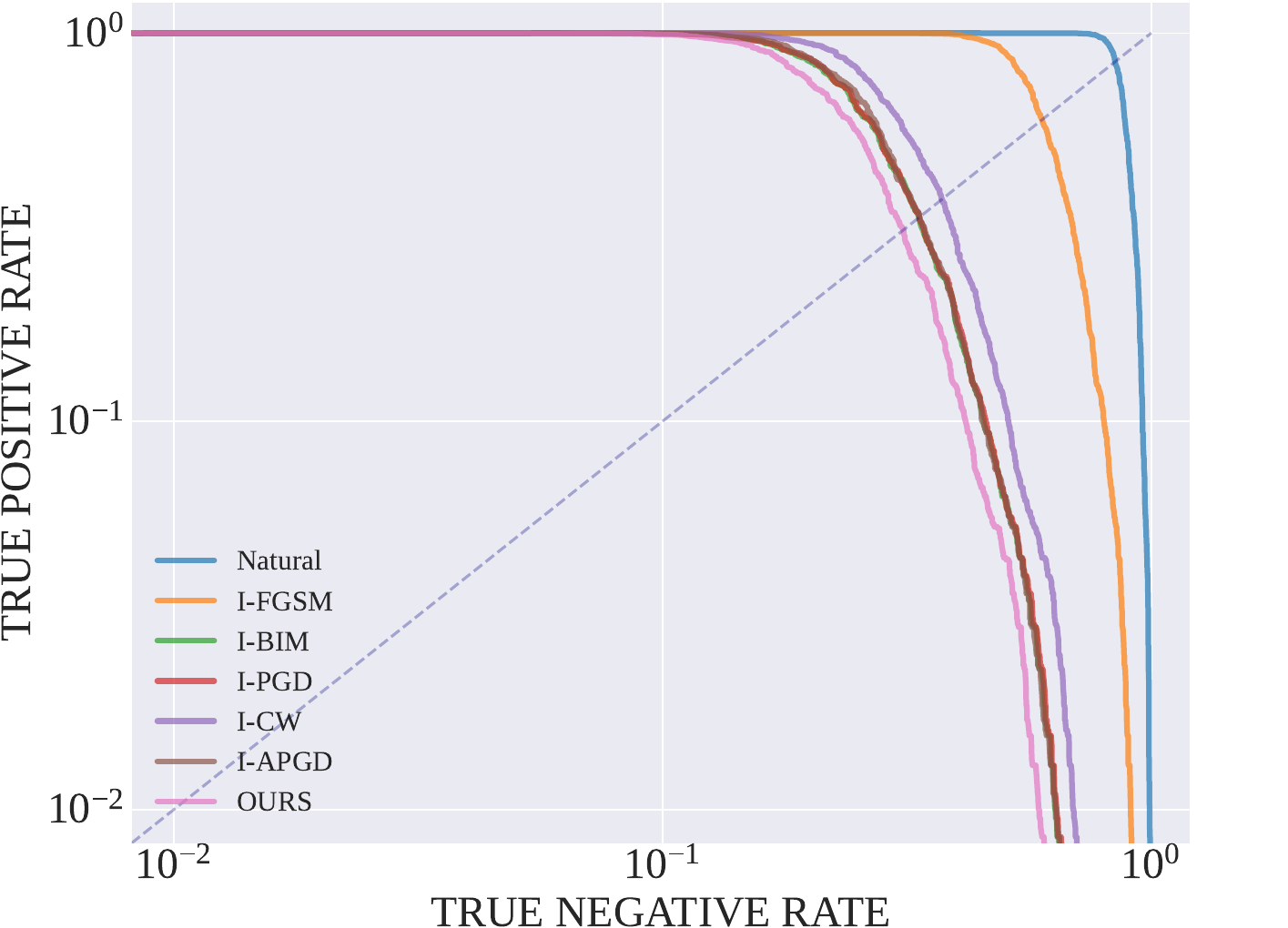}
        \subcaption{$\|\delta\|_{\infty} \leq 5.0/255$}
    \end{minipage}
    \hfill
    \begin{minipage}{0.49\textwidth}
        \centering
        \includegraphics[width=\textwidth]{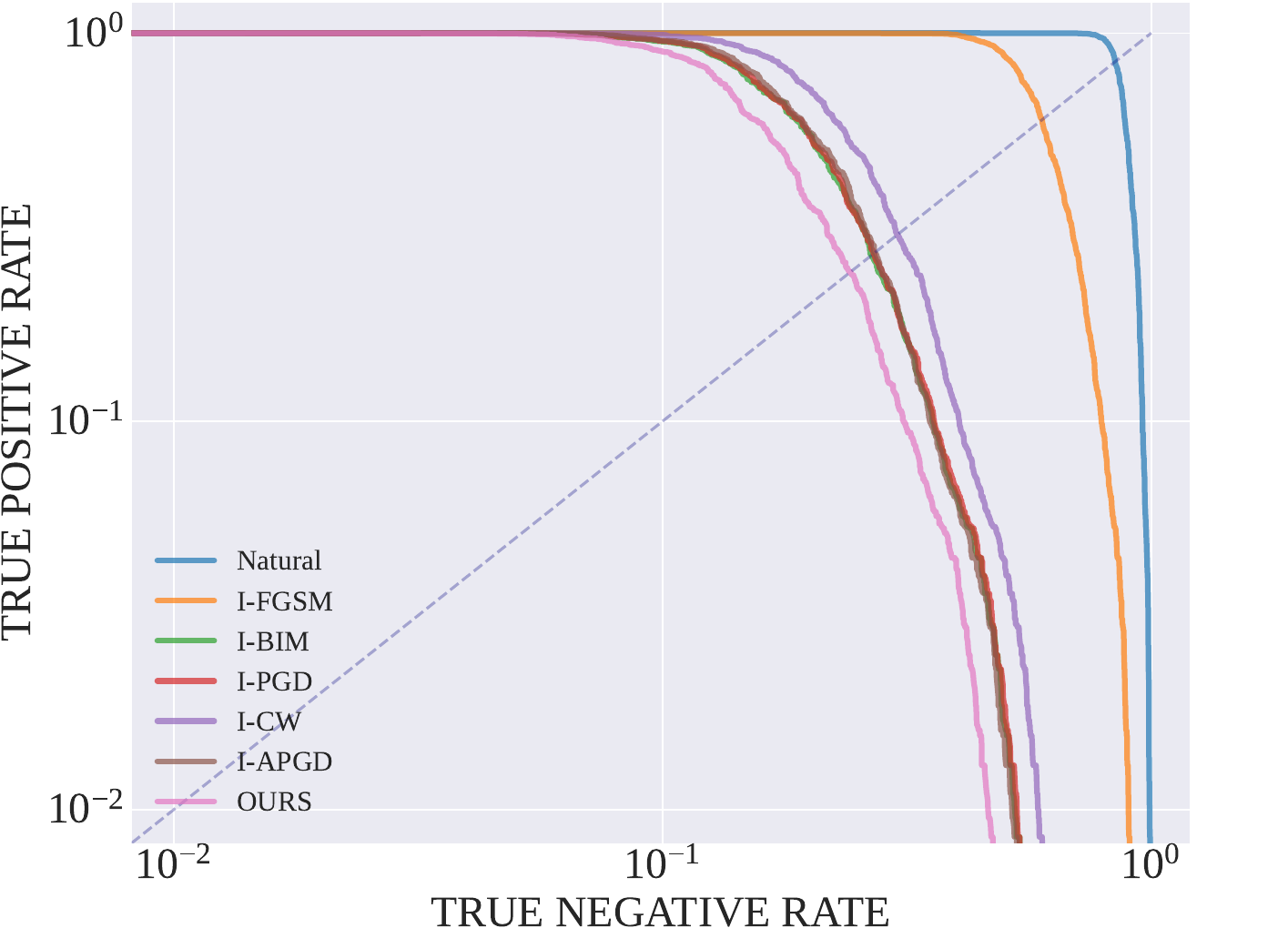}
        \subcaption{$\|\delta\|_{\infty} \leq 6.0/255$}
    \end{minipage}

    \begin{minipage}{0.49\textwidth}
        \centering
        \includegraphics[width=\textwidth]{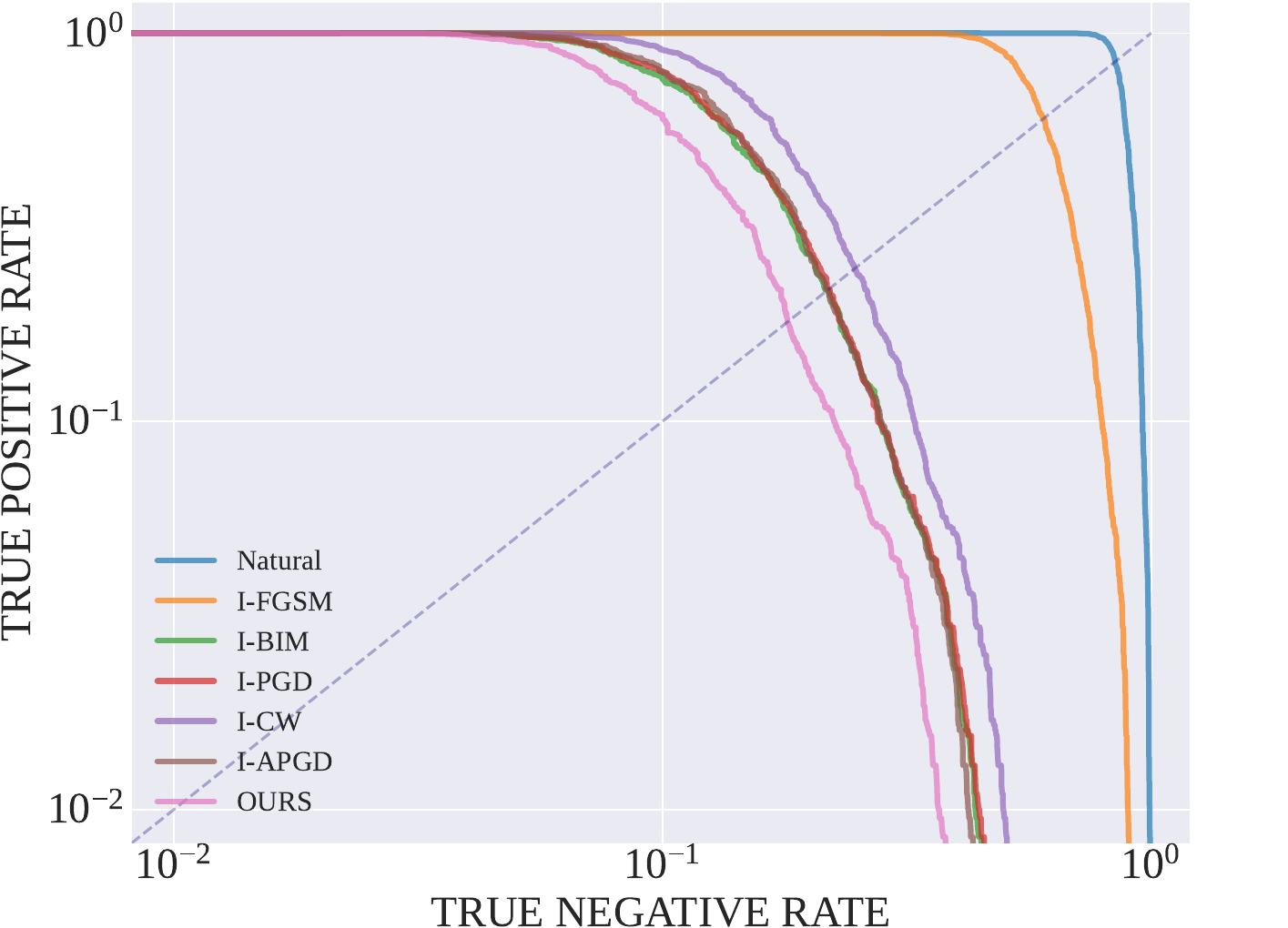}
        \subcaption{$\|\delta\|_{\infty} \leq 7.0/255$}
    \end{minipage}
    \hfill
    \begin{minipage}{0.49\textwidth}
        \centering
        \includegraphics[width=\textwidth]{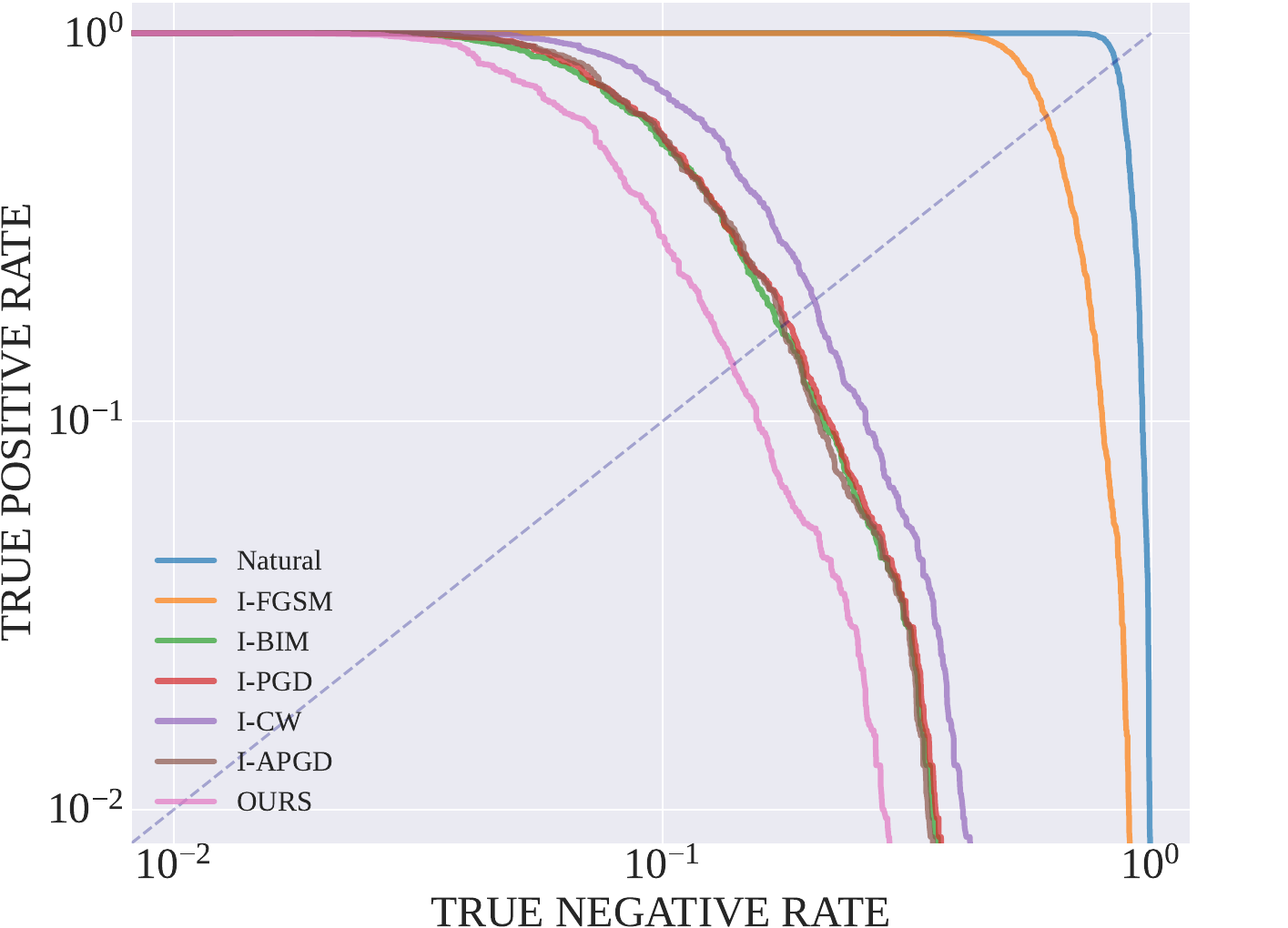}
        \subcaption{$\|\delta\|_{\infty} \leq 8.0/255$}
    \end{minipage}
    \caption{\footnotesize Comparison of the \textbf{Error Area} Between Our Member Fabrication Attack and Baselines Across Diverse Perturbation Bounds on \textbf{CIFAR-100}.}
    \label{Fabric_figure_cifar100}
    \end{center}
    \vspace{-1em}
\end{figure*}

\begin{figure*}[!t]
    \begin{center}
    \begin{minipage}{0.49\textwidth}
        \centering
        \includegraphics[width=\textwidth]{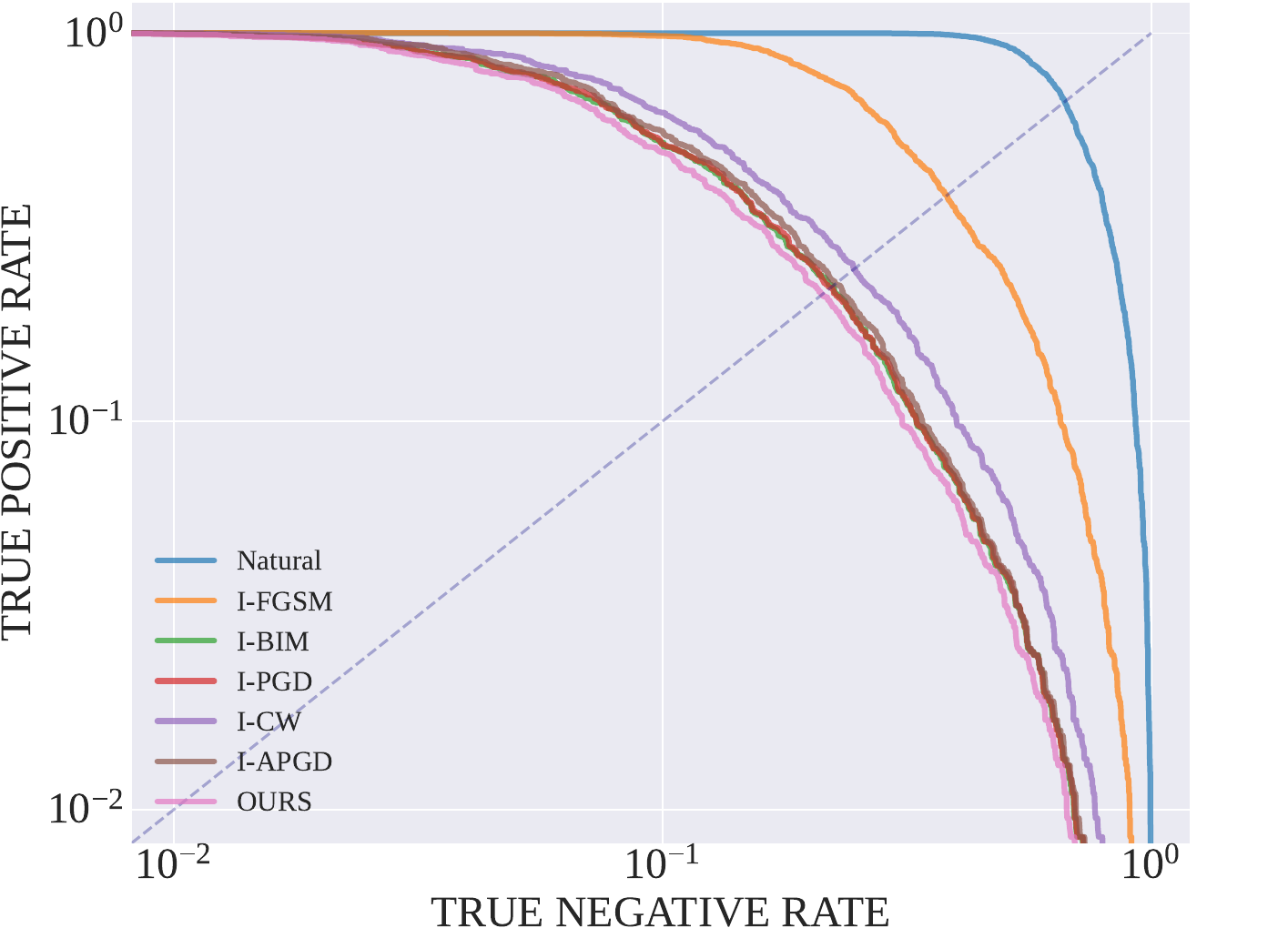}
        \subcaption{$\|\delta\|_{\infty} \leq 3.0/255$}
    \end{minipage}
    \hfill
    \begin{minipage}{0.49\textwidth}
        \centering
        \includegraphics[width=\textwidth]{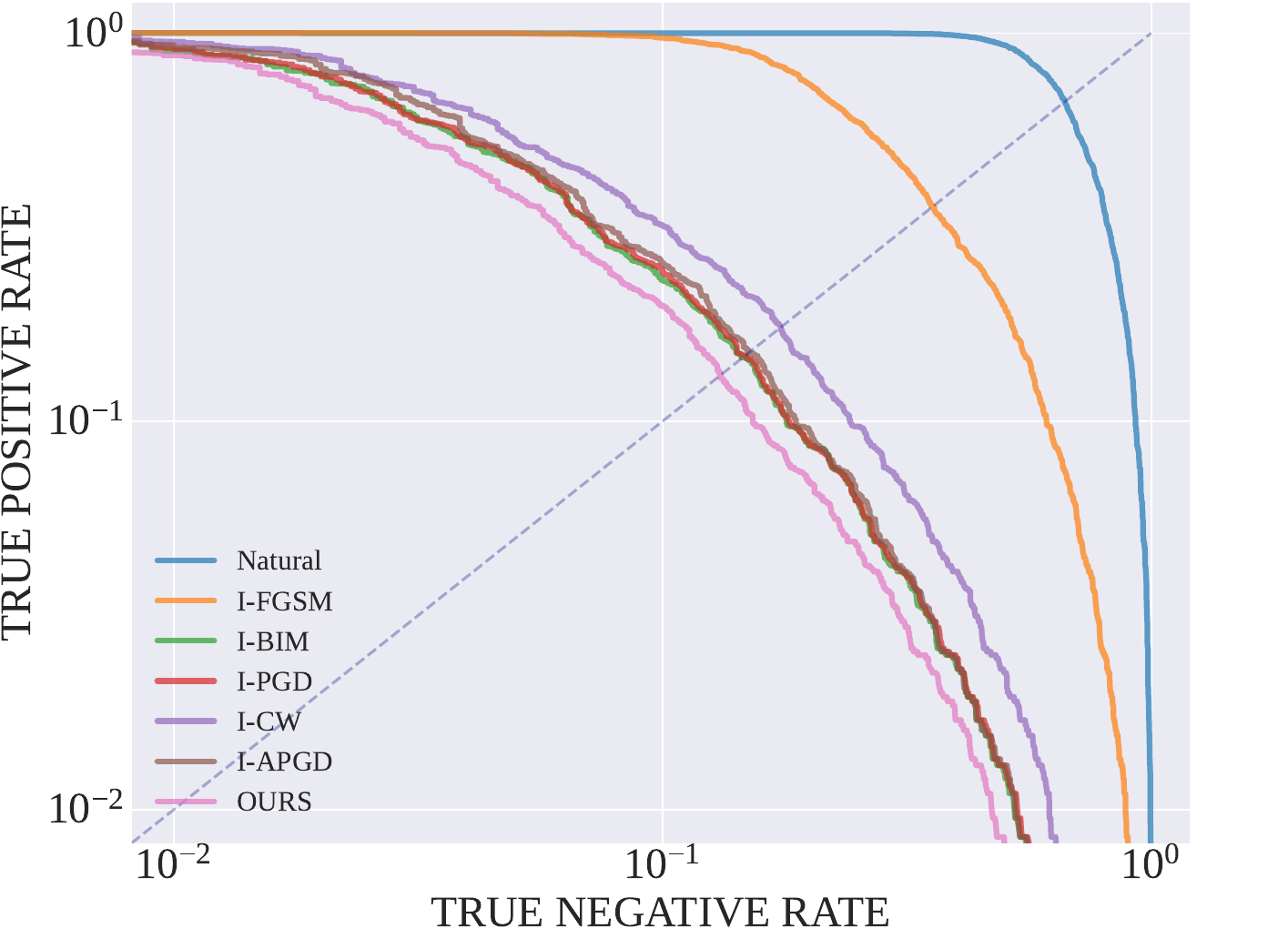}
        \subcaption{$\|\delta\|_{\infty} \leq 4.0/255$}
    \end{minipage}

    \begin{minipage}{0.49\textwidth}
        \centering
        \includegraphics[width=\textwidth]{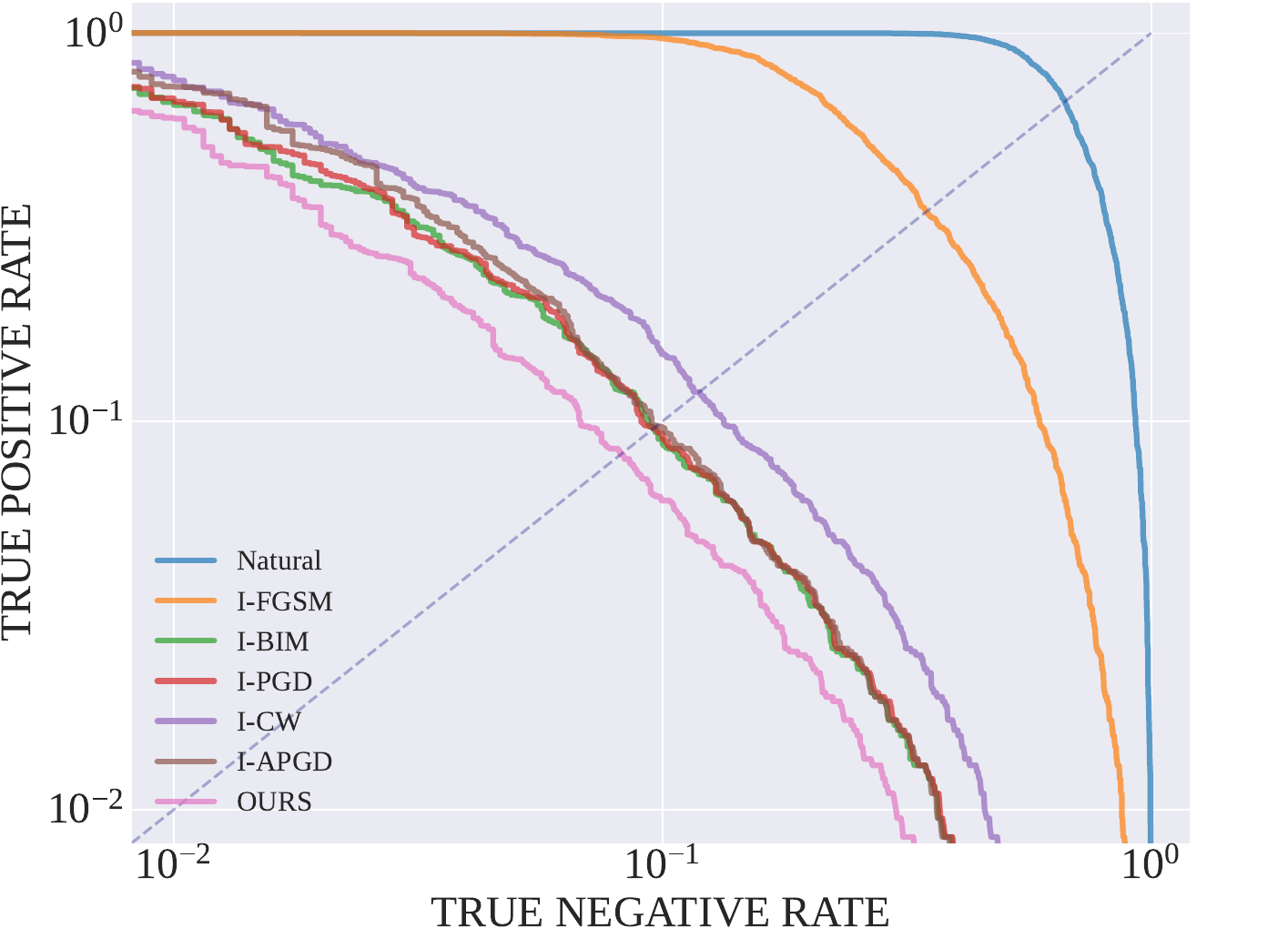}
        \subcaption{$\|\delta\|_{\infty} \leq 5.0/255$}
    \end{minipage}
    \hfill
    \begin{minipage}{0.49\textwidth}
        \centering
        \includegraphics[width=\textwidth]{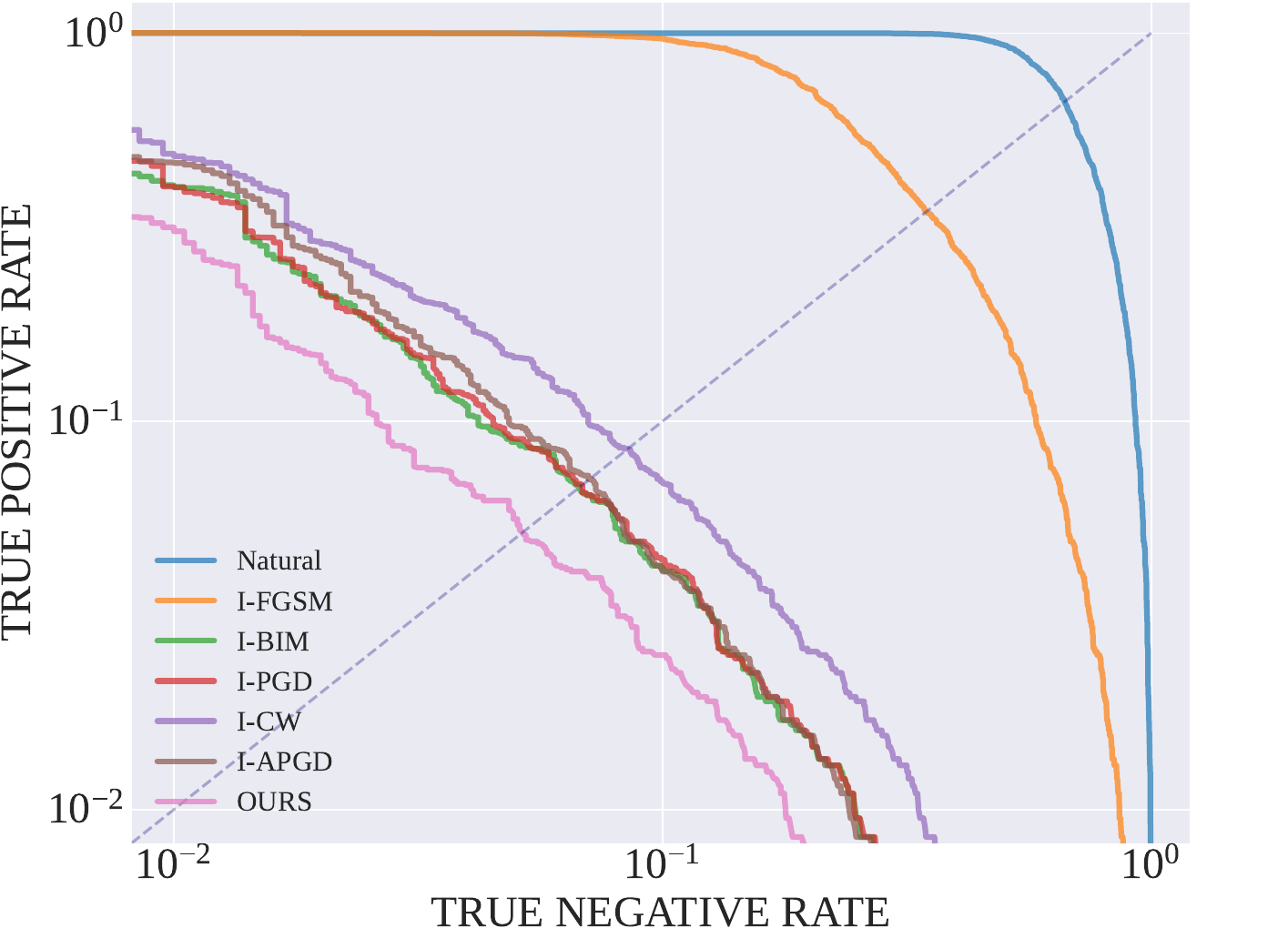}
        \subcaption{$\|\delta\|_{\infty} \leq 6.0/255$}
    \end{minipage}

    \begin{minipage}{0.49\textwidth}
        \centering
        \includegraphics[width=\textwidth]{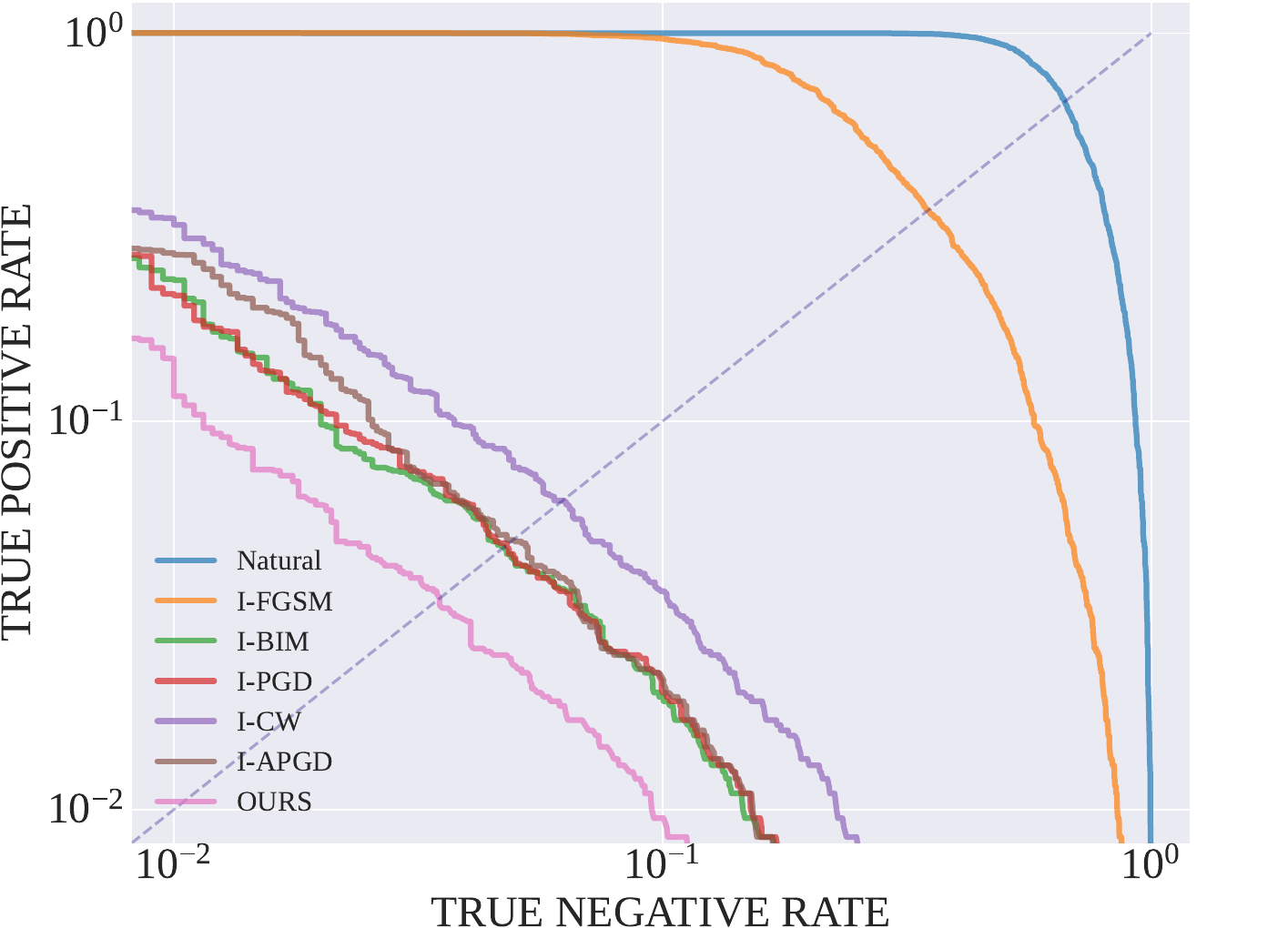}
        \subcaption{$\|\delta\|_{\infty} \leq 7.0/255$}
    \end{minipage}
    \hfill
    \begin{minipage}{0.49\textwidth}
        \centering
        \includegraphics[width=\textwidth]{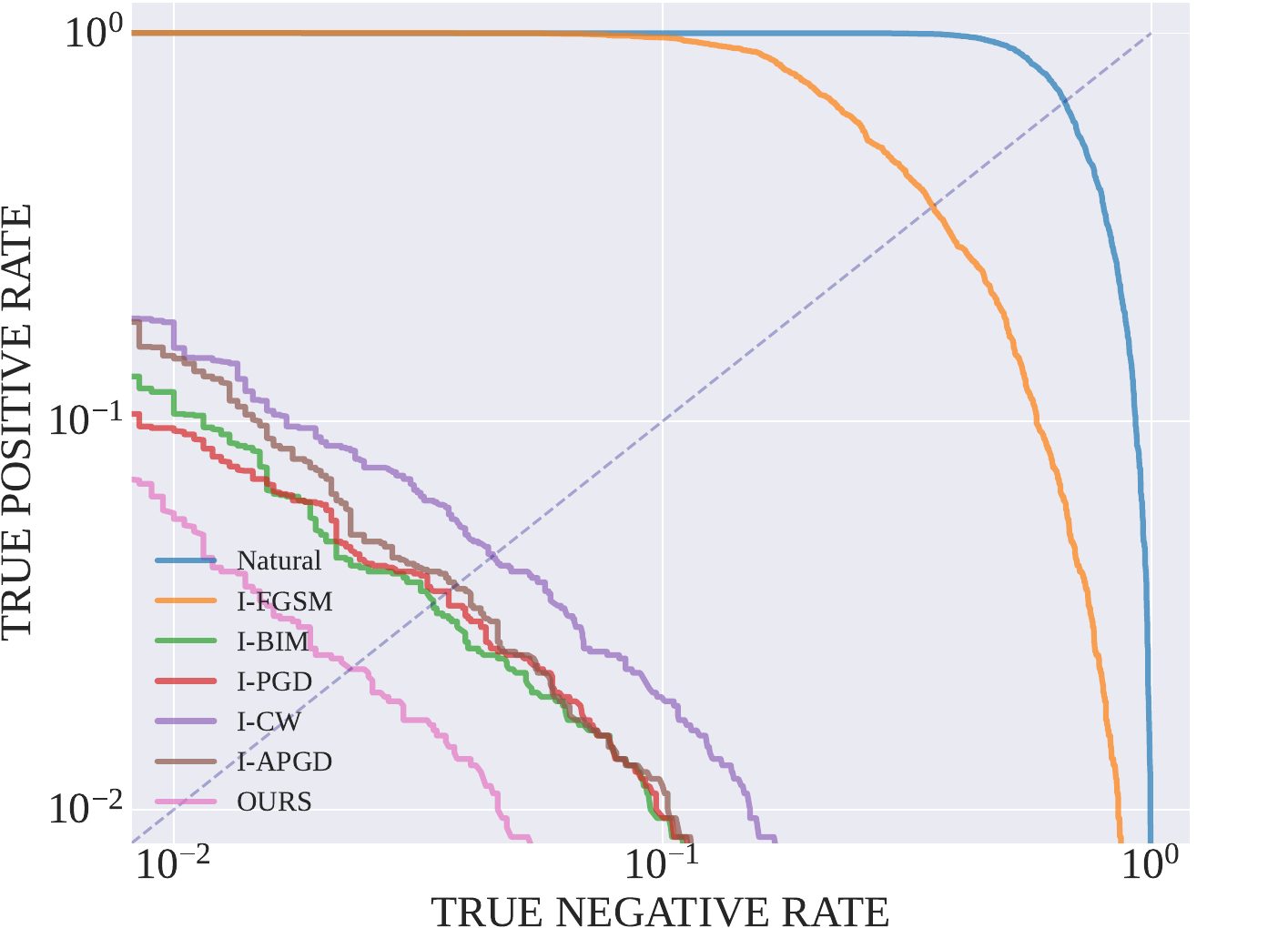}
        \subcaption{$\|\delta\|_{\infty} \leq 8.0/255$}
    \end{minipage}
    \caption{\footnotesize Comparison of the \textbf{Error Area} Between Our Member Fabrication Attack and Baselines Across Diverse Perturbation Bounds on \textbf{CINIC-10}.}
    \label{Fabric_figure_cinic}
    \end{center}
    \vspace{-1em}
\end{figure*}

\begin{figure*}[!t]
    \begin{center}
    \begin{minipage}{0.49\textwidth}
        \centering
        \includegraphics[width=\textwidth]{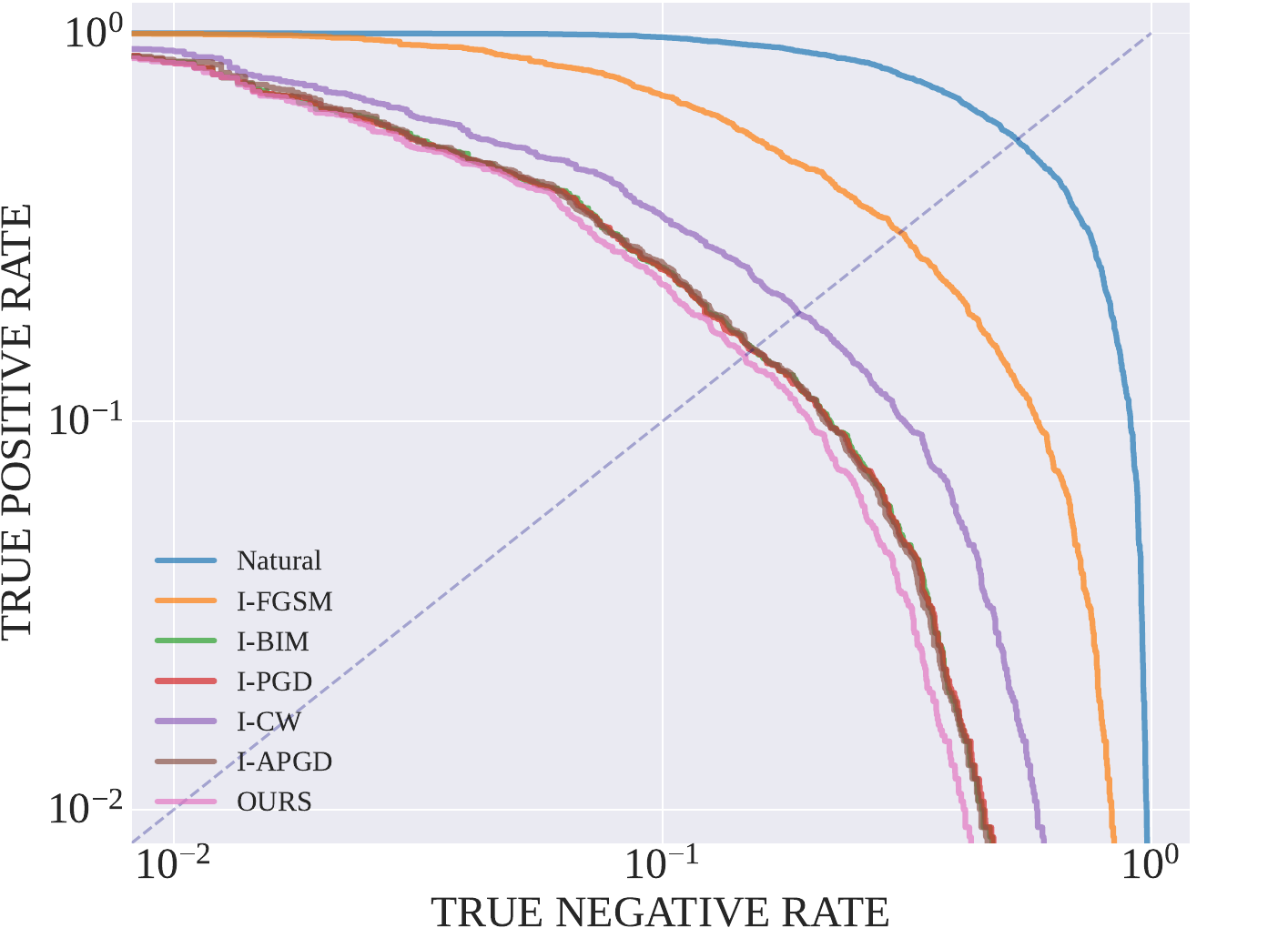}
        \subcaption{$\|\delta\|_{\infty} \leq 3.0/255$}
    \end{minipage}
    \hfill
    \begin{minipage}{0.49\textwidth}
        \centering
        \includegraphics[width=\textwidth]{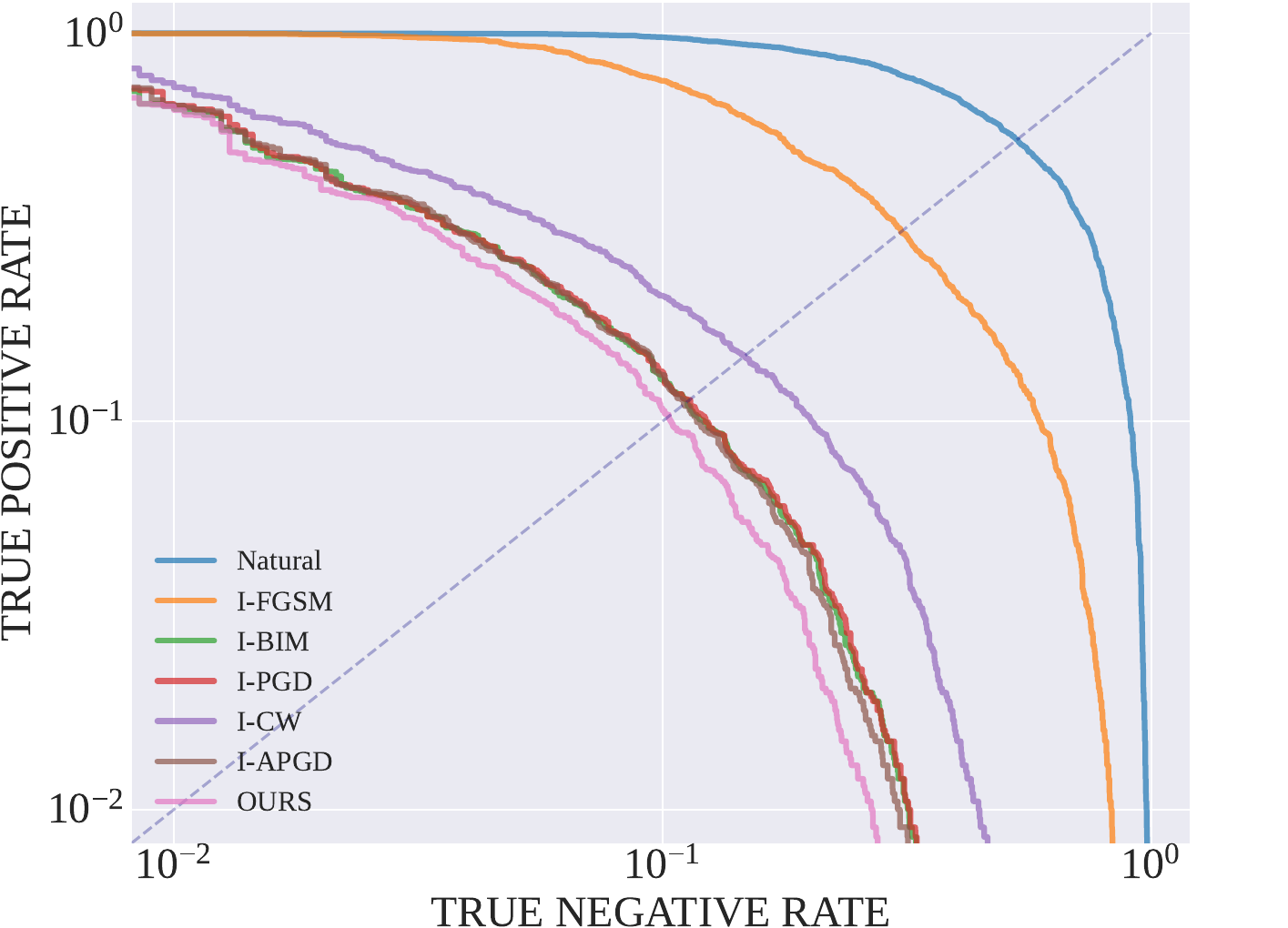}
        \subcaption{$\|\delta\|_{\infty} \leq 4.0/255$}
    \end{minipage}

    \begin{minipage}{0.49\textwidth}
        \centering
        \includegraphics[width=\textwidth]{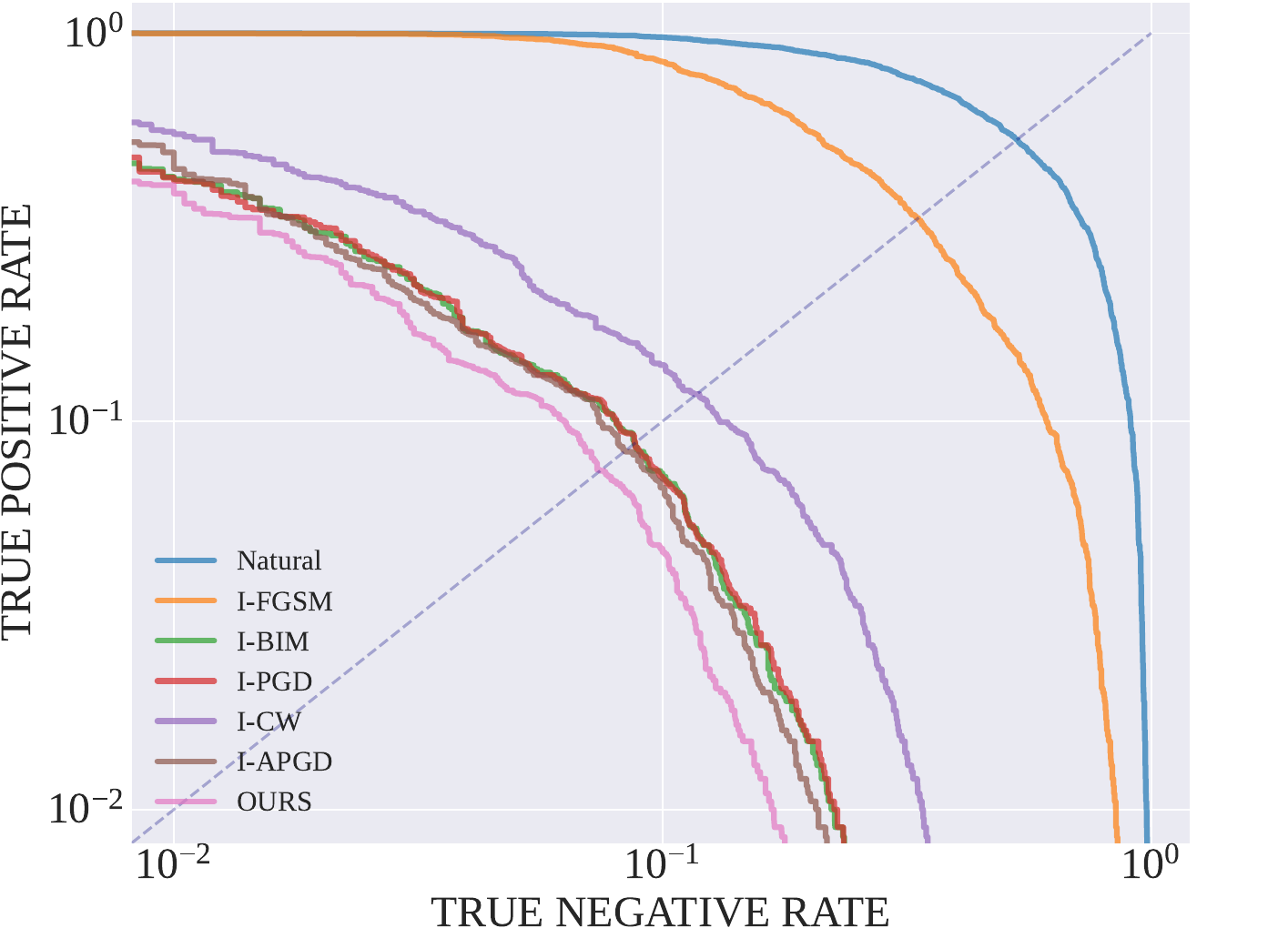}
        \subcaption{$\|\delta\|_{\infty} \leq 5.0/255$}
    \end{minipage}
    \hfill
    \begin{minipage}{0.49\textwidth}
        \centering
        \includegraphics[width=\textwidth]{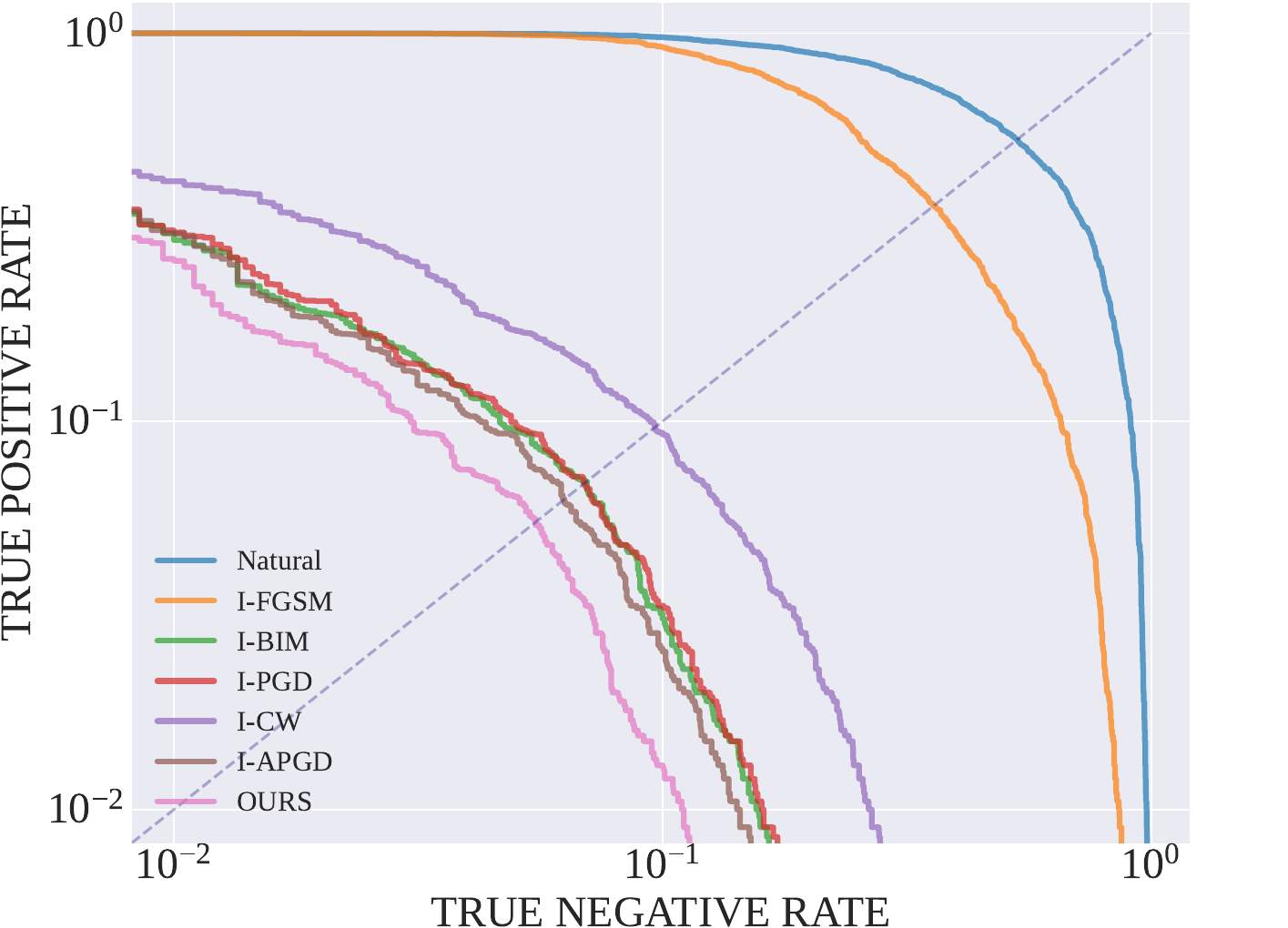}
        \subcaption{$\|\delta\|_{\infty} \leq 6.0/255$}
    \end{minipage}

    \begin{minipage}{0.49\textwidth}
        \centering
        \includegraphics[width=\textwidth]{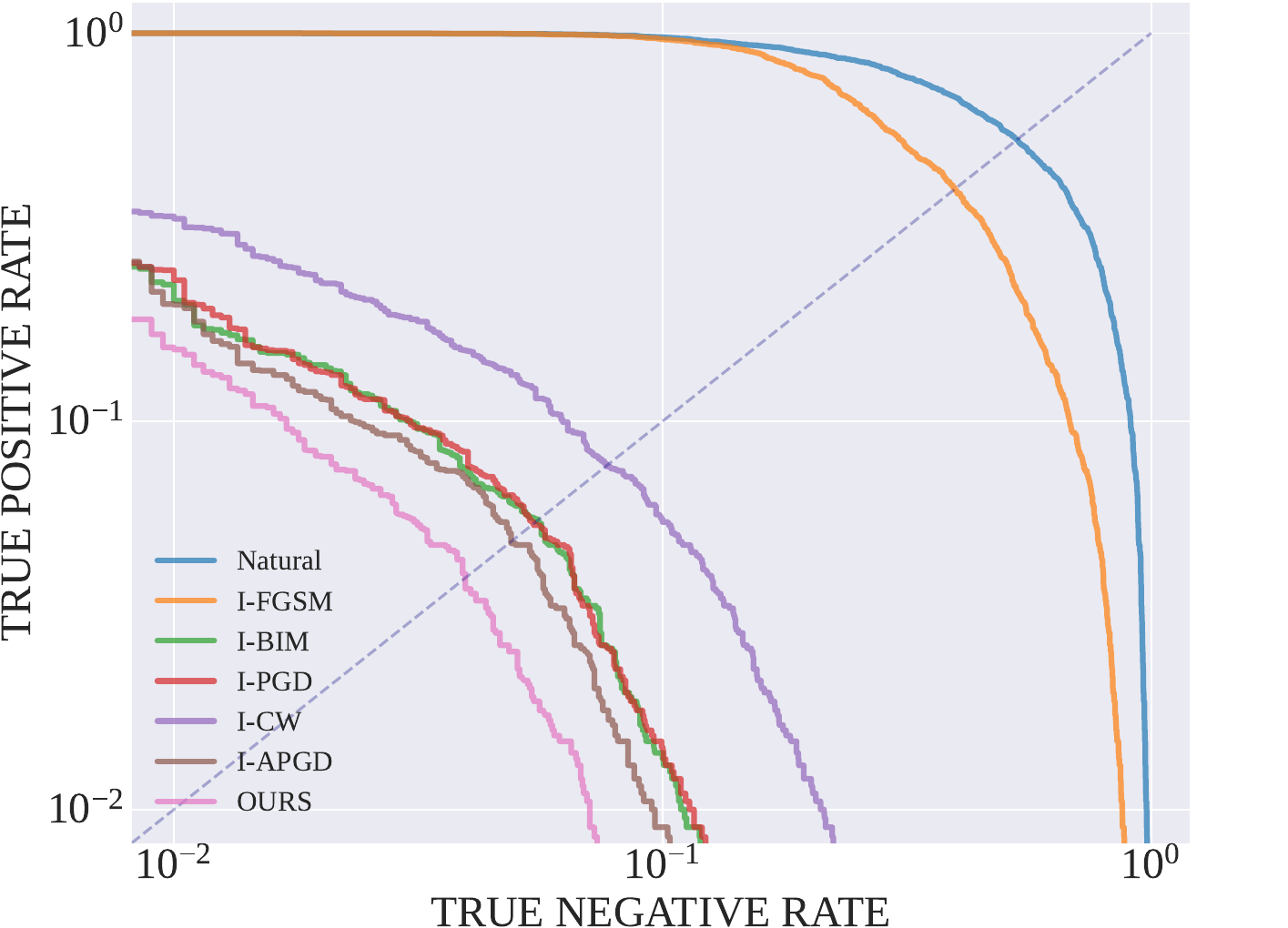}
        \subcaption{$\|\delta\|_{\infty} \leq 7.0/255$}
    \end{minipage}
    \hfill
    \begin{minipage}{0.49\textwidth}
        \centering
        \includegraphics[width=\textwidth]{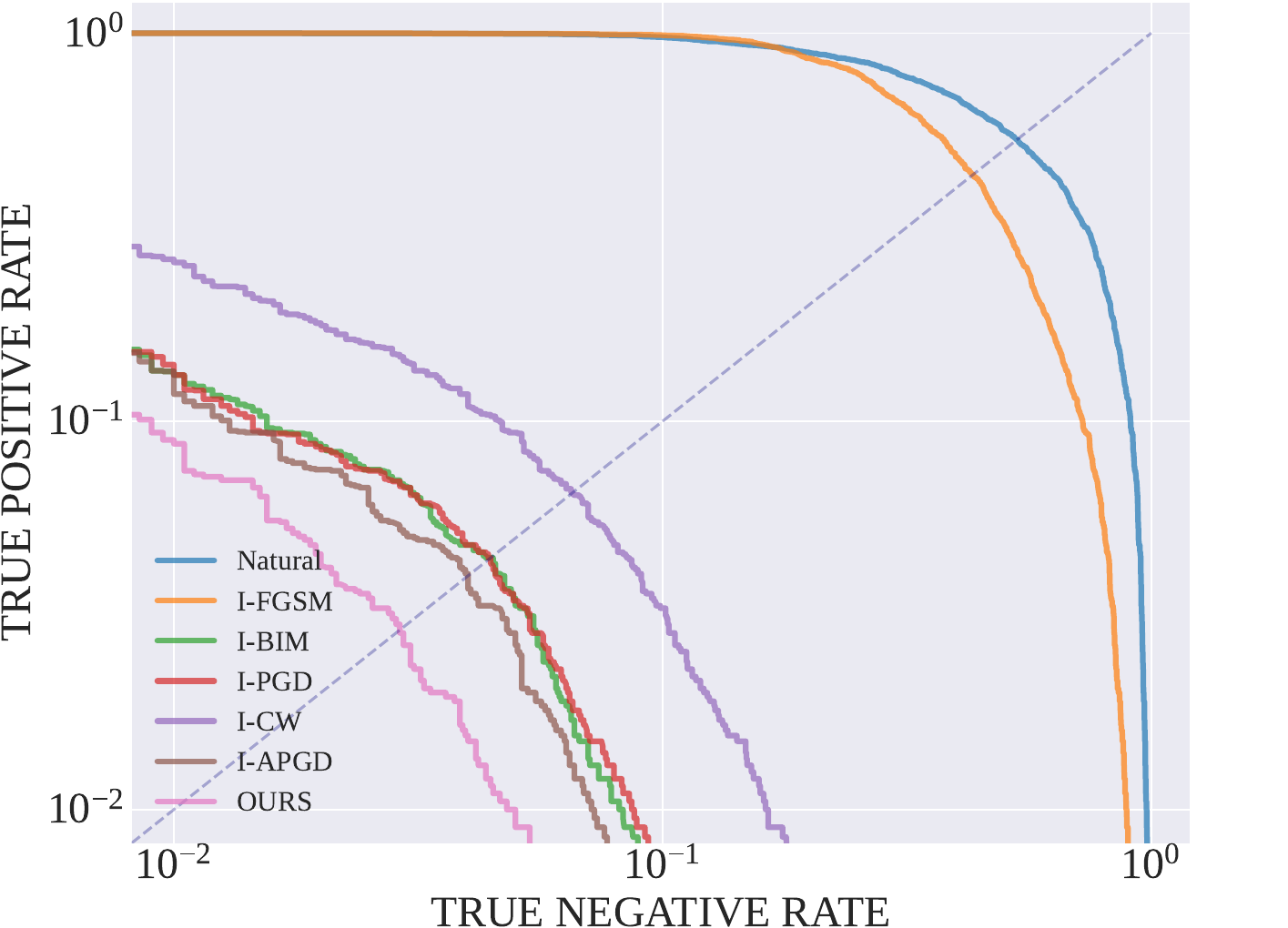}
        \subcaption{$\|\delta\|_{\infty} \leq 8.0/255$}
    \end{minipage}
    \caption{\footnotesize  Comparison of the \textbf{Error Area} Between Our Member Fabrication Attack and Baselines Across Diverse Perturbation Bounds on \textbf{SVHN}.}
    \label{Fabric_figure_svhn}
    \end{center}
    \vspace{-1em}
\end{figure*}

\begin{figure*}[!t]
    \begin{center}
    \begin{minipage}{0.49\textwidth}
        \centering
        \includegraphics[width=\textwidth]{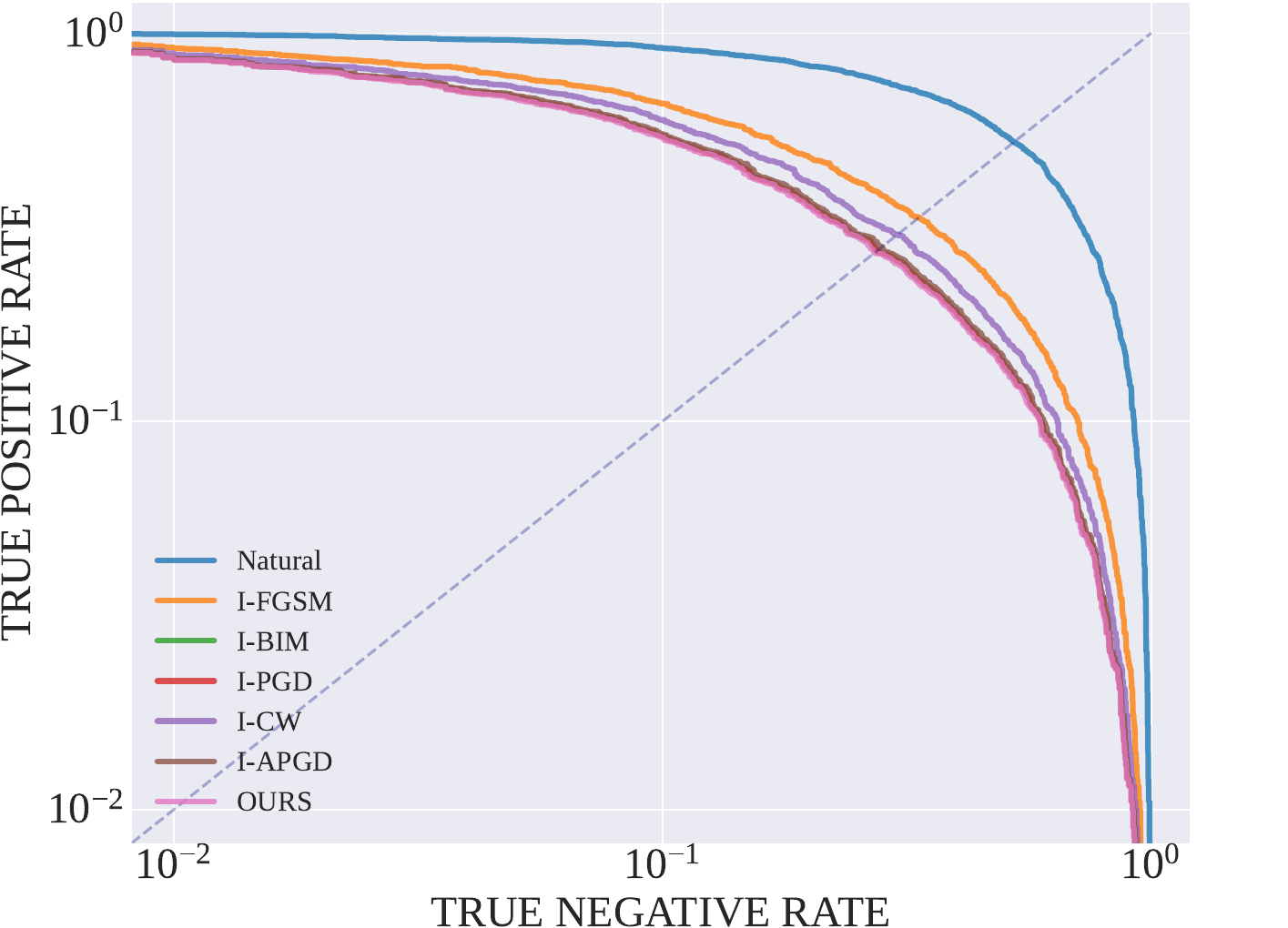}
        \subcaption{$\|\delta\|_{\infty} \leq 0.5/255$}
    \end{minipage}
    \hfill
    \begin{minipage}{0.49\textwidth}
        \centering
        \includegraphics[width=\textwidth]{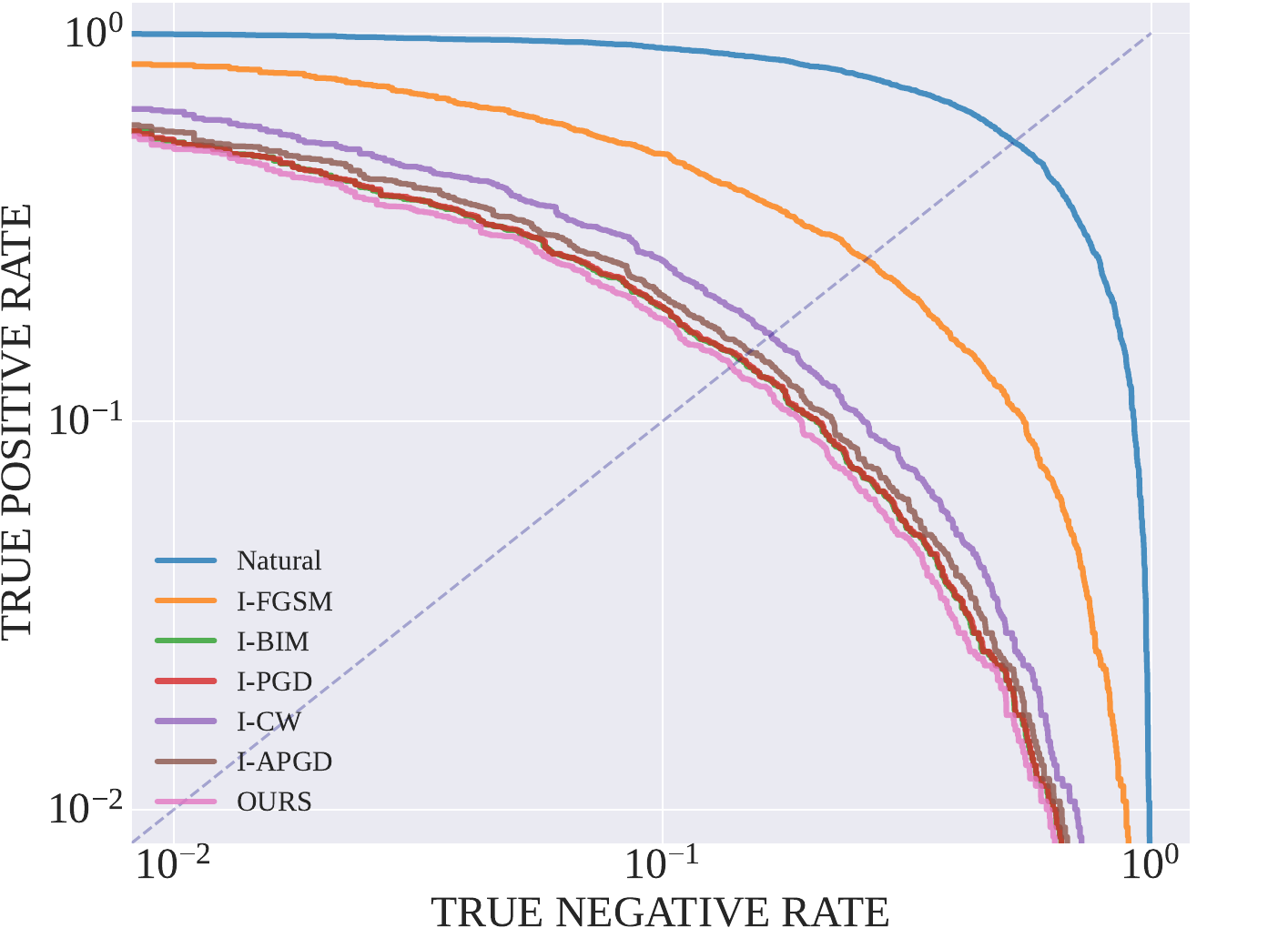}
        \subcaption{$\|\delta\|_{\infty} \leq 1.0/255$}
    \end{minipage}

    \begin{minipage}{0.49\textwidth}
        \centering
        \includegraphics[width=\textwidth]{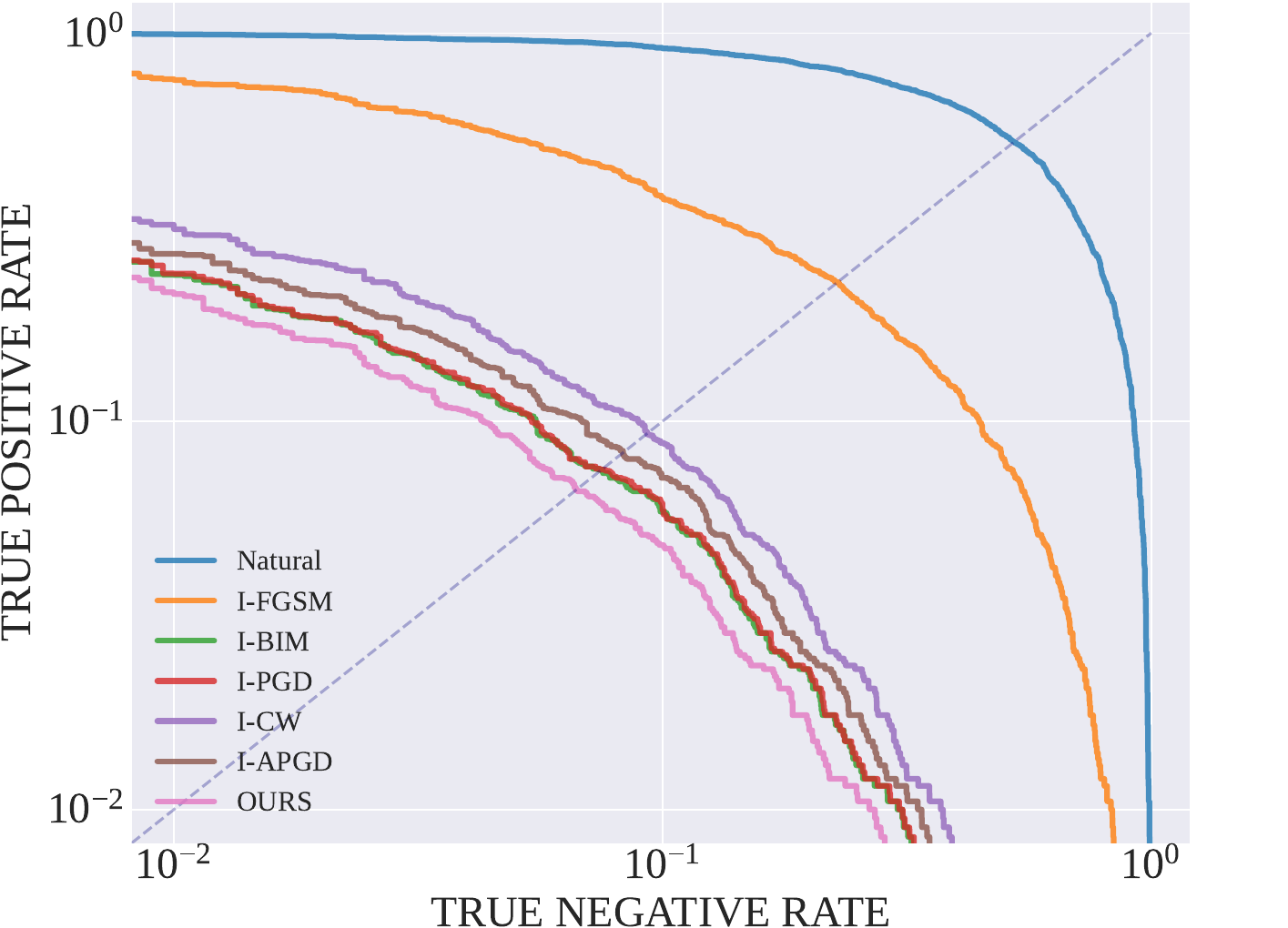}
        \subcaption{$\|\delta\|_{\infty} \leq 1.5/255$}
    \end{minipage}
    \hfill
    \begin{minipage}{0.49\textwidth}
        \centering
        \includegraphics[width=\textwidth]{crop_Figure/Fabric_figure/crop_roc_curve_comparison_loss_Imagenet_2.pdf}
        \subcaption{$\|\delta\|_{\infty} \leq 2.0/255$}
    \end{minipage}
    \caption{\footnotesize  Comparison of the \textbf{Error Area} Between Our Member Fabrication Attack and Baselines Across Diverse Perturbation Bounds on \textbf{ImageNet-100}. }
    \label{Fabric_figure_Imagenet}
    \end{center}
    \vspace{-1em}
\end{figure*}

\vspace{0.2em}
\noindent\textbf{Experimental Results of \textbf{MFA}.} In \Cref{Fabric_figure_4_attack}, we show that our \textbf{MFA} can effectively deceive different MIAs. Four subfigures depict four representative MIAs: loss attack, Attack R, LiRA, and RMIA. In each subfigure, we present four datasets: \textbf{CIFAR-10} \citep{krizhevsky2009learning}, \textbf{CIFAR-100} \citep{krizhevsky2009learning}, \textbf{SVHN} \citep{krizhevsky2009learning}, and \textbf{CINIC-10} \citep{darlow2018cinic}.
We observe that our Member Fabrication Attack (\textbf{MFA}) leads these MIAs to exhibit a low TPR, low TNR, high Error Area, and high Equal Error Rate (EER) in the TNR-TPR curve. The TNR-TPR curve is close to the bottom left. In \Cref{Fabric_figure_cifar10,Fabric_figure_cifar100,Fabric_figure_cinic,Fabric_figure_svhn,Fabric_figure_Imagenet}, we compare our \textbf{MFA} against the loss attack, as well as five adapted adversarial attacks across different perturbation levels and datasets: \emph{(i)} Inverted Fast Gradient Sign Method (I-FGSM)~\citep{goodfellow2014explaining}, \emph{(ii)} Inverted Basic Iterative Method (I-BIM)~\citep{kurakin2016adversarial}, \emph{(iii)} Inverted Projected Gradient Descent (I-PGD)~\citep{Madry18PGD}, \emph{(iv)} Inverted Carlini and Wagner Attack (I-CW)~\citep{carlini2017towards}, and \emph{(v)} Inverted Adaptive PGD (I-APGD)~\citep{croce2020reliable}. The results show that our methods perform the best among these techniques in the above metrics. The quantitative comparison results can be found in \Cref{tab:fabric_error_area,tab:fabric_error_area_more_MIAs,tab:fabric_eer_more_MIAs,tab:fabric_eer}, where the best results are highlighted in red.

\begin{figure*}[!t]
    \begin{center}
    \begin{minipage}{0.49\textwidth}
        \centering
        \includegraphics[width=\textwidth]{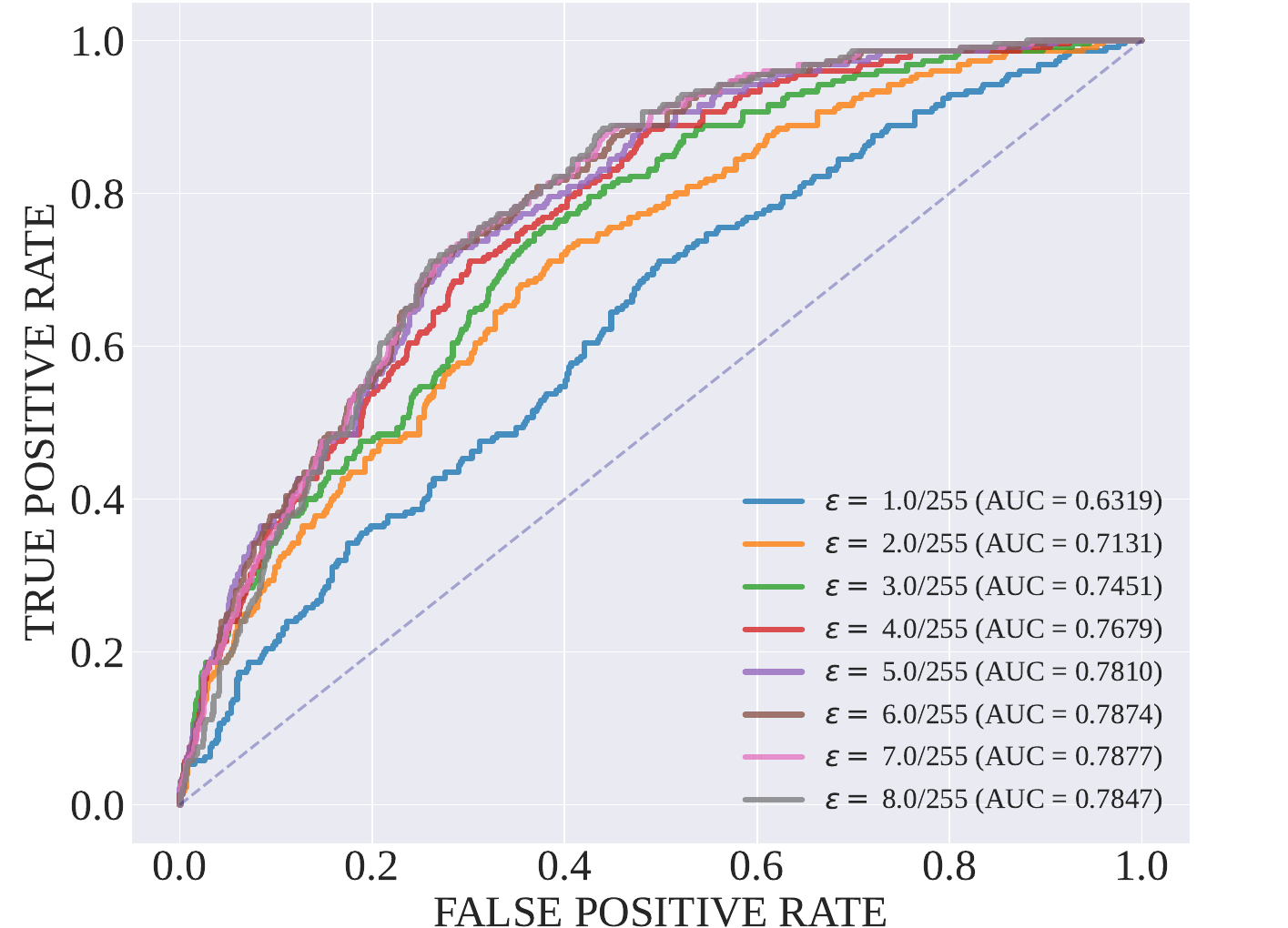}
        \subcaption{Fabricated Members by I-FGSM}
    \end{minipage}
    \hfill
    \begin{minipage}{0.49\textwidth}
        \centering
        \includegraphics[width=\textwidth]{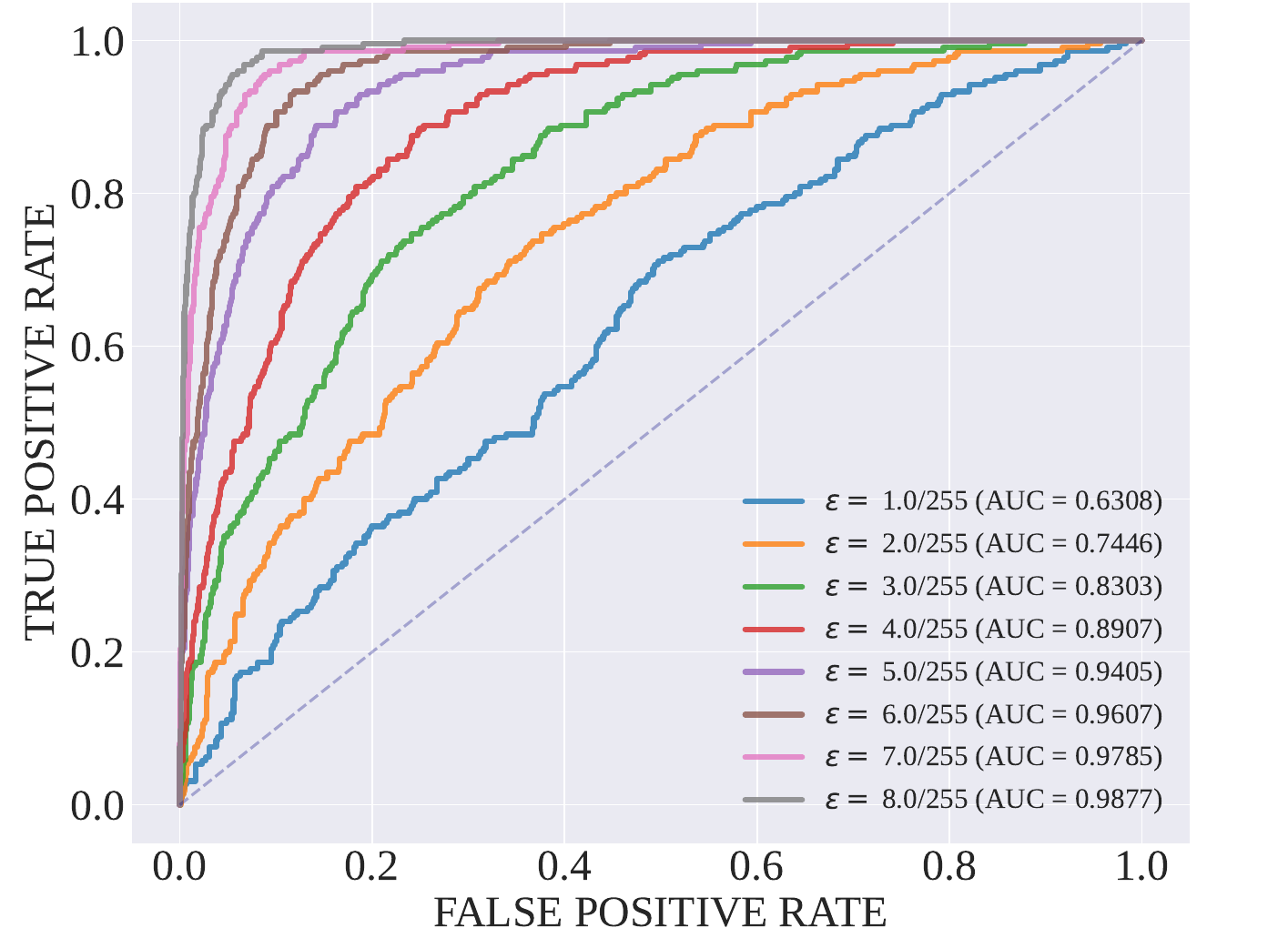}
        \subcaption{Fabricated Members by I-BIM}
    \end{minipage}

    \begin{minipage}{0.49\textwidth}
        \centering
        \includegraphics[width=\textwidth]{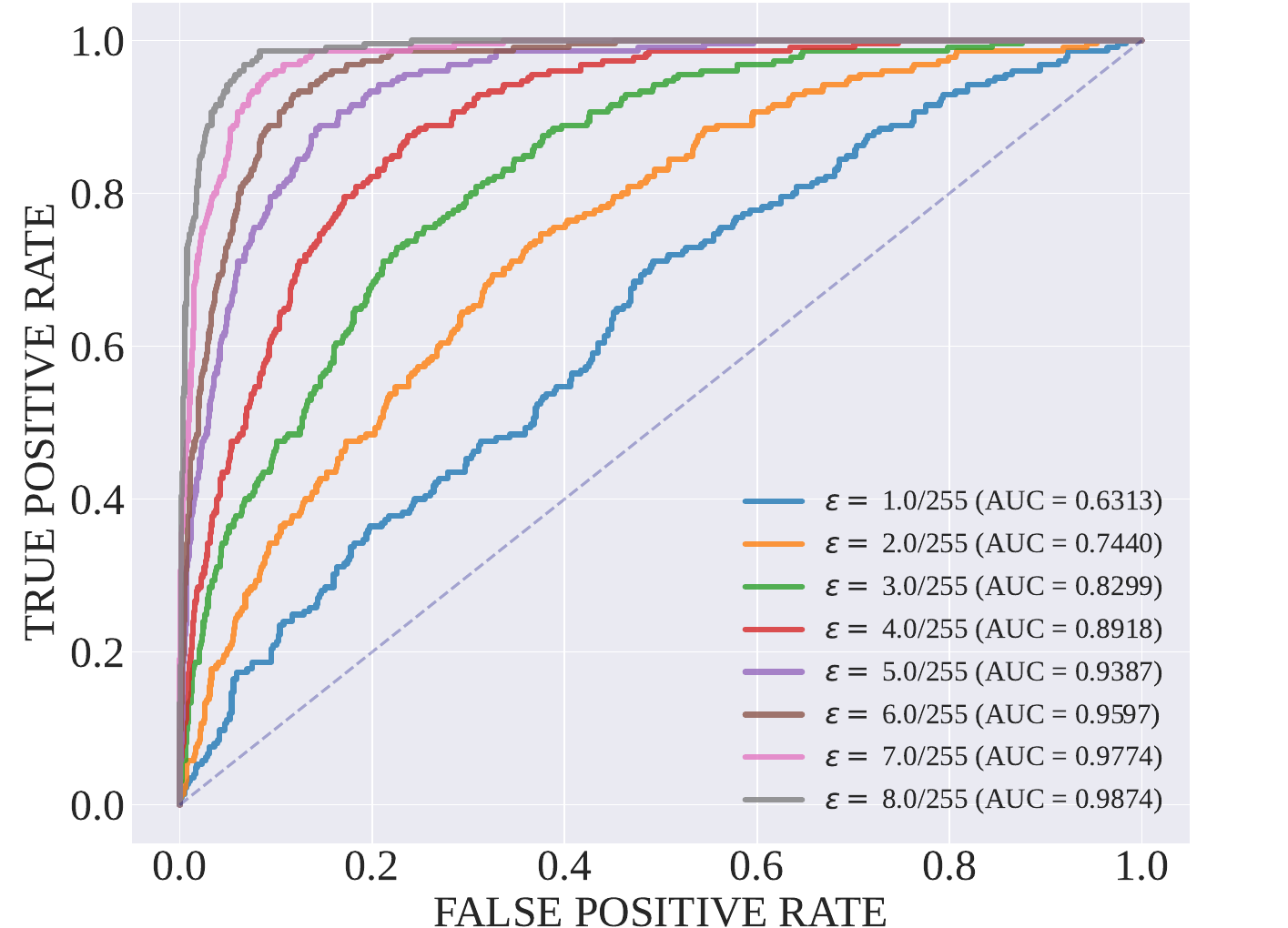}
        \subcaption{Fabricated Members by I-PGD}
    \end{minipage}
    \hfill
    \begin{minipage}{0.49\textwidth}
        \centering
        \includegraphics[width=\textwidth]{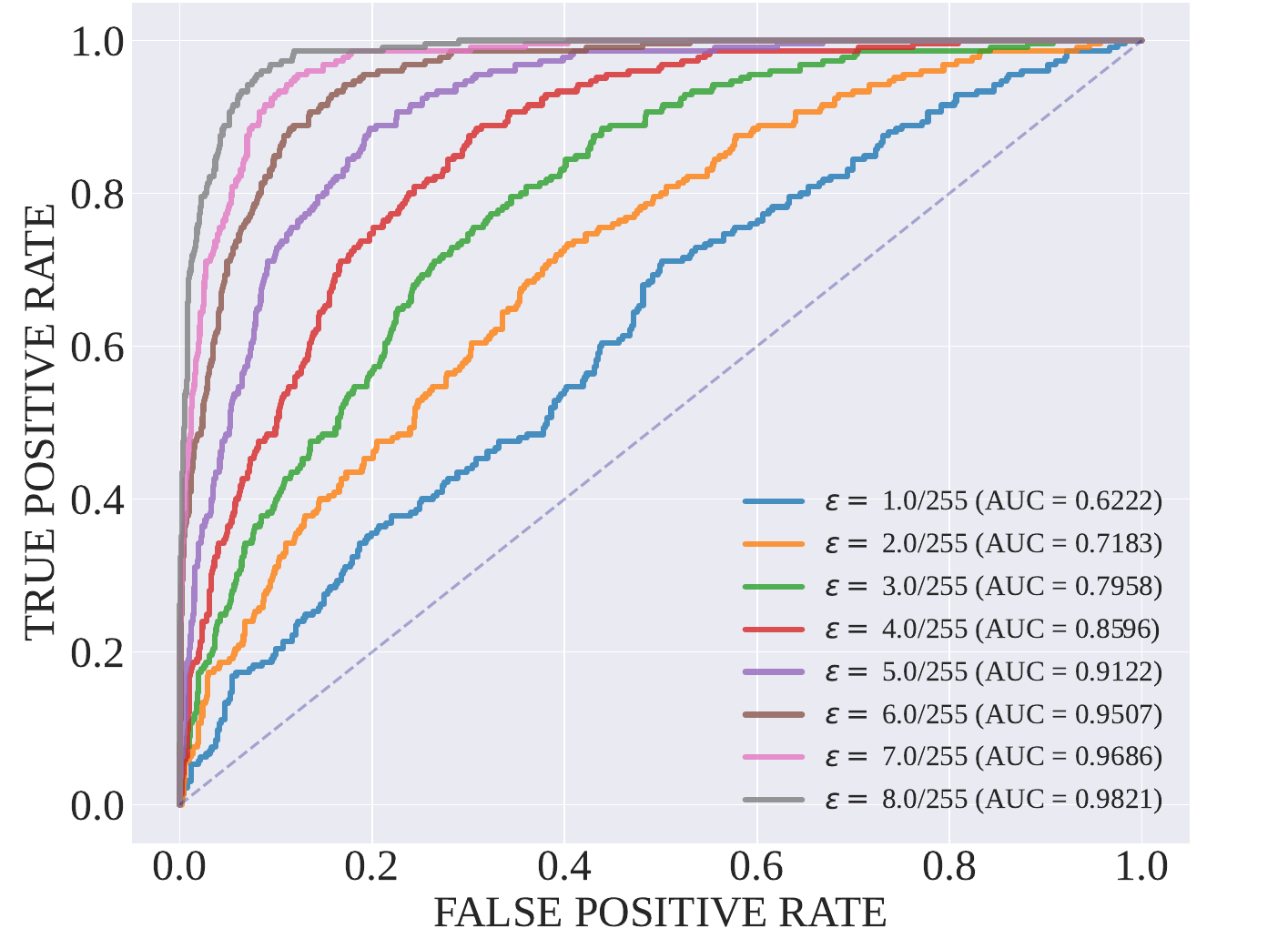}
        \subcaption{Fabricated Members by I-CW}
    \end{minipage}

    \begin{minipage}{0.49\textwidth}
        \centering
        \includegraphics[width=\textwidth]{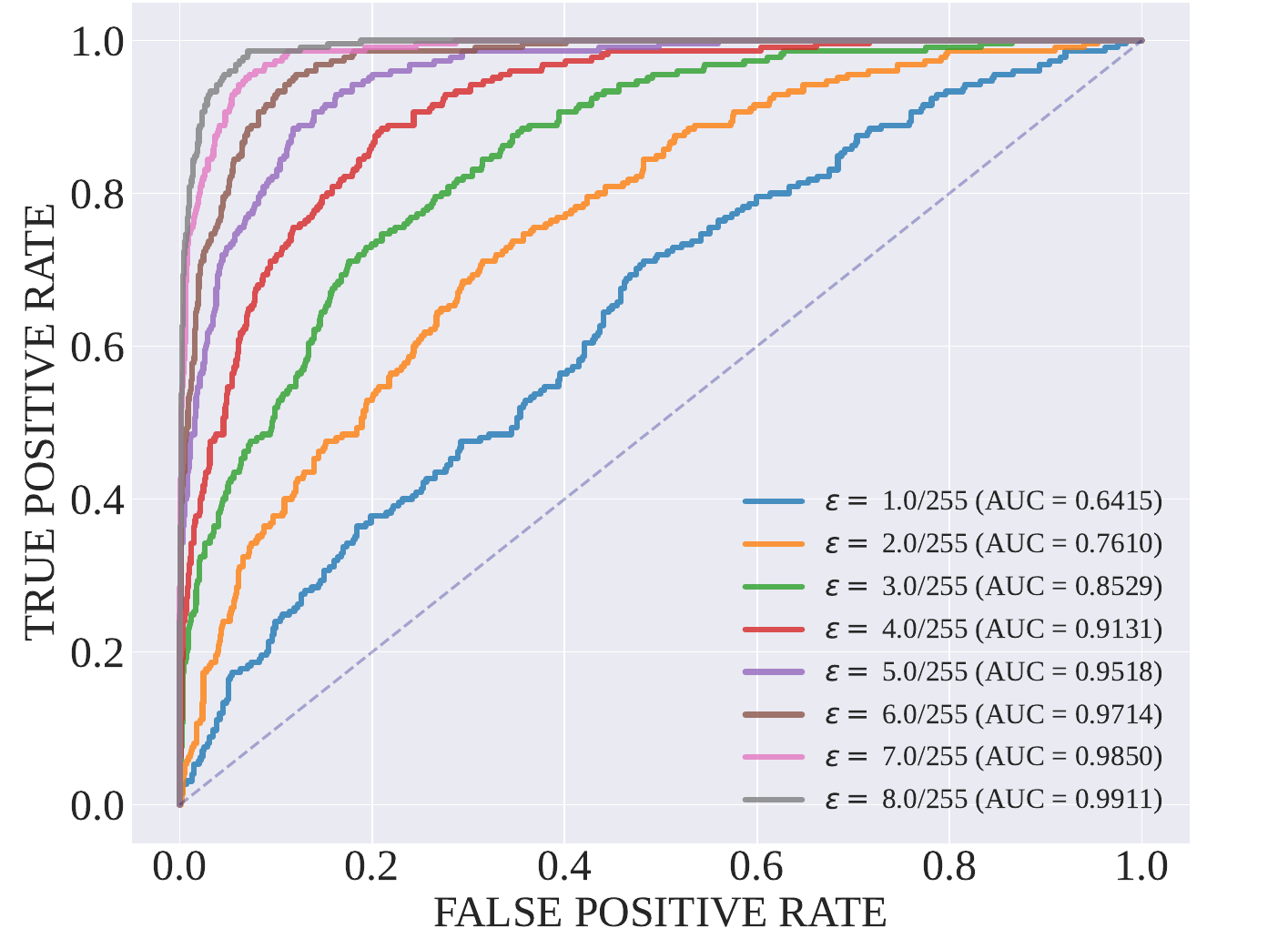}
        \subcaption{Fabricated Members by I-APGD}
    \end{minipage}
    \hfill
    \begin{minipage}{0.49\textwidth}
        \centering
        \includegraphics[width=\textwidth]{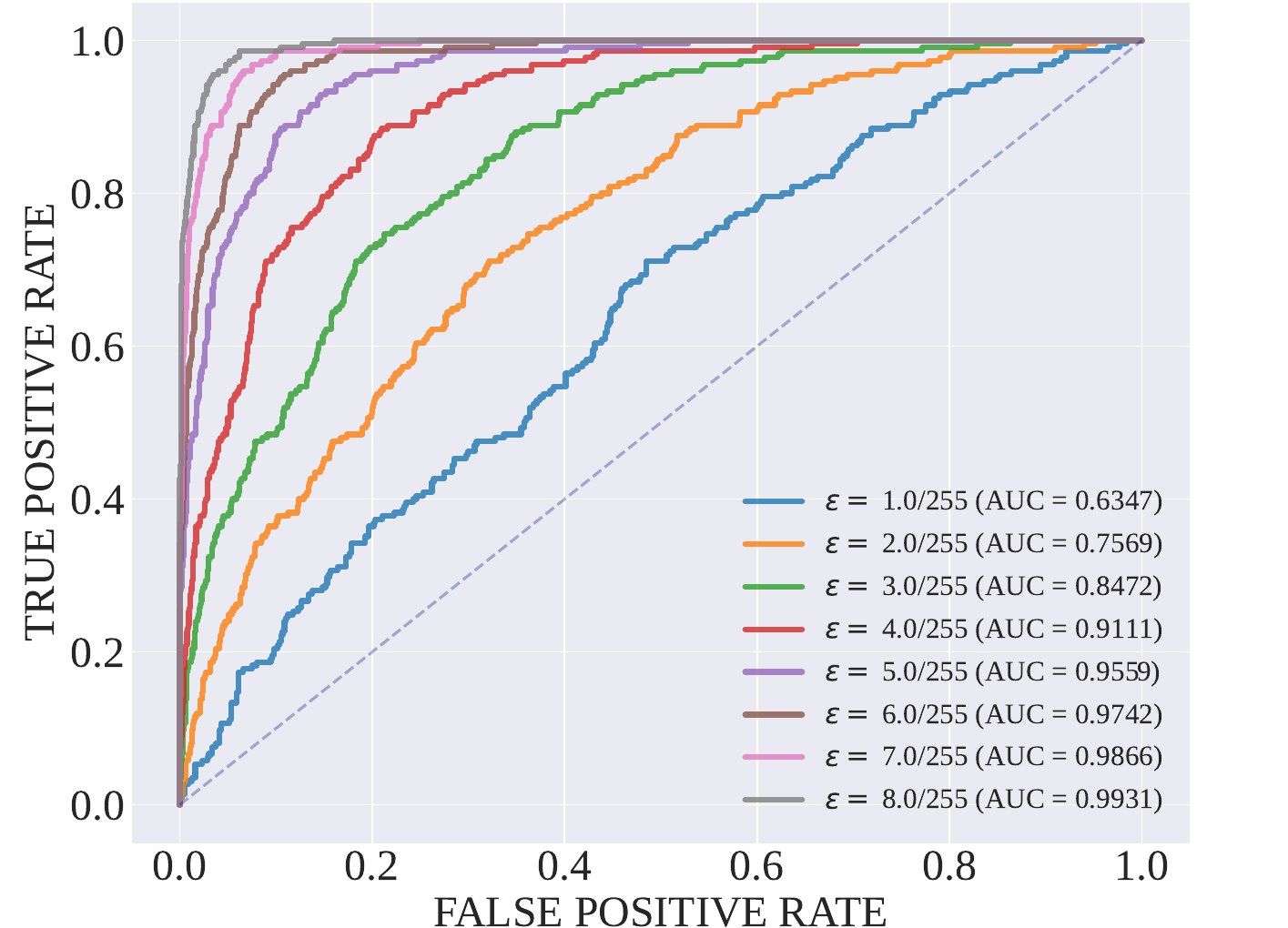}
        \subcaption{Fabricated Members by OURS Attack}
    \end{minipage}
    \caption{\footnotesize  Comparison of the ROC Curve for Our Member Fabrication Detection Across Diverse Perturbation Bounds on \textbf{CIFAR-10}.}
    \label{Detection_figure_cifar10}
    \end{center}
    \vspace{-1em}
\end{figure*}

\begin{figure*}[!t]
    \begin{center}
    \begin{minipage}{0.49\textwidth}
        \centering
        \includegraphics[width=\textwidth]{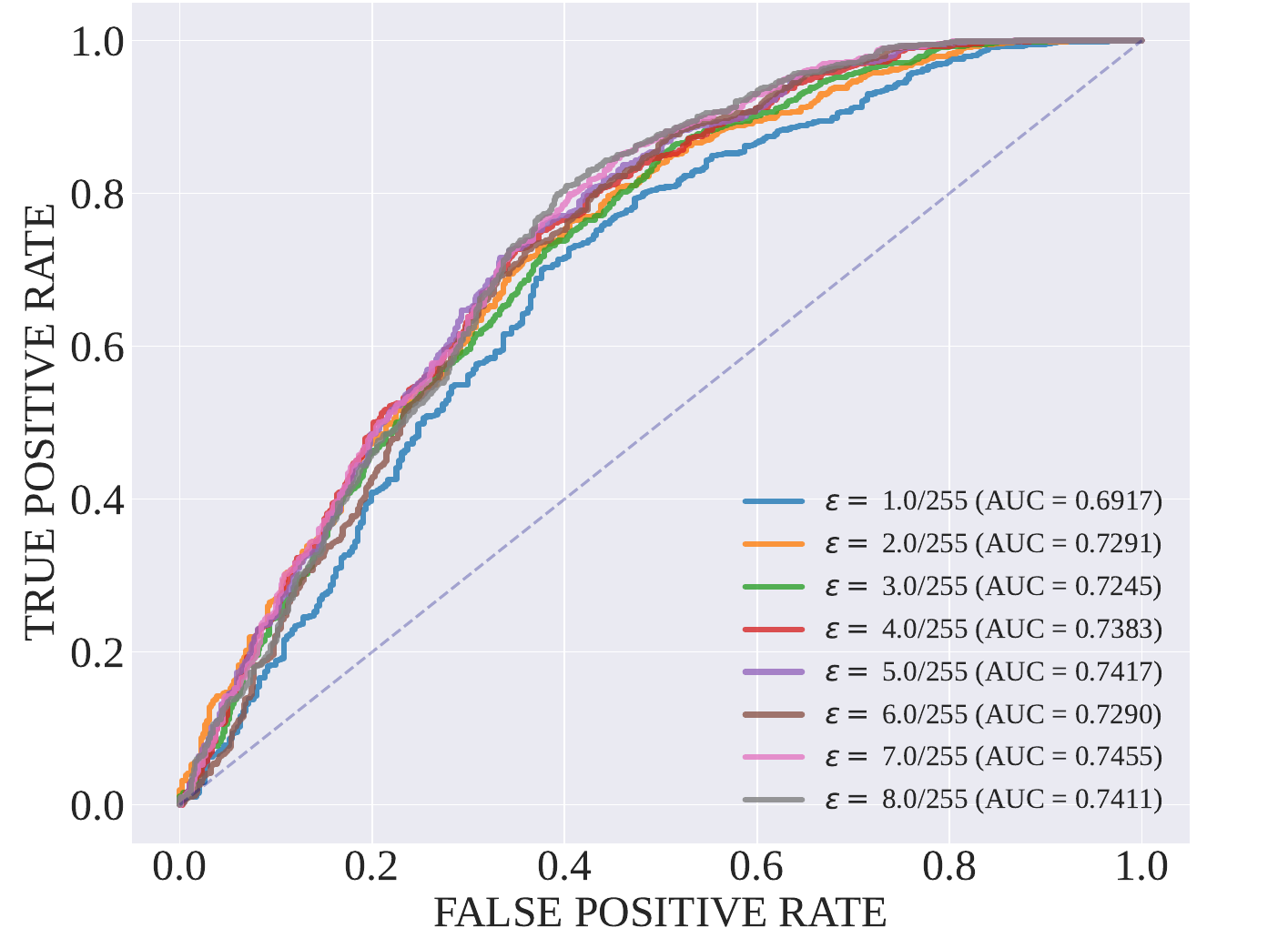}
        \subcaption{Fabricated Members by I-FGSM}
    \end{minipage}
    \hfill
    \begin{minipage}{0.49\textwidth}
        \centering
        \includegraphics[width=\textwidth]{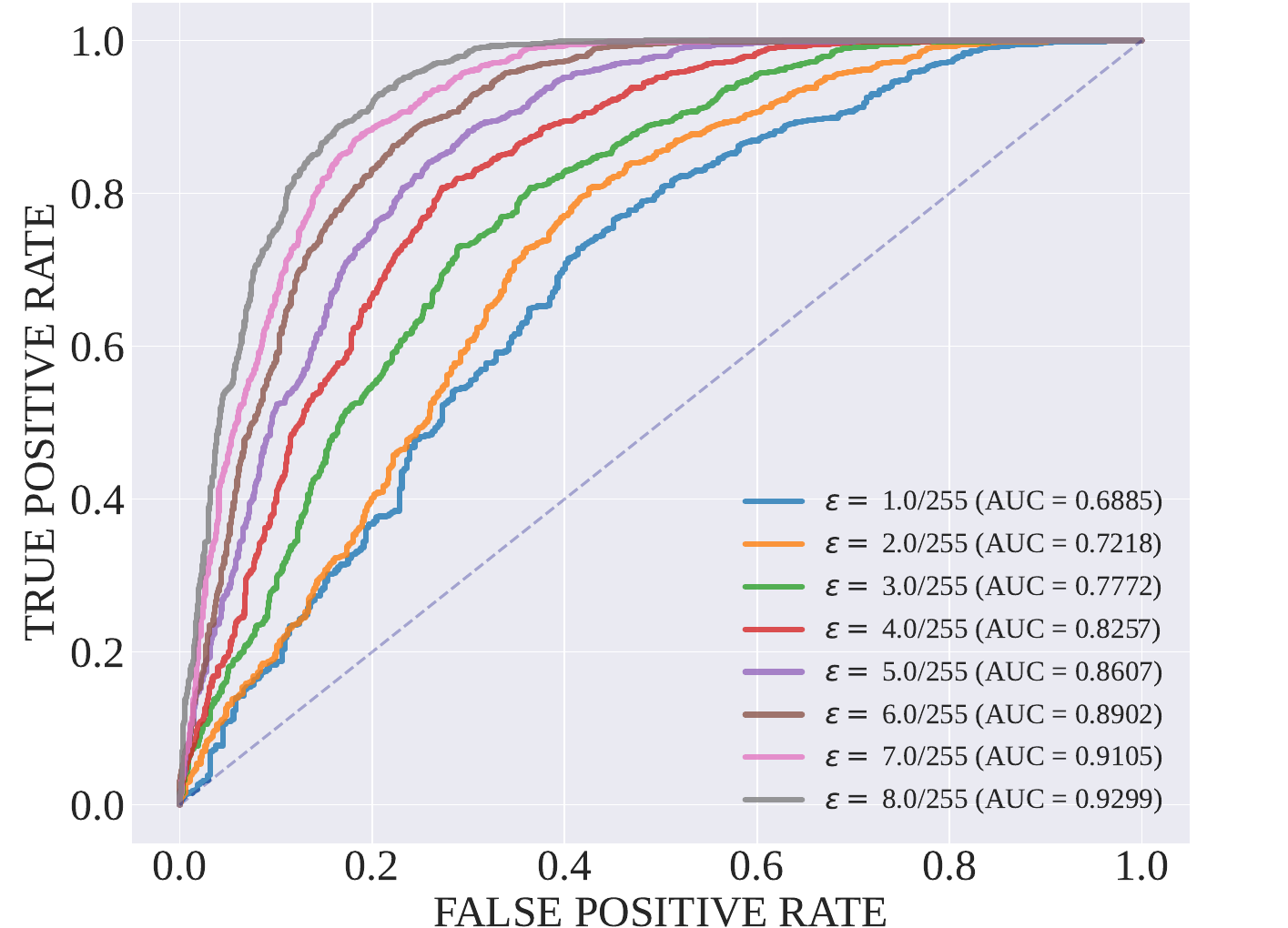}
        \subcaption{Fabricated Members by I-BIM}
    \end{minipage}

    \begin{minipage}{0.49\textwidth}
        \centering
        \includegraphics[width=\textwidth]{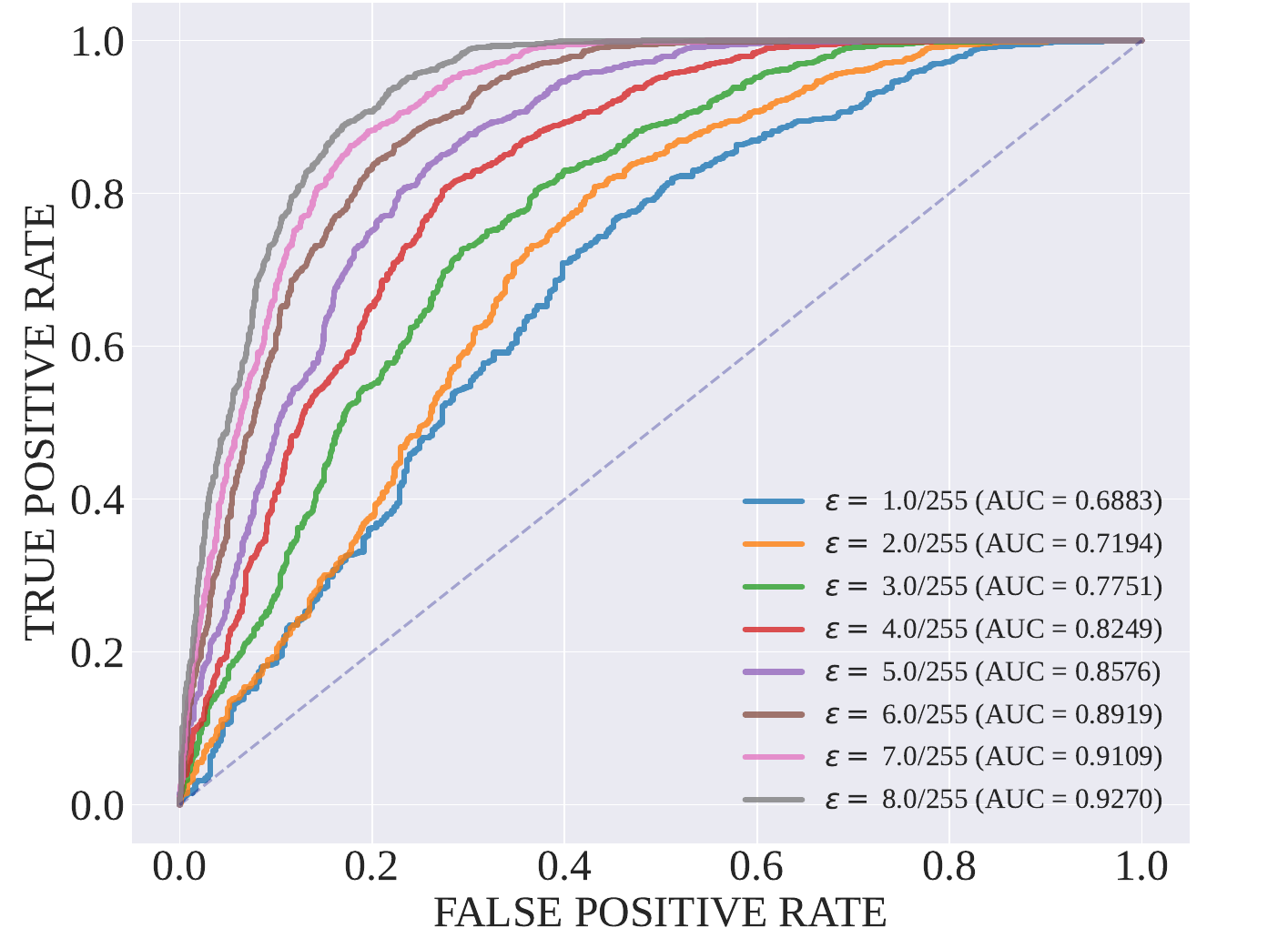}
        \subcaption{Fabricated Members by I-PGD}
    \end{minipage}
    \hfill
    \begin{minipage}{0.49\textwidth}
        \centering
        \includegraphics[width=\textwidth]{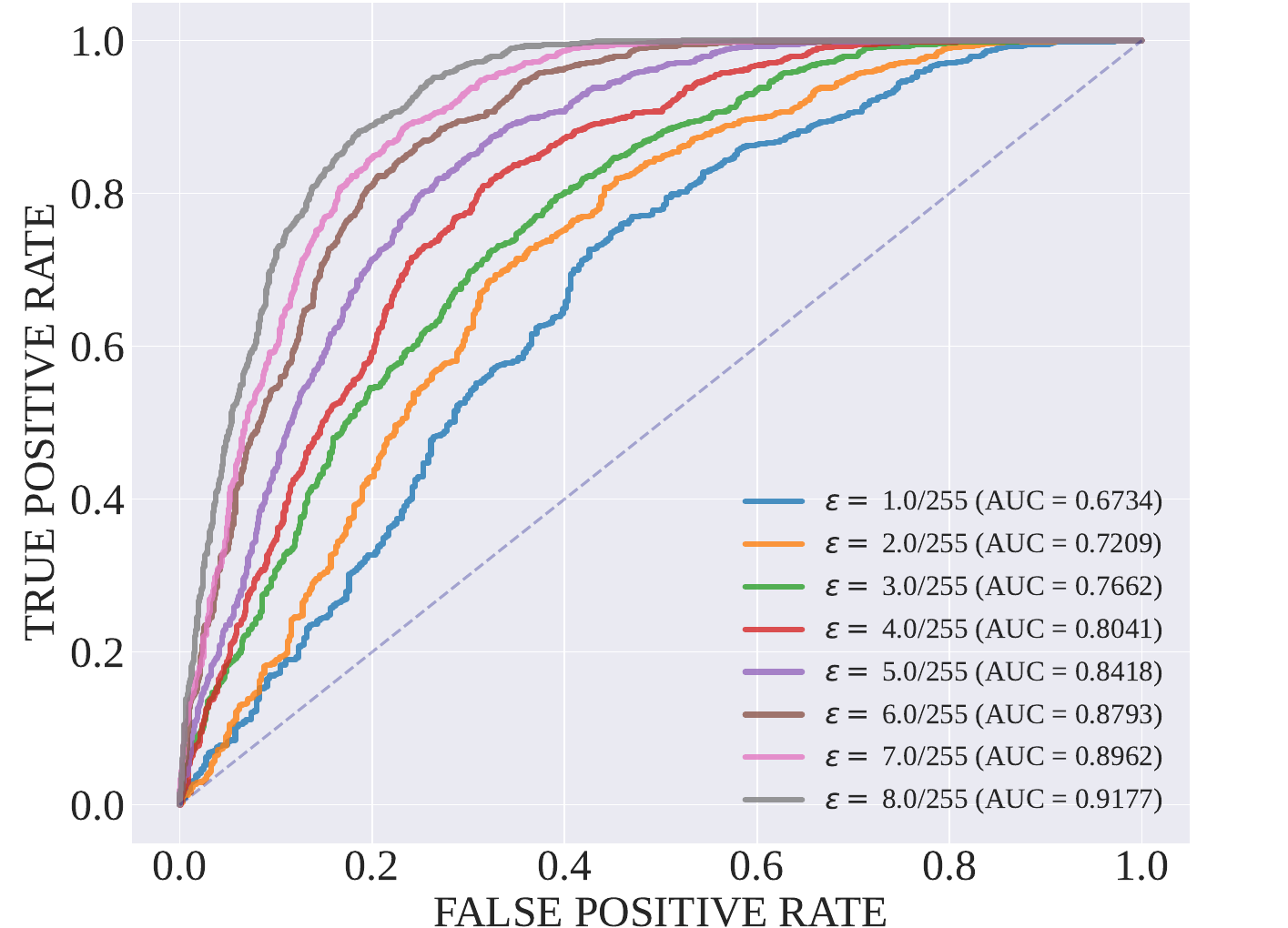}
        \subcaption{Fabricated Members by I-CW}
    \end{minipage}

    \begin{minipage}{0.49\textwidth}
        \centering
        \includegraphics[width=\textwidth]{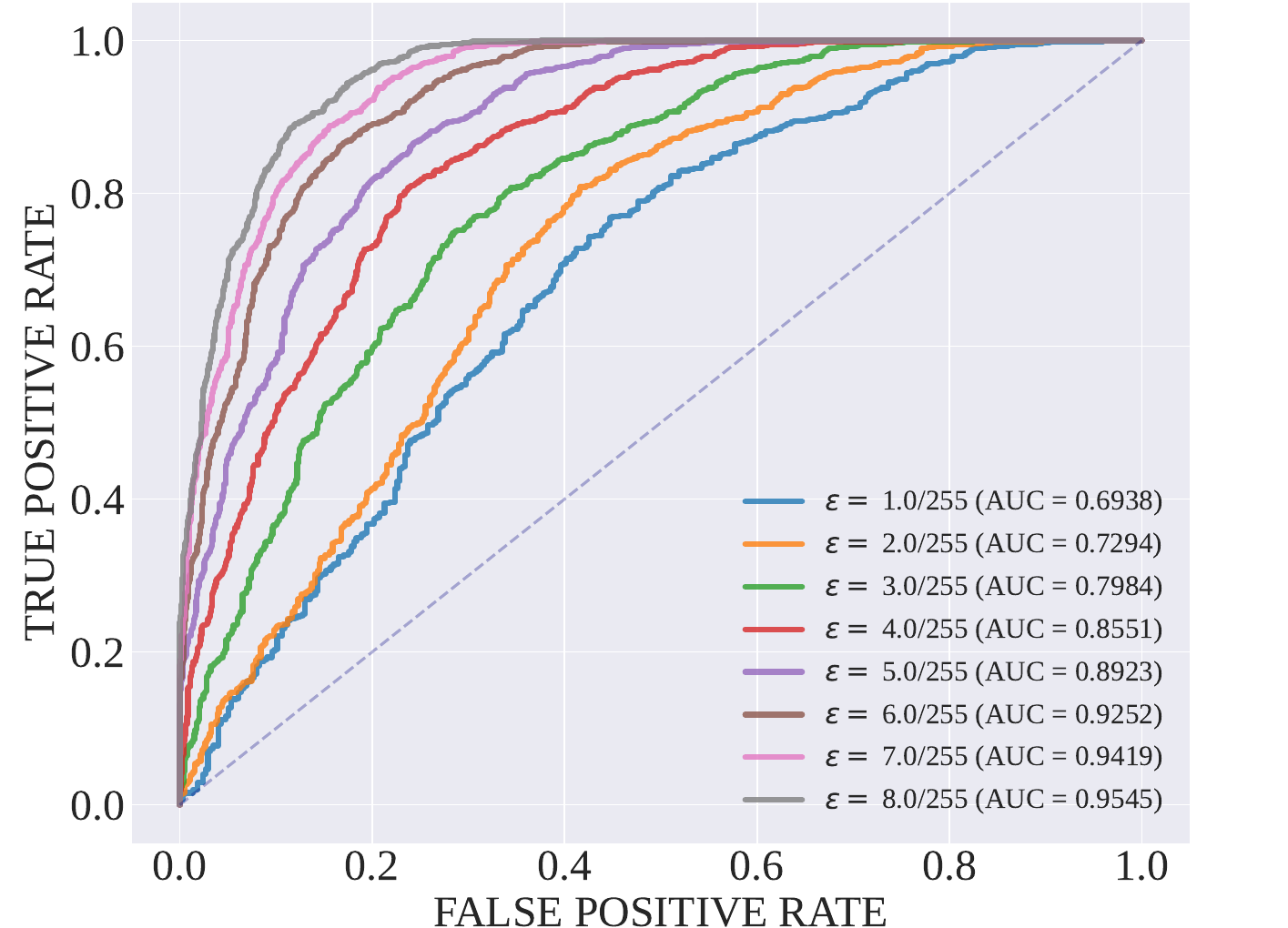}
        \subcaption{Fabricated Members by I-APGD}
    \end{minipage}
    \hfill
    \begin{minipage}{0.49\textwidth}
        \centering
        \includegraphics[width=\textwidth]{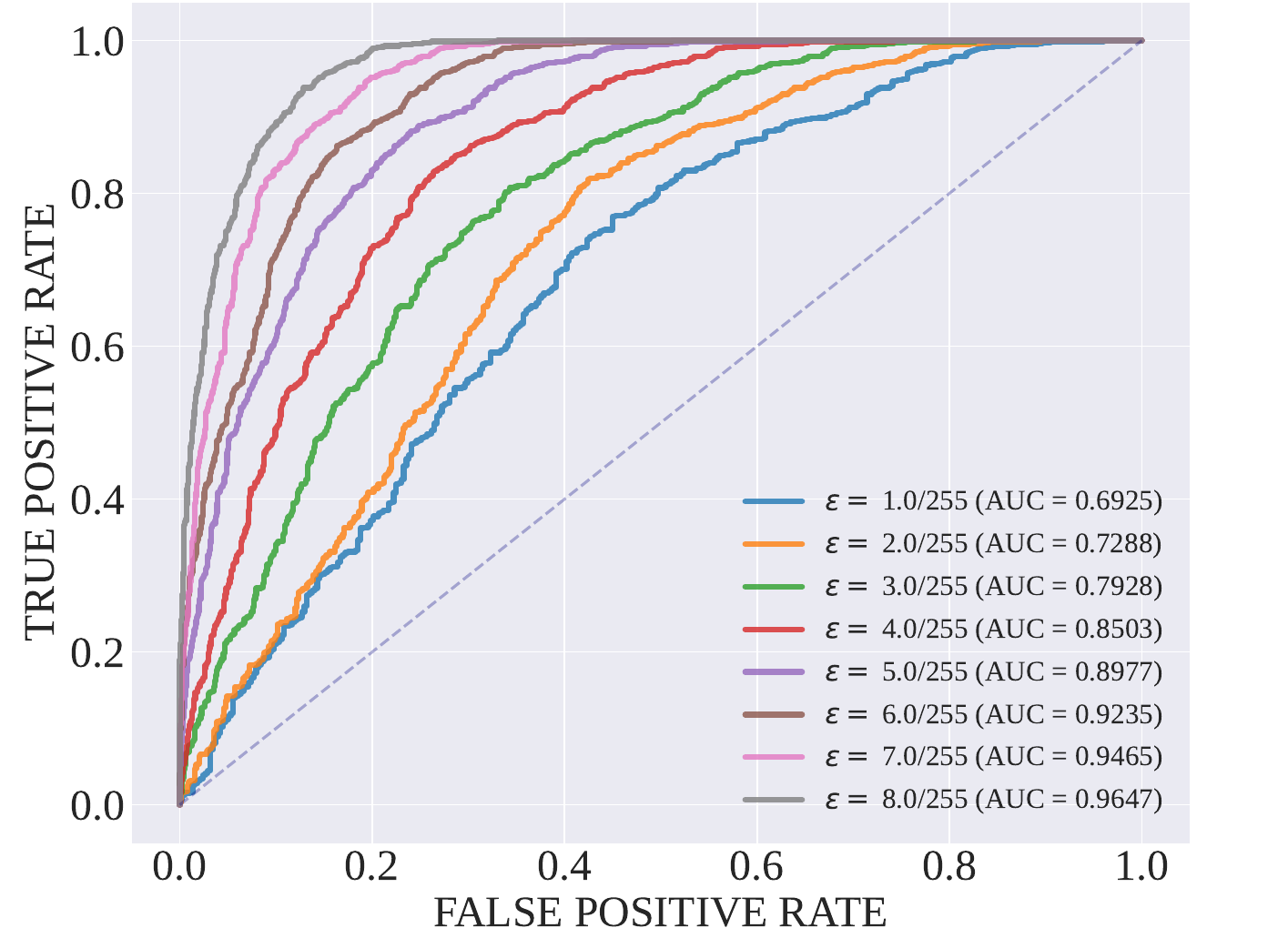}
        \subcaption{Fabricated Members by OURS Attack}
    \end{minipage}
    \caption{\footnotesize  Comparison of the ROC Curve for Our Member Fabrication Detection Across Diverse Perturbation Bounds on \textbf{CIFAR-100}.}
    \label{Detection_figure_cifar100}
    \end{center}
    \vspace{-1em}
\end{figure*}

\begin{figure*}[!t]
    \begin{center}
    \begin{minipage}{0.49\textwidth}
        \centering
        \includegraphics[width=\textwidth]{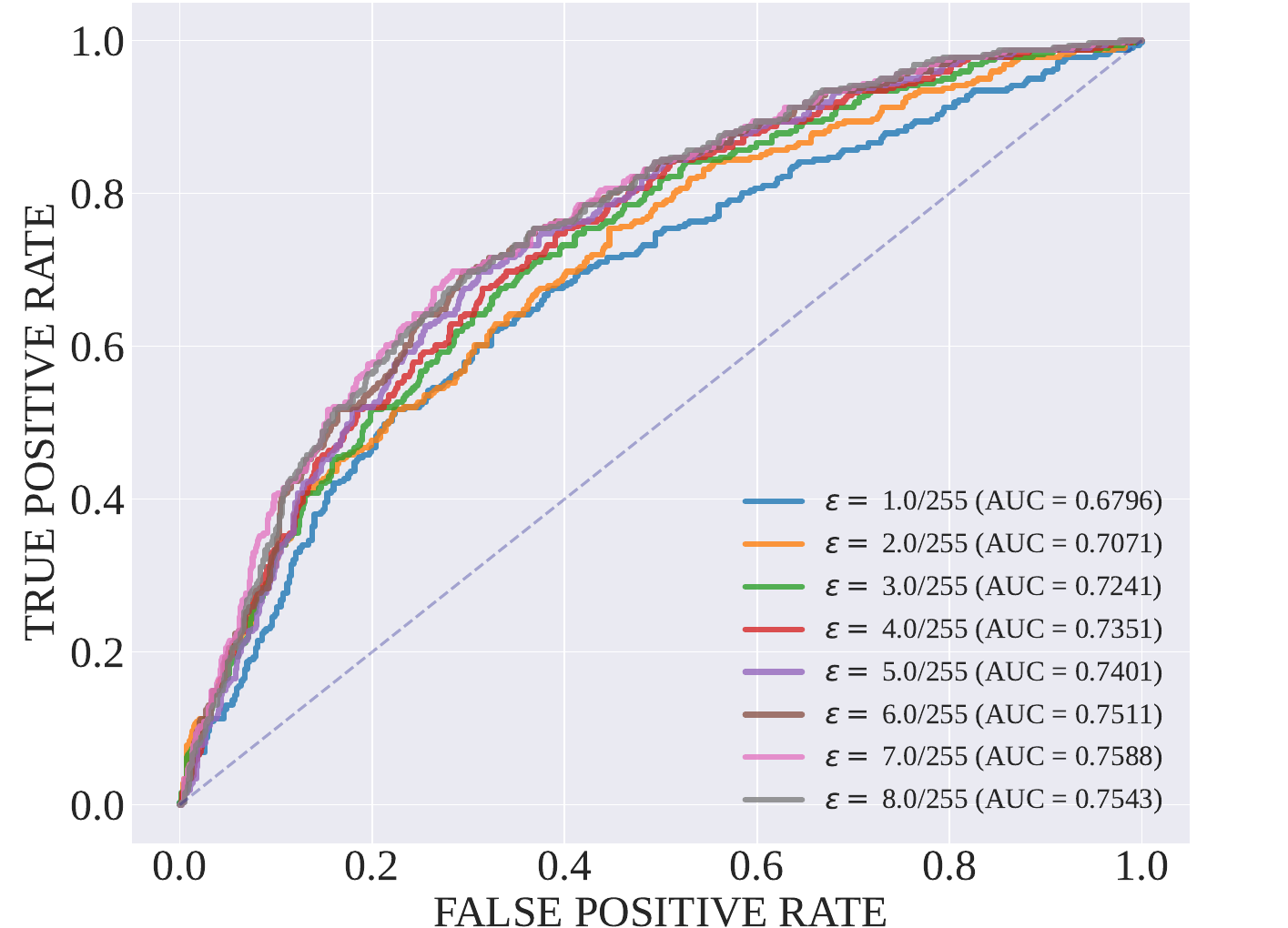}
        \subcaption{Fabricated Members by I-FGSM}
    \end{minipage}
    \hfill
    \begin{minipage}{0.49\textwidth}
        \centering
        \includegraphics[width=\textwidth]{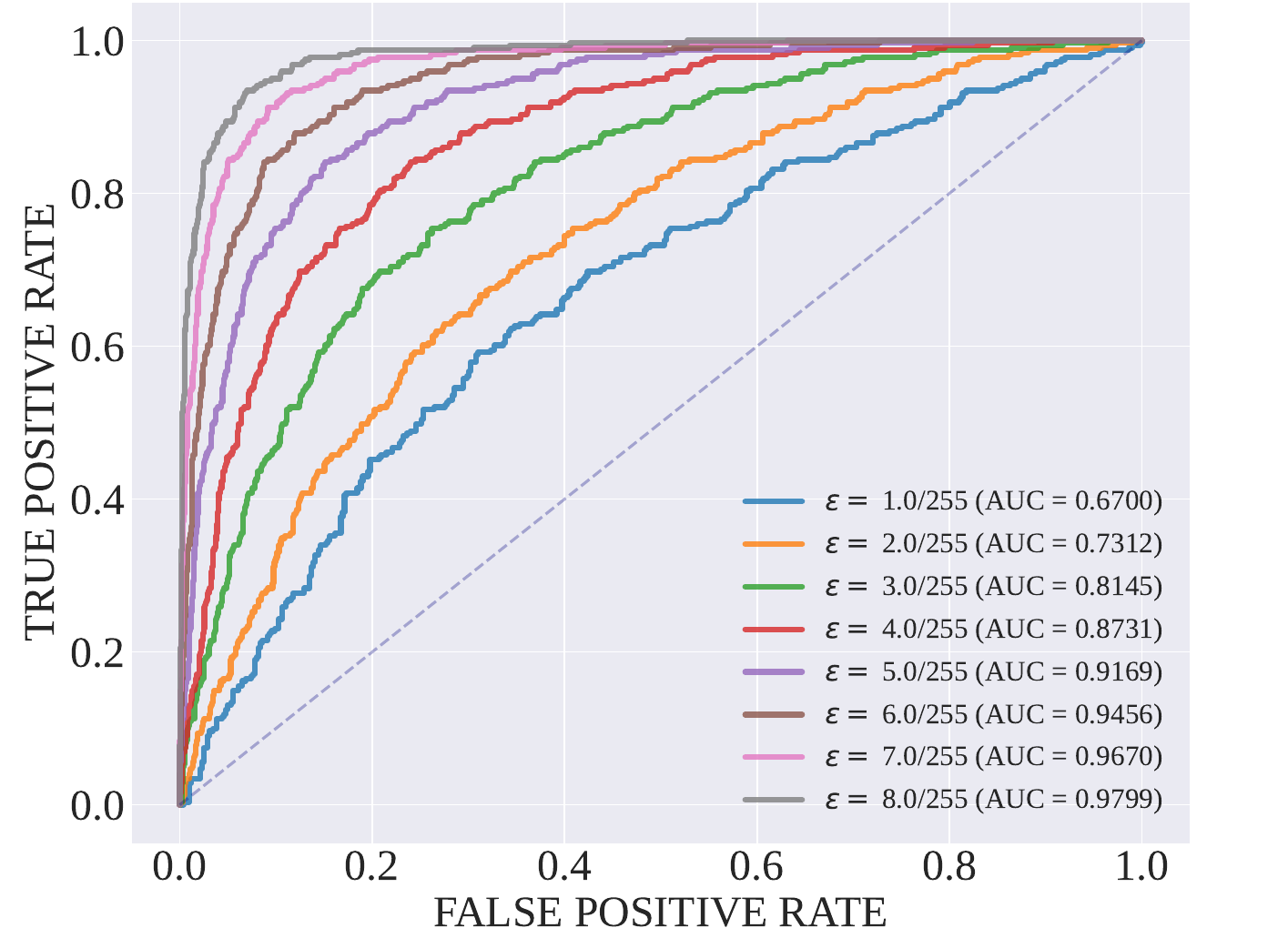}
        \subcaption{Fabricated Members by I-BIM}
    \end{minipage}

    \begin{minipage}{0.49\textwidth}
        \centering
        \includegraphics[width=\textwidth]{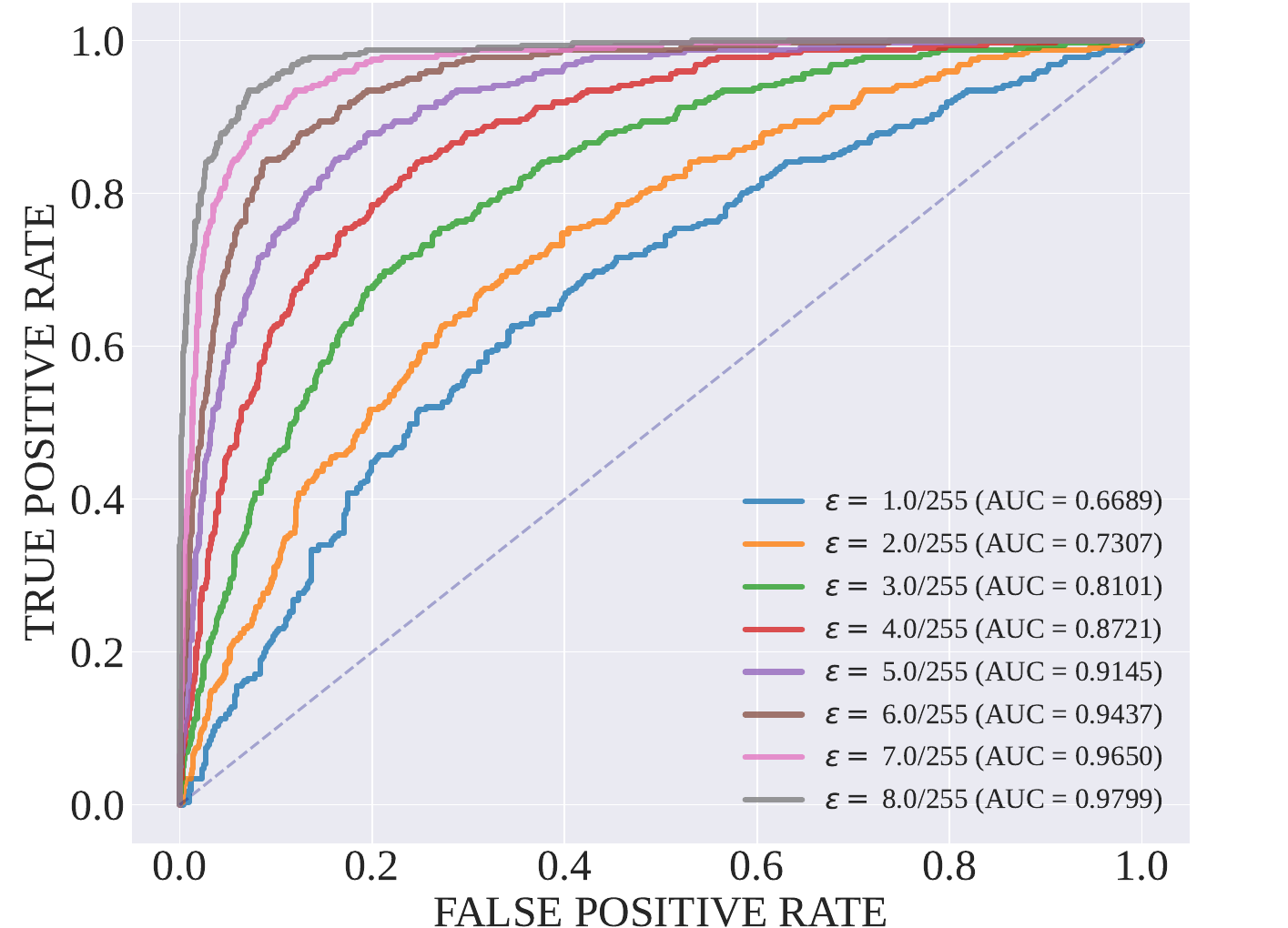}
        \subcaption{Fabricated Members by I-PGD}
    \end{minipage}
    \hfill
    \begin{minipage}{0.49\textwidth}
        \centering
        \includegraphics[width=\textwidth]{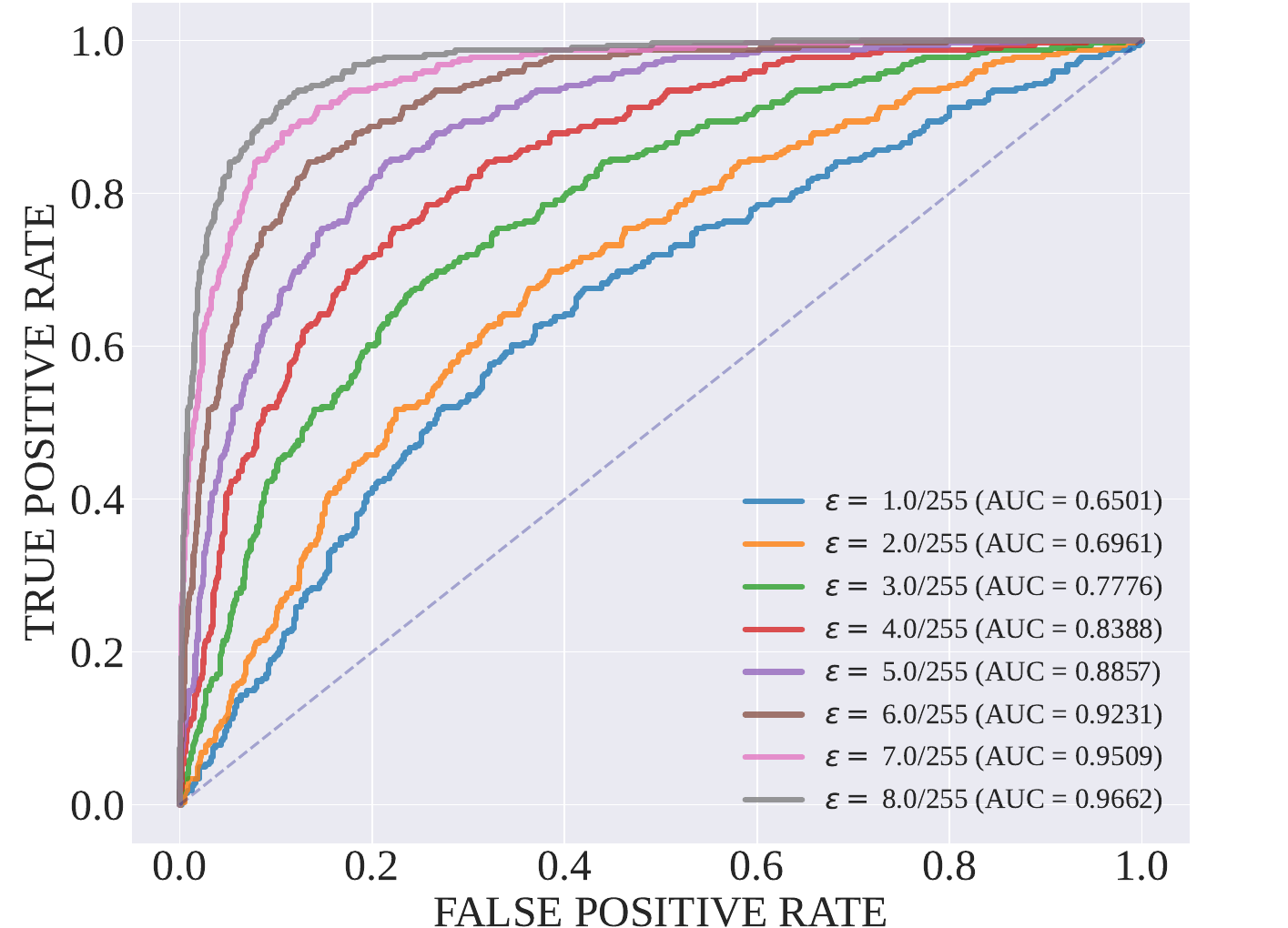}
        \subcaption{Fabricated Members by I-CW}
    \end{minipage}

    \begin{minipage}{0.49\textwidth}
        \centering
        \includegraphics[width=\textwidth]{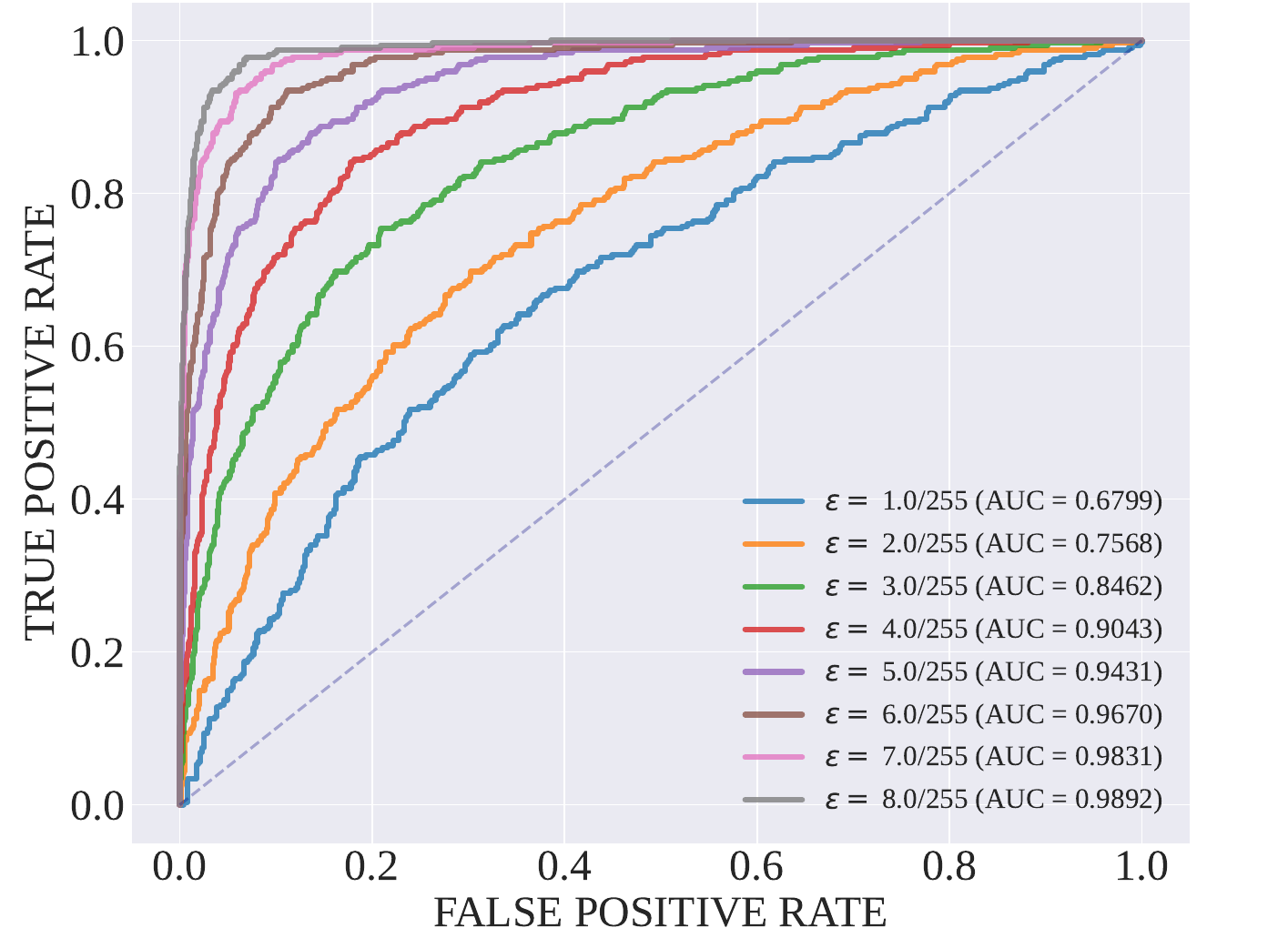}
        \subcaption{Fabricated Members by I-APGD}
    \end{minipage}
    \hfill
    \begin{minipage}{0.49\textwidth}
        \centering
        \includegraphics[width=\textwidth]{crop_Figure/Detection_figure/crop_roc_curve_cinic_stable_adaptive_pgd.pdf}
        \subcaption{Fabricated Members by OURS Attack}
    \end{minipage}
    \caption{\footnotesize  Comparison of the ROC Curve for Our Member Fabrication Detection Across Diverse Perturbation Bounds on \textbf{CINIC-10}.}
    \label{Detection_figure_cinic}
    \end{center}
    \vspace{-1em}
\end{figure*}

\begin{figure*}[!t]
    \begin{center}
    \begin{minipage}{0.49\textwidth}
        \centering
        \includegraphics[width=\textwidth]{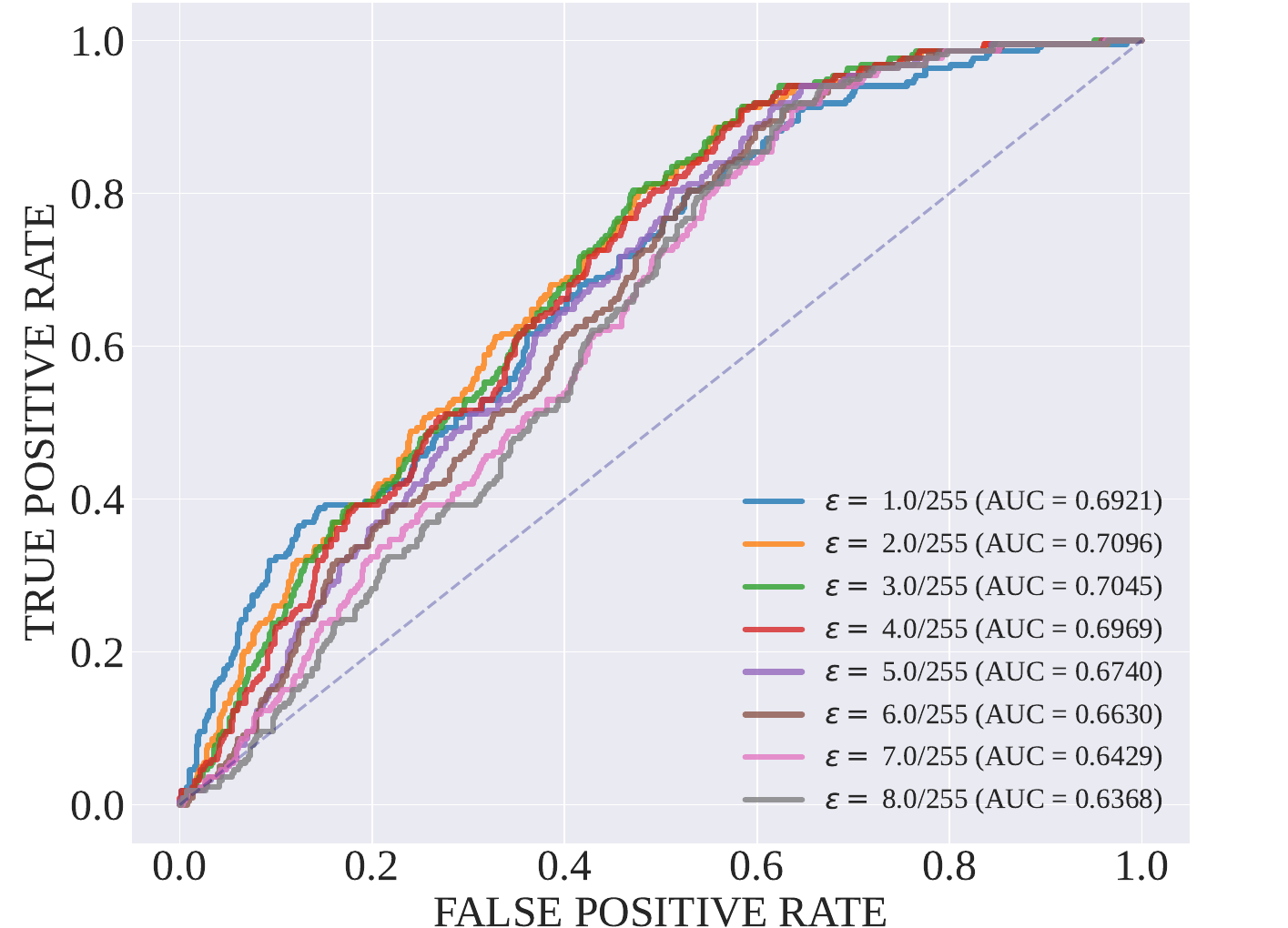}
        \subcaption{Fabricated Members by I-FGSM}
    \end{minipage}
    \hfill
    \begin{minipage}{0.49\textwidth}
        \centering
        \includegraphics[width=\textwidth]{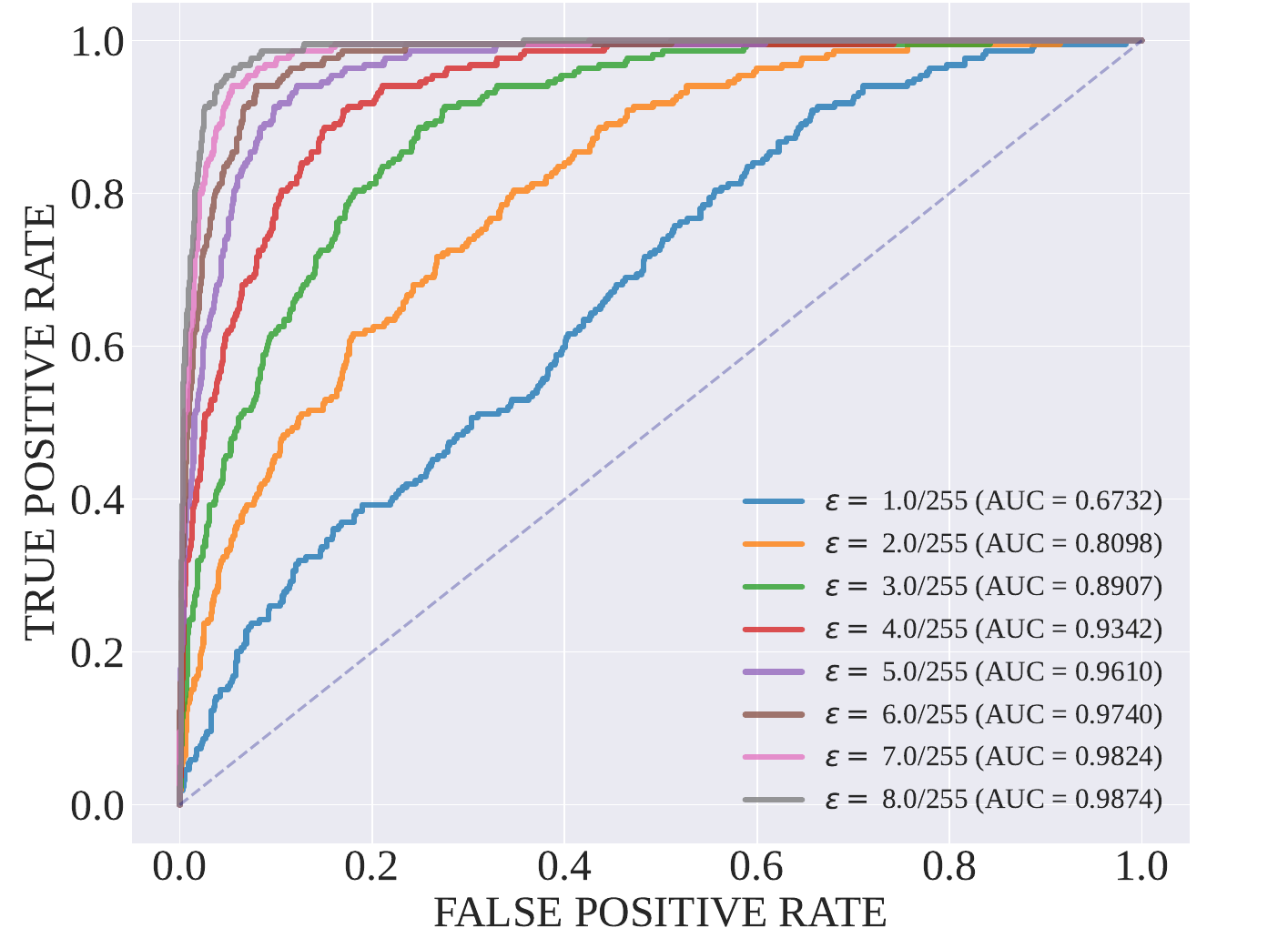}
        \subcaption{Fabricated Members by I-BIM}
    \end{minipage}

    \begin{minipage}{0.49\textwidth}
        \centering
        \includegraphics[width=\textwidth]{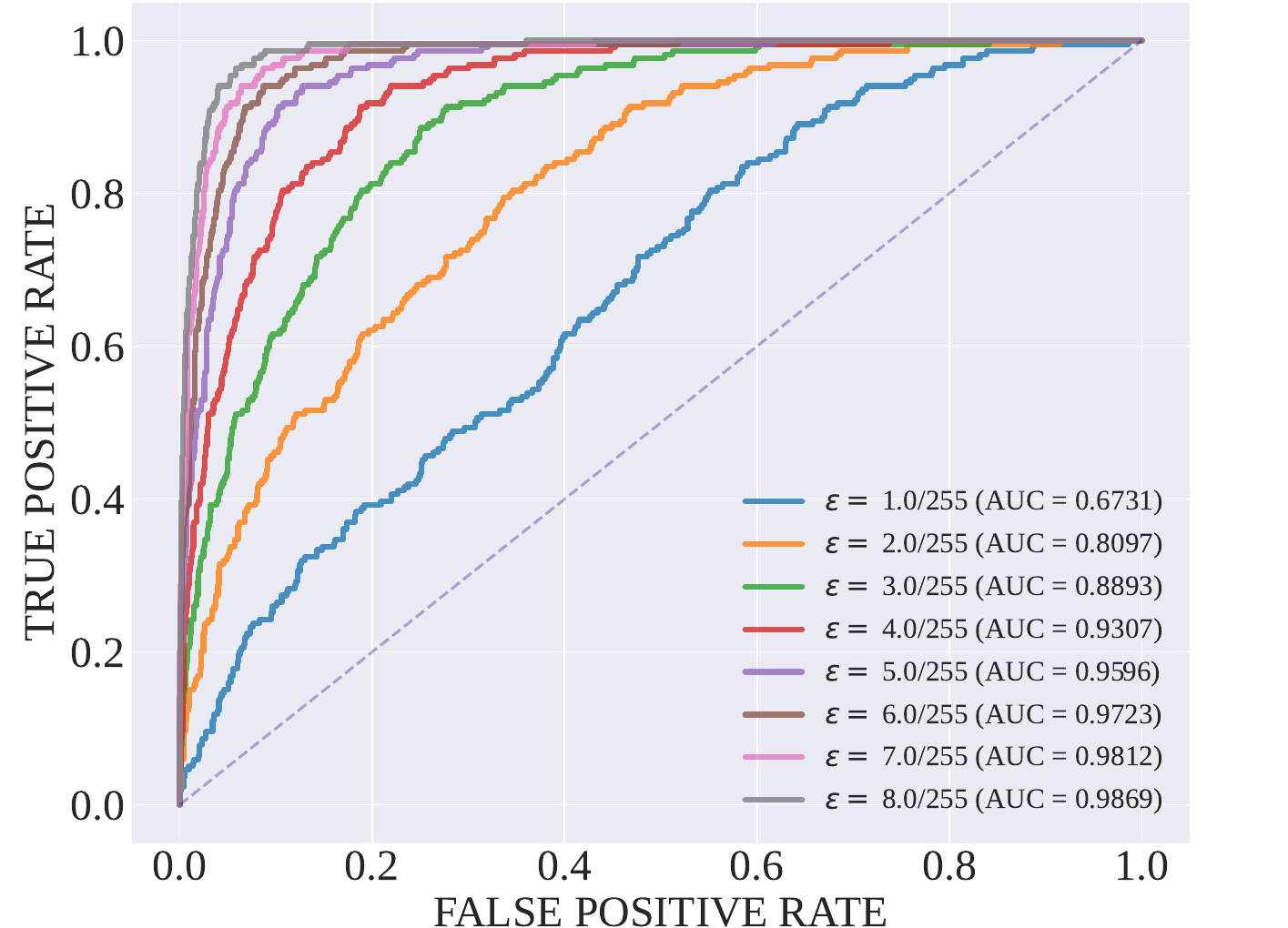}
        \subcaption{Fabricated Members by I-PGD}
    \end{minipage}
    \hfill
    \begin{minipage}{0.49\textwidth}
        \centering
        \includegraphics[width=\textwidth]{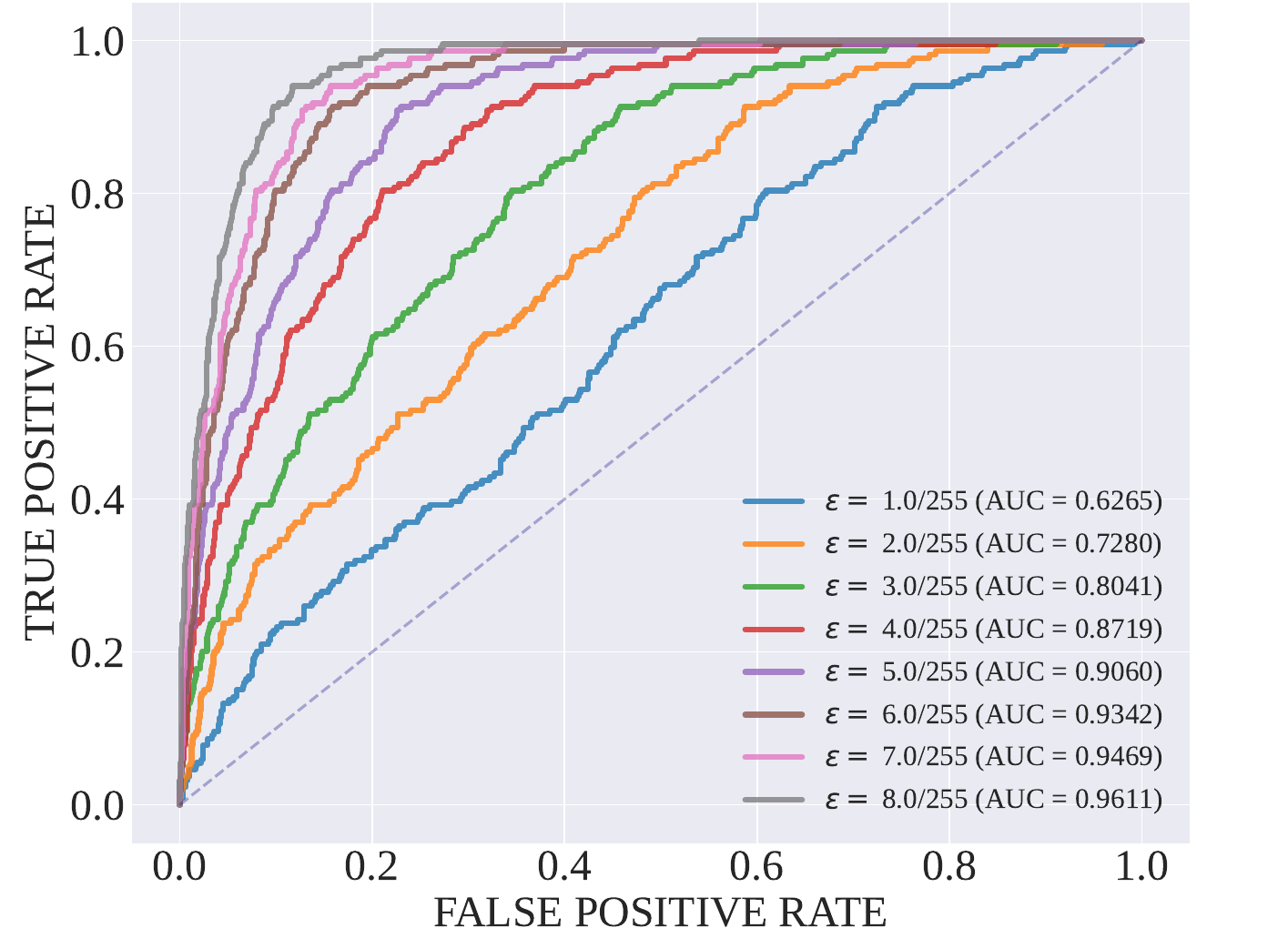}
        \subcaption{Fabricated Members by I-CW}
    \end{minipage}

    \begin{minipage}{0.49\textwidth}
        \centering
        \includegraphics[width=\textwidth]{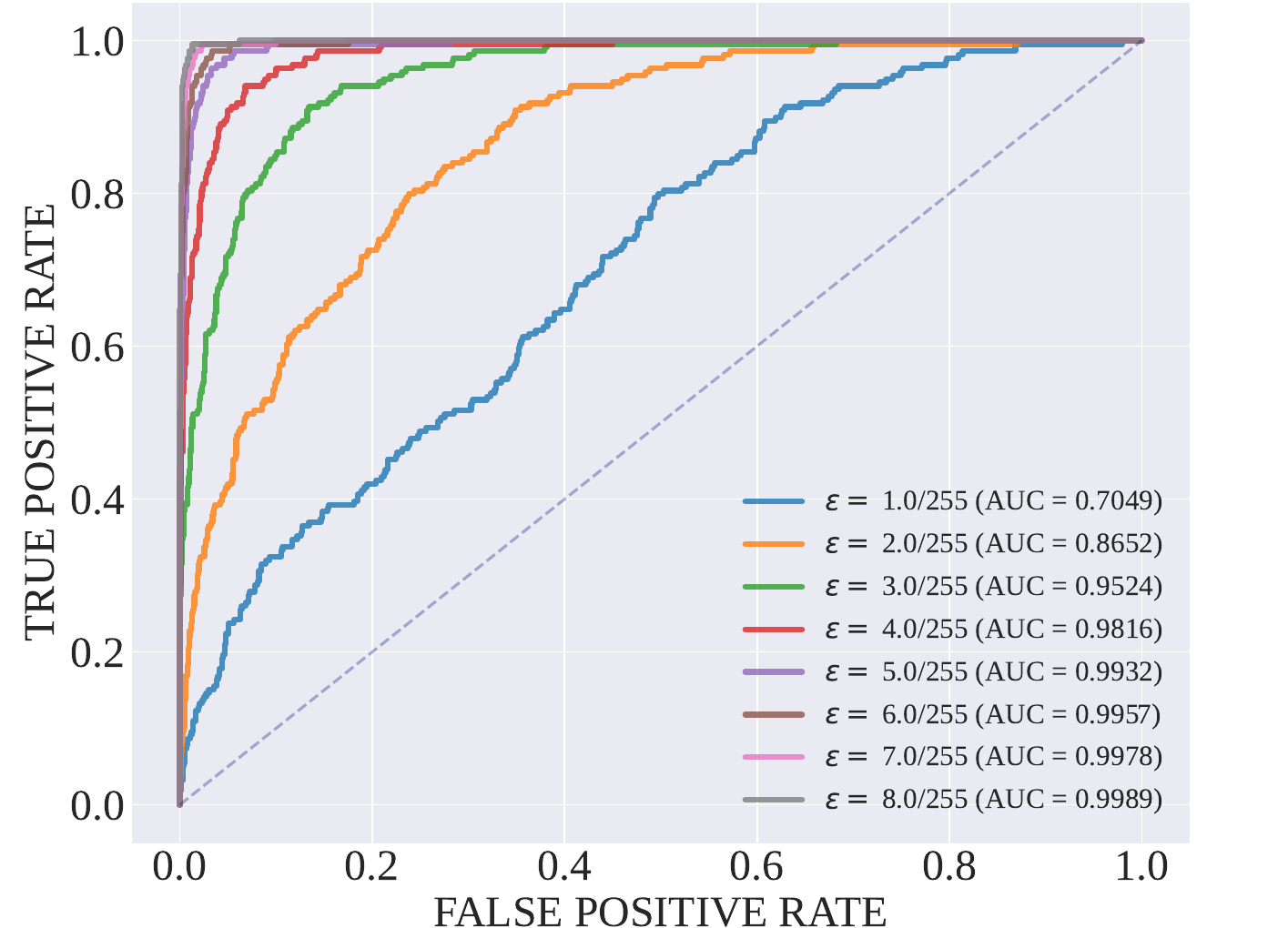}
        \subcaption{Fabricated Members by I-APGD}
    \end{minipage}
    \hfill
    \begin{minipage}{0.49\textwidth}
        \centering
        \includegraphics[width=\textwidth]{crop_Figure/Detection_figure/crop_roc_curve_svhn_stable_adaptive_pgd.pdf}
        \subcaption{Fabricated Members by OURS Attack}
    \end{minipage}
    \caption{\footnotesize  Comparison of the ROC Curve for Our Member Fabrication Detection Across Diverse Perturbation Bounds on \textbf{SVHN}.}
    \label{Detection_figure_svhn}
    \end{center}
    \vspace{-1em}
\end{figure*}

\begin{figure*}[!t]
    \begin{center}
    \begin{minipage}{0.49\textwidth}
        \centering
        \includegraphics[width=\textwidth]{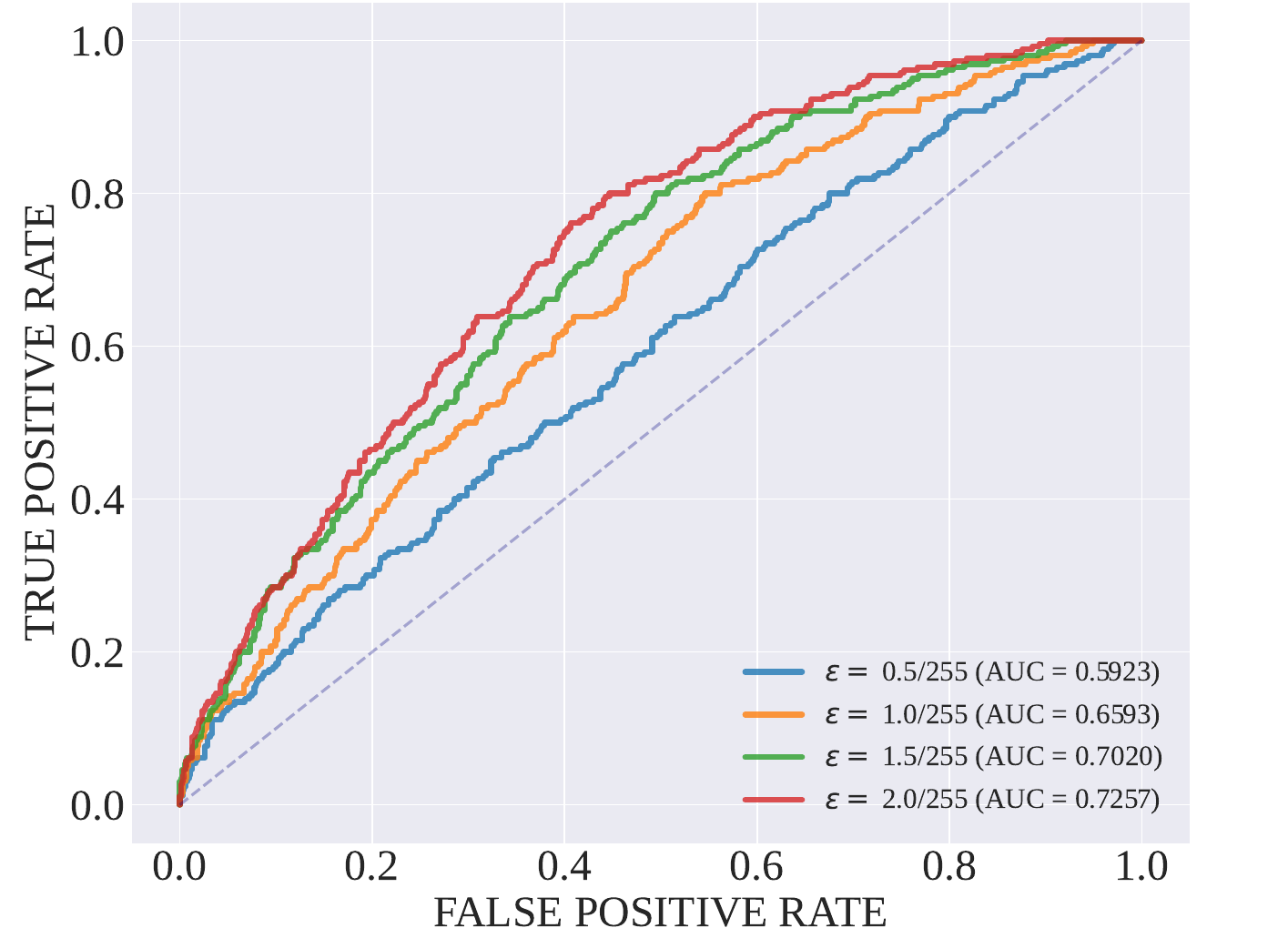}
        \subcaption{Fabricated Members by I-FGSM}
    \end{minipage}
    \hfill
    \begin{minipage}{0.49\textwidth}
        \centering
        \includegraphics[width=\textwidth]{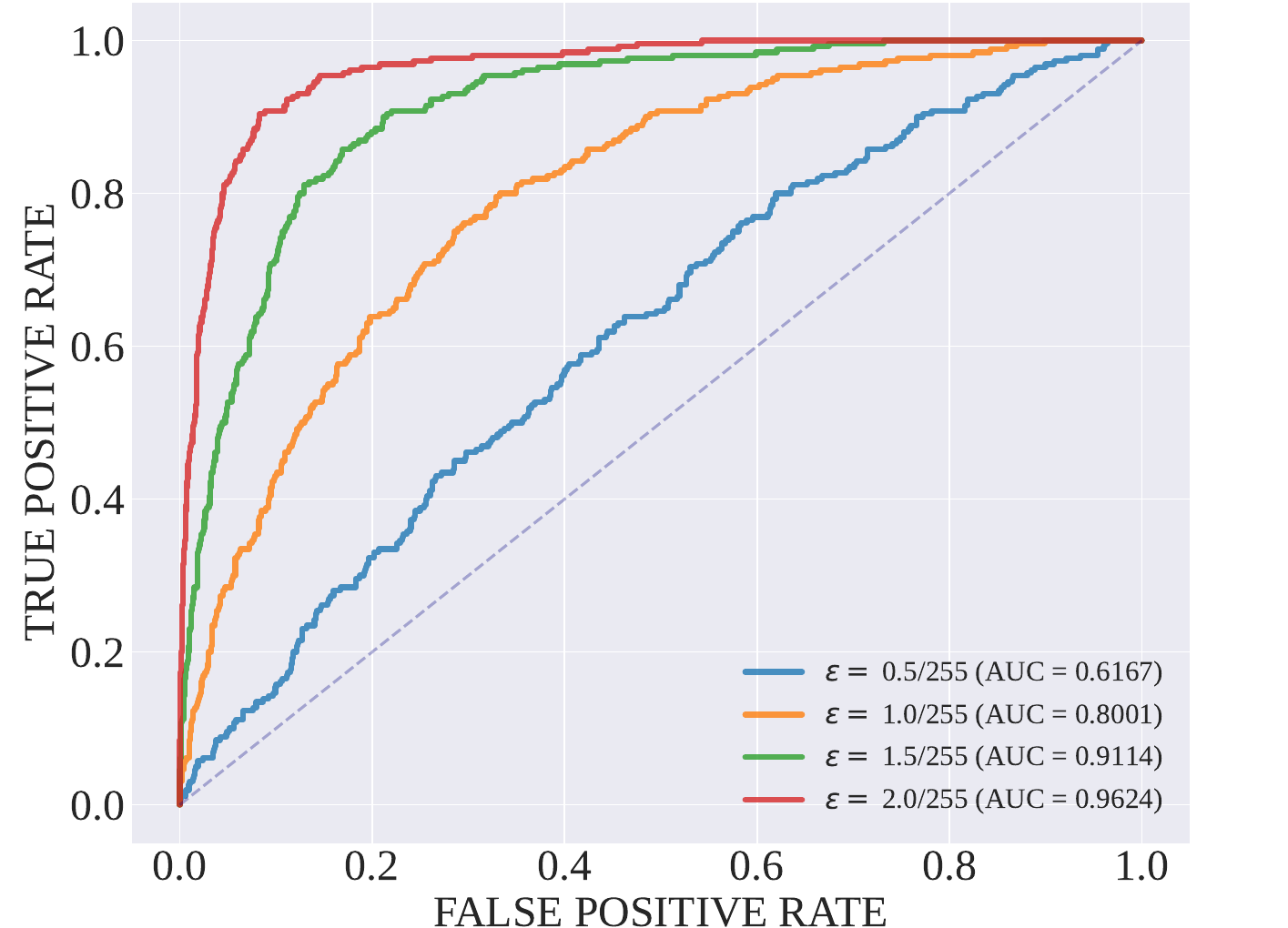}
        \subcaption{Fabricated Members by I-BIM}
    \end{minipage}

    \begin{minipage}{0.49\textwidth}
        \centering
        \includegraphics[width=\textwidth]{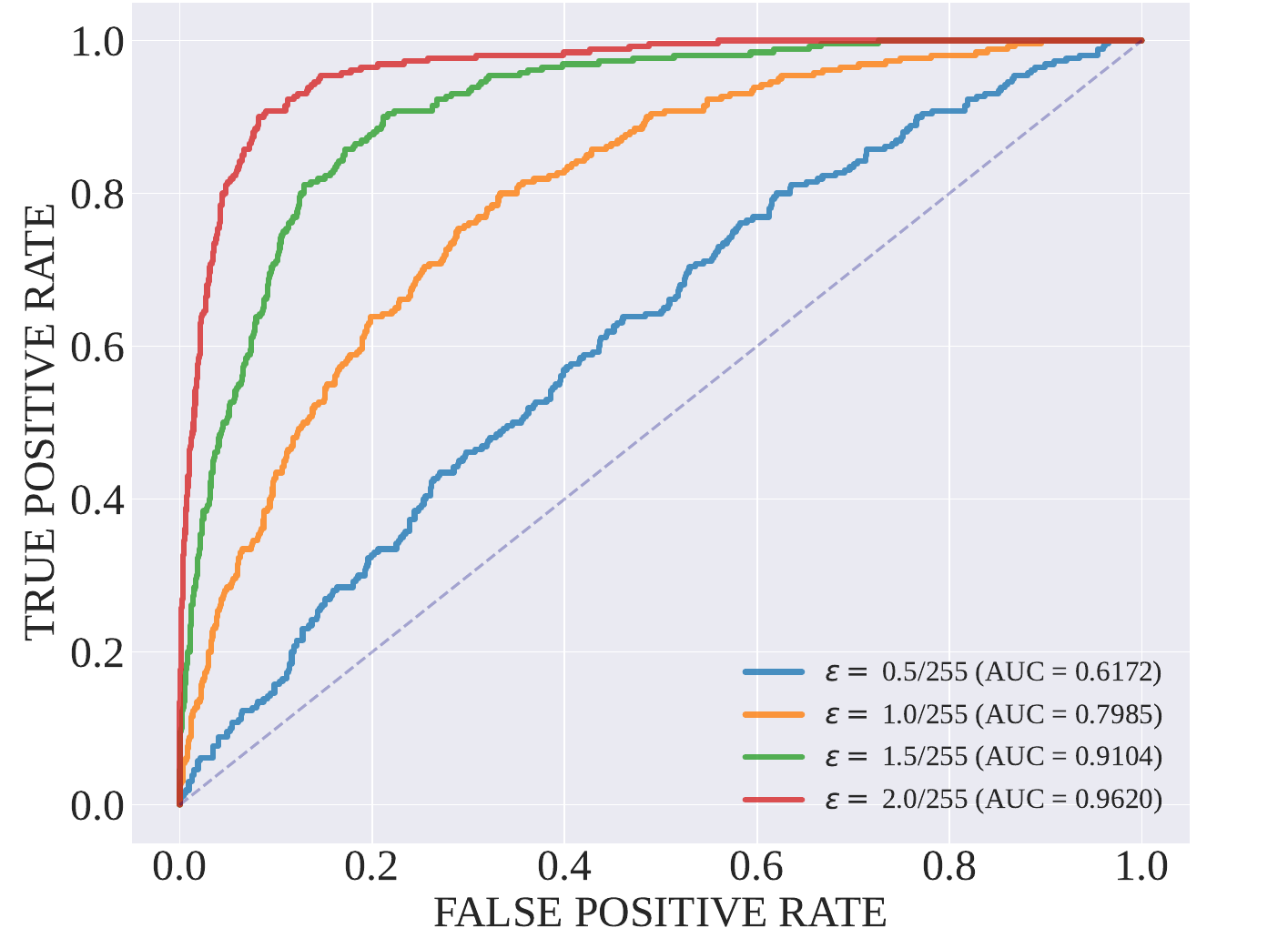}
        \subcaption{Fabricated Members by I-PGD}
    \end{minipage}
    \hfill
    \begin{minipage}{0.49\textwidth}
        \centering
        \includegraphics[width=\textwidth]{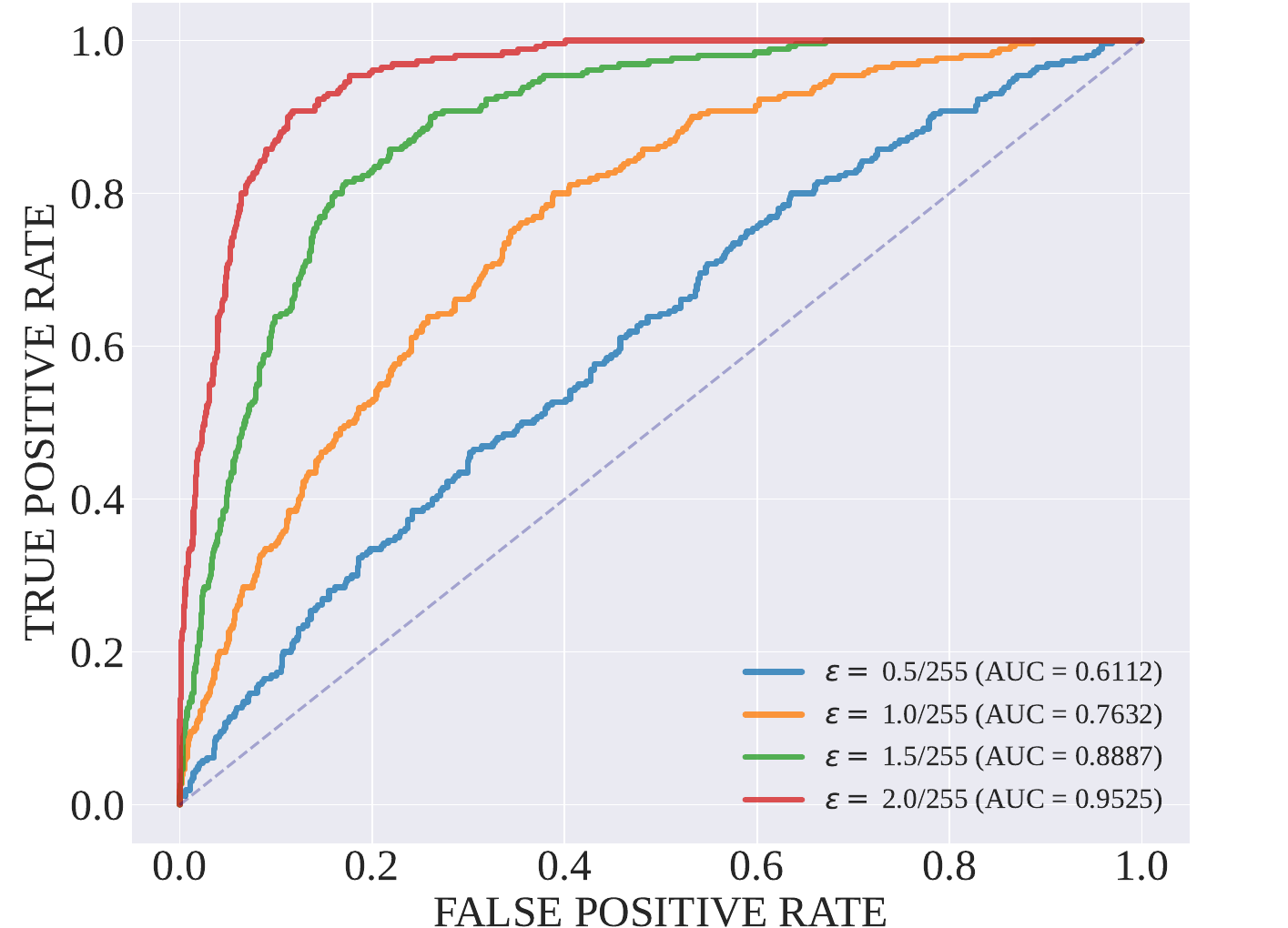}
        \subcaption{Fabricated Members by I-CW}
    \end{minipage}

    \begin{minipage}{0.49\textwidth}
        \centering
        \includegraphics[width=\textwidth]{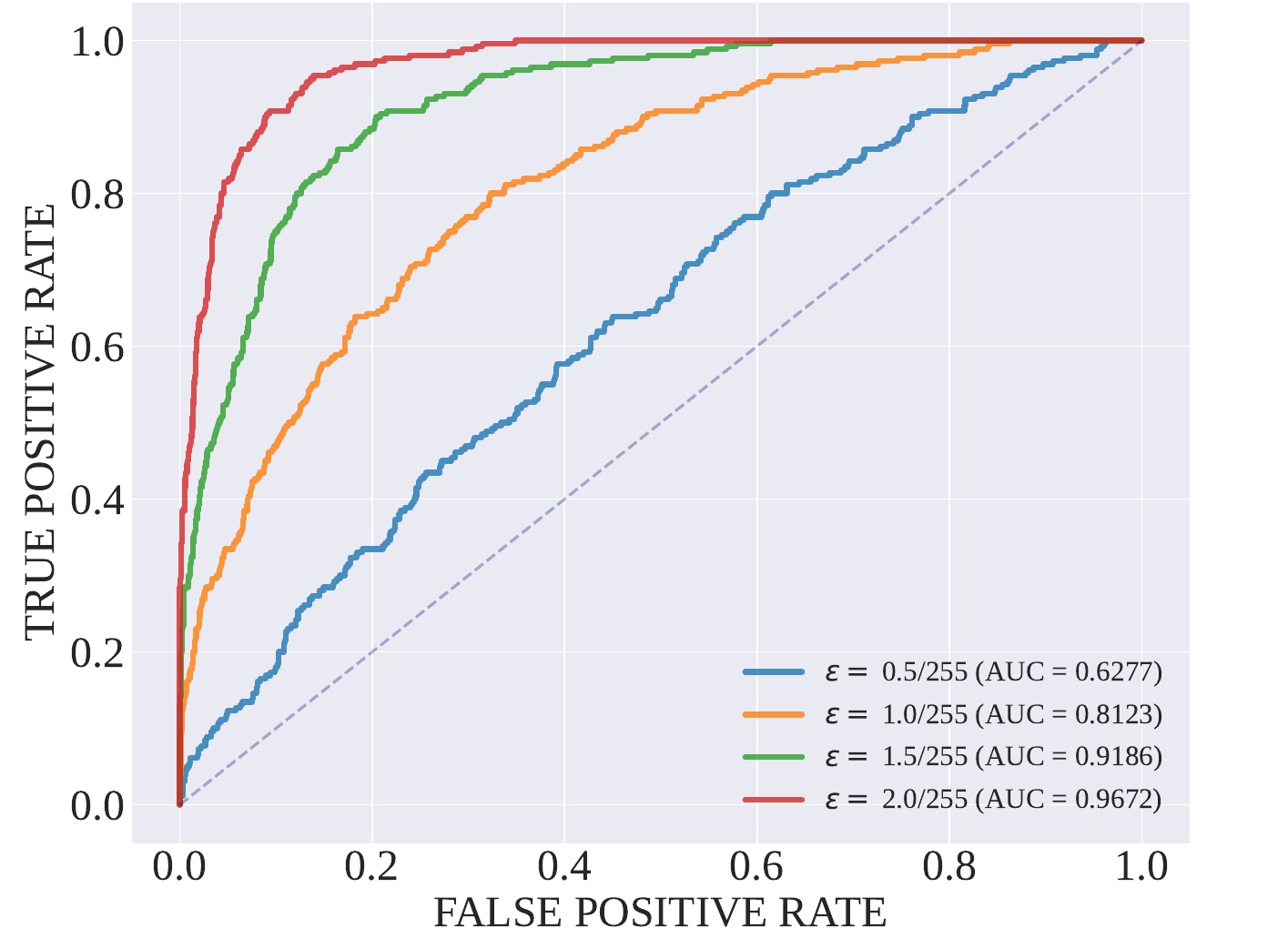}
        \subcaption{Fabricated Members by I-APGD}
    \end{minipage}
    \hfill
    \begin{minipage}{0.49\textwidth}
        \centering
        \includegraphics[width=\textwidth]{crop_Figure/Detection_figure/crop_roc_curve_Imagenet_stable_adaptive_pgd.pdf}
        \subcaption{Fabricated Members by OURS Attack}
    \end{minipage}
    \caption{\footnotesize  Comparison of the ROC Curve for Our Member Fabrication Detection Across Diverse Perturbation Bounds on \textbf{ImageNet-100}.}
    \label{Detection_figure_Imagenet}
    \end{center}
    \vspace{-1em}
\end{figure*}

\vspace{0.2em}
\noindent\textbf{Experimental Results of \textbf{MFD}.} In \Cref{Detection_figure_cifar10,Detection_figure_cifar100,Detection_figure_cinic,Detection_figure_svhn,Detection_figure_Imagenet}, we show that our \textbf{MFD} can effectively distinguish true members from fabricated members across different perturbation levels and datasets. The AUC is greater than 50\% and increases as the perturbation level increases.

\begin{figure*}[!t]
    \begin{center}
    \begin{minipage}{0.49\textwidth}
        \centering
        \includegraphics[width=\textwidth]{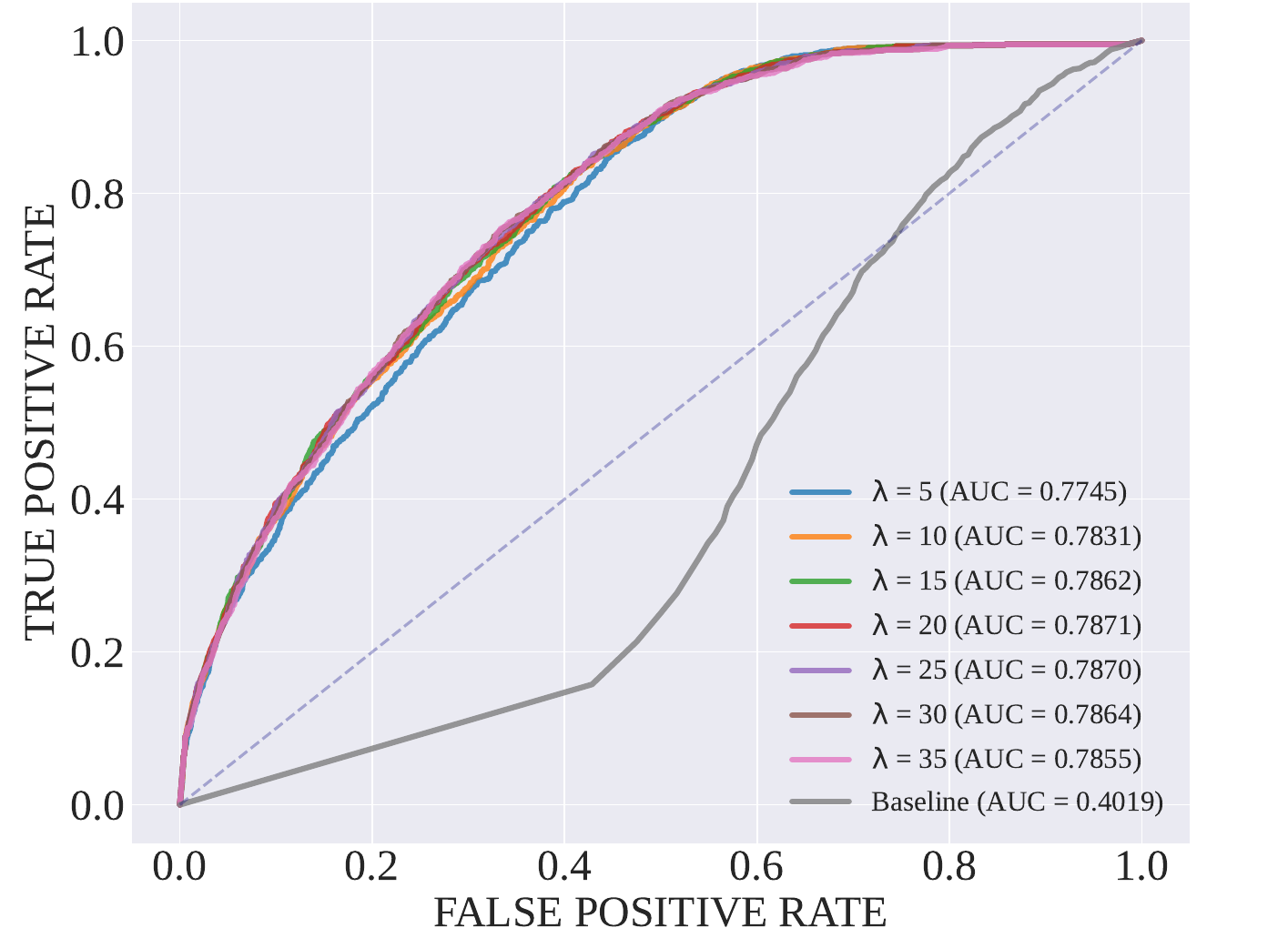}
        \subcaption{Adversarially Robust Attack R}
    \end{minipage}
    \hfill
    \begin{minipage}{0.49\textwidth}
        \centering
        \includegraphics[width=\textwidth]{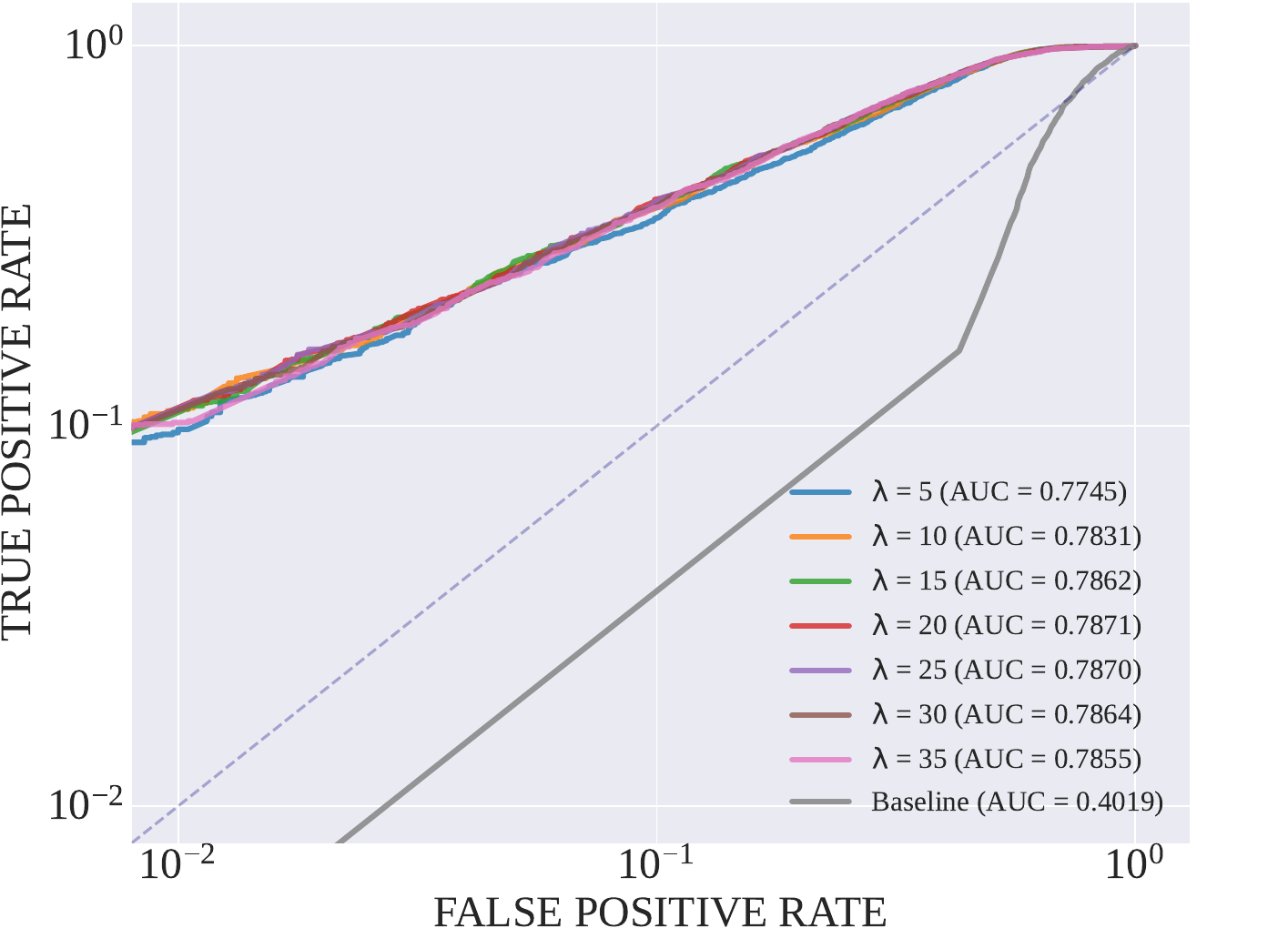}
        \subcaption{Adversarially Robust Attack R (log scale)}
    \end{minipage}

    \begin{minipage}{0.49\textwidth}
        \centering
        \includegraphics[width=\textwidth]{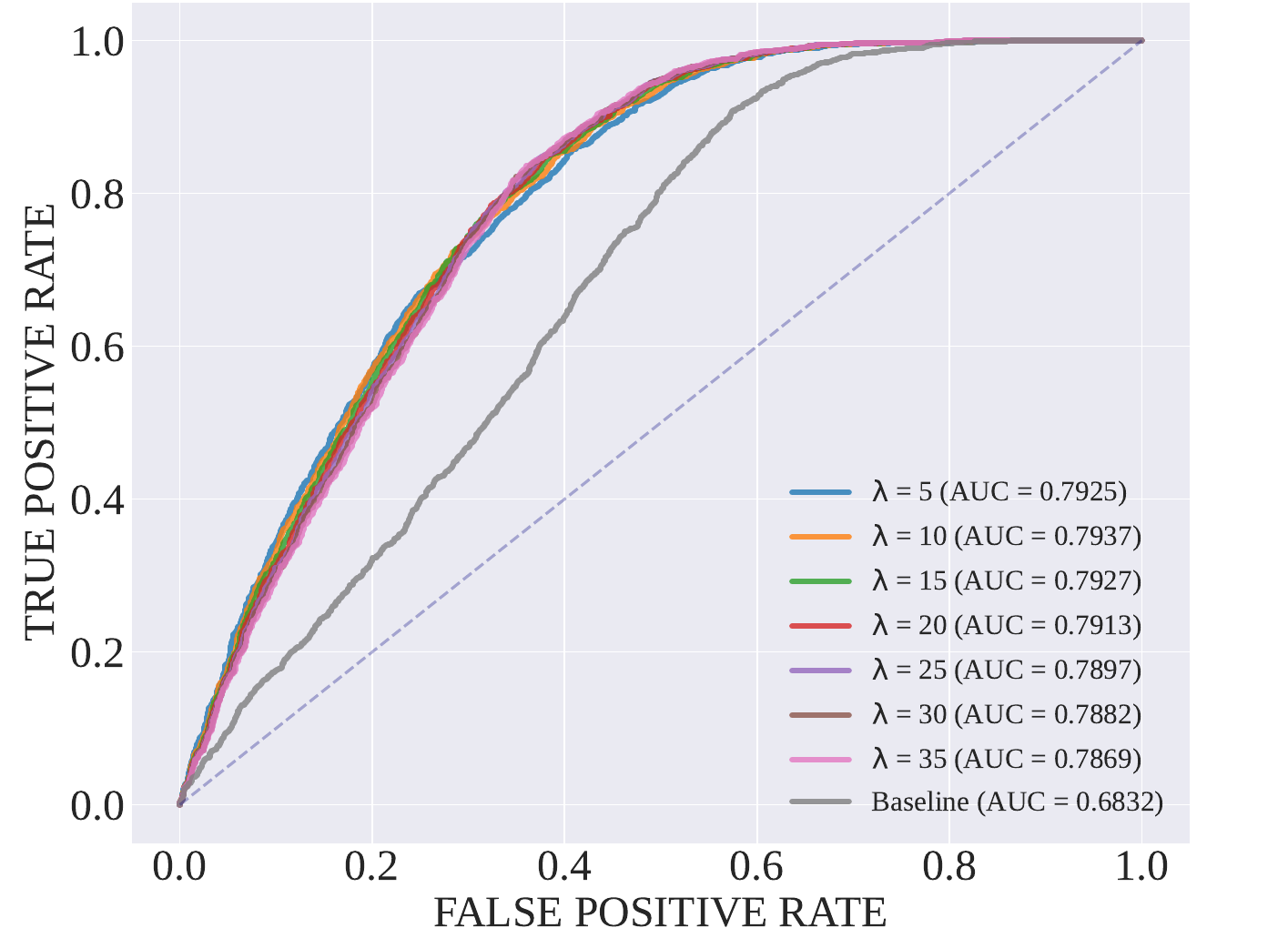}
        \subcaption{Adversarially Robust LiRA}
    \end{minipage}
    \hfill
    \begin{minipage}{0.49\textwidth}
        \centering
        \includegraphics[width=\textwidth]{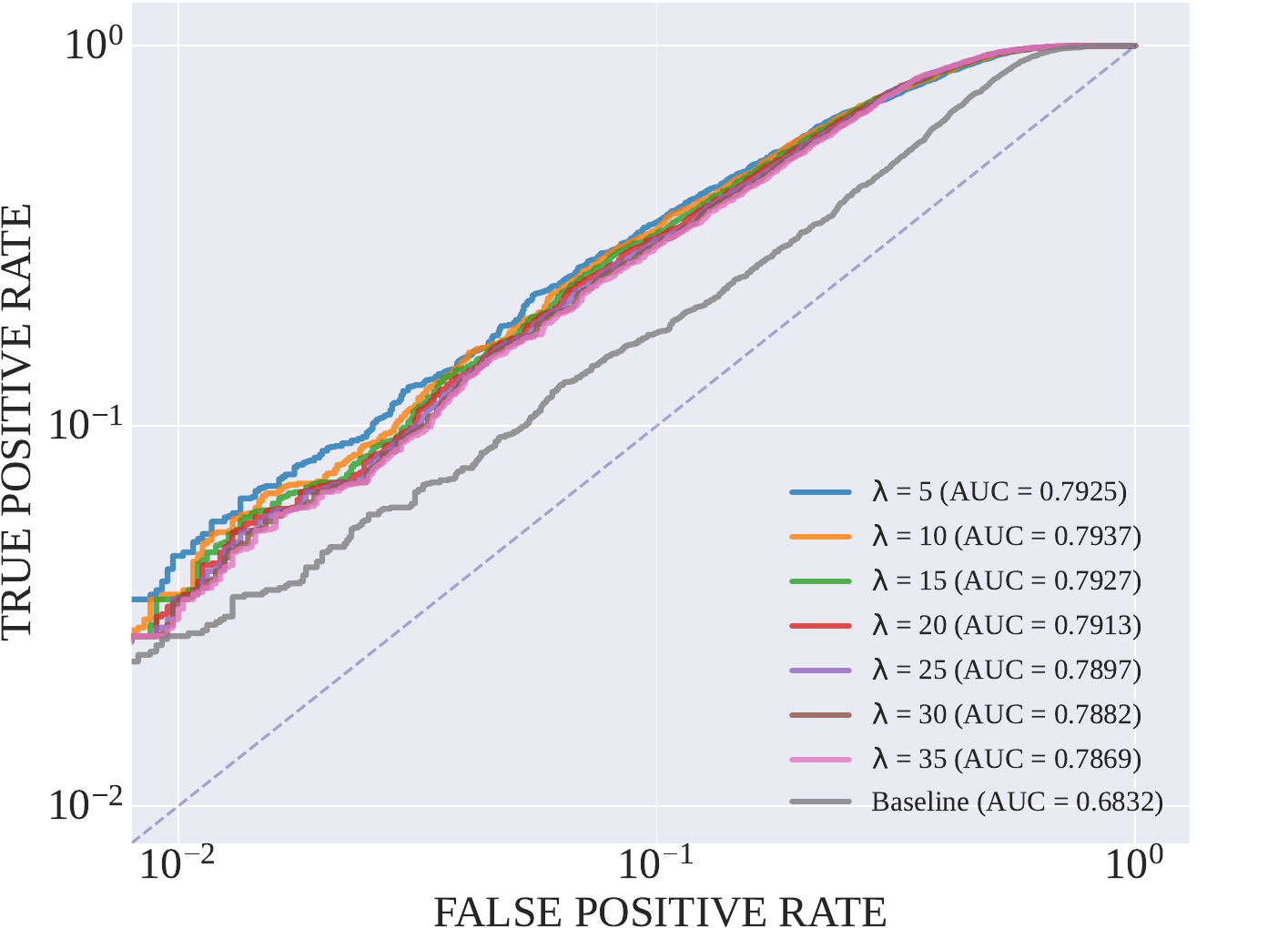}
        \subcaption{Adversarially Robust LiRA (log scale)}
    \end{minipage}

    \begin{minipage}{0.49\textwidth}
        \centering
        \includegraphics[width=\textwidth]{robust_MIA/crop_rmia_cifar10_roc_plot.pdf}
        \subcaption{Adversarially Robust RMIA}
    \end{minipage}
    \hfill
    \begin{minipage}{0.49\textwidth}
        \centering
        \includegraphics[width=\textwidth]{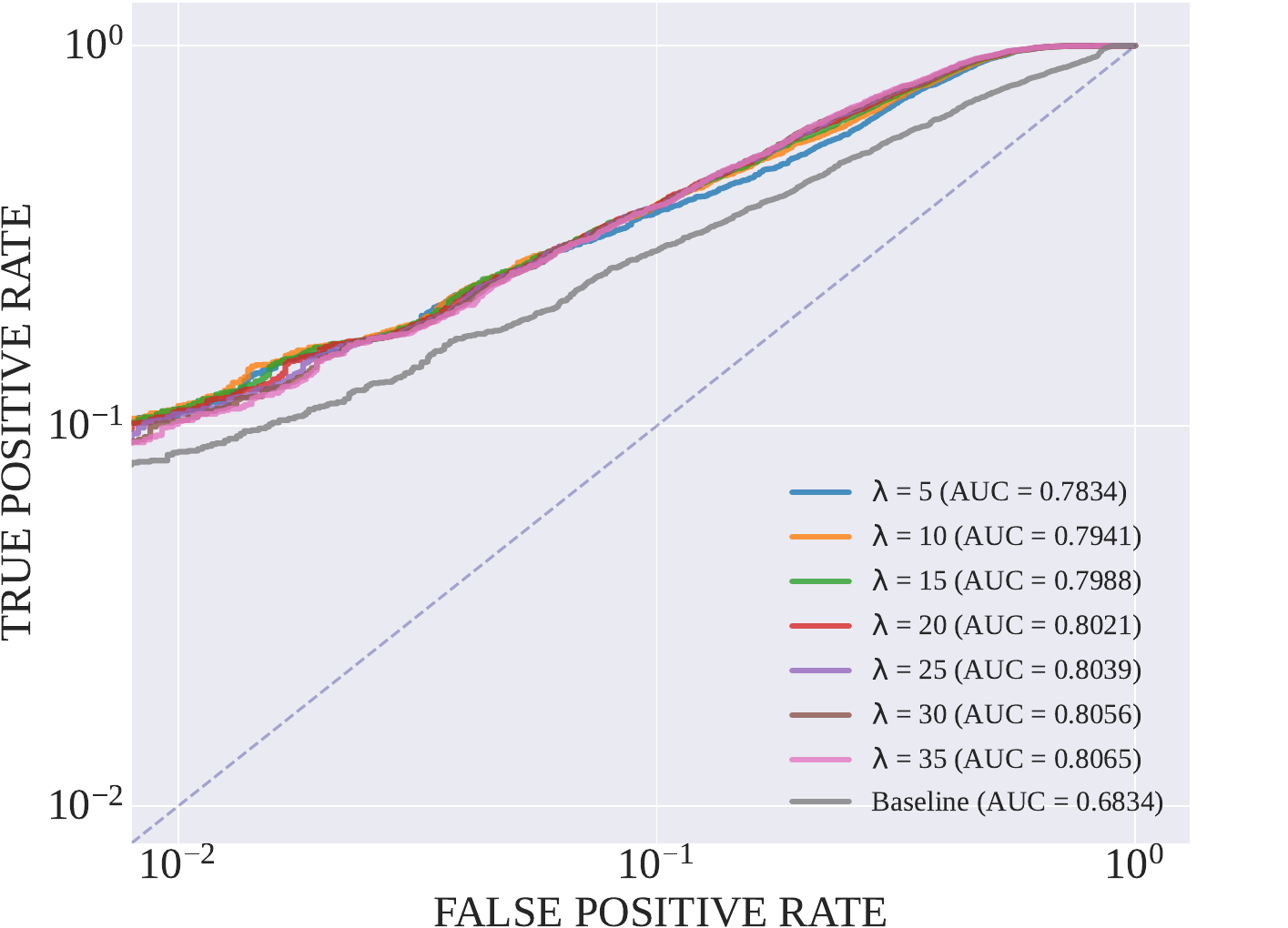}
        \subcaption{Adversarially Robust RMIA (log scale)}
    \end{minipage}
    \caption{\footnotesize Comparison of ROC Curves for Our Adversarially Robust MIAs and Baselines on \textbf{CIFAR-10}.}
    \label{Robust_MIA_cifar10}
    \end{center}
    \vspace{-1em}
\end{figure*}

\begin{figure*}[!t]
    \begin{center}
    \begin{minipage}{0.49\textwidth}
        \centering
        \includegraphics[width=\textwidth]{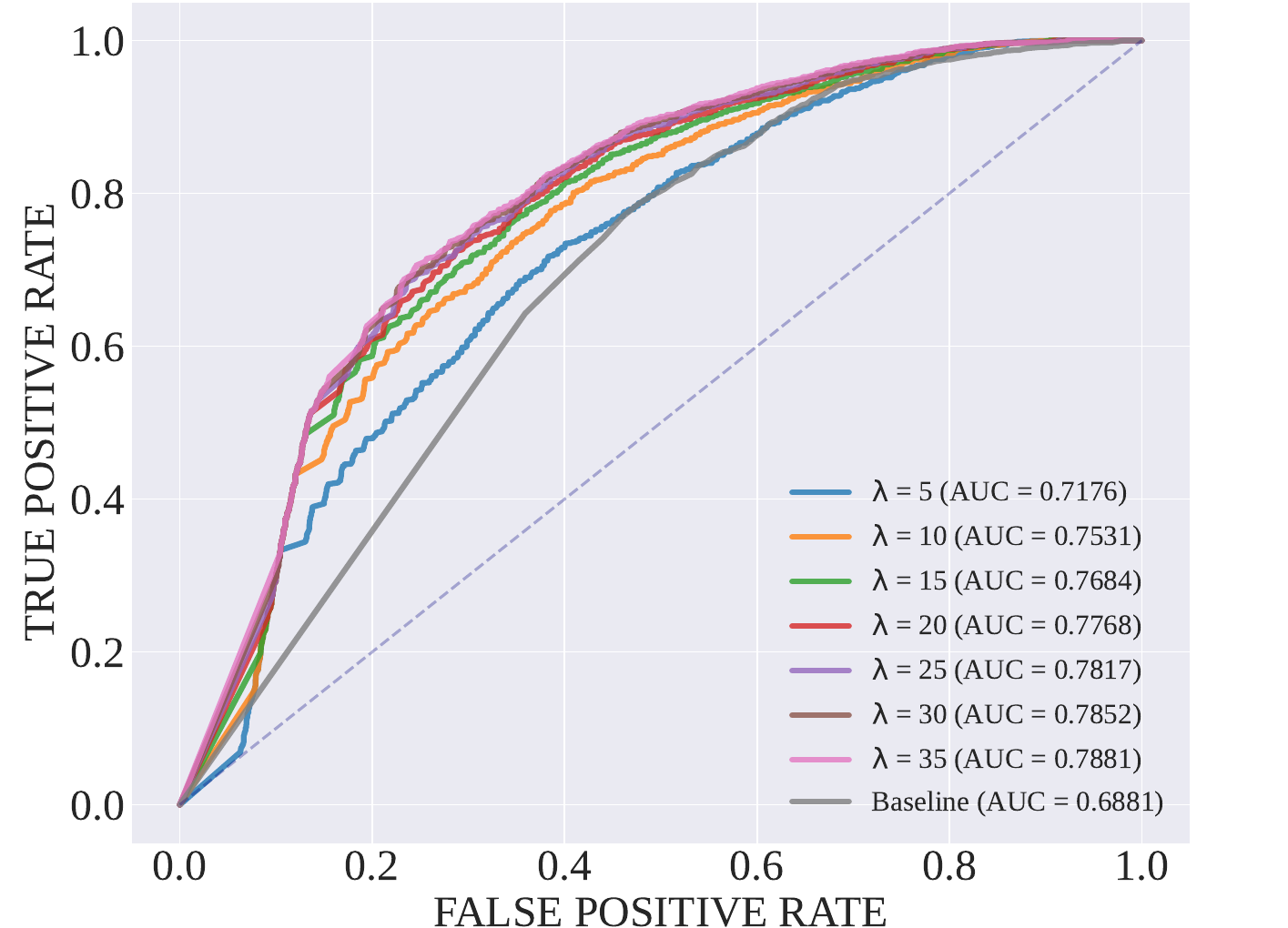}
        \subcaption{Adversarially Robust Attack R}
    \end{minipage}
    \hfill
    \begin{minipage}{0.49\textwidth}
        \centering
        \includegraphics[width=\textwidth]{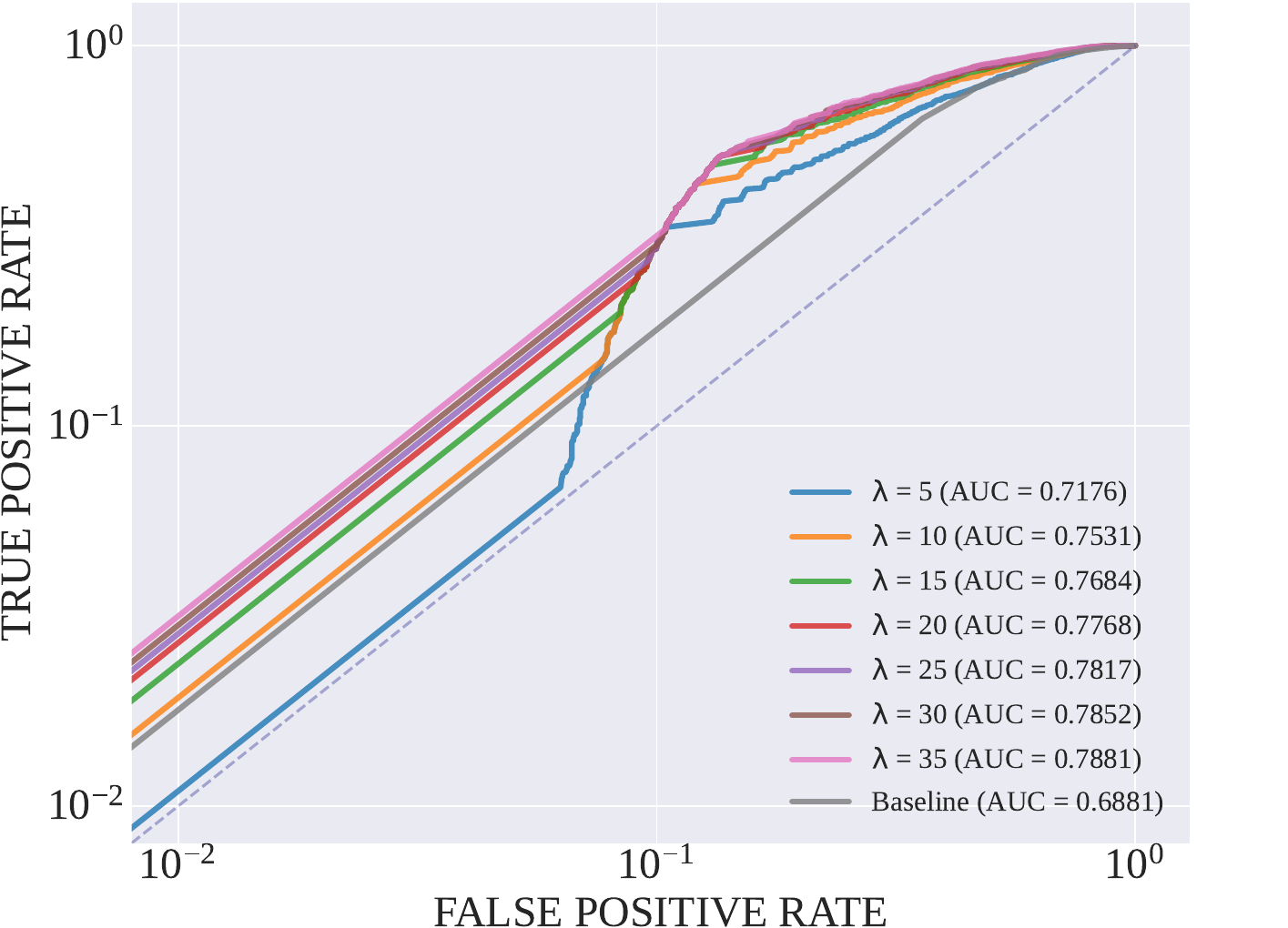}
        \subcaption{Adversarially Robust Attack R (log scale)}
    \end{minipage}

    \begin{minipage}{0.49\textwidth}
        \centering
        \includegraphics[width=\textwidth]{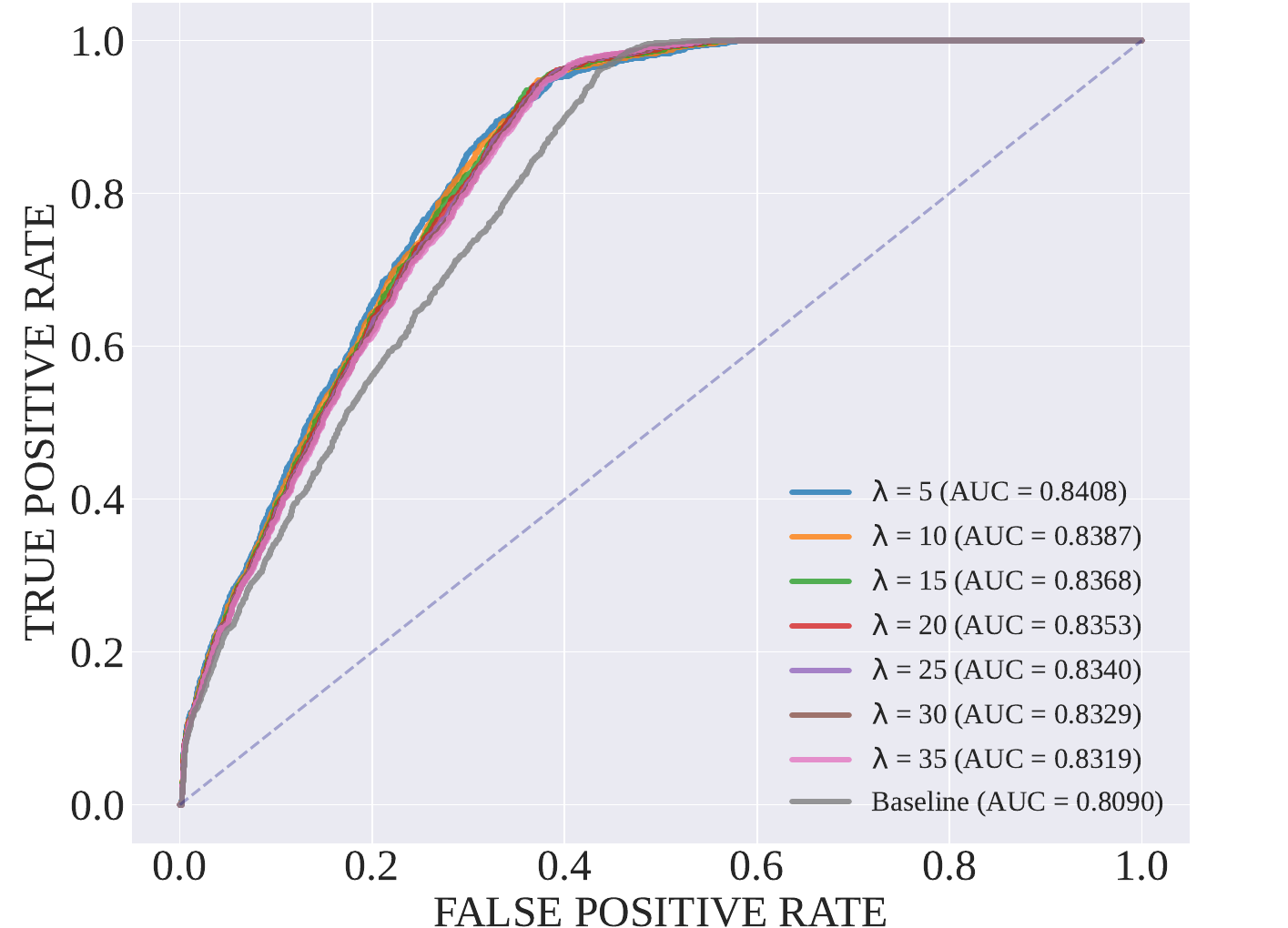}
        \subcaption{Adversarially Robust LiRA}
    \end{minipage}
    \hfill
    \begin{minipage}{0.49\textwidth}
        \centering
        \includegraphics[width=\textwidth]{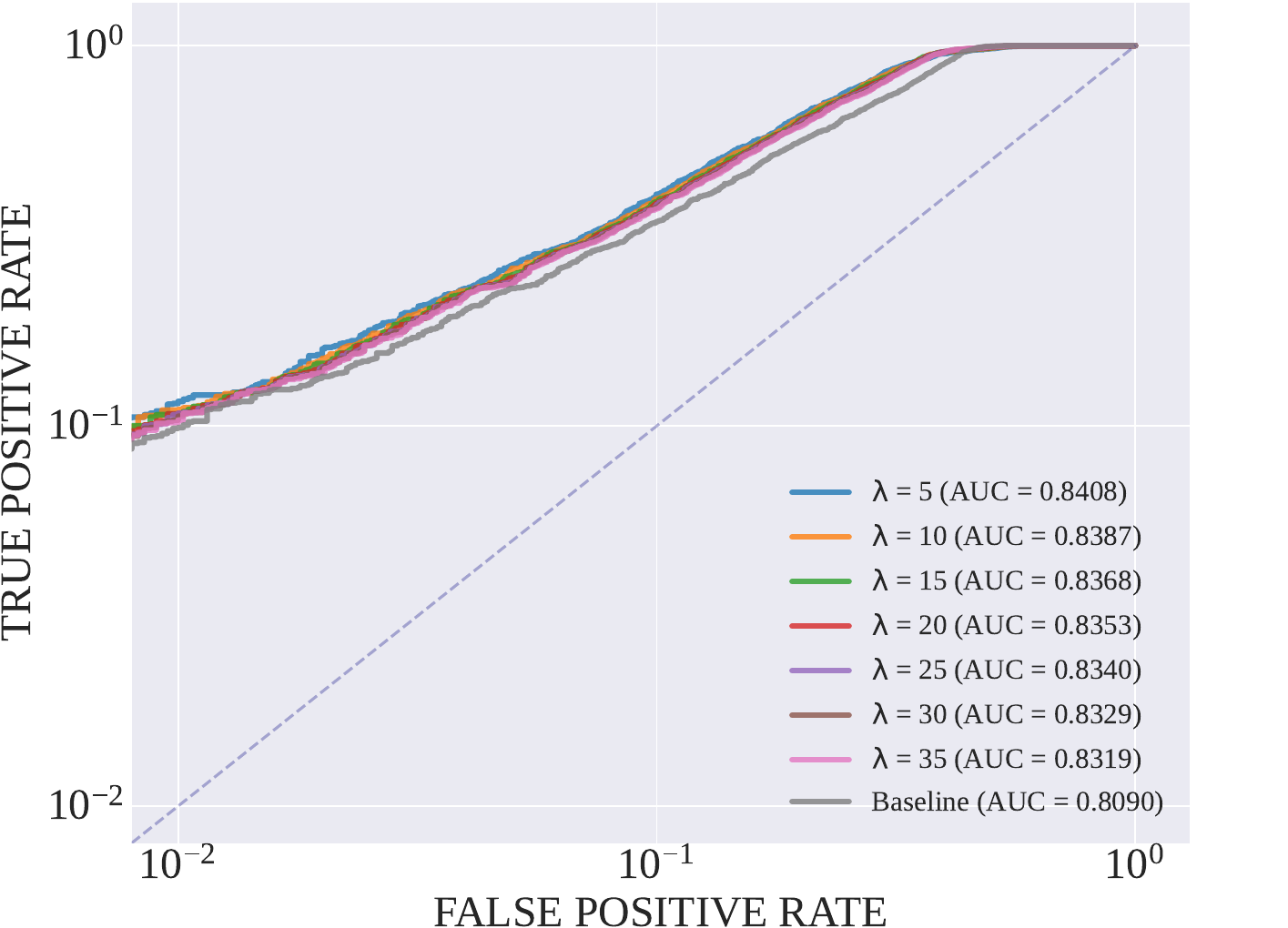}
        \subcaption{Adversarially Robust LiRA (log scale)}
    \end{minipage}

    \begin{minipage}{0.49\textwidth}
        \centering
        \includegraphics[width=\textwidth]{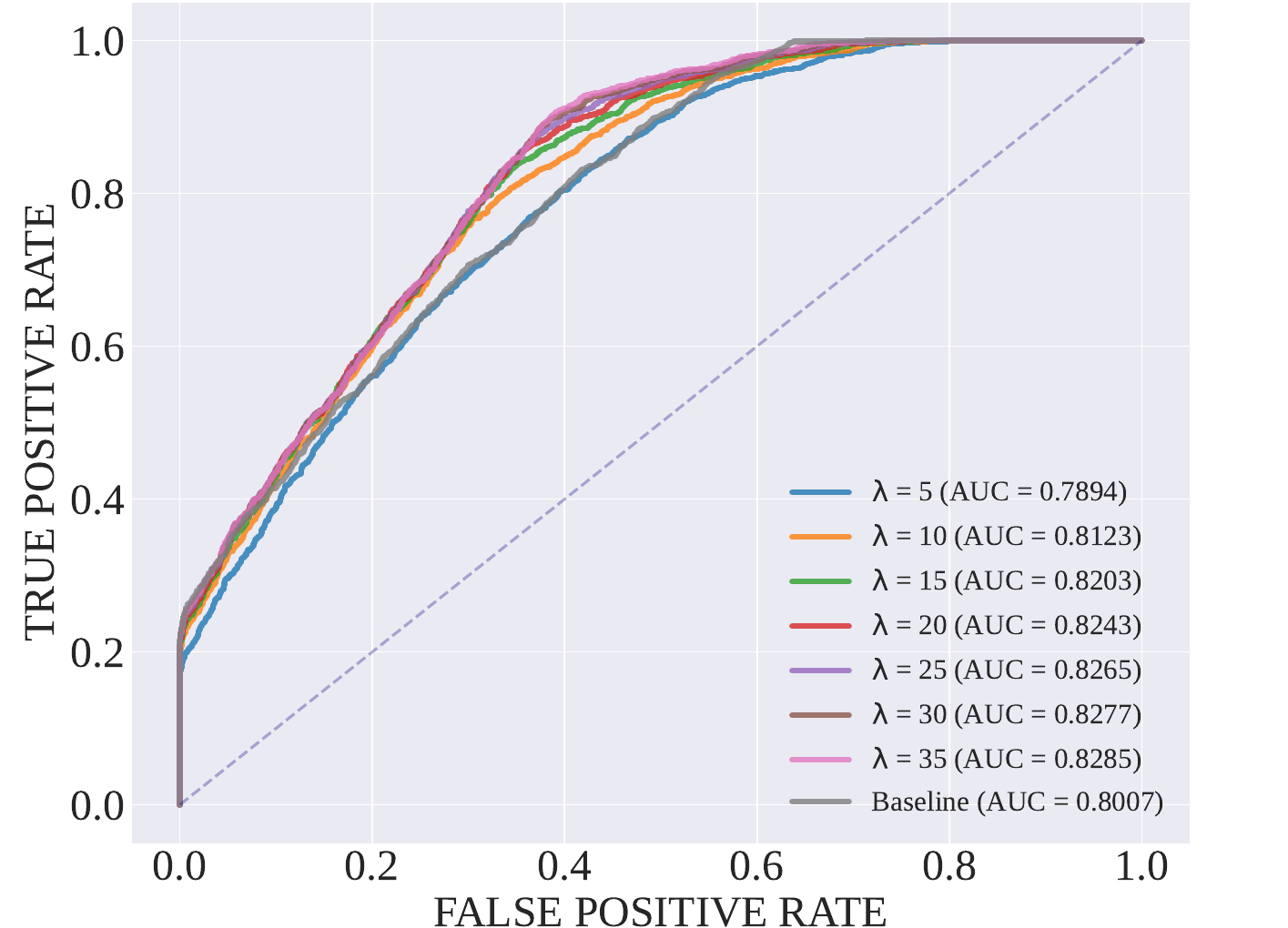}
        \subcaption{Adversarially Robust RMIA}
    \end{minipage}
    \hfill
    \begin{minipage}{0.49\textwidth}
        \centering
        \includegraphics[width=\textwidth]{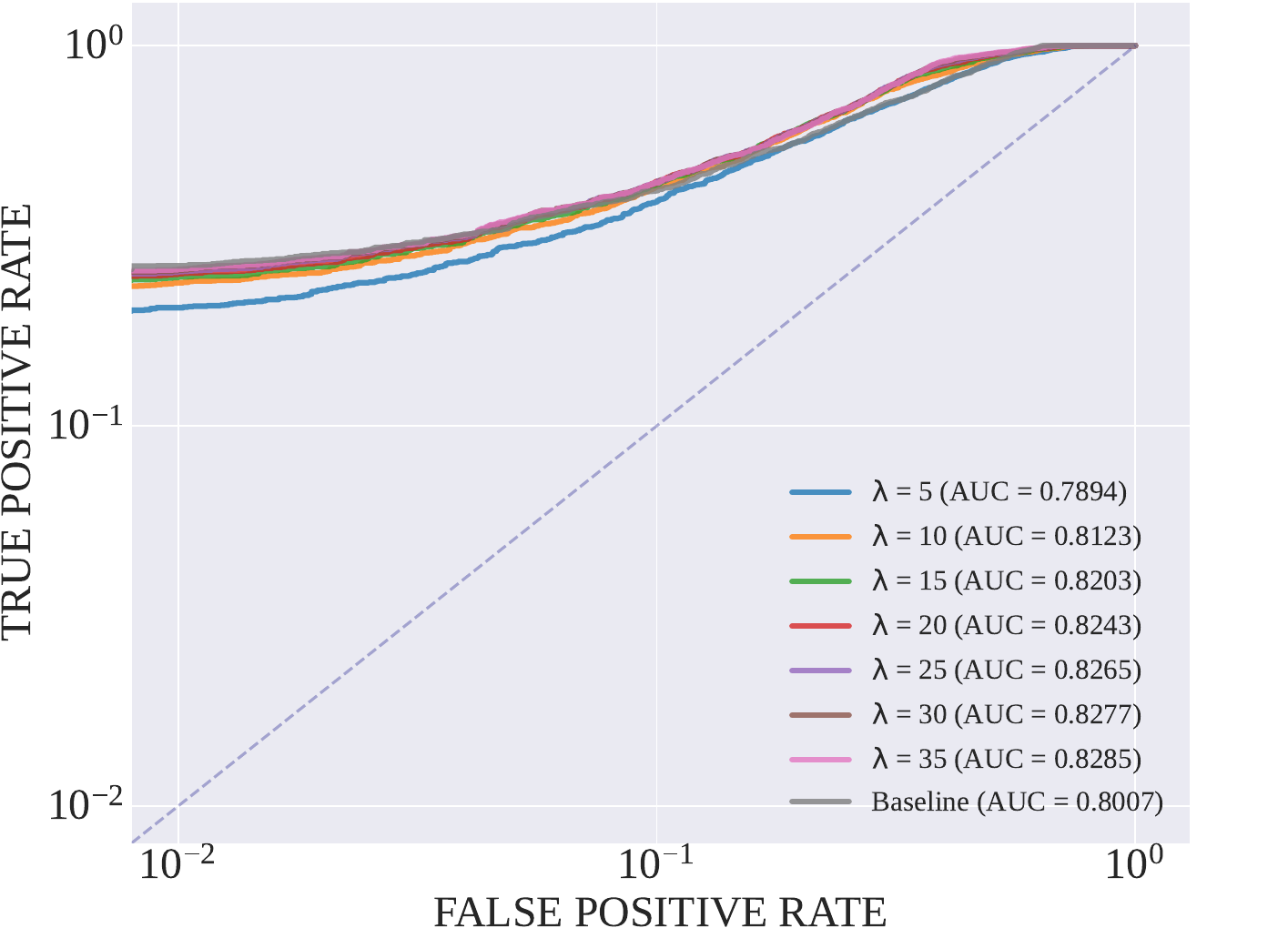}
        \subcaption{Adversarially Robust RMIA (log scale)}
    \end{minipage}
    \caption{\footnotesize Comparison of ROC Curves for Our Adversarially Robust MIAs and Baselines on \textbf{CIFAR-100}.}
    \label{Robust_MIA_cifar100}
    \end{center}
    \vspace{-1em}
\end{figure*}

\begin{figure*}[!t]
    \begin{center}
    \begin{minipage}{0.49\textwidth}
        \centering
        \includegraphics[width=\textwidth]{robust_MIA/crop_attack_r_svhn_roc_plot.pdf}
        \subcaption{Adversarially Robust Attack R}
    \end{minipage}
    \hfill
    \begin{minipage}{0.49\textwidth}
        \centering
        \includegraphics[width=\textwidth]{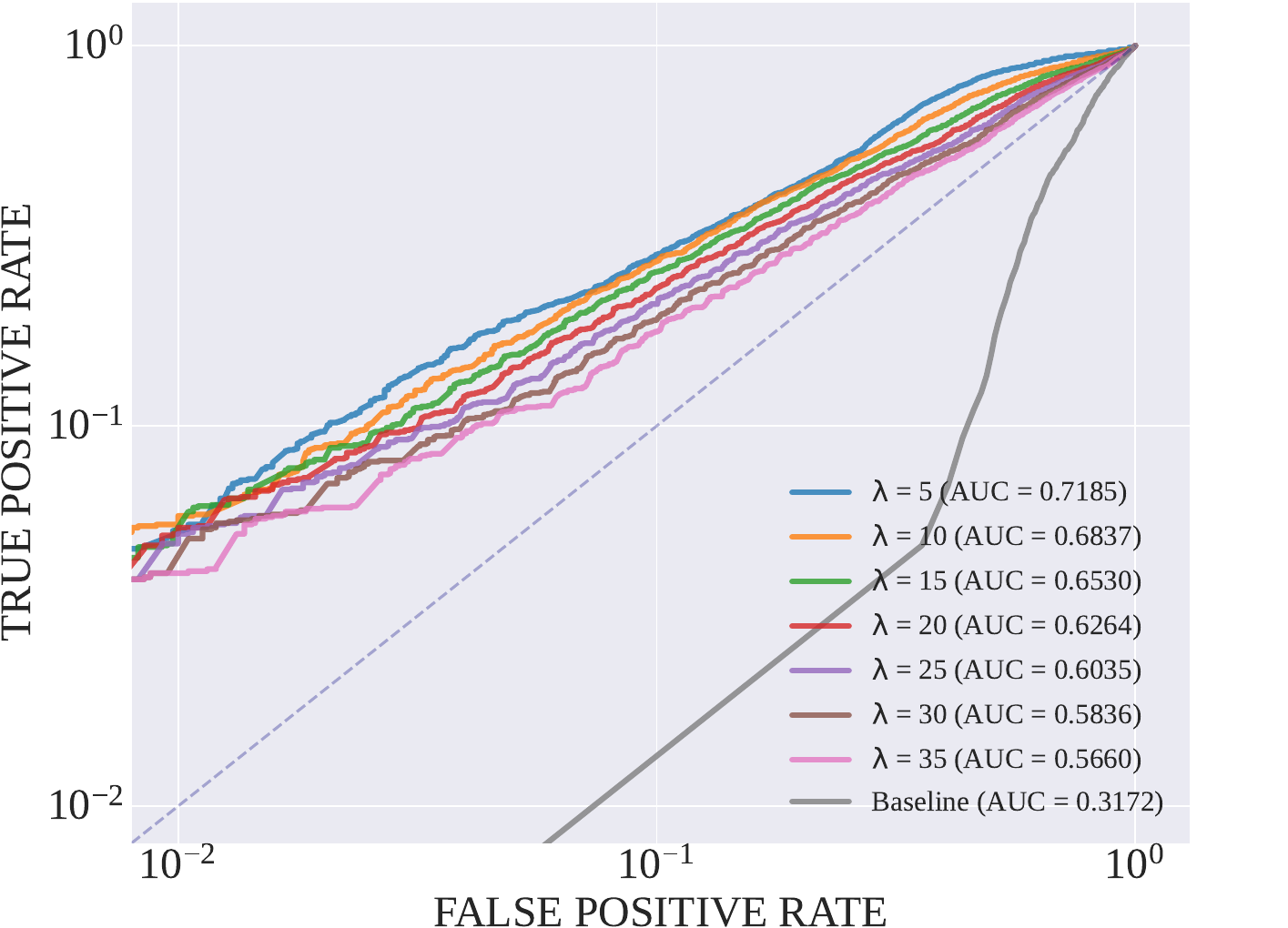}
        \subcaption{Adversarially Robust Attack R (log scale)}
    \end{minipage}

    \begin{minipage}{0.49\textwidth}
        \centering
        \includegraphics[width=\textwidth]{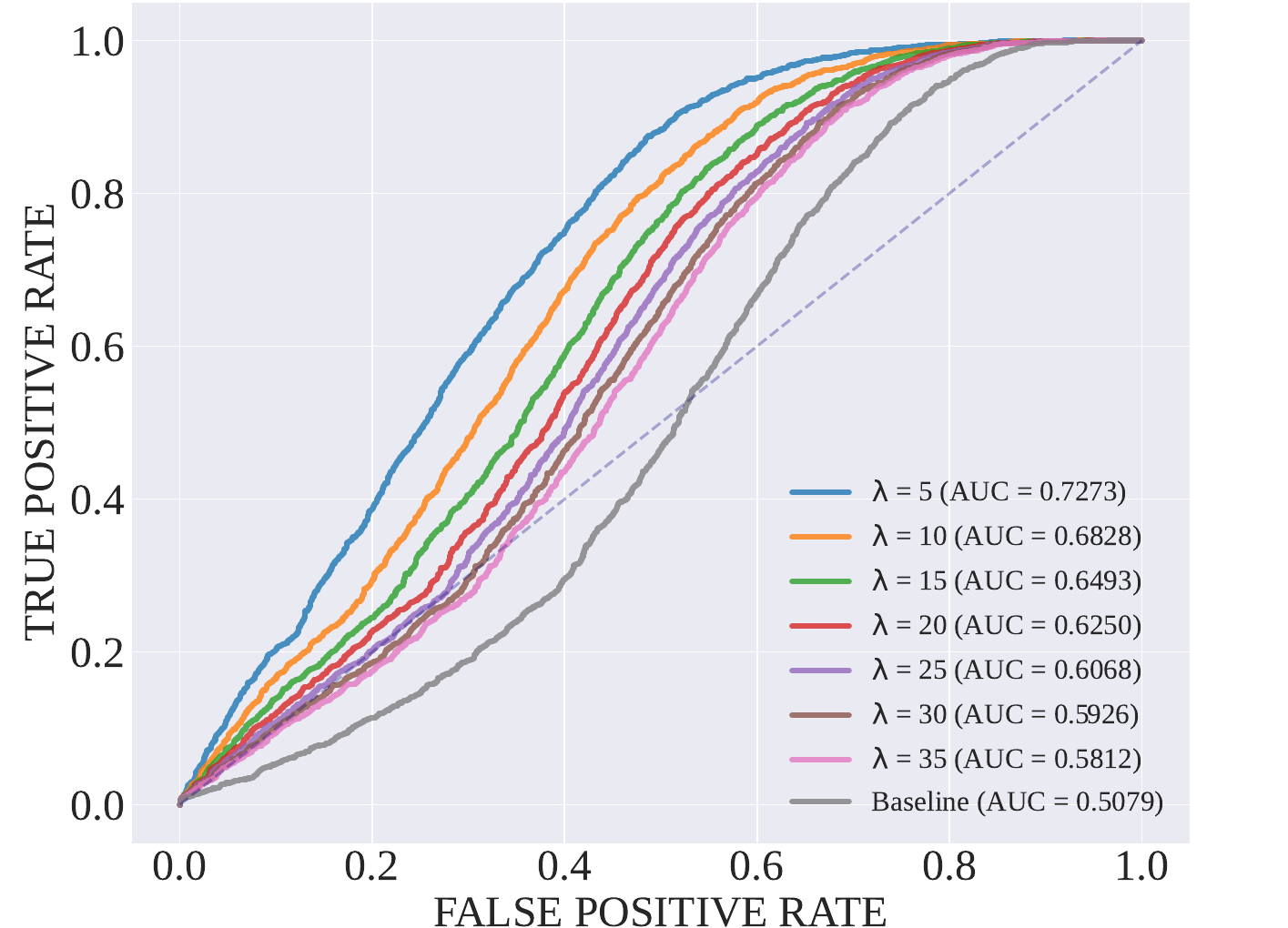}
        \subcaption{Adversarially Robust LiRA}
    \end{minipage}
    \hfill
    \begin{minipage}{0.49\textwidth}
        \centering
        \includegraphics[width=\textwidth]{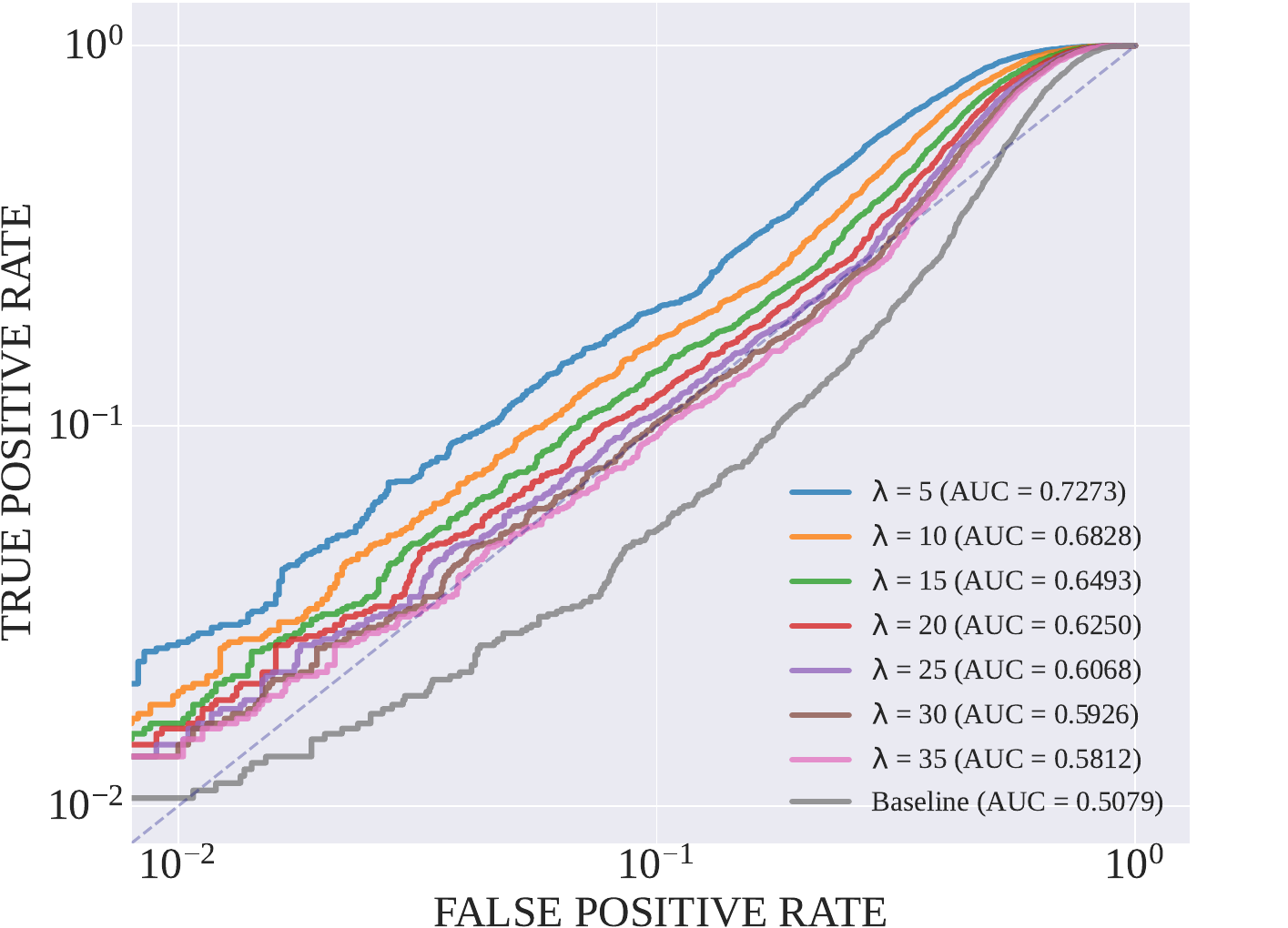}
        \subcaption{Adversarially Robust LiRA (log scale)}
    \end{minipage}

    \begin{minipage}{0.49\textwidth}
        \centering
        \includegraphics[width=\textwidth]{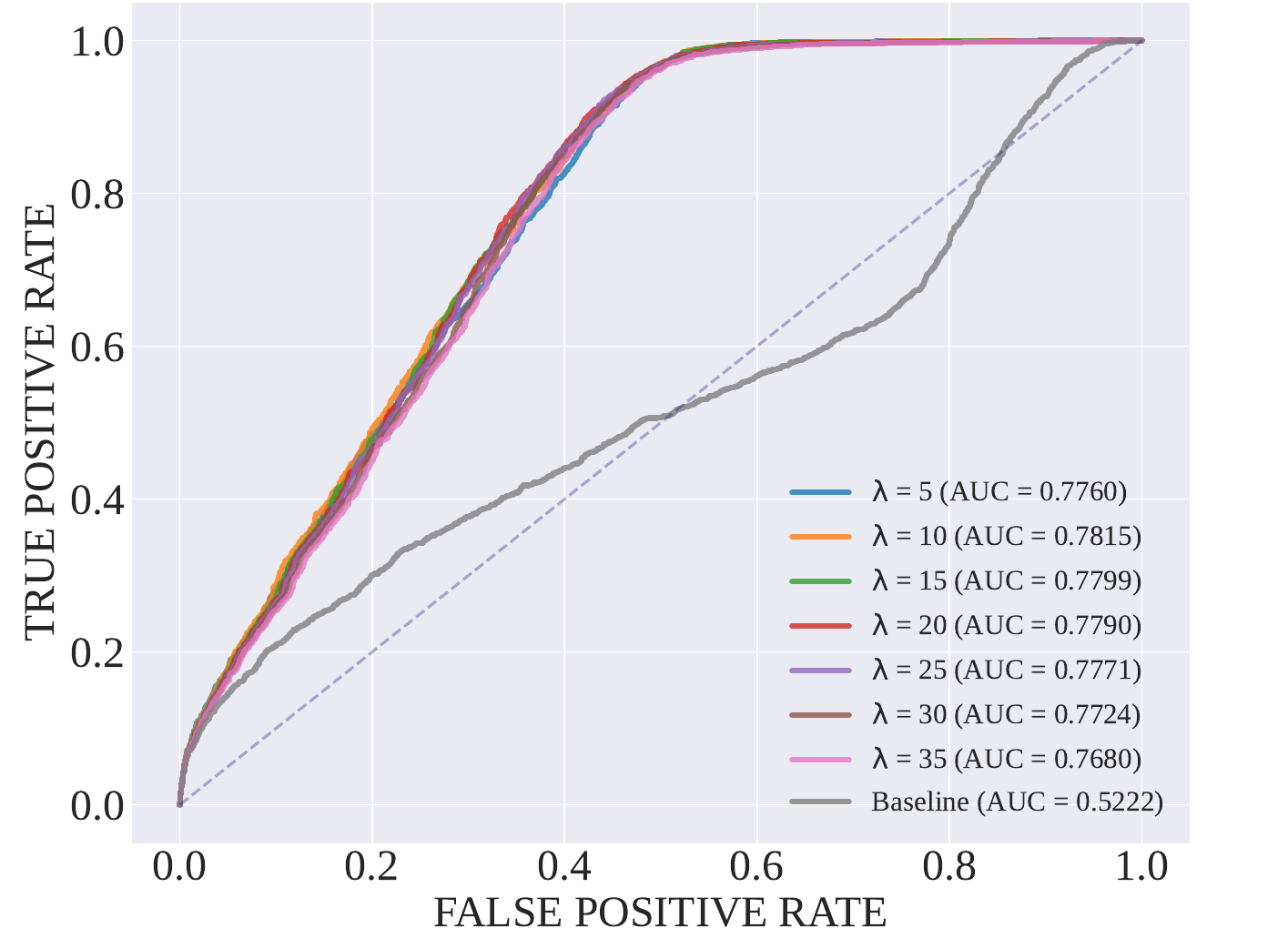}
        \subcaption{Adversarially Robust RMIA}
    \end{minipage}
    \hfill
    \begin{minipage}{0.49\textwidth}
        \centering
        \includegraphics[width=\textwidth]{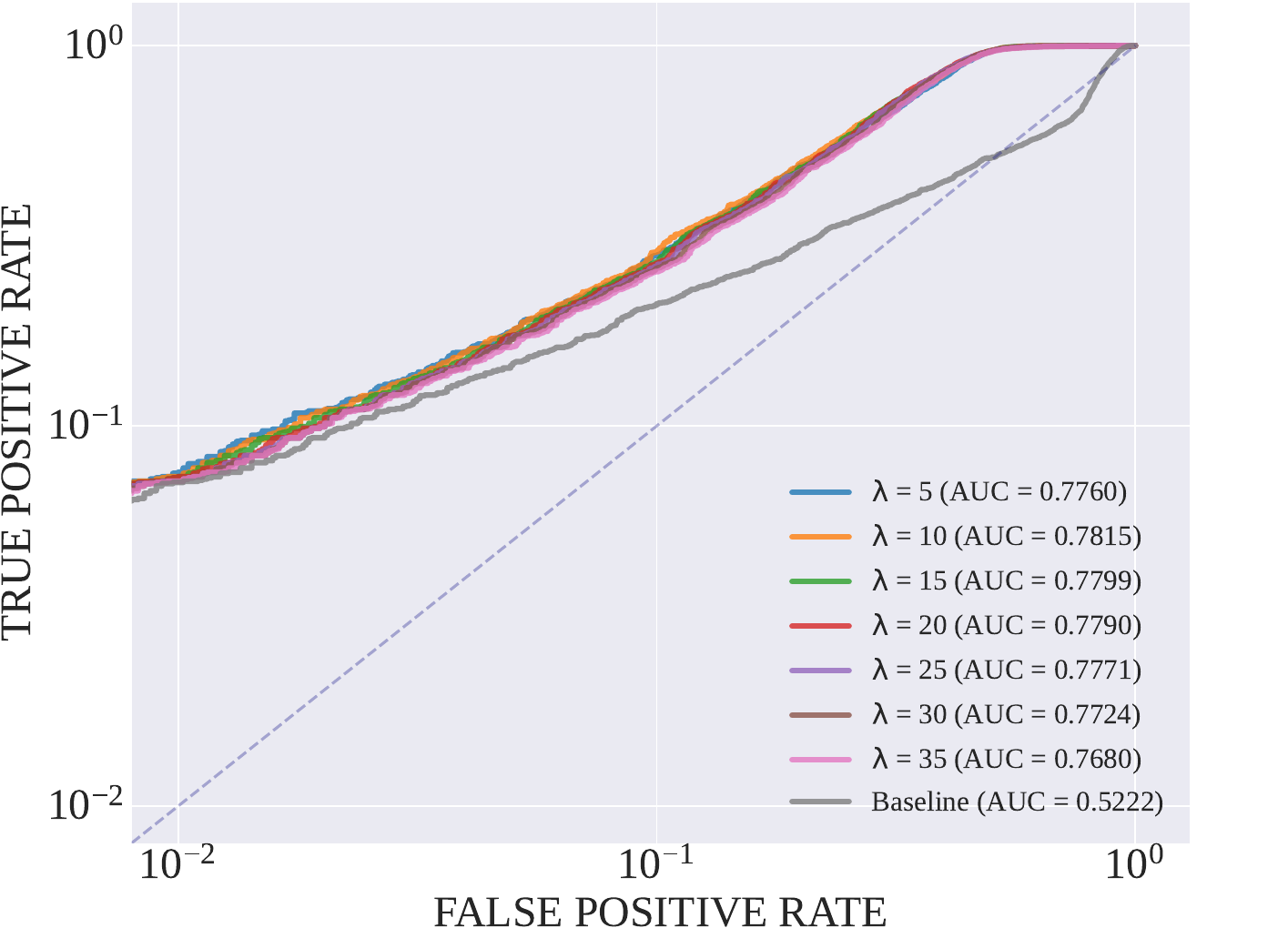}
        \subcaption{Adversarially Robust RMIA (log scale)}
    \end{minipage}
    \caption{\footnotesize Comparison of ROC Curves for Our Adversarially Robust MIAs and Baselines on \textbf{SVHN}.}
    \label{Robust_MIA_svhn}
    \end{center}
    \vspace{-1em}
\end{figure*}

\begin{figure*}[!t]
    \begin{center}
    \begin{minipage}{0.49\textwidth}
        \centering
        \includegraphics[width=\textwidth]{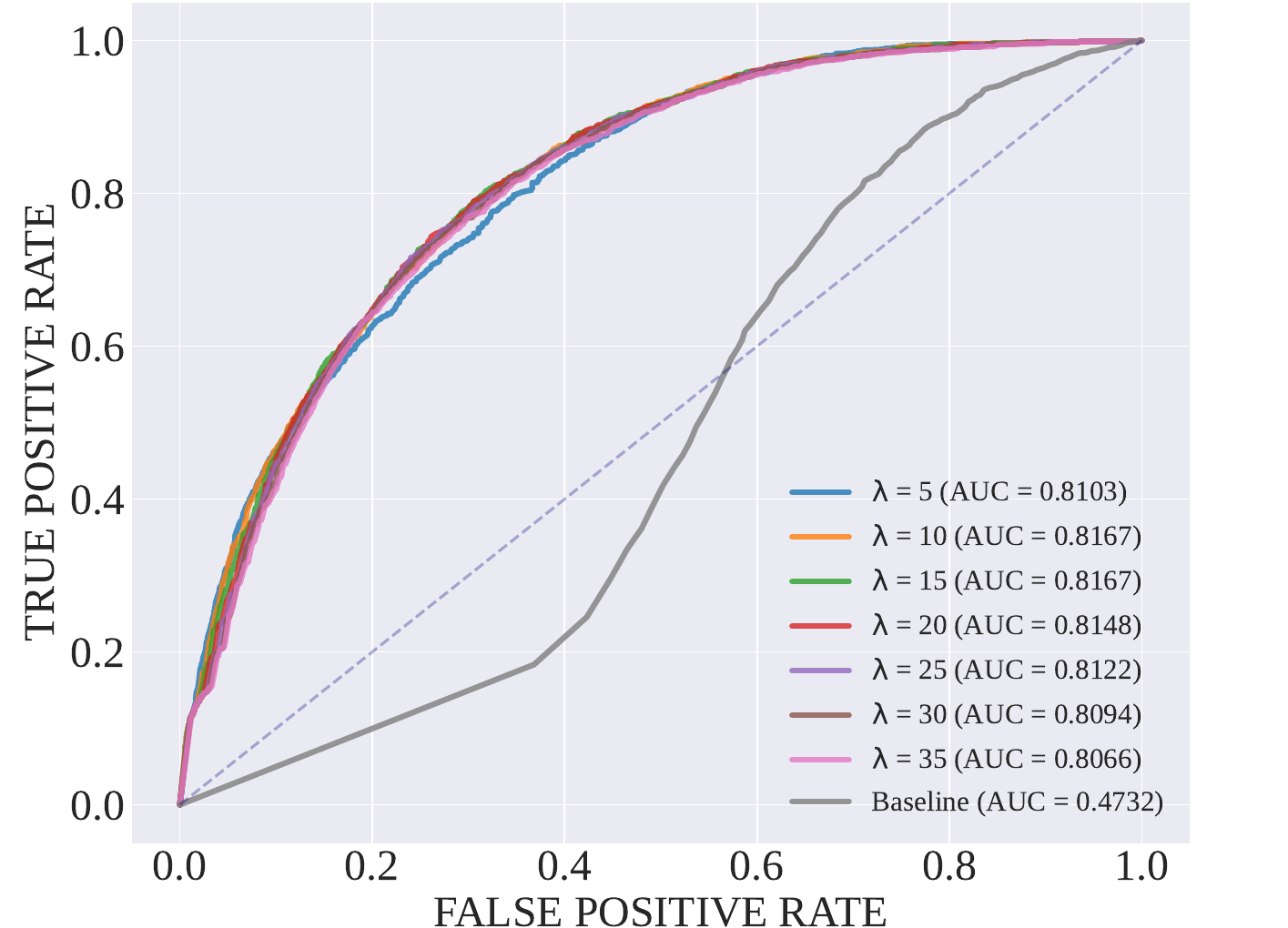}
        \subcaption{Adversarially Robust Attack R}
    \end{minipage}
    \hfill
    \begin{minipage}{0.49\textwidth}
        \centering
        \includegraphics[width=\textwidth]{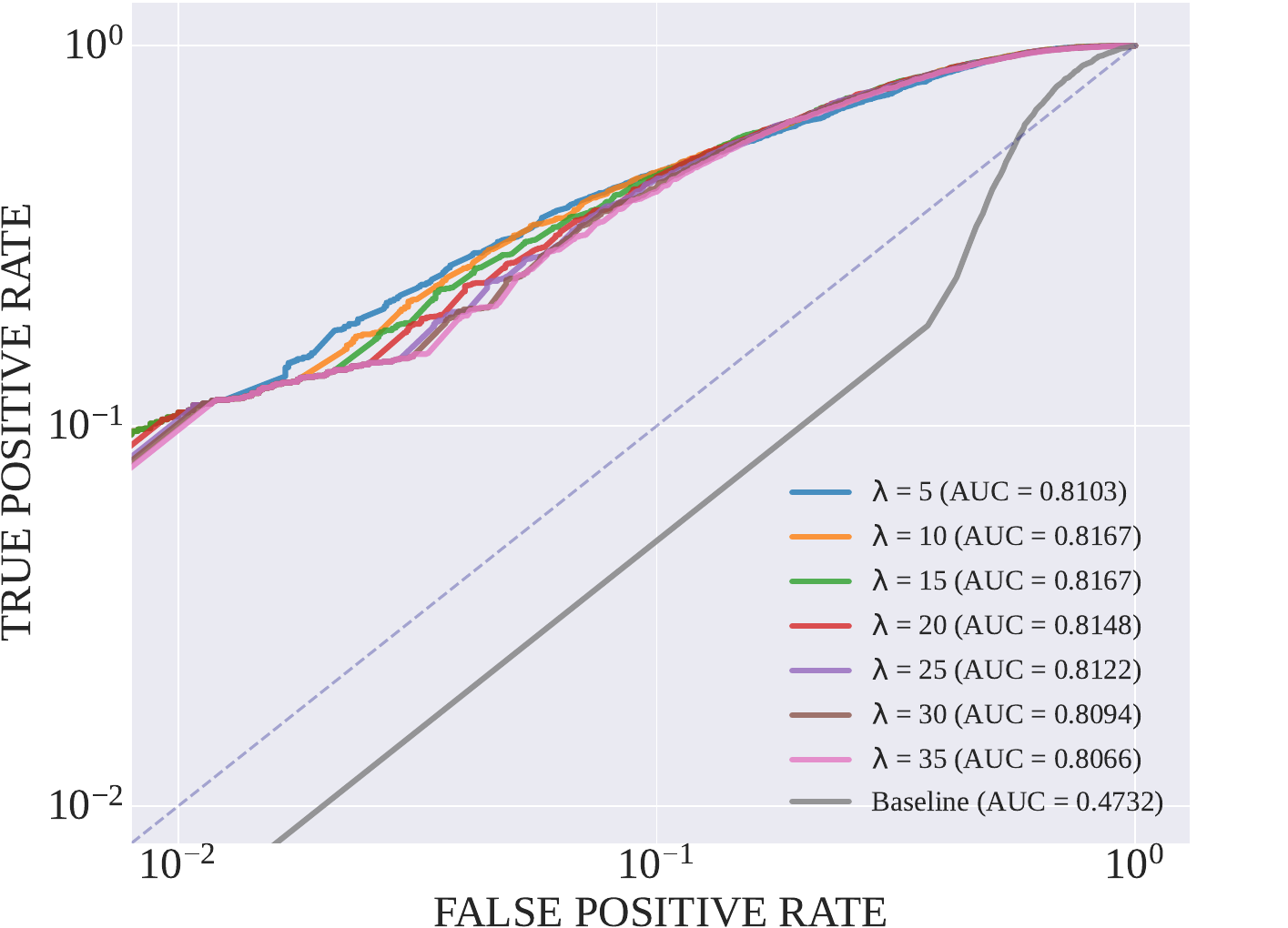}
        \subcaption{Adversarially Robust Attack R (log scale)}
    \end{minipage}

    \begin{minipage}{0.49\textwidth}
        \centering
        \includegraphics[width=\textwidth]{robust_MIA/crop_lira_cinic_roc_plot.pdf}
        \subcaption{Adversarially Robust LiRA}
    \end{minipage}
    \hfill
    \begin{minipage}{0.49\textwidth}
        \centering
        \includegraphics[width=\textwidth]{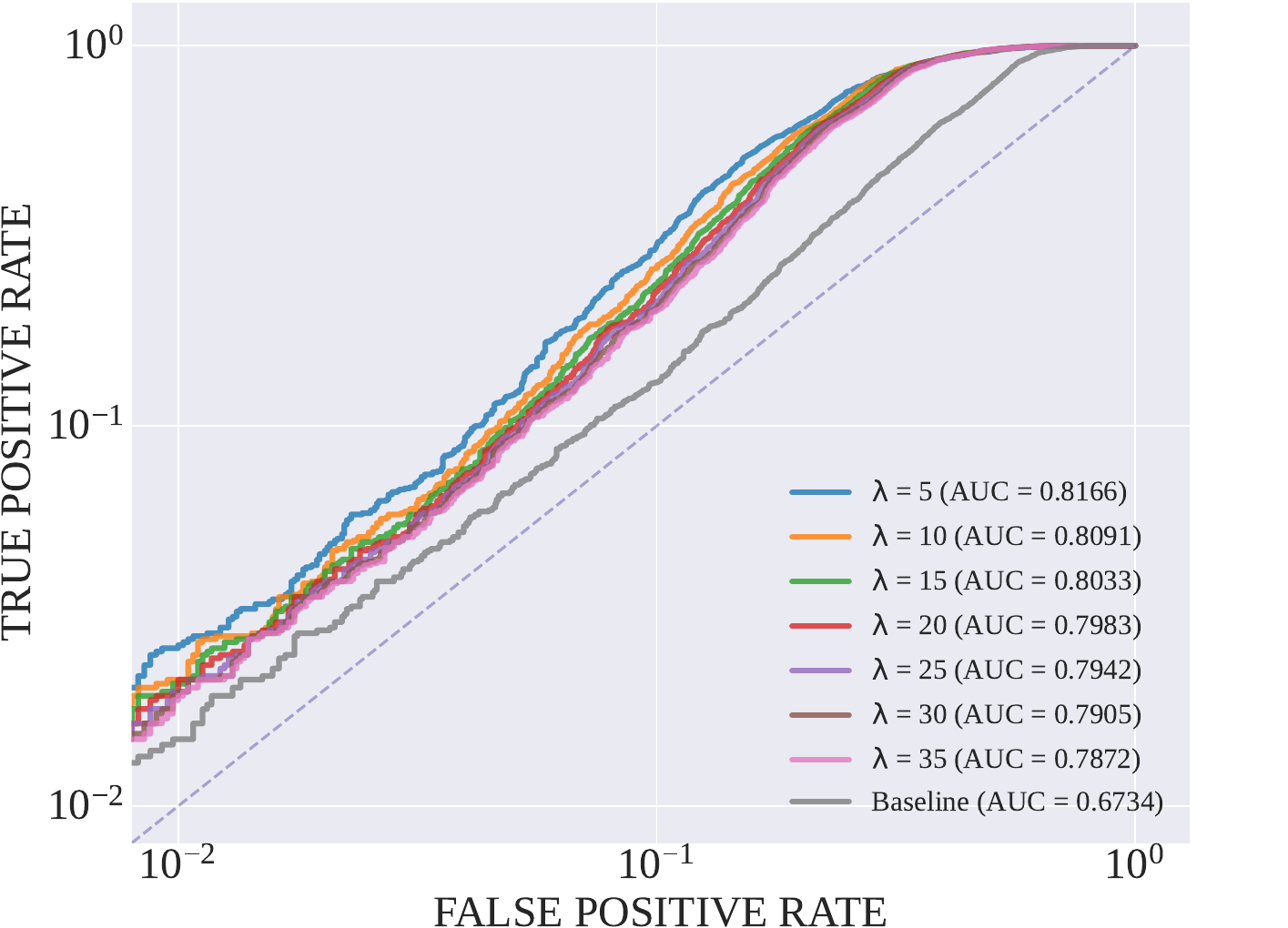}
        \subcaption{Adversarially Robust LiRA (log scale)}
    \end{minipage}

    \begin{minipage}{0.49\textwidth}
        \centering
        \includegraphics[width=\textwidth]{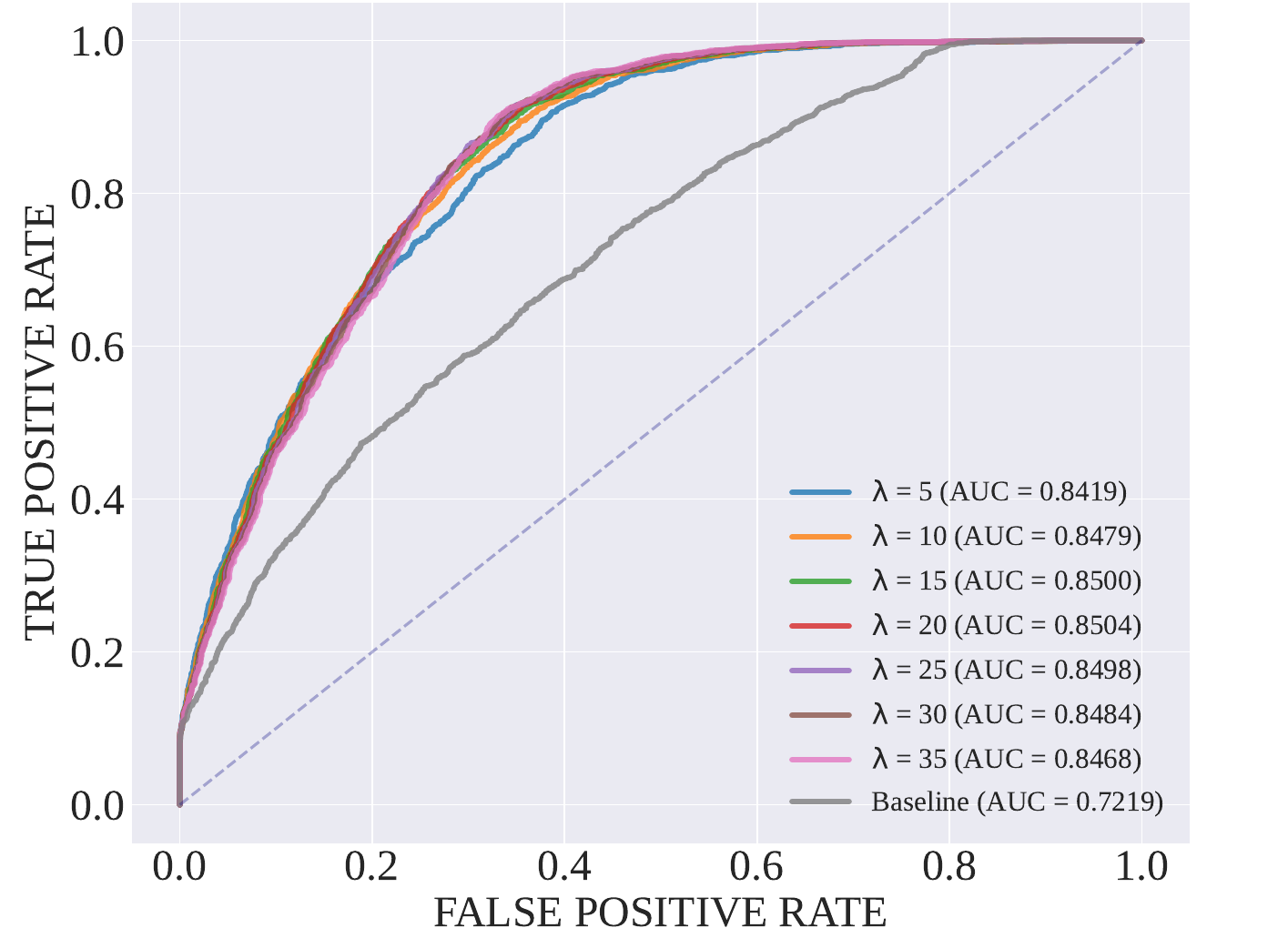}
        \subcaption{Adversarially Robust RMIA}
    \end{minipage}
    \hfill
    \begin{minipage}{0.49\textwidth}
        \centering
        \includegraphics[width=\textwidth]{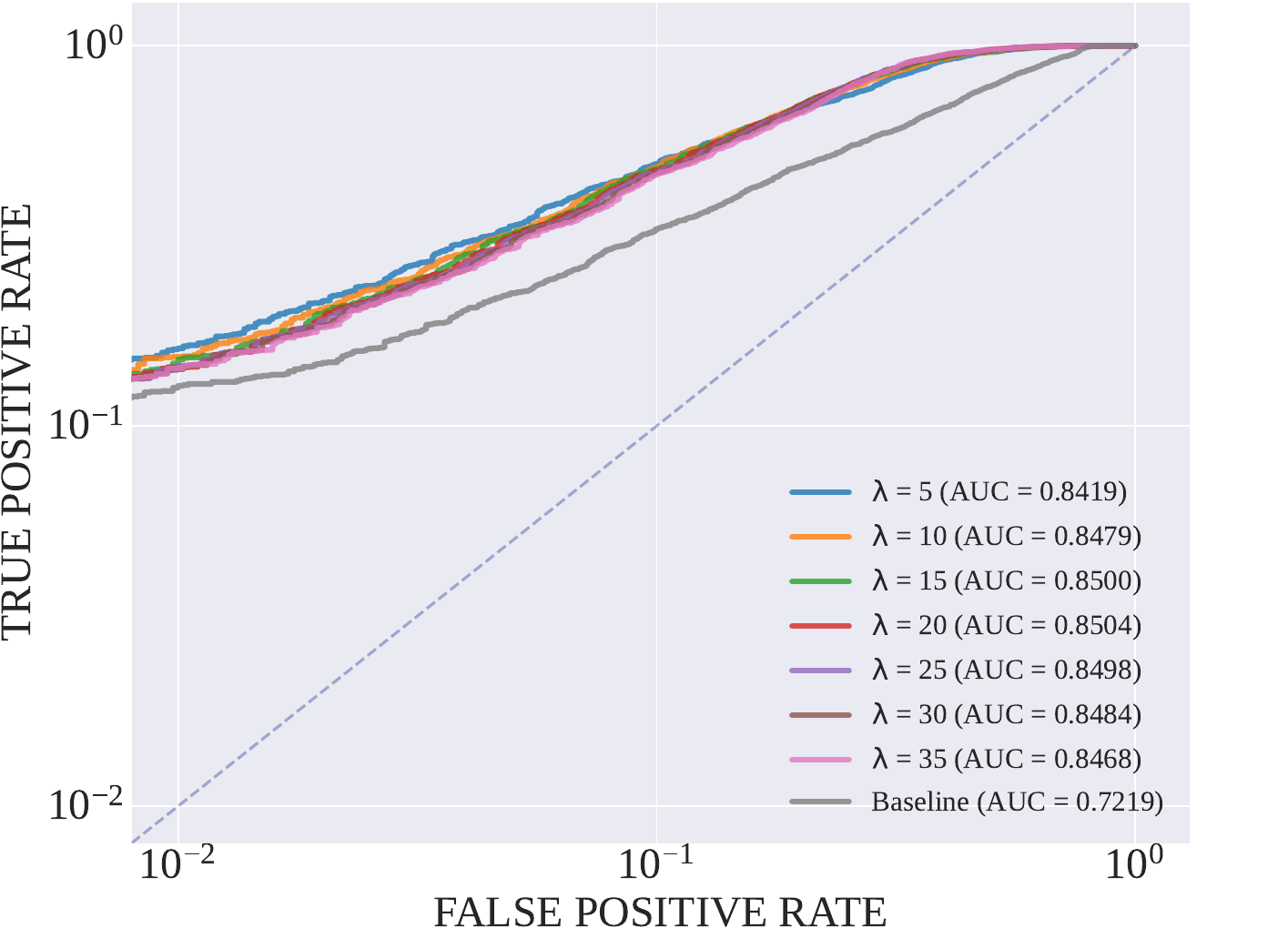}
        \subcaption{Adversarially Robust RMIA (log scale)}
    \end{minipage}
    \caption{\footnotesize Comparison of ROC Curves for Our Adversarially Robust MIAs and Baselines on \textbf{CINIC-10}.}
    \label{Robust_MIA_cinic}
    \end{center}
    \vspace{-1em}
\end{figure*}

\vspace{0.2em}
\noindent\textbf{Experimental Results of \textbf{AR-MIAs}.} In \Cref{Robust_MIA_cifar10,Robust_MIA_cifar100,Robust_MIA_svhn,Robust_MIA_cinic}, we show that our \textbf{AR-MIAs} can be effectively combined with three existing strong MIAs—Attack R, LiRA, and RMIA—and significantly improve the baseline AUC value and TPR @ low FPR across different datasets. The quantitative comparison results can be found in \Cref{tab:robust_MIA_Attack_R,tab:robust_MIA_LiRA,tab:robust_MIA_RMIA}, where the best results are highlighted in red.

\clearpage

\clearpage

\clearpage

\clearpage

\clearpage

\clearpage

\clearpage

\clearpage

\clearpage

\clearpage

\clearpage

\clearpage

\clearpage

\clearpage

\clearpage

\begin{table*}[!t]
\vspace{-0.1in}
\setlength{\tabcolsep}{5mm}
\scriptsize\sc
\renewcommand\arraystretch{1}
\centering
\caption{The Comparison of \textbf{Error Area} between Our Member Fabrication Attack with Baselines across Diverse Datasets. }
\label{tab:fabric_error_area}
\resizebox{1.0\textwidth}{!}{
\begin{tabular}{lccccccc}
\toprule[1.5pt]
\rowcolor[HTML]{D9D9D9}
Perturbation & Natural & I-FGSM & I-BIM & I-PGD & I-CW & I-APGD & OURS\\
\midrule[0.1pt]
\midrule[0.1pt]
\rowcolor[HTML]{F2F2F2}
\multicolumn{8}{c}{\textbf{CIFAR-10}} \\
\midrule[0.1pt]
\midrule[0.1pt]
 $\|\delta\|_{\infty} \leq 1.0/255$ & 0.3685 & 0.5649 & 0.5955 & 0.5953 & 0.5818 & 0.5942 & \textcolor{red}{0.5969} \\
 $\|\delta\|_{\infty} \leq 2.0/255$ & 0.3685 & 0.6773 & 0.7723 & 0.7719 & 0.7477 & 0.7679 & \textcolor{red}{0.7761} \\
 $\|\delta\|_{\infty} \leq 3.0/255$ & 0.3685 & 0.7374 & 0.8833 & 0.8827 & 0.8593 & 0.8783 & \textcolor{red}{0.8887} \\
 $\|\delta\|_{\infty} \leq 4.0/255$ & 0.3685 & 0.7706 & 0.9403 & 0.9399 & 0.9228 & 0.9369 & \textcolor{red}{0.9451} \\
 $\|\delta\|_{\infty} \leq 5.0/255$ & 0.3685 & 0.7883 & 0.9684 & 0.9680 & 0.9568 & 0.9657 & \textcolor{red}{0.9719} \\
 $\|\delta\|_{\infty} \leq 6.0/255$ & 0.3685 & 0.7962 & 0.9828 & 0.9826 & 0.9755 & 0.9805 & \textcolor{red}{0.9852} \\
 $\|\delta\|_{\infty} \leq 7.0/255$ & 0.3685 & 0.7973 & 0.9903 & 0.9903 & 0.9854 & 0.9886 & \textcolor{red}{0.9922} \\
 $\|\delta\|_{\infty} \leq 8.0/255$ & 0.3685 & 0.7934 & 0.9946 & 0.9943 & 0.9913 & 0.9931 & \textcolor{red}{0.9958} \\
\midrule[0.1pt]
\midrule[0.1pt]
\rowcolor[HTML]{F2F2F2}
\multicolumn{8}{c}{\textbf{CIFAR-100}} \\
\midrule[0.1pt]
\midrule[0.1pt]
 $\|\delta\|_{\infty} \leq 1.0/255$ & 0.1039 & 0.1814 & 0.1948 & 0.1947 & 0.1857 & 0.1943 & \textcolor{red}{0.1955} \\
 $\|\delta\|_{\infty} \leq 2.0/255$ & 0.1039 & 0.2547 & 0.3238 & 0.3235 & 0.2971 & 0.3211 & \textcolor{red}{0.3276} \\
 $\|\delta\|_{\infty} \leq 3.0/255$ & 0.1039 & 0.3092 & 0.4664 & 0.4657 & 0.4255 & 0.4616 & \textcolor{red}{0.4771} \\
 $\|\delta\|_{\infty} \leq 4.0/255$ & 0.1039 & 0.3445 & 0.5935 & 0.5917 & 0.5466 & 0.5880 & \textcolor{red}{0.6136} \\
 $\|\delta\|_{\infty} \leq 5.0/255$ & 0.1039 & 0.3626 & 0.6942 & 0.6925 & 0.6489 & 0.6905 & \textcolor{red}{0.7225} \\
 $\|\delta\|_{\infty} \leq 6.0/255$ & 0.1039 & 0.3670 & 0.7730 & 0.7712 & 0.7316 & 0.7695 & \textcolor{red}{0.8069} \\
 $\|\delta\|_{\infty} \leq 7.0/255$ & 0.1039 & 0.3622 & 0.8346 & 0.8317 & 0.7979 & 0.8309 & \textcolor{red}{0.8682} \\
 $\|\delta\|_{\infty} \leq 8.0/255$ & 0.1039 & 0.3512 & 0.8781 & 0.8748 & 0.8472 & 0.8763 & \textcolor{red}{0.9101} \\
\midrule[0.1pt]
\midrule[0.1pt]
\rowcolor[HTML]{F2F2F2}
\multicolumn{8}{c}{\textbf{SVHN}} \\
\midrule[0.1pt]
\midrule[0.1pt]
 $\|\delta\|_{\infty} \leq 1.0/255$ & 0.4448 & 0.6587 & 0.7042 & 0.7042 & 0.6763 & 0.7023 & \textcolor{red}{0.7056} \\
 $\|\delta\|_{\infty} \leq 2.0/255$ & 0.4448 & 0.7382 & 0.8511 & 0.8512 & 0.8146 & 0.8489 & \textcolor{red}{0.8555} \\
 $\|\delta\|_{\infty} \leq 3.0/255$ & 0.4448 & 0.7539 & 0.9209 & 0.9212 & 0.8898 & 0.9210 & \textcolor{red}{0.9279} \\
 $\|\delta\|_{\infty} \leq 4.0/255$ & 0.4448 & 0.7407 & 0.9554 & 0.9548 & 0.9318 & 0.9560 & \textcolor{red}{0.9618} \\
 $\|\delta\|_{\infty} \leq 5.0/255$ & 0.4448 & 0.7109 & 0.9734 & 0.9733 & 0.9550 & 0.9745 & \textcolor{red}{0.9794} \\
 $\|\delta\|_{\infty} \leq 6.0/255$ & 0.4448 & 0.6709 & 0.9839 & 0.9833 & 0.9699 & 0.9848 & \textcolor{red}{0.9886} \\
 $\|\delta\|_{\infty} \leq 7.0/255$ & 0.4448 & 0.6261 & 0.9895 & 0.9893 & 0.9793 & 0.9906 & \textcolor{red}{0.9933} \\
 $\|\delta\|_{\infty} \leq 8.0/255$ & 0.4448 & 0.5792 & 0.9934 & 0.9934 & 0.9851 & 0.9938 & \textcolor{red}{0.9963} \\
\midrule[0.1pt]
\midrule[0.1pt]
\rowcolor[HTML]{F2F2F2}
\multicolumn{8}{c}{\textbf{CINIC-10}} \\
\midrule[0.1pt]
\midrule[0.1pt]
 $\|\delta\|_{\infty} \leq 1.0/255$ & 0.2688 & 0.4522 & 0.4921 & 0.4919 & 0.4770 & 0.4900 & \textcolor{red}{0.4936} \\
 $\|\delta\|_{\infty} \leq 2.0/255$ & 0.2688 & 0.5698 & 0.7054 & 0.7051 & 0.6749 & 0.6995 & \textcolor{red}{0.7113} \\
 $\|\delta\|_{\infty} \leq 3.0/255$ & 0.2688 & 0.6316 & 0.8450 & 0.8442 & 0.8138 & 0.8383 & \textcolor{red}{0.8552} \\
 $\|\delta\|_{\infty} \leq 4.0/255$ & 0.2688 & 0.6619 & 0.9207 & 0.9198 & 0.8971 & 0.9162 & \textcolor{red}{0.9324} \\
 $\|\delta\|_{\infty} \leq 5.0/255$ & 0.2688 & 0.6750 & 0.9597 & 0.9590 & 0.9438 & 0.9565 & \textcolor{red}{0.9694} \\
 $\|\delta\|_{\infty} \leq 6.0/255$ & 0.2688 & 0.6779 & 0.9788 & 0.9783 & 0.9696 & 0.9768 & \textcolor{red}{0.9857} \\
 $\|\delta\|_{\infty} \leq 7.0/255$ & 0.2688 & 0.6751 & 0.9885 & 0.9880 & 0.9821 & 0.9868 & \textcolor{red}{0.9931} \\
 $\|\delta\|_{\infty} \leq 8.0/255$ & 0.2688 & 0.6679 & 0.9934 & 0.9931 & 0.9896 & 0.9922 & \textcolor{red}{0.9966} \\
\midrule[0.1pt]
\midrule[0.1pt]
\rowcolor[HTML]{F2F2F2}
\multicolumn{8}{c}{\textbf{ImageNet-100}} \\
\midrule[0.1pt]
\midrule[0.1pt]
 $\|\delta\|_{\infty} \leq 0.5/255$ & 0.4664 & 0.7231 & 0.7890 & 0.7888 & 0.7623 &0.7850 & \textcolor{red}{0.7920} \\
 $\|\delta\|_{\infty} \leq 1.0/255$ & 0.4664 & 0.8151 & 0.9334 & 0.9331 & 0.9135 & 0.9274 & \textcolor{red}{0.9381} \\
 $\|\delta\|_{\infty} \leq 1.5/255$ & 0.4664 & 0.8552 & 0.9799 & 0.9797 & 0.9712 & 0.9763 & \textcolor{red}{0.9834} \\
 $\|\delta\|_{\infty} \leq 2.0/255$ & 0.4664 & 0.8757 & 0.9932 & 0.9931 & 0.9902 & 0.9918 & \textcolor{red}{0.9954} \\
\bottomrule[1.5pt]
\end{tabular}
}
\end{table*}

\clearpage

\begin{table*}[!t]
\vspace{-0.1in}
\setlength{\tabcolsep}{10mm}
\scriptsize\sc
\renewcommand\arraystretch{1}
\centering
\caption{The \textbf{Error Area} of Our Member Fabrication Attack across Diverse MIAs and Datasets. }
\label{tab:fabric_error_area_more_MIAs}
\resizebox{1.0\textwidth}{!}{
\begin{tabular}{lcccc}
\toprule[1.5pt]
\rowcolor[HTML]{D9D9D9}
Perturbation & Loss & LiRA & Attack R & RMIA \\
\midrule[0.1pt]
\midrule[0.1pt]
\rowcolor[HTML]{F2F2F2}
\multicolumn{5}{c}{\textbf{CIFAR-10}} \\
\midrule[0.1pt]
\midrule[0.1pt]
 Natural & 0.3685 & 0.2814 & 0.3110 & 0.2830 \\
 $\|\delta\|_{\infty} \leq 4.0/255$ & \textcolor{red}{0.9451} & \textcolor{red}{0.3523} & \textcolor{red}{0.8852} & \textcolor{red}{0.3502} \\
\midrule[0.1pt]
\midrule[0.1pt]
\rowcolor[HTML]{F2F2F2}
\multicolumn{5}{c}{\textbf{CIFAR-100}} \\
\midrule[0.1pt]
\midrule[0.1pt]
 Natural & 0.1040 & 0.0423 & 0.1021 & 0.0808 \\
 $\|\delta\|_{\infty} \leq 4.0/255$ & \textcolor{red}{0.6136} & \textcolor{red}{0.3398} & \textcolor{red}{0.5217} & \textcolor{red}{0.3178} \\
\midrule[0.1pt]
\midrule[0.1pt]
\rowcolor[HTML]{F2F2F2}
\multicolumn{5}{c}{\textbf{SVHN}} \\
\midrule[0.1pt]
\midrule[0.1pt]
 Natural & 0.4448 & 0.3936 & 0.4284 & 0.3885 \\
 $\|\delta\|_{\infty} \leq 4.0/255$ & \textcolor{red}{0.9618} & \textcolor{red}{0.5905} & \textcolor{red}{0.9372} & \textcolor{red}{0.5671} \\
\midrule[0.1pt]
\midrule[0.1pt]
\rowcolor[HTML]{F2F2F2}
\multicolumn{5}{c}{\textbf{CINIC-10}} \\
\midrule[0.1pt]
\midrule[0.1pt]
 Natural & 0.2688 & 0.1532 & 0.2164 & 0.1704 \\
 $\|\delta\|_{\infty} \leq 4.0/255$ & \textcolor{red}{0.9324} & \textcolor{red}{0.5001} & \textcolor{red}{0.8371} & \textcolor{red}{0.3859} \\
\bottomrule[1.5pt]
\end{tabular}
}
\end{table*}

\begin{table*}[!t]
\vspace{-0.1in}
\setlength{\tabcolsep}{10mm}
\scriptsize\sc
\renewcommand\arraystretch{1}
\centering
\caption{The \textbf{Equal Error Rate} of Our Member Fabrication Attack across Diverse MIAs and Datasets. }
\label{tab:fabric_eer_more_MIAs}
\resizebox{1.0\textwidth}{!}{
\begin{tabular}{lcccc}
\toprule[1.5pt]
\rowcolor[HTML]{D9D9D9}
Perturbation & Loss & LiRA & Attack R & RMIA \\
\midrule[0.1pt]
\midrule[0.1pt]
\rowcolor[HTML]{F2F2F2}
\multicolumn{5}{c}{\textbf{CIFAR-10}} \\
\midrule[0.1pt]
\midrule[0.1pt]
 Natural & 42.30\% & 36.70\% & 36.35\% & 36.40\% \\
 $\|\delta\|_{\infty} \leq 4.0/255$ & \textcolor{red}{87.60\%} & \textcolor{red}{41.85\%} & \textcolor{red}{82.25\%} & \textcolor{red}{38.20\%} \\
\midrule[0.1pt]
\midrule[0.1pt]
\rowcolor[HTML]{F2F2F2}
\multicolumn{5}{c}{\textbf{CIFAR-100}} \\
\midrule[0.1pt]
\midrule[0.1pt]
 Natural & 15.60\% & 12.10\% & 17.40\% & 18.30\% \\
 $\|\delta\|_{\infty} \leq 4.0/255$ & \textcolor{red}{59.80\%} & \textcolor{red}{41.00\%} & \textcolor{red}{67.20\%} & \textcolor{red}{39.05\%} \\
\midrule[0.1pt]
\midrule[0.1pt]
\rowcolor[HTML]{F2F2F2}
\multicolumn{5}{c}{\textbf{SVHN}} \\
\midrule[0.1pt]
\midrule[0.1pt]
 Natural & 46.90\% & 43.90\% & 46.40\% & 43.75\% \\
 $\|\delta\|_{\infty} \leq 4.0/255$ & \textcolor{red}{89.70\%} & \textcolor{red}{60.45\%} & \textcolor{red}{88.20\%} & \textcolor{red}{53.25\%} \\
\midrule[0.1pt]
\midrule[0.1pt]
\rowcolor[HTML]{F2F2F2}
\multicolumn{5}{c}{\textbf{CINIC-10}} \\
\midrule[0.1pt]
\midrule[0.1pt]
 Natural & 33.50\% & 26.00\% & 29.70\% & 27.55\% \\
 $\|\delta\|_{\infty} \leq 4.0/255$ & \textcolor{red}{86.85\%} & \textcolor{red}{51.95\%} & \textcolor{red}{81.95\%} & \textcolor{red}{42.00\%} \\
\bottomrule[1.5pt]
\end{tabular}
}
\end{table*}

\begin{table*}[!t]
\vspace{-0.1in}
\setlength{\tabcolsep}{5mm}
\scriptsize\sc
\renewcommand\arraystretch{1}
\centering
\caption{The Comparison of \textbf{Equal Error Rate} between Our Member Fabrication Attack with Baselines across Diverse Datasets.}
\label{tab:fabric_eer}
\resizebox{1.0\textwidth}{!}{
\begin{tabular}{lccccccc}
\toprule[1.5pt]
\rowcolor[HTML]{D9D9D9}
Perturbation & Natural & I-FGSM & I-BIM & I-PGD & I-CW & I-APGD & OURS\\
\midrule[0.1pt]
\midrule[0.1pt]
\rowcolor[HTML]{F2F2F2}
\multicolumn{8}{c}{\textbf{CIFAR-10}} \\
\midrule[0.1pt]
\midrule[0.1pt]
 $\|\delta\|_{\infty} \leq 1.0/255$ & 42.30\% & 56.60\% & 58.80\% & 58.75\% & 57.90\% & 58.65\% & \textcolor{red}{58.90\%} \\
 $\|\delta\|_{\infty} \leq 2.0/255$ & 42.30\% & 65.00\% & 72.10\% & 72.10\% & 70.00\% & 71.85\% & \textcolor{red}{72.40\%} \\
 $\|\delta\|_{\infty} \leq 3.0/255$ & 42.30\% & 69.40\% & 80.55\% & 80.55\% & 77.95\% & 80.10\% & \textcolor{red}{81.00\%} \\
 $\|\delta\|_{\infty} \leq 4.0/255$ & 42.30\% & 71.80\% & 86.80\% & 86.75\% & 84.50\% & 86.30\% & \textcolor{red}{87.60\%} \\
 $\|\delta\|_{\infty} \leq 5.0/255$ & 42.30\% & 73.20\% & 90.85\% & 90.75\% & 89.05\% & 90.50\% & \textcolor{red}{91.45\%} \\
 $\|\delta\|_{\infty} \leq 6.0/255$ & 42.30\% & 73.75\% & 93.20\% & 93.25\% & 91.80\% & 92.85\% & \textcolor{red}{93.85\%} \\
 $\|\delta\|_{\infty} \leq 7.0/255$ & 42.30\% & 73.95\% & 95.15\% & 95.05\% & 93.85\% & 94.80\% & \textcolor{red}{95.55\%} \\
 $\|\delta\|_{\infty} \leq 8.0/255$ & 42.30\% & 73.60\% & 96.40\% & 96.30\% & 95.45\% & 96.00\% & \textcolor{red}{96.75\%} \\
\midrule[0.1pt]
\midrule[0.1pt]
\rowcolor[HTML]{F2F2F2}
\multicolumn{8}{c}{\textbf{CIFAR-100}} \\
\midrule[0.1pt]
\midrule[0.1pt]
 $\|\delta\|_{\infty} \leq 1.0/255$ & 15.60\% & 23.65\% & \textcolor{red}{24.80\%} & \textcolor{red}{24.80\%} & 24.35\% & \textcolor{red}{24.80\%} & \textcolor{red}{24.80\%} \\
 $\|\delta\|_{\infty} \leq 2.0/255$ & 15.60\% & 30.75\% & 36.80\% & 36.75\% & 34.55\% & 36.50\% & \textcolor{red}{37.10\%} \\
 $\|\delta\|_{\infty} \leq 3.0/255$ & 15.60\% & 35.30\% & 48.20\% & 48.25\% & 45.00\% & 48.20\% & \textcolor{red}{49.05\%} \\
 $\|\delta\|_{\infty} \leq 4.0/255$ & 15.60\% & 38.55\% & 58.05\% & 58.05\% & 54.75\% & 57.85\% & \textcolor{red}{59.80\%} \\
 $\|\delta\|_{\infty} \leq 5.0/255$ & 15.60\% & 40.20\% & 66.50\% & 66.45\% & 62.75\% & 66.40\% & \textcolor{red}{68.95\%} \\
 $\|\delta\|_{\infty} \leq 6.0/255$ & 15.60\% & 40.20\% & 73.20\% & 72.95\% & 69.75\% & 72.80\% & \textcolor{red}{75.80\%} \\
 $\|\delta\|_{\infty} \leq 7.0/255$ & 15.60\% & 39.75\% & 78.25\% & 78.15\% & 75.25\% & 78.25\% & \textcolor{red}{82.00\%} \\
 $\|\delta\|_{\infty} \leq 8.0/255$ & 15.60\% & 39.00\% & 82.55\% & 82.05\% & 79.55\% & 82.30\% & \textcolor{red}{86.05\%} \\
\midrule[0.1pt]
\midrule[0.1pt]
\rowcolor[HTML]{F2F2F2}
\multicolumn{8}{c}{\textbf{SVHN}} \\
\midrule[0.1pt]
\midrule[0.1pt]
 $\|\delta\|_{\infty} \leq 1.0/255$ & 46.90\% & 61.75\% & 65.30\% & 65.30\% & 63.20\% & 65.10\% & \textcolor{red}{65.35\%} \\
 $\|\delta\|_{\infty} \leq 2.0/255$ & 46.90\% & 68.40\% & 77.60\% & 77.65\% & 73.90\% & 77.40\% & \textcolor{red}{77.95\%} \\
 $\|\delta\|_{\infty} \leq 3.0/255$ & 46.90\% & 69.40\% & 84.70\% & 84.75\% & 81.05\% & 84.65\% & \textcolor{red}{85.25\%} \\
 $\|\delta\|_{\infty} \leq 4.0/255$ & 46.90\% & 69.20\% & 88.75\% & 88.60\% & 85.30\% & 88.95\% & \textcolor{red}{89.70\%} \\
 $\|\delta\|_{\infty} \leq 5.0/255$ & 46.90\% & 66.65\% & 91.25\% & 91.20\% & 88.40\% & 91.60\% & \textcolor{red}{92.55\%} \\
 $\|\delta\|_{\infty} \leq 6.0/255$ & 46.90\% & 64.10\% & 93.00\% & 93.10\% & 90.35\% & 93.70\% & \textcolor{red}{94.50\%} \\
 $\|\delta\|_{\infty} \leq 7.0/255$ & 46.90\% & 60.35\% & 94.45\% & 94.55\% & 92.30\% & 95.10\% & \textcolor{red}{96.05\%} \\
 $\|\delta\|_{\infty} \leq 8.0/255$ & 46.90\% & 56.80\% & 95.50\% & 95.55\% & 93.45\% & 96.00\% & \textcolor{red}{97.10\%} \\
\midrule[0.1pt]
\midrule[0.1pt]
\rowcolor[HTML]{F2F2F2}
\multicolumn{8}{c}{\textbf{CINIC-10}} \\
\midrule[0.1pt]
\midrule[0.1pt]
 $\|\delta\|_{\infty} \leq 1.0/255$ & 33.50\% & 48.40\% & 51.15\% & 51.10\% & 50.30\% & 50.90\% & \textcolor{red}{51.35\%} \\
 $\|\delta\|_{\infty} \leq 2.0/255$ & 33.50\% & 57.45\% & 66.85\% & 66.75\% & 64.80\% & 66.35\% & \textcolor{red}{67.20\%} \\
 $\|\delta\|_{\infty} \leq 3.0/255$ & 33.50\% & 61.90\% & 77.70\% & 77.95\% & 75.40\% & 77.45\% & \textcolor{red}{78.80\%} \\
 $\|\delta\|_{\infty} \leq 4.0/255$ & 33.50\% & 64.25\% & 85.35\% & 85.25\% & 82.55\% & 85.00\% & \textcolor{red}{86.85\%} \\
 $\|\delta\|_{\infty} \leq 5.0/255$ & 33.50\% & 65.35\% & 90.45\% & 90.45\% & 88.10\% & 90.30\% & \textcolor{red}{91.80\%} \\
 $\|\delta\|_{\infty} \leq 6.0/255$ & 33.50\% & 65.35\% & 93.25\% & 93.15\% & 91.50\% & 92.95\% & \textcolor{red}{94.85\%} \\
 $\|\delta\|_{\infty} \leq 7.0/255$ & 33.50\% & 65.10\% & 95.25\% & 95.20\% & 93.70\% & 95.05\% & \textcolor{red}{96.50\%} \\
 $\|\delta\|_{\infty} \leq 8.0/255$ & 33.50\% & 64.20\% & 96.60\% & 96.35\% & 95.50\% & 96.25\% & \textcolor{red}{97.70\%} \\
\midrule[0.1pt]
\midrule[0.1pt]
\rowcolor[HTML]{F2F2F2}
\multicolumn{8}{c}{\textbf{ImageNet-100}} \\
\midrule[0.1pt]
\midrule[0.1pt]
 $\|\delta\|_{\infty} \leq 0.5/255$ & 47.55\% & 66.45\% & 72.55\% & 72.55\% & 69.75\% & 71.80\% & \textcolor{red}{72.90\%} \\
 $\|\delta\|_{\infty} \leq 1.0/255$ & 47.55\% & 73.90\% & 85.60\% & 85.45\% & 83.30\% & 84.90\% & \textcolor{red}{86.15\%} \\
 $\|\delta\|_{\infty} \leq 1.5/255$ & 47.55\% & 77.30\% & 92.45\% & 92.55\% & 90.70\% & 91.70\% & \textcolor{red}{93.35\%} \\
 $\|\delta\|_{\infty} \leq 2.0/255$ & 47.55\% & 79.20\% & 95.95\% & 96.05\% & 94.75\% & 95.25\% & \textcolor{red}{96.85\%} \\
\bottomrule[1.5pt]
\end{tabular}
}
\end{table*}

\begin{table*}[!t]
\vspace{-0.1in}
\setlength{\tabcolsep}{3.5mm}
\scriptsize\sc
\renewcommand\arraystretch{1}
\centering
\caption{Comparison of Attack R and Our Adversarially Robust Attack R.}
\label{tab:robust_MIA_Attack_R}
\resizebox{1.0\textwidth}{!}{
\begin{tabular}{lcccccc}
\toprule[1.5pt]
\rowcolor[HTML]{D9D9D9}
Methods & \textbf{AUC} value & EER value & TPR@1\%FPR & TPR@5\%FPR & TPR@10\%FPR &TPR@20\%FPR \\
\midrule[0.1pt]
\midrule[0.1pt]
\rowcolor[HTML]{F2F2F2}
\multicolumn{7}{c}{\textbf{CIFAR-10} $(\|\delta\|_{\infty} \leq 4.0/255)$} \\
\midrule[0.1pt]
\midrule[0.1pt]
Attack R Baseline  & 0.4019 & 58.20\% & 0.37\% & 1.84\% & 3.68\% & 7.35\% \\
\textcolor{red}{[OURS]} $\lambda = 5$   & 0.7745 & 31.30\% & 9.80\% & 25.16\% & 35.25\% & 52.25\% \\
\textcolor{red}{[OURS]} $\lambda = 10$  & 0.7831 & 31.08\% & 11.05\% & 26.15\% & 37.55\% & 55.55\% \\
\textcolor{red}{[OURS]} $\lambda = 15$  & 0.7862 & 30.15\% & 10.85\% & \textcolor{red}{26.75\%} & 37.55\% & 56.08\% \\
\textcolor{red}{[OURS]} $\lambda = 20$  & \textcolor{red}{0.7871} & 29.80\% & \textcolor{red}{11.13\%} & 25.80\% & \textcolor{red}{39.46\%} & 56.10\% \\
\textcolor{red}{[OURS]} $\lambda = 25$  & 0.7870 & 29.78\% & 11.05\% & 25.30\% & 39.05\% & 55.70\% \\ 
\textcolor{red}{[OURS]} $\lambda = 30$  & 0.7864 & 29.68\% & 10.95\% & 24.85\% & 37.95\% & 55.82\% \\
\textcolor{red}{[OURS]} $\lambda = 35$  & 0.7855 & \textcolor{red}{29.58\%} & 10.20\% & 24.85\% & 37.57\% & \textcolor{red}{56.43\%} \\
\midrule[0.1pt]
\midrule[0.1pt]
\rowcolor[HTML]{F2F2F2}
\multicolumn{7}{c}{\textbf{CIFAR-100} $(\|\delta\|_{\infty} \leq 4.0/255)$} \\
\midrule[0.1pt]
\midrule[0.1pt]
Attack R Baseline  & 0.6881 & 35.90\% & 1.79\% & 8.96\% & 17.91\% & 35.82\% \\
\textcolor{red}{[OURS]} $\lambda = 5$   & 0.7176 & 33.67\% & 1.10\% & 5.48\% & 29.75\% & 47.95\% \\
\textcolor{red}{[OURS]} $\lambda = 10$  & 0.7531 & 31.32\% & 1.93\% & 9.64\% & 29.75\% & 55.85\% \\
\textcolor{red}{[OURS]} $\lambda = 15$  & 0.7684 & 29.23\% & 2.37\% & 11.85\% & 29.75\% & 58.90\% \\
\textcolor{red}{[OURS]} $\lambda = 20$  & 0.7768 & 28.40\% & 2.69\% & 13.45\% & 29.75\% & 60.93\% \\
\textcolor{red}{[OURS]} $\lambda = 25$  & 0.7817 & 28.45\% & 2.84\% & 14.21\% & 29.75\% & 61.38\% \\ 
\textcolor{red}{[OURS]} $\lambda = 30$  & 0.7852 & 27.52\% & 2.99\% & 14.96\% & 29.93\% & 62.71\% \\
\textcolor{red}{[OURS]} $\lambda = 35$  & \textcolor{red}{0.7881} & \textcolor{red}{27.12\%} & \textcolor{red}{3.16\%} & \textcolor{red}{15.80\%} & \textcolor{red}{31.61\%} & \textcolor{red}{63.26\%} \\
\midrule[0.1pt]
\midrule[0.1pt]
\rowcolor[HTML]{F2F2F2}
\multicolumn{7}{c}{\textbf{SVHN} $(\|\delta\|_{\infty} \leq 4.0/255)$} \\
\midrule[0.1pt]
\midrule[0.1pt]
Attack R Baseline  & 0.3172 & 62.10\% & 0.14\% & 0.68\% & 1.35\% & 2.71\% \\
\textcolor{red}{[OURS]} $\lambda = 5$   & \textcolor{red}{0.7185} & \textcolor{red}{33.75\%} & 5.30\% & \textcolor{red}{18.97\%} & \textcolor{red}{28.20\%} & \textcolor{red}{43.60\%} \\
\textcolor{red}{[OURS]} $\lambda = 10$  & 0.6837 & 36.02\% & \textcolor{red}{5.80\%} & 16.70\% & 27.20\% & 42.75\% \\
\textcolor{red}{[OURS]} $\lambda = 15$  & 0.6530 & 39.17\% & 5.42\% & 15.40\% & 25.45\% & 40.45\% \\
\textcolor{red}{[OURS]} $\lambda = 20$  & 0.6264 & 41.05\% & 5.40\% & 14.25\% & 23.05\% & 37.40\% \\
\textcolor{red}{[OURS]} $\lambda = 25$  & 0.6035 & 43.33\% & 5.20\% & 12.55\% & 20.90\% & 34.75\% \\ 
\textcolor{red}{[OURS]} $\lambda = 30$  & 0.5836 & 45.00\% & 4.58\% & 11.42\% & 19.03\% & 32.08\% \\
\textcolor{red}{[OURS]} $\lambda = 35$  & 0.5660 & 45.95\% & 4.10\% & 10.95\% & 17.70\% & 29.35\% \\
\midrule[0.1pt]
\midrule[0.1pt]
\rowcolor[HTML]{F2F2F2}
\multicolumn{7}{c}{\textbf{CINIC-10} $(\|\delta\|_{\infty} \leq 4.0/255)$} \\
\midrule[0.1pt]
\midrule[0.1pt]
Attack R Baseline  & 0.4732 & 53.02\% & 0.50\% & 2.49\% & 4.98\% & 9.97\% \\
\textcolor{red}{[OURS]} $\lambda = 5$   & 0.8103 & 27.68\% & \textcolor{red}{10.85\%} & \textcolor{red}{31.23\%} & 46.37\% & 62.66\% \\
\textcolor{red}{[OURS]} $\lambda = 10$  & \textcolor{red}{0.8167} & 26.75\% & \textcolor{red}{10.85\%} & 31.15\% & \textcolor{red}{46.50\%} & 64.35\% \\
\textcolor{red}{[OURS]} $\lambda = 15$  & \textcolor{red}{0.8167} & 26.45\% & \textcolor{red}{10.85\%} & 28.45\% & 45.58\% & 64.51\% \\
\textcolor{red}{[OURS]} $\lambda = 20$  & 0.8148 & \textcolor{red}{26.10\%} & \textcolor{red}{10.85\%} & 26.85\% & 45.10\% & \textcolor{red}{64.80\%} \\
\textcolor{red}{[OURS]} $\lambda = 25$  & 0.8122 & 26.38\% & 10.38\% & 25.37\% & 44.55\% & 64.47\% \\ 
\textcolor{red}{[OURS]} $\lambda = 30$  & 0.8094 & 26.60\% & 10.13\% & 24.40\% & 42.43\% & 64.55\% \\
\textcolor{red}{[OURS]} $\lambda = 35$  & 0.8066 & 26.90\% & 9.75\% & 23.60\% & 41.25\% & 64.35\% \\
\bottomrule[1.5pt]
\end{tabular}
}
\end{table*}

\begin{table*}[!t]
\vspace{-0.1in}
\setlength{\tabcolsep}{3.5mm}
\scriptsize\sc
\renewcommand\arraystretch{1}
\centering
\caption{Comparison of LiRA and Our Adversarially Robust LiRA.}
\label{tab:robust_MIA_LiRA}
\resizebox{1.0\textwidth}{!}{
\begin{tabular}{lcccccc}
\toprule[1.5pt]
\rowcolor[HTML]{D9D9D9}
Methods & \textbf{AUC} value & EER value & TPR@1\%FPR & TPR@5\%FPR & TPR@10\%FPR &TPR@20\%FPR \\
\midrule[0.1pt]
\midrule[0.1pt]
\rowcolor[HTML]{F2F2F2}
\multicolumn{7}{c}{\textbf{CIFAR-10} $(\|\delta\|_{\infty} \leq 4.0/255)$} \\
\midrule[0.1pt]
\midrule[0.1pt]
LiRA Baseline  & 0.6832 & 38.48\% & 2.80\% & 9.55\% & 17.55\% & 32.25\% \\
\textcolor{red}{[OURS]} $\lambda = 5$  & 0.7925 & 28.95\% & \textcolor{red}{4.55\%} & \textcolor{red}{18.45\%} & \textcolor{red}{34.35\%} & \textcolor{red}{57.05\%} \\
\textcolor{red}{[OURS]} $\lambda = 10$  & \textcolor{red}{0.7937} & \textcolor{red}{28.23\%} & 3.60\% & 17.95\% & 32.95\% & 57.00\% \\
\textcolor{red}{[OURS]} $\lambda = 15$  & 0.7927 & 28.43\% & 3.55\% & 17.05\% & 31.95\% & 55.50\% \\
\textcolor{red}{[OURS]} $\lambda = 20$  & 0.7913 & 28.60\% & 3.50\% & 16.95\% & 31.25\% & 54.25\% \\
\textcolor{red}{[OURS]} $\lambda = 25$  & 0.7897 & 28.62\% & 3.50\% & 16.60\% & 30.95\% & 54.05\% \\ 
\textcolor{red}{[OURS]} $\lambda = 30$  & 0.7882 & 28.85\% & 3.50\% & 16.50\% & 29.90\% & 52.40\% \\
\textcolor{red}{[OURS]} $\lambda = 35$  & 0.7869 & 29.03\% & 3.30\% & 16.20\% & 29.65\% & 52.00\% \\
\midrule[0.1pt]
\midrule[0.1pt]
\rowcolor[HTML]{F2F2F2}
\multicolumn{7}{c}{\textbf{CIFAR-100} $(\|\delta\|_{\infty} \leq 4.0/255)$} \\
\midrule[0.1pt]
\midrule[0.1pt]
LiRA Baseline  & 0.8090 & 28.78\% & 9.90\% & 23.00\% & 34.40\% & 56.25\% \\
\textcolor{red}{[OURS]} $\lambda = 5$   & \textcolor{red}{0.8408} & \textcolor{red}{24.73\%} & \textcolor{red}{11.60\%} & \textcolor{red}{26.55\%} & \textcolor{red}{40.65\%} & \textcolor{red}{65.60\%} \\
\textcolor{red}{[OURS]} $\lambda = 10$  & 0.8387 & 25.45\% & 11.05\% & 25.95\% & 39.50\% & 64.30\% \\
\textcolor{red}{[OURS]} $\lambda = 15$  & 0.8368 & 25.50\% & 10.75\% & 25.00\% & 38.80\% & 63.95\% \\
\textcolor{red}{[OURS]} $\lambda = 20$  & 0.8353 & 25.65\% & 10.80\% & 24.55\% & 38.50\% & 63.50\% \\
\textcolor{red}{[OURS]} $\lambda = 25$  & 0.8340 & 25.77\% & 10.80\% & 23.90\% & 38.20\% & 62.65\% \\ 
\textcolor{red}{[OURS]} $\lambda = 30$  & 0.8329 & 26.05\% & 10.60\% & 23.90\% & 37.60\% & 61.95\% \\
\textcolor{red}{[OURS]} $\lambda = 35$  & 0.8319 & 26.35\% & 10.35\% & 23.80\% & 37.25\% & 61.25\% \\
\midrule[0.1pt]
\midrule[0.1pt]
\rowcolor[HTML]{F2F2F2}
\multicolumn{7}{c}{\textbf{SVHN} $(\|\delta\|_{\infty} \leq 4.0/255)$} \\
\midrule[0.1pt]
\midrule[0.1pt]
LiRA Baseline  & 0.5079 & 51.30\% & 1.05\% & 2.85\% & 5.30\% & 11.35\%\\
\textcolor{red}{[OURS]} $\lambda = 5$   & \textcolor{red}{0.7273} & \textcolor{red}{34.00\%} & \textcolor{red}{2.70\%} & \textcolor{red}{11.50\%} & \textcolor{red}{20.35\%} & \textcolor{red}{38.65\%} \\
\textcolor{red}{[OURS]} $\lambda = 10$  & 0.6828 & 37.60\% & 2.00\% & 8.75\% & 16.55\% & 29.50\% \\
\textcolor{red}{[OURS]} $\lambda = 15$  & 0.6493 & 40.35\% & 1.65\% & 7.40\% & 13.90\% & 24.50\% \\
\textcolor{red}{[OURS]} $\lambda = 20$  & 0.6250 & 42.40\% & 1.60\% & 6.40\% & 12.00\% & 22.70\% \\
\textcolor{red}{[OURS]} $\lambda = 25$  & 0.6068 & 43.65\% & 1.45\% & 5.95\% & 10.80\% & 20.20\% \\ 
\textcolor{red}{[OURS]} $\lambda = 30$  & 0.5926 & 44.65\% & 1.45\% & 5.40\% & 10.15\% & 18.60\% \\
\textcolor{red}{[OURS]} $\lambda = 35$  & 0.5812 & 45.50\% & 1.35\% & 5.15\% & 9.40\% & 17.50\% \\
\midrule[0.1pt]
\midrule[0.1pt]
\rowcolor[HTML]{F2F2F2}
\multicolumn{7}{c}{\textbf{CINIC-10} $(\|\delta\|_{\infty} \leq 4.0/255)$} \\
\midrule[0.1pt]
\midrule[0.1pt]
LiRA Baseline  & 0.6734 & 38.50\% & 1.50\% & 6.90\% & 13.00\% & 29.10\% \\
\textcolor{red}{[OURS]} $\lambda = 5$   & \textcolor{red}{0.8166} & \textcolor{red}{24.82\%} & \textcolor{red}{2.65\%} & \textcolor{red}{12.05\%} & \textcolor{red}{29.95\%} & \textcolor{red}{62.30\%} \\
\textcolor{red}{[OURS]} $\lambda = 10$  & 0.8091 & 25.85\% & 2.15\% & 10.90\% & 26.00\% & 59.40\% \\
\textcolor{red}{[OURS]} $\lambda = 15$  & 0.8033 & 26.47\% & 2.10\% & 10.35\% & 23.60\% & 57.30\% \\
\textcolor{red}{[OURS]} $\lambda = 20$  & 0.7983 & 27.20\% & 2.15\% & 9.85\% & 22.55\% & 55.10\% \\
\textcolor{red}{[OURS]} $\lambda = 25$  & 0.7942 & 27.70\% & 2.00\% & 9.45\% & 21.20\% & 53.10\% \\ 
\textcolor{red}{[OURS]} $\lambda = 30$  & 0.7905 & 28.18\% & 2.00\% & 9.30\% & 20.40\% & 51.80\% \\
\textcolor{red}{[OURS]} $\lambda = 35$  & 0.7872 & 28.50\% & 1.95\% & 9.15\% & 20.15\% & 51.10\% \\
\bottomrule[1.5pt]
\end{tabular}
}
\end{table*}

\begin{table*}[!t]
\vspace{-0.1in}
\setlength{\tabcolsep}{3.5mm}
\scriptsize\sc
\renewcommand\arraystretch{1}
\centering
\caption{Comparison of RMIA and Our Adversarially Robust RMIA.}
\label{tab:robust_MIA_RMIA}
\resizebox{1.0\textwidth}{!}{
\begin{tabular}{lcccccc}
\toprule[1.5pt]
\rowcolor[HTML]{D9D9D9}
Methods & \textbf{AUC} value & EER value & TPR@1\%FPR & TPR@5\%FPR & TPR@10\%FPR &TPR@20\%FPR \\
\midrule[0.1pt]
\midrule[0.1pt]
\rowcolor[HTML]{F2F2F2}
\multicolumn{7}{c}{\textbf{CIFAR-10} $(\|\delta\|_{\infty} \leq 4.0/255)$} \\
\midrule[0.1pt]
\midrule[0.1pt]
RMIA Baseline  & 0.6834 & 37.30\% & 8.55\% & 18.50\% & 28.95\% & 42.80\% \\
\textcolor{red}{[OURS]} $\lambda = 5$   & 0.7834 & 30.88\% & 10.90\% & 25.60\% & 36.40\% & 51.65\% \\
\textcolor{red}{[OURS]} $\lambda = 10$  & 0.7941 & 29.98\% & \textcolor{red}{11.30\%} & \textcolor{red}{25.85\%} & \textcolor{red}{38.15\%} & 55.65\% \\
\textcolor{red}{[OURS]} $\lambda = 15$  & 0.7988 & 29.23\% & 11.10\% & 25.70\% & 37.90\% & 56.90\% \\
\textcolor{red}{[OURS]} $\lambda = 20$  & 0.8021 & 28.82\% & 11.00\% & 25.45\% & 37.95\% & 58.25\% \\
\textcolor{red}{[OURS]} $\lambda = 25$  & 0.8039 & 28.50\% & 10.75\% & 25.30\% & 37.75\% & 58.50\% \\ 
\textcolor{red}{[OURS]} $\lambda = 30$  & 0.8056 & 28.05\% & 10.35\% & 25.05\% & 37.80\% & 59.20\% \\
\textcolor{red}{[OURS]} $\lambda = 35$  & \textcolor{red}{0.8065} & \textcolor{red}{27.85\%} & 10.30\% & 25.30\% & 37.85\% & \textcolor{red}{59.45\%} \\
\midrule[0.1pt]
\midrule[0.1pt]
\rowcolor[HTML]{F2F2F2}
\multicolumn{7}{c}{\textbf{CIFAR-100} $(\|\delta\|_{\infty} \leq 4.0/255)$} \\
\midrule[0.1pt]
\midrule[0.1pt]
RMIA Baseline  & 0.8007 & 29.80\% & 26.35\% & 34.15\% & 41.55\% & 56.15\% \\
\textcolor{red}{[OURS]} $\lambda = 5$   & 0.7894 & 30.18\% & 20.45\% & 29.65\% & 38.85\% & 56.10\% \\
\textcolor{red}{[OURS]} $\lambda = 10$  & 0.8123 & 27.73\% & 23.80\% & 32.75\% & 42.55\% & 59.85\% \\
\textcolor{red}{[OURS]} $\lambda = 15$  & 0.8203 & 27.57\% & 24.65\% & 33.65\% & 43.00\% & \textcolor{red}{60.80\%} \\
\textcolor{red}{[OURS]} $\lambda = 20$  & 0.8243 & \textcolor{red}{27.50\%} & 25.15\% & 34.30\% & \textcolor{red}{43.95\%} & 60.55\% \\
\textcolor{red}{[OURS]} $\lambda = 25$  & 0.8265 & \textcolor{red}{27.50\%} & 25.45\% & 34.45\% & 43.70\% & 60.50\% \\ 
\textcolor{red}{[OURS]} $\lambda = 30$  & 0.8277 & 27.52\% & 25.70\% & 34.75\% & 43.40\% & 60.35\% \\
\textcolor{red}{[OURS]} $\lambda = 35$  & \textcolor{red}{0.8285} & 27.62\% & \textcolor{red}{25.85\%} & \textcolor{red}{34.90\%} & 43.55\% & 60.15\% \\
\midrule[0.1pt]
\midrule[0.1pt]
\rowcolor[HTML]{F2F2F2}
\multicolumn{7}{c}{\textbf{SVHN} $(\|\delta\|_{\infty} \leq 4.0/255)$} \\
\midrule[0.1pt]
\midrule[0.1pt]
RMIA Baseline  & 0.5222 & 49.65\% & 7.10\% & 14.60\% & 20.95\% & 30.05\% \\
\textcolor{red}{[OURS]} $\lambda = 5$   & 0.7760 & 31.75\% & \textcolor{red}{7.55\%} & 17.70\% & 28.00\% & 48.45\% \\
\textcolor{red}{[OURS]} $\lambda = 10$  & \textcolor{red}{0.7815} & 30.55\% & 7.35\% & \textcolor{red}{17.80\%} & \textcolor{red}{28.70\%} & \textcolor{red}{48.90\%} \\
\textcolor{red}{[OURS]} $\lambda = 15$  & 0.7799 & \textcolor{red}{30.45\%} & 7.35\% & 17.25\% & 27.00\% & 47.80\% \\
\textcolor{red}{[OURS]} $\lambda = 20$  & 0.7790 & 30.50\% & 7.35\% & 17.30\% & 26.75\% & 46.75\% \\
\textcolor{red}{[OURS]} $\lambda = 25$  & 0.7771 & 30.95\% & 7.15\% & 17.15\% & 26.30\% & 46.75\% \\ 
\textcolor{red}{[OURS]} $\lambda = 30$  & 0.7724 & 31.27\% & 7.15\% & 16.90\% & 26.00\% & 46.55\% \\
\textcolor{red}{[OURS]} $\lambda = 35$  & 0.7680 & 31.95\% & 7.15\% & 16.15\% & 25.45\% & 45.00\% \\
\midrule[0.1pt]
\midrule[0.1pt]
\rowcolor[HTML]{F2F2F2}
\multicolumn{7}{c}{\textbf{CINIC-10} $(\|\delta\|_{\infty} \leq 4.0/255)$} \\
\midrule[0.1pt]
\midrule[0.1pt]
RMIA Baseline  & 0.7219 & 35.43\% & 12.75\% & 22.35\% & 32.85\% & 48.20\% \\
\textcolor{red}{[OURS]} $\lambda = 5$   & 0.8419 & 25.67\% & \textcolor{red}{16.00\%} & \textcolor{red}{33.60\%} & \textcolor{red}{48.90\%} & 67.50\% \\
\textcolor{red}{[OURS]} $\lambda = 10$  & 0.8479 & 24.32\% & 15.20\% & 32.00\% & 47.95\% & 69.35\% \\
\textcolor{red}{[OURS]} $\lambda = 15$  & 0.8500 & 23.93\% & 14.95\% & 31.95\% & 47.30\% & \textcolor{red}{69.85\%} \\
\textcolor{red}{[OURS]} $\lambda = 20$  & \textcolor{red}{0.8504} & \textcolor{red}{23.70\%} & 14.25\% & 31.30\% & 46.95\% & 69.50\% \\
\textcolor{red}{[OURS]} $\lambda = 25$  & 0.8498 & 23.80\% & 14.40\% & 31.20\% & 46.45\% & 68.95\% \\ 
\textcolor{red}{[OURS]} $\lambda = 30$  & 0.8484 & 24.00\% & 14.10\% & 30.70\% & 46.25\% & 67.15\% \\
\textcolor{red}{[OURS]} $\lambda = 35$  & 0.8468 & 24.15\% & 14.20\% & 29.30\% & 45.80\% & 66.50\% \\
\bottomrule[1.5pt]
\end{tabular}
}
\end{table*}

\end{document}